\documentclass[a4paper,fleqn]{cas-dc}

\usepackage[authoryear]{natbib}

\AtBeginDocument{\setlength{\FullWidth}{\textwidth}}

\usepackage{subcaption}
\usepackage{multirow}
\usepackage{alphalph}

\def\tsc#1{\csdef{#1}{\textsc{\lowercase{#1}}\xspace}}
\tsc{WGM}
\tsc{QE}
\tsc{EP}
\tsc{PMS}
\tsc{BEC}
\tsc{DE}

\usepackage{amsthm}
\usepackage{amsmath}
\usepackage{xspace}
\usepackage{amssymb}
\usepackage{bbm}
\usepackage{color}
\usepackage{pgfplots, pgfplotstable}
\usepackage[colorinlistoftodos]{todonotes}

\renewcommand{\vec}[1]{\boldsymbol{\mathbf{#1}}}

\def\bx{\vec{x}}

\def\mP{\mathcal{P}}
\def\Real{\mathbb{R}}

\def\lon{\lambda}
\def\lat{\phi}
\def\lons{\lon^{(s)}}
\def\lats{\lat^{(s)}}
\def\ths{\th^{(s)}}

\newcommand\latz[1]{\lat^{(#1)}}
\newcommand\lonz[1]{\lon^{(#1)}}

\def\s{s}

\def\fs{r^{(s)}}
\def\fm{r^{(m)}}
\def\fn{r^{(n)}}
\def\th{\vec{x}}
\def\sd{\Delta D} \def\ca{\Delta\delta}
\def\radius{R}

\newcommand{\dataset}{\mathcal{M}}

\newcommand{\nscale}{S}

\newcommand{\lr}{lr}

\newcommand{\numresnet}{h}
\newcommand{\numneuron}{k}

\newcommand{\maxscale}{r_{max}}
\newcommand{\minscale}{r_{min}}

\newcommand{\kernelsize}{\sigma}
\newcommand{\numkernel}{m}

\newcommand{\imgclsloss}{\mathcal{L}^{image}}
\newcommand{\lossweight}{\beta}
\newcommand{\sampleset}{\mathbb{X}}

\newcommand{\imgenc}{\mathbf{F}}
\newcommand{\imgcls}{\mathbf{Q}}

\newcommand{\embdim}{d}

\newcommand{\peemb}{\mathbf{p}[\th]}

\newcommand{\negsamp}{\mathcal{N}}

\newcommand{\freq}{S}

\newcommand{\enc}{Enc}

\newcommand{\unit}{\mathbf{a}}

\newcommand{\pemlp}{\mathbf{NN}}
\newcommand{\peffn}{\pemlp^{ffn}}

\newcommand{\coordspasphere}{\mathbb{S}}
\newcommand{\params}{\theta}

\newcommand{\image}{\mathbf{I}}

\newcommand{\classemb}{\mathbf{T}}
\newcommand{\numclass}{c}
\newcommand{\classy}{y}

\newcommand{\act}{\sigma}

\newcommand{\pefunc}{f}

\newcommand{\tile}{tile}
\newcommand{\aodha}{wrap}
\newcommand{\aodhaffn}{wrap + ffn}

\newcommand{\rbf}{rbf}
\newcommand{\rff}{rff}
\newcommand{\grid}{grid}
\newcommand{\theory}{theory}
\newcommand{\xyz}{xyz}
\newcommand{\nerf}{NeRF}

\newcommand{\spe}[1]{PE^{#1}} 

\newcommand{\sphere}{sphereC}
\newcommand{\spheregrid}{sphereC+}
\newcommand{\spheremixscale}{sphereM}
\newcommand{\spheregridmixscale}{sphereM+}
\newcommand{\dft}{dfs}

\newcommand{\shift}{b}
\newcommand{\dirvec}{\omega}
\newcommand{\rffenc}{\varphi}
\newcommand{\rffdim}{D}
\newcommand{\rffcov}{\delta}
\newcommand{\iidsim}{\overset{i.i.d}{\sim}}

\newcommand{\modelname}{Sphere2Vec}

\newcommand{\spacevec}{Space2Vec}

\newcommand{\vspaceclustering}{\vspace*{-0.6cm}}
\newcommand{\vspacepred}{\vspace*{-0.5cm}}

\newcommand{\vspacecomp}{\vspace*{-0.4cm}}

\newcommand{\numvmfcla}{\mathcal{C}}
\newcommand{\numvmfsamplepercla}{\mathcal{SP}}
\newcommand{\latlontoxyz}{\chi}
\newcommand{\vmf}{vMF}
\newcommand{\vmfmu}{\mu}
\newcommand{\numvmfMUinterval}{N_{\mu}}
\newcommand{\numvmfMUclaperinterval}{\numvmfcla_{\mu}}
\newcommand{\vmfkappa}{\kappa}
\newcommand{\vmfkappamin}{\kappa_{min}}
\newcommand{\vmfkappamax}{\kappa_{max}}

\newcommand\mai[1]{\textcolor{black}{{#1}}}

\newcommand\maigch[1]{\textcolor{black}{{#1}}} 

\begin{document}
\let\WriteBookmarks\relax
\def\floatpagepagefraction{1}
\def\textpagefraction{.001}

\shorttitle{\modelname}

\shortauthors{Mai et~al.}

\title [mode = title]{\modelname: A General-Purpose Location Representation Learning
over a Spherical Surface for Large-Scale Geospatial Predictions}

\author[1,4,5,6]{Gengchen Mai}[type=editor,
                        auid=000,
                        bioid=1,
orcid=0000-0002-7818-7309]

\cormark[1]
\fnmark[1]
\ead{gengchen.mai25@uga.edu}
\ead[url]{https://gengchenmai.github.io/}

\author[2]{Yao Xuan}[]
\ead{yxuan@ucsb.edu}
\fnmark[1]

\author[3]{Wenyun Zuo}[]

\author[4]{Yutong He}[]
\author[4]{Jiaming Song}[]
\author[4]{Stefano Ermon}[]

\author[5,6,7]
{Krzysztof Janowicz}

\author[8]
{Ni Lao}
\fnmark[2]

\address[1]{Spatially Explicit Artificial Intelligence Lab, Department of Geography, University of Georgia, Athens, Georgia, 30602, USA}
\address[2]{Department of Mathematics, University of California Santa Barbara, Santa Barbara, California, 93106, USA}
\address[3]{Department of Biology, Stanford University, Stanford, California, 94305, USA}
\address[4]{Department of Computer Science, Stanford University, Stanford, California, 94305, USA}
\address[5]{STKO Lab, Department of Geography, University of California Santa Barbara,  Santa Barbara, California, 93106, USA}
\address[6]{Center for Spatial Studies, University of California Santa Barbara,  Santa Barbara, California, 93106, USA}
\address[7]{Department of Geography and Regional Research, University of Vienna,  Vienna, 1040, Austria}
\address[8]{Mosaix.ai, Palo Alto, California, 94043, USA}

\cortext[cor1]{Corresponding author}

\fntext[fn1]{Both authors contribute equally to this work.}

\begin{abstract}
Generating learning-friendly representations for points in space 
is a fundamental and long-standing problem in machine learning. 
Recently, multi-scale encoding schemes (such as 
\spacevec \mai{~and \nerf}) were proposed to
directly encode any point in 2D \mai{or 3D Euclidean} space as a high-dimensional vector, 
and has been successfully applied to various (geo)spatial prediction \mai{and generative} tasks.
However, \mai{all current 2D and 3D location encoders are designed to model point distances in Euclidean space. So when applied to large-scale real-world GPS coordinate datasets (e.g., species or satellite images taken all over the world), which require distance metric learning on the spherical surface, both types of models can fail due to the \textit{map projection distortion problem} (2D) and the \textit{spherical-to-Euclidean distance approximation error} (3D). }
To solve \mai{these problems}, we propose a multi-scale location encoder called \emph{\modelname} which 
\mai{ can preserve spherical distances when encoding point coordinates on a spherical surface. }
\mai{We developed a unified view of distance-reserving encoding on spheres based on the Double Fourier Sphere (DFS).}
We \mai{also} provide theoretical proof that the 
\emph{\modelname} encoding preserves the spherical surface distance between any two points\mai{, while existing encoding schemes such as 
\spacevec~and \nerf~do not}. 
Experiments on 20 synthetic datasets show that \emph{\modelname} can outperform all baseline models including the state-of-the-art (SOTA) 
\mai{2D location encoder (i.e., \spacevec) and 3D encoder \nerf } 
on all these datasets with up to 30.8\% error rate reduction.
We then apply \emph{\modelname} to three geo-aware image classification tasks - fine-grained species recognition, Flickr image recognition, and remote sensing image classification.
Results on 7 real-world datasets show the superiority of \emph{\modelname} over multiple 2D \mai{and 3D} location encoders on all three tasks. 
Further analysis shows that \emph{\modelname} outperforms other location encoder models, especially in the polar regions and data-sparse areas because of its nature for spherical surface distance preservation. \maigch{Code and data of this work are available at \url{https://gengchenmai.github.io/sphere2vec-website/}.}

\end{abstract}

\begin{highlights}
    \item We propose a general-purpose spherical location encoder, \emph{\modelname}, which, as far as we know, is the first location encoder which aims at preserving spherical distance.

    \item We provide a theoretical proof about the spherical-distance-kept nature of \emph{\modelname}. 

    \item \mai{We provide theoretical proof to show why the previous 2D location encoders and NeRF-style 3D location encoders cannot model spherical distance correctly. }
    
    \item We first construct 20 synthetic datasets based on the mixture of von Mises-Fisher (MvMF) distributions and show that \emph{\modelname} can outperform all baseline models including the state-of-the-art (SOTA) 2D location encoders \mai{and NeRF-style 3D location encoders} on all these datasets with an up to 30.8\% error rate reduction.
        
    \item Next, we conduct extensive experiments on seven real-world datasets for three geo-aware image classification tasks.
    Results show that \emph{\modelname} outperforms all baseline models on all datasets. 

    \item Further analysis shows that \emph{\modelname} is able to produce finer-grained and compact spatial distributions, and does significantly better than 2D \mai{and 3D} Euclidean location encoders in the polar regions and areas with sparse training samples.

\end{highlights} 

\maketitle

\section{Introduction}
\label{sec:intro}

The fact that the Earth is round but not planar should surprise nobody \citep{chrisman2017calculating}. However, studying geospatial problems on a flat map with the plane analytical geometry \citep{boyer2012history} is still the common practice adopted by most of the geospatial community and well supported by all the softwares and technology of geographic information systems (GIS). 
Moreover, over the years, certain programmers and researchers have blurred the distinction between a (spherical) geographic coordinate system and a (planar) projected coordinate system \citep{chrisman2017calculating}, and directly treated latitude-longitude pairs as 2D Cartesian coordinates for analytical purpose. 
This distorted pseudo-projection results, so-called Plate Carrée, although remaining meaningless, have been unconsciously used in many scientific work across different disciplines. This blindness to the obvious round Earth and ignorance of the distortion brought by various map projections have led to tremendous negative effects and major mistakes. 
For example, 
typical mistakes brought by the Mercator projection are that it leads people to believe that Greenland is in the same size of Africa or Alaska looms larger than Mexico \citep{nyt2021map}. In fact, Greenland is no bigger than the Democratic Republic of Congo \citep{cnn2017map} and Alaska is smaller than Mexico.
A more extreme case about France was documented by 
\cite{harmel2009nouveau} during the period of the single area payment. After converting from the old national coordinate system (a Lambert conformal conic projection) to the new coordinate system (RGF 93), subsidies to the agriculture sector were reduced by 17 million euros because of the reduced scale error in the map projection.

\begin{figure*}
	\centering \tiny
\includegraphics[width=1\textwidth]{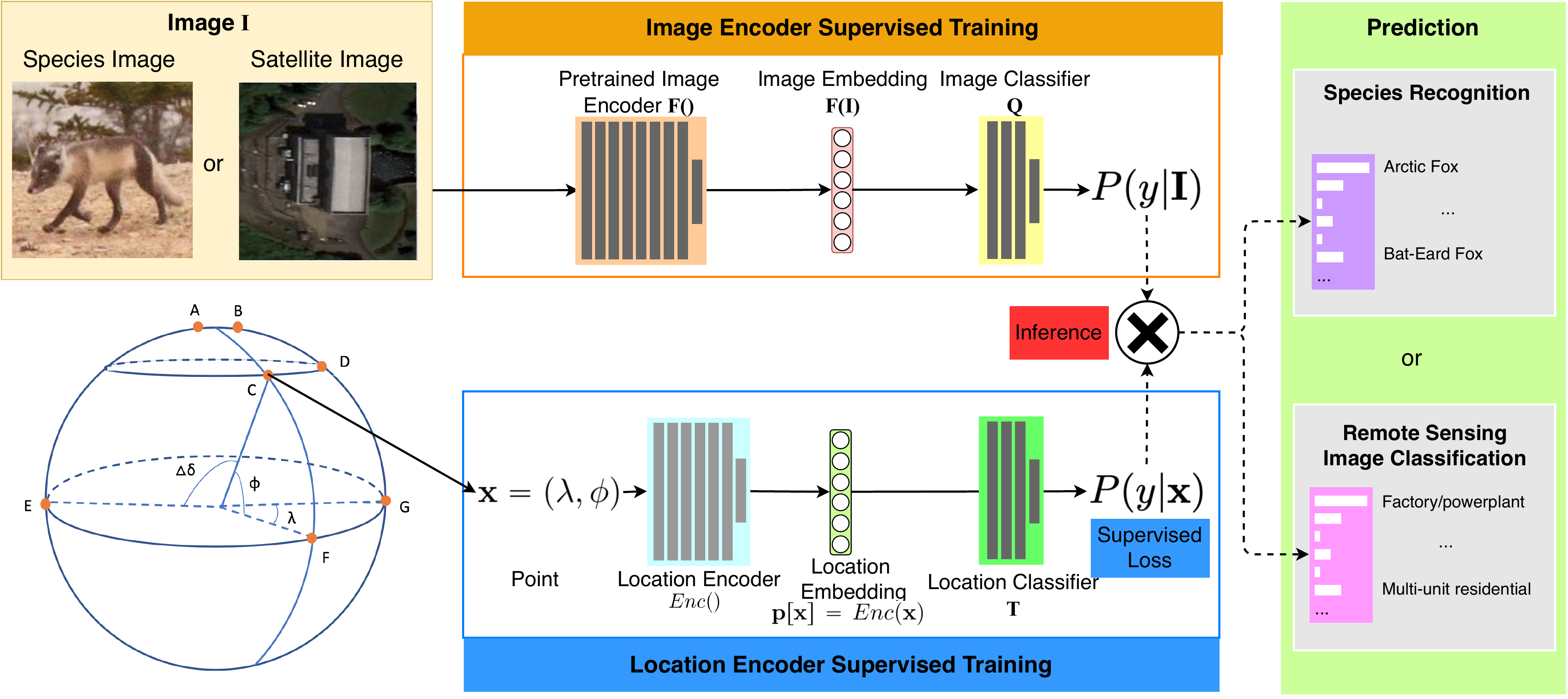}\vspace*{-0.1cm}
	\caption{
	Applying \emph{\modelname} to geo-aware image classification task. Here, we use the fine-grained species recognition and remote sensing (RS) image classification as examples. 
Given a species image $\image$, it is very difficult to decide whether it is an Arctic fox or \mai{a} gray fox just based on the appearance information. However, if we know this image is taken from the Arctic area, then we have more confidence to say this is an Arctic fox. 
	Similarly, an overhead remote sensing image of factories and multi-unit residential buildings might look similar. However, they locate in different neighborhoods with different land use types which can be estimated as geographic priors by a location encoder. 
	So the idea of geo-aware image classification is to combine (the red box) the predictions from an image encoder (the orange box) and a location encoder (the blue box).
	The image encoder \mai{(the orange box) can be a pretrained model} such as an  InceptionV3 network \citep{mac2019presence} for species recognition or a MoCo-V2+TP \citep{ayush2020selfsup} for the RS image classification. 
    \mai{We can append a separated image classifier $\imgcls$ at the end of the image encoder $\imgenc()$ and supervised fine-tune the whole image classification model on the corresponding training dataset to obtain the probability distribution of image labels for a given image $\image$, i.e., $P(\classy|\image)$. }
The location encoder (the blue box) can be \emph{\modelname} or \mai{any other inductive location encoders \citep{chu2019geo,mac2019presence,mai2020multiscale,mildenhall2020nerf}}. Supervised training \mai{of the location encoder $\enc()$ together} with a location classifier $\classemb$ can yield the geographic prior distributions of image labels $P(\classy|\th)$.
	The predictions from both components are combined (multiplied) to make the final prediction (the red box).
    The dotted lines indicates that there is no back-propagation through these lines.
	} 
	\label{fig:pos_enc}
	\vspace*{-0.15cm}
\end{figure*}

\begin{figure*}
	\centering \tiny
	\vspace*{-0.2cm}
	\begin{subfigure}[b]{0.10\textwidth}  
		\centering 
		\includegraphics[width=\textwidth]{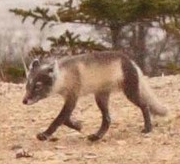}\vspace*{-0.2cm}
		\caption[]{{\small 
		Image
		}}    
		\label{fig:fox1}
	\end{subfigure}
	\hfill
	\begin{subfigure}[b]{0.22\textwidth}  
		\centering 
		\includegraphics[width=\textwidth]{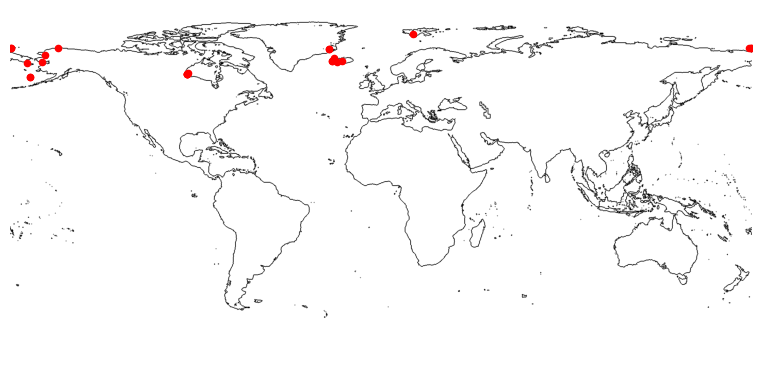}\vspacepred
		\caption[]{{\small 
		Arctic Fox }}    
		\label{fig:4084_dist_intro}
	\end{subfigure}
	\hfill
	\begin{subfigure}[b]{0.22\textwidth}  
		\centering 
		\includegraphics[width=\textwidth]{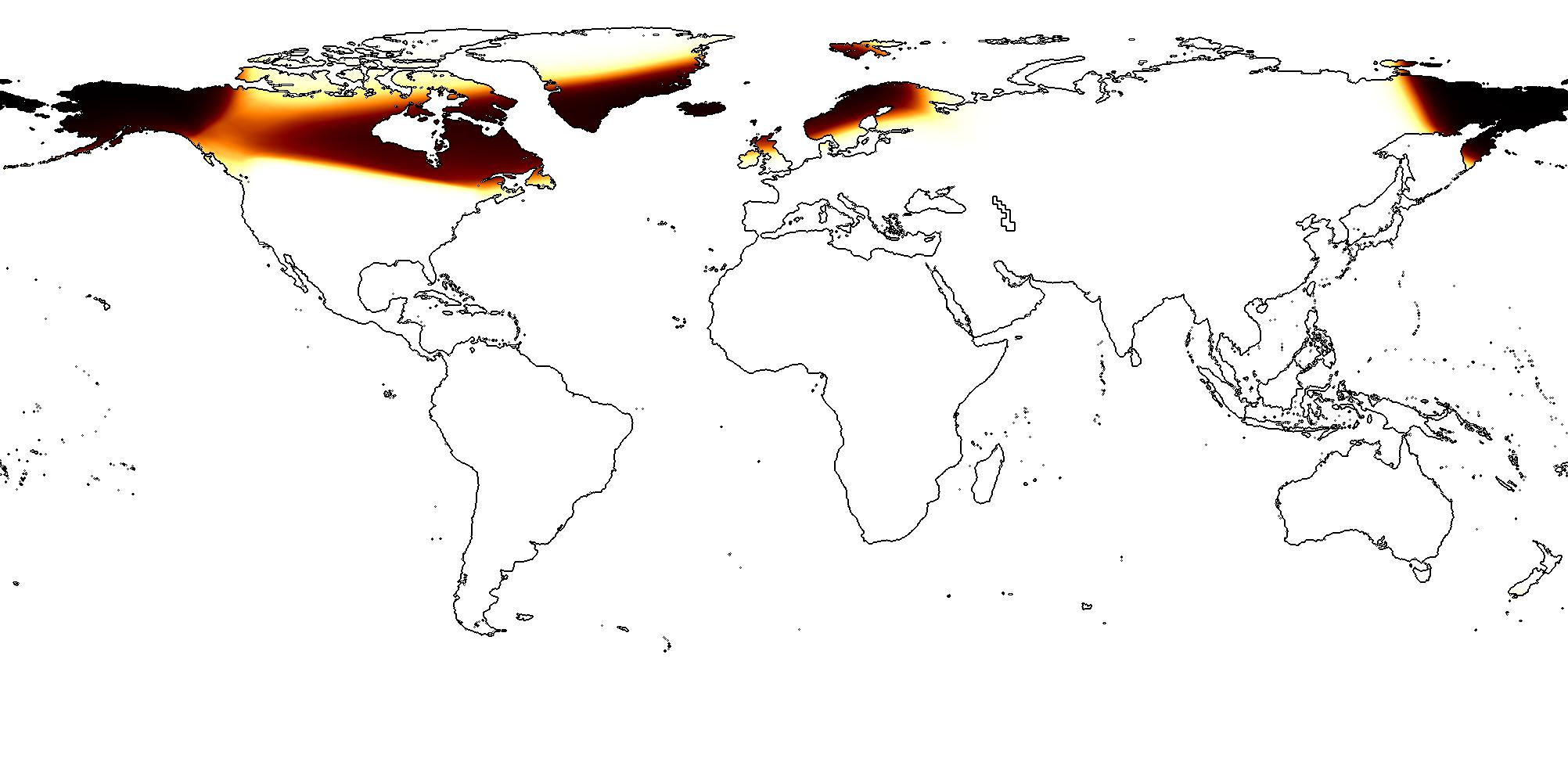}\vspacepred
		\caption[]{{\small 
		$\aodha$
		}}    
		\label{fig:4084_aodha_intro}
	\end{subfigure}
	\hfill
	\begin{subfigure}[b]{0.22\textwidth}  
		\centering 
		\includegraphics[width=\textwidth]{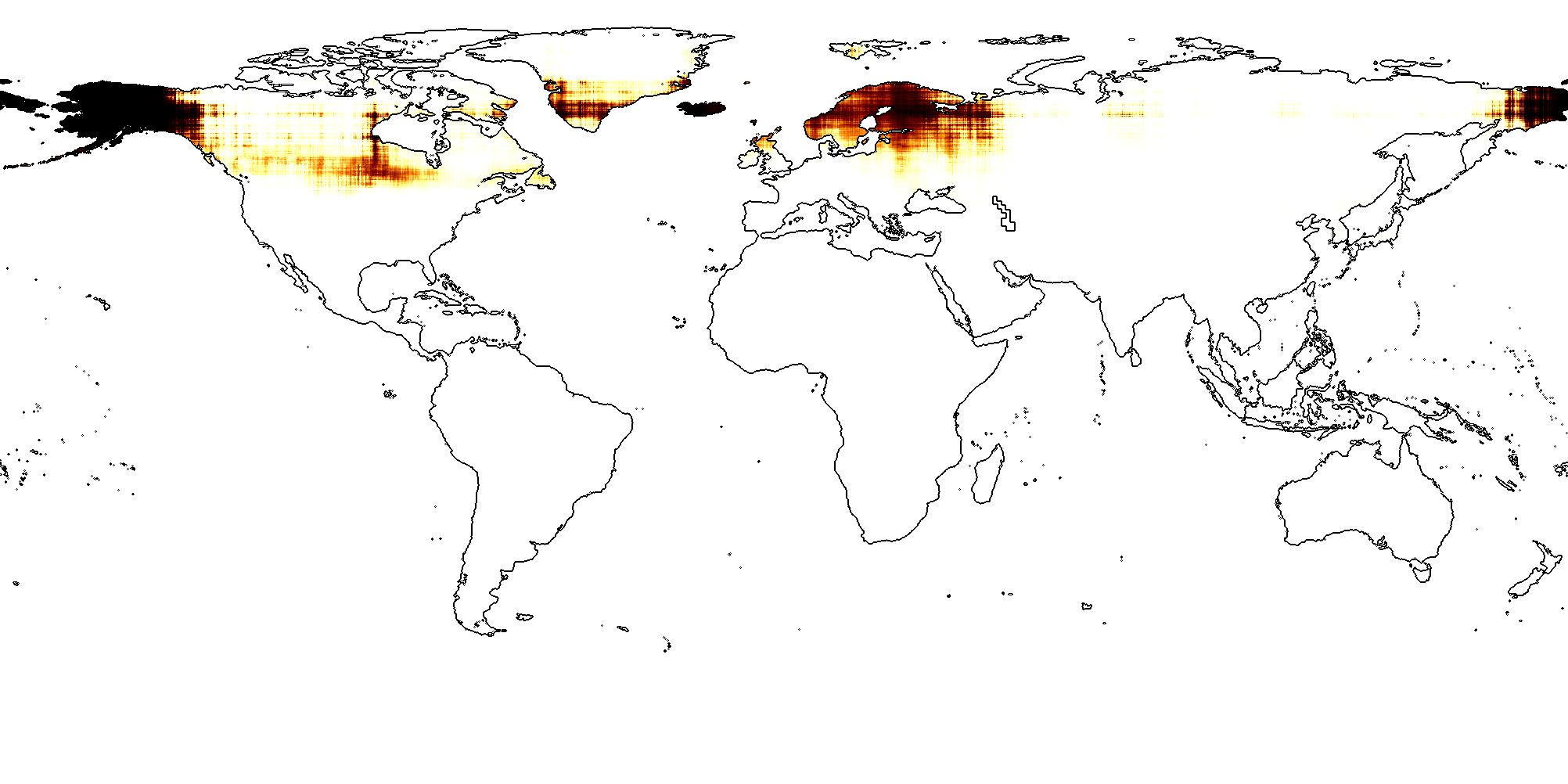}\vspacepred
		\caption[]{{\small 
		$\grid$
		}}    
		\label{fig:4084_grid_intro}
	\end{subfigure}
	\hfill
	\begin{subfigure}[b]{0.22\textwidth}  
		\centering 
		\includegraphics[width=\textwidth]{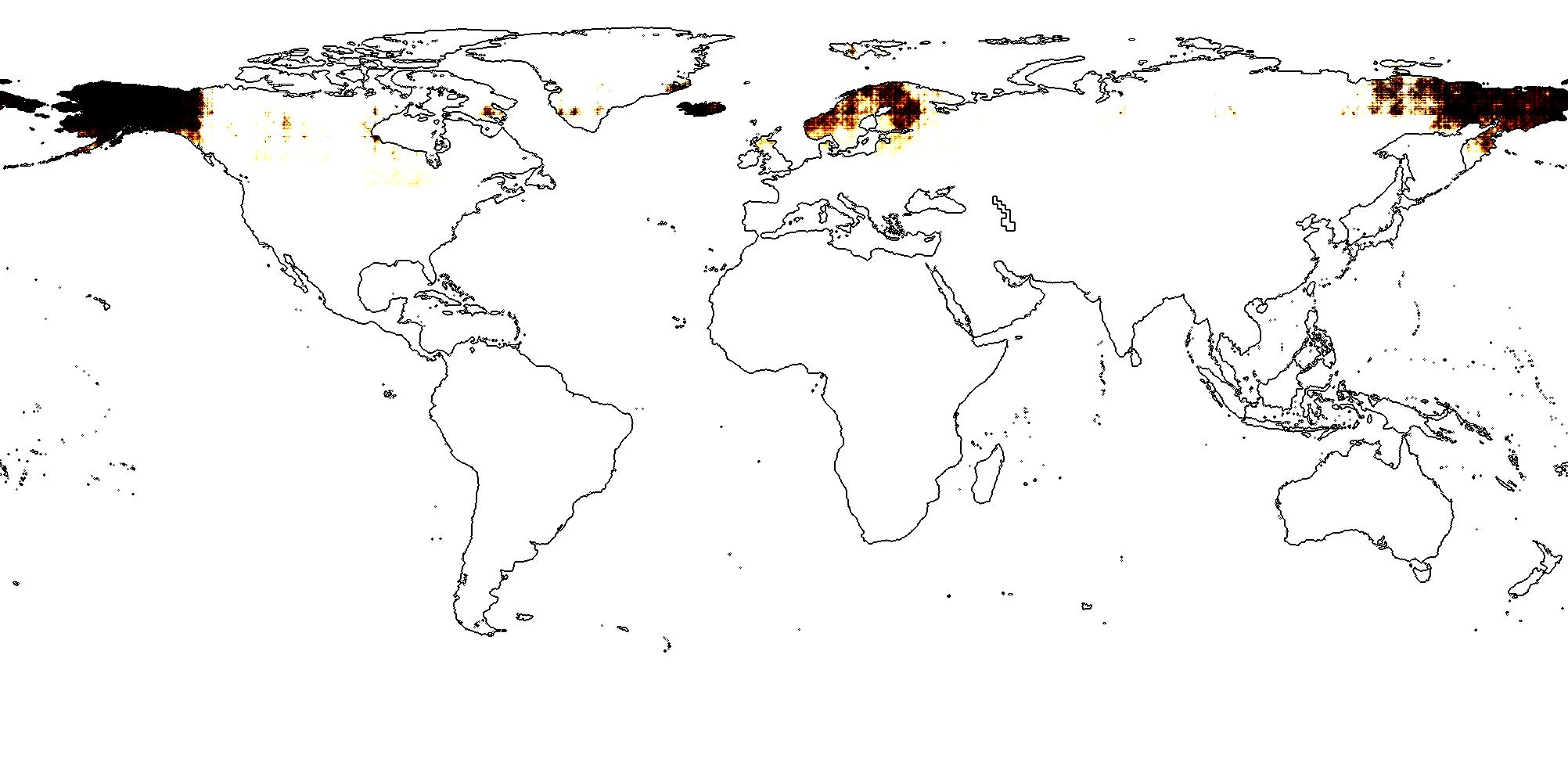}\vspacepred
		\caption[]{{\small 
		$\spheregrid$
		}}    
		\label{fig:4084_spheregrid_intro}
	\end{subfigure}
	
	\begin{subfigure}[b]{0.10\textwidth}  
		\centering 
		\includegraphics[width=\textwidth]{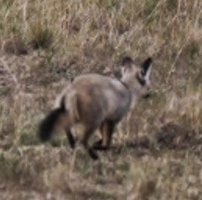}\vspace*{-0.2cm}
		\caption[]{{\small 
		Image
		}}    
		\label{fig:fox2}
	\end{subfigure}
	\hfill
	\begin{subfigure}[b]{0.22\textwidth}  
		\centering 
		\includegraphics[width=\textwidth]{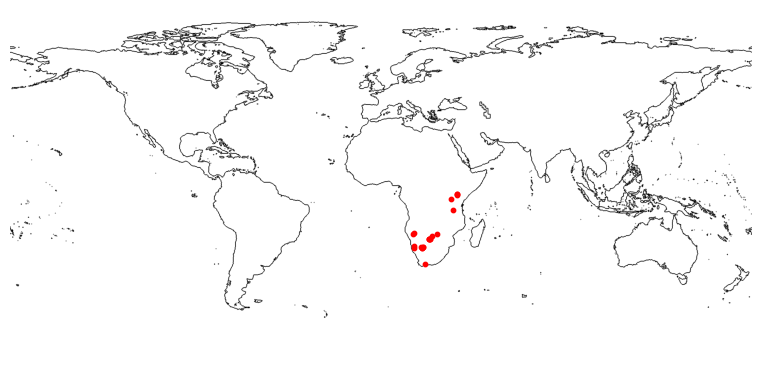}\vspacepred
		\caption[]{{\small 
		Bat-Eared Fox }}    
		\label{fig:4081_dist_intro}
	\end{subfigure}
	\hfill
	\begin{subfigure}[b]{0.22\textwidth}  
		\centering 
		\includegraphics[width=\textwidth]{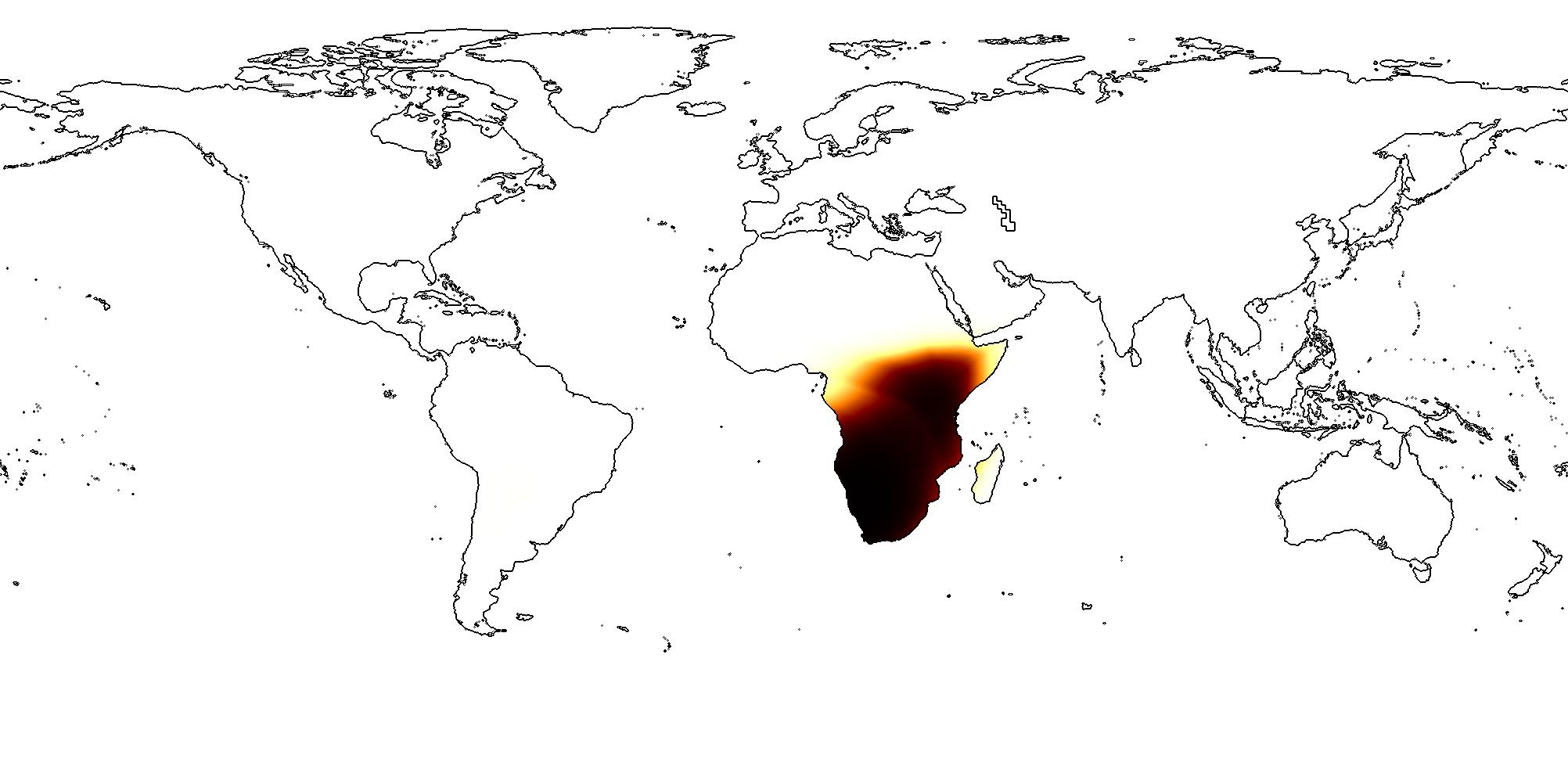}\vspacepred
		\caption[]{{\small 
		$\aodha$
		}}    
		\label{fig:4081_aodha_intro}
	\end{subfigure}
	\hfill
	\begin{subfigure}[b]{0.22\textwidth}  
		\centering 
		\includegraphics[width=\textwidth]{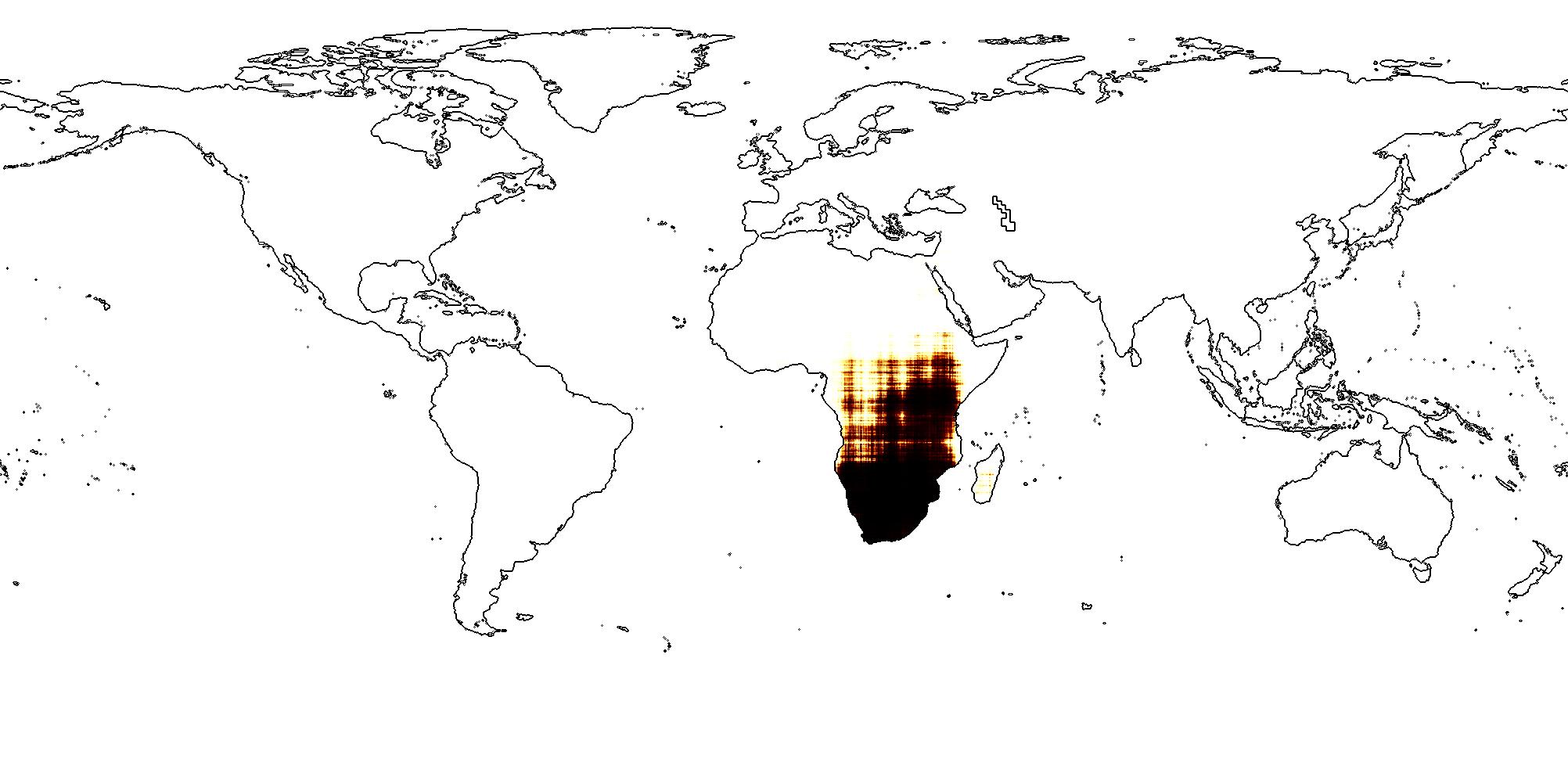}\vspacepred
		\caption[]{{\small 
		$\grid$
		}}    
		\label{fig:4081_grid_intro}
	\end{subfigure}
	\hfill
	\begin{subfigure}[b]{0.22\textwidth}  
		\centering 
		\includegraphics[width=\textwidth]{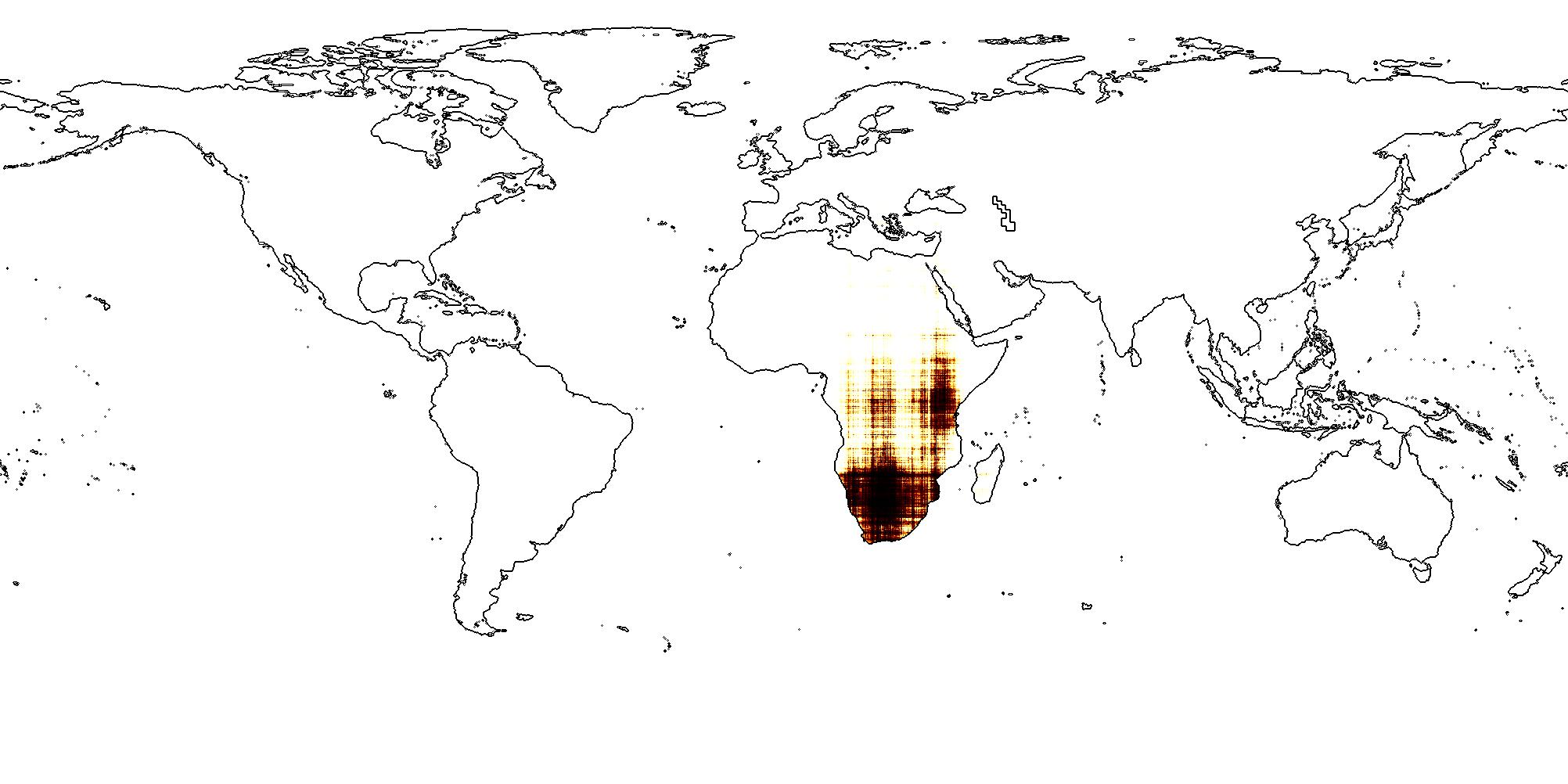}\vspacepred
		\caption[]{{\small 
		$\spheregrid$
		}}    
		\label{fig:4081_spheregrid_intro}
	\end{subfigure}
	
    \hfill
    \begin{subfigure}[b]{0.10\textwidth}  
		\centering 
		\includegraphics[width=\textwidth]{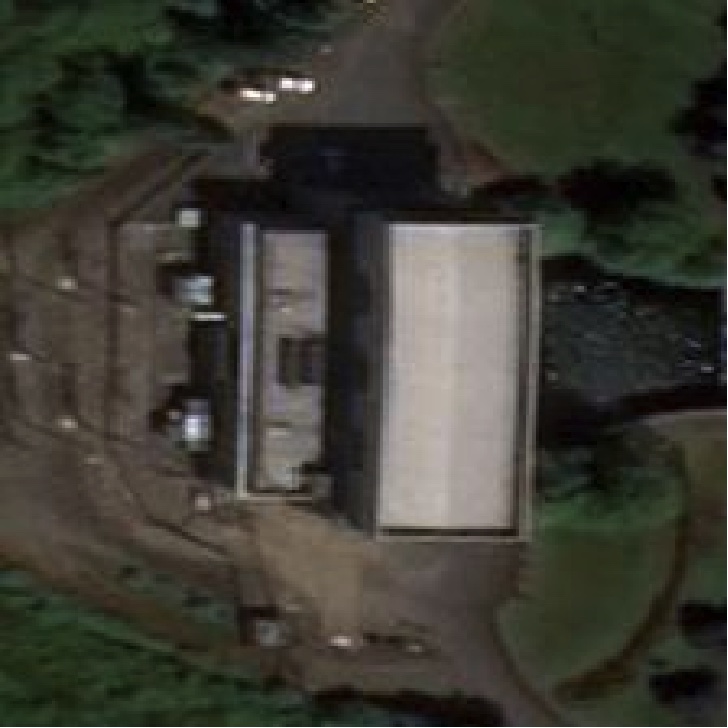}\vspace*{-0.2cm}
		\caption[]{{\small 
		Image
		}}    
		\label{fig:factory_or_powerplant_img_intro}
	\end{subfigure}
	\hfill
	\begin{subfigure}[b]{0.22\textwidth}  
		\centering 
		\includegraphics[width=\textwidth]{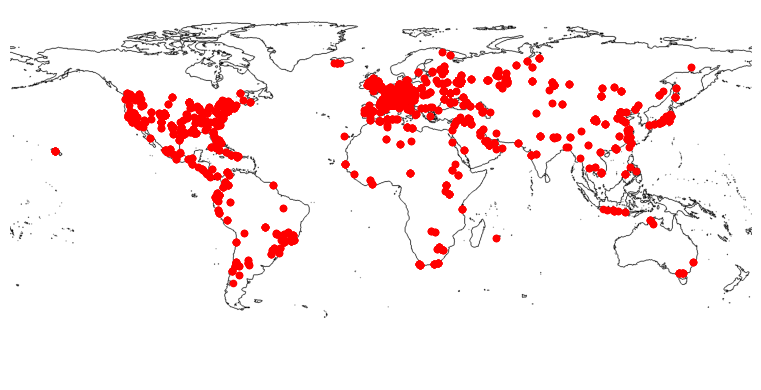}\vspacepred
		\caption[]{{\small 
		Factory or powerplant
		}}    
		\label{fig:factory_or_powerplant_loc_intro}
	\end{subfigure}
	\hfill
	\begin{subfigure}[b]{0.22\textwidth}  
		\centering 
		\includegraphics[width=\textwidth]{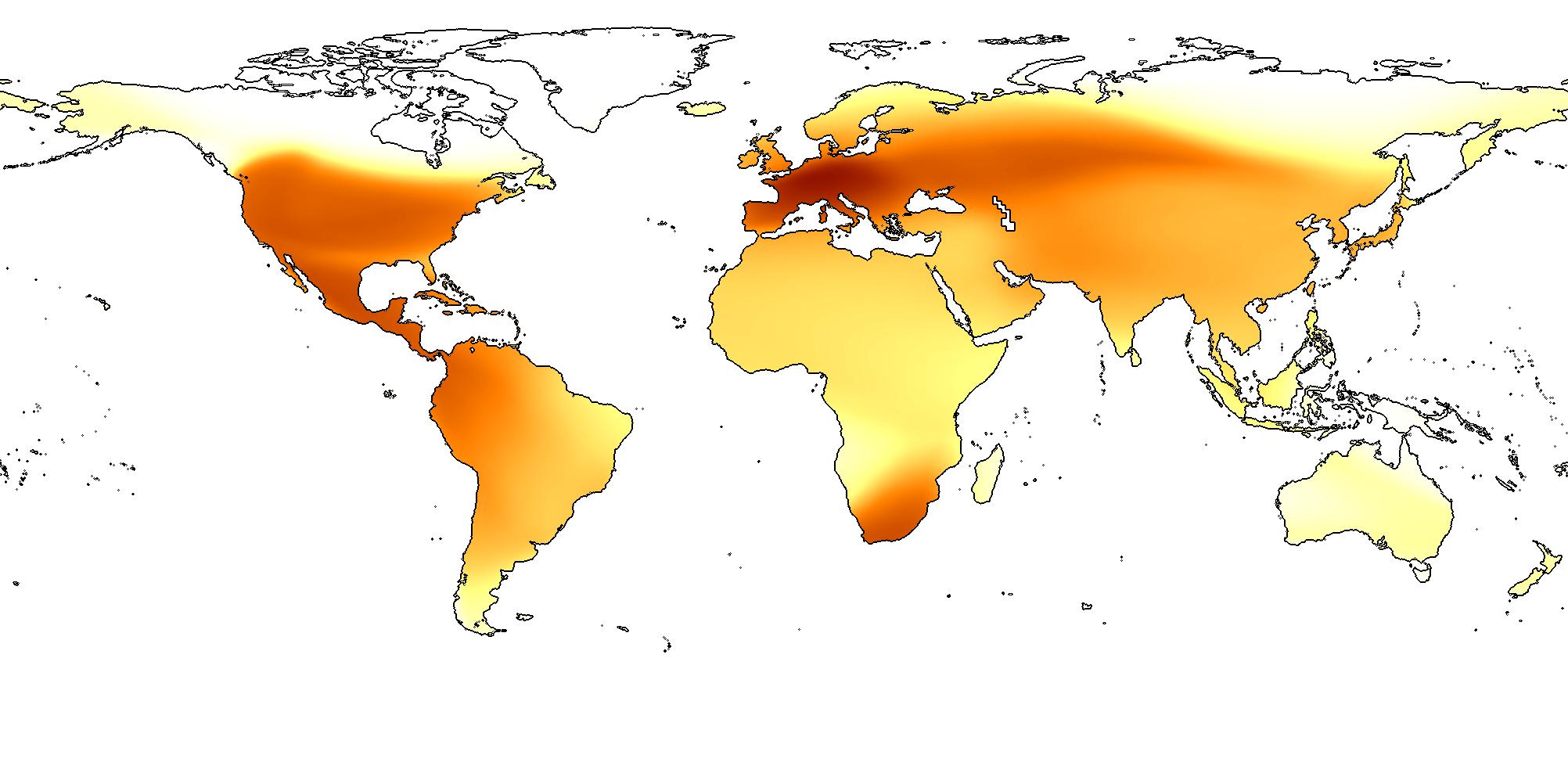}\vspacepred
		\caption[]{{\small 
		$\aodha$
		}}    
		\label{fig:factory_or_powerplant_aodha_pred_intro}
	\end{subfigure}
	\hfill
	\begin{subfigure}[b]{0.22\textwidth}  
		\centering 
		\includegraphics[width=\textwidth]{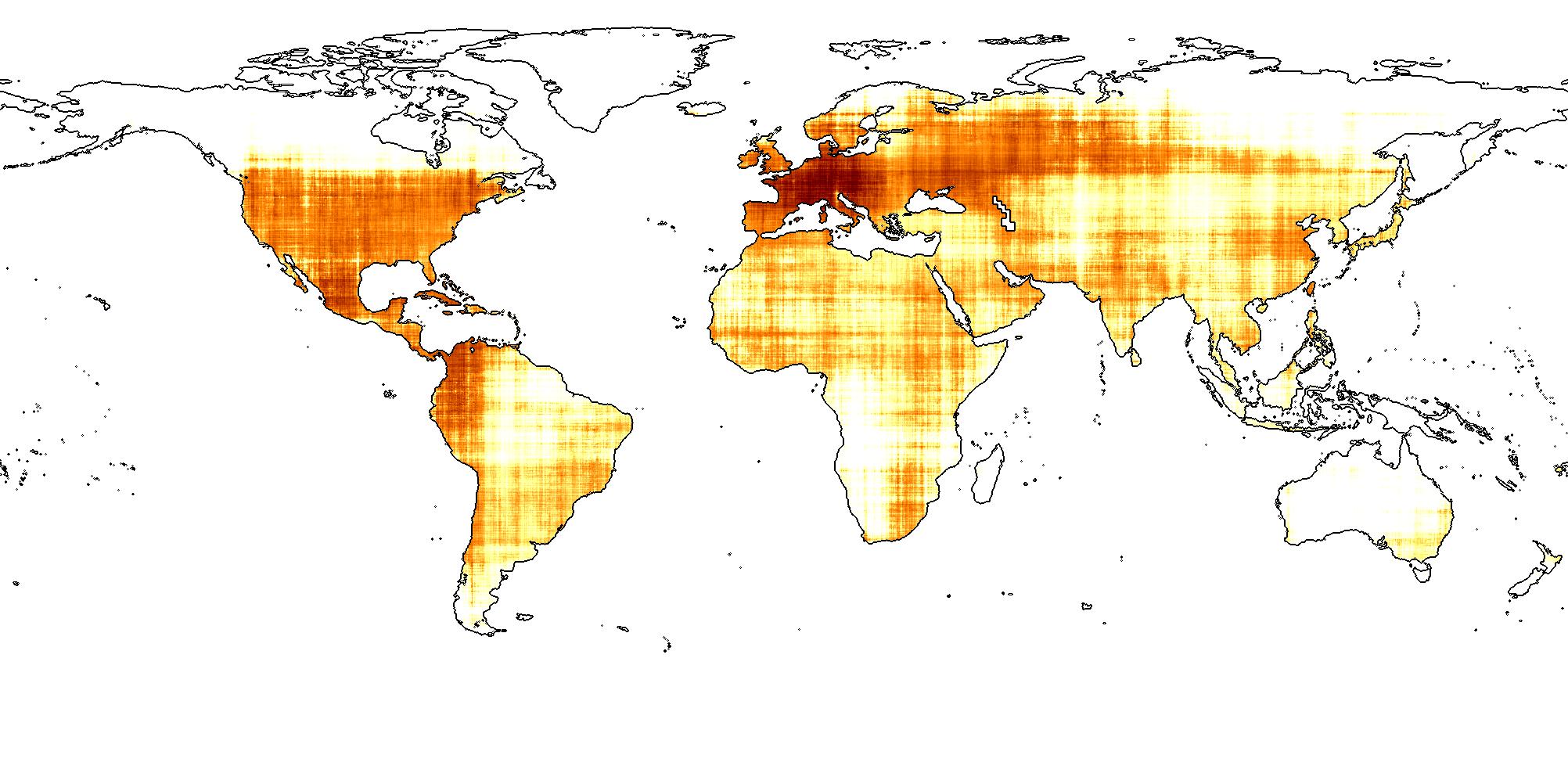}\vspacepred
		\caption[]{{\small 
		$\grid$
		}}    
		\label{fig:factory_or_powerplant_grid_pred_intro}
	\end{subfigure}
	\hfill
	\begin{subfigure}[b]{0.22\textwidth}  
		\centering 
		\includegraphics[width=\textwidth]{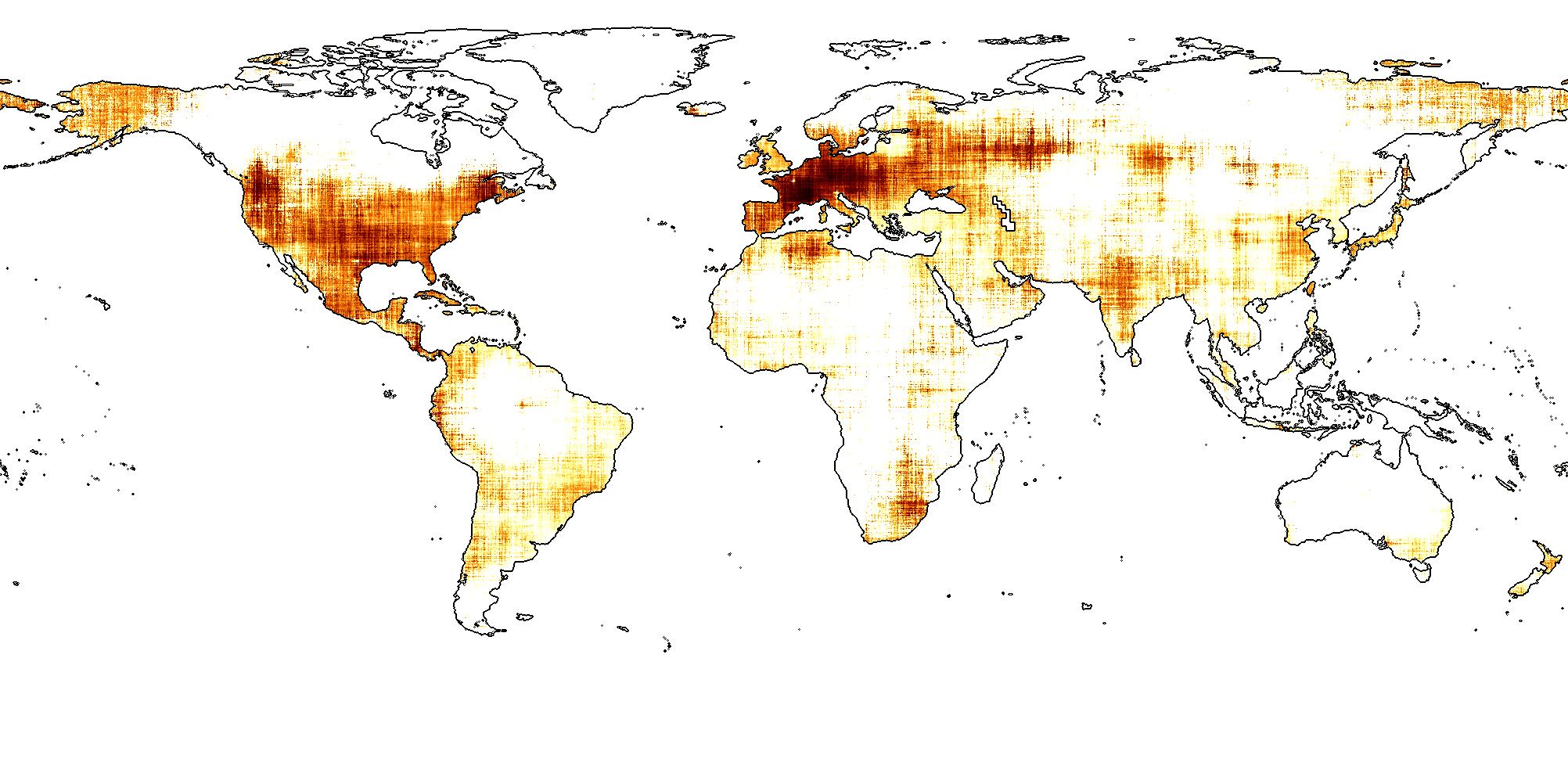}\vspacepred
		\caption[]{{\small 
		$\dft$
		}}    
		\label{fig:factory_or_powerplant_dft_pred_intro}
	\end{subfigure}

	\hfill
	\begin{subfigure}[b]{0.10\textwidth}  
		\centering 
		\includegraphics[width=\textwidth]{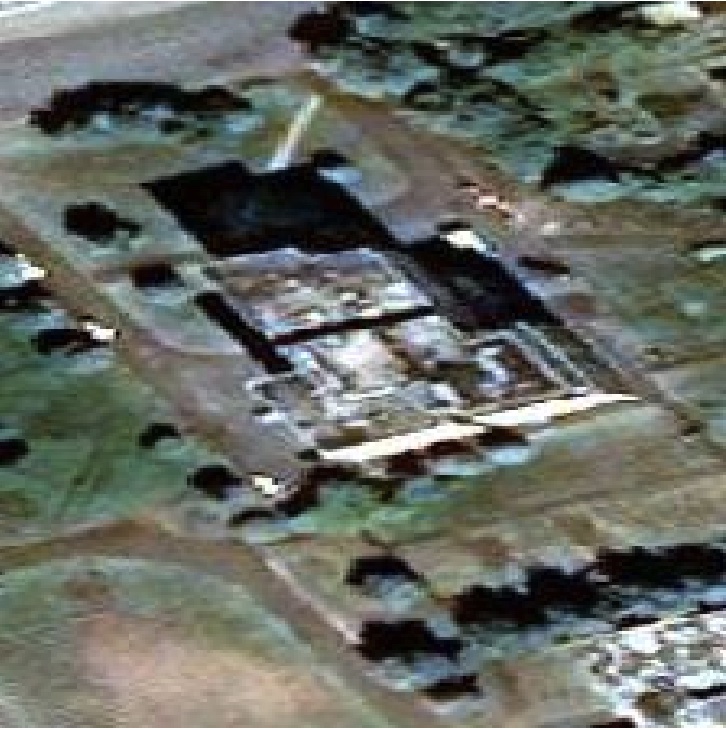}\vspace*{-0.2cm}
		\caption[]{{\small 
		Image
		}}    
		\label{fig:multi-unit_residential_img_intro}
	\end{subfigure}
	\hfill
	\begin{subfigure}[b]{0.22\textwidth}  
		\centering 
		\includegraphics[width=\textwidth]{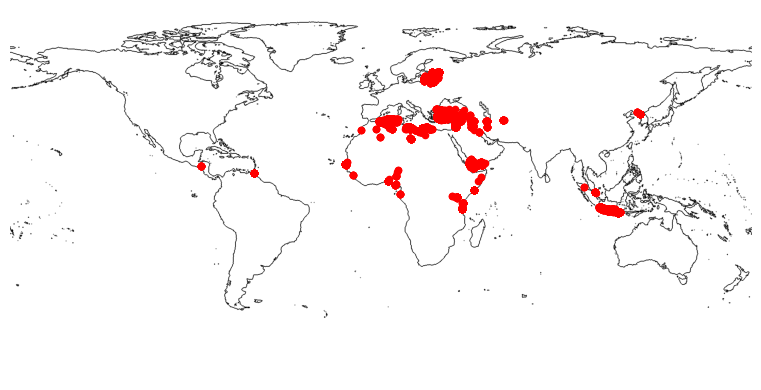}\vspacepred
		\caption[]{{\small 
		Multi-unit residential
		}}    
		\label{fig:multi-unit_residential_loc_intro}
	\end{subfigure}
	\hfill
	\begin{subfigure}[b]{0.22\textwidth}  
		\centering 
		\includegraphics[width=\textwidth]{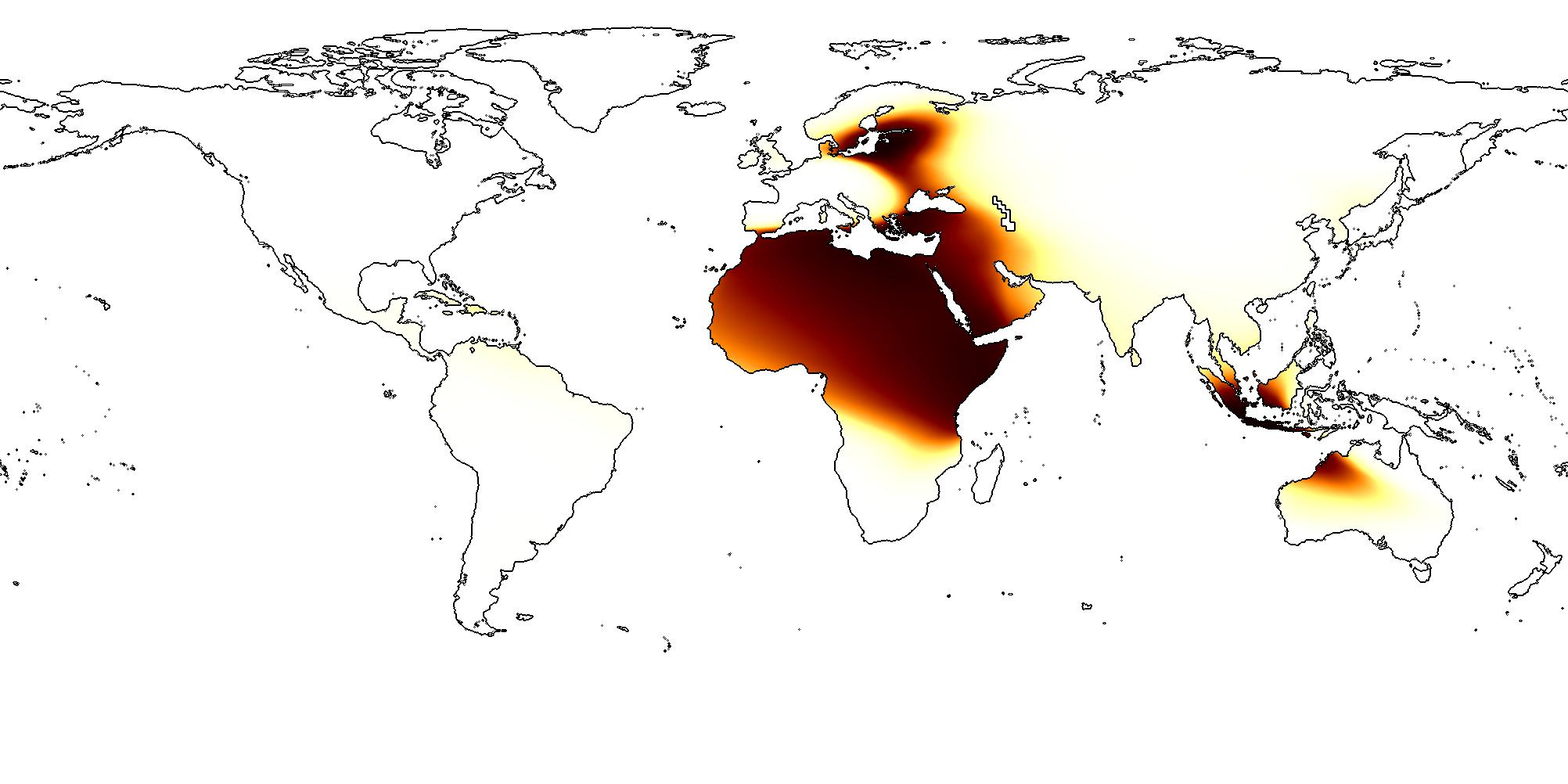}\vspacepred
		\caption[]{{\small 
		$\aodha$
		}}    
		\label{fig:multi-unit_residential_aodha_pred_intro}
	\end{subfigure}
	\hfill
	\begin{subfigure}[b]{0.22\textwidth}  
		\centering 
		\includegraphics[width=\textwidth]{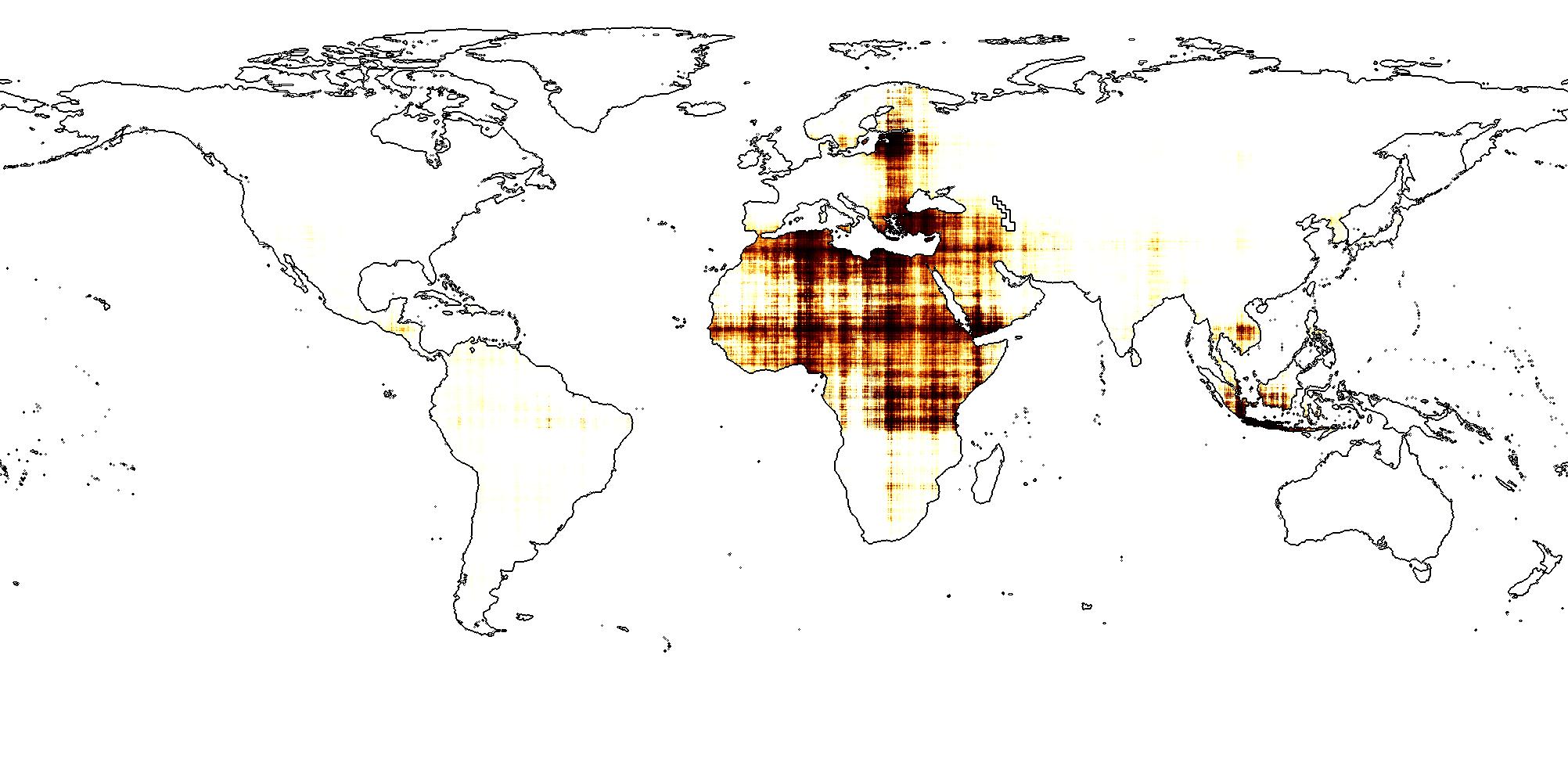}\vspacepred
		\caption[]{{\small 
		$\grid$
		}}    
		\label{fig:multi-unit_residential_grid_pred_intro}
	\end{subfigure}
	\hfill
	\begin{subfigure}[b]{0.22\textwidth}  
		\centering 
		\includegraphics[width=\textwidth]{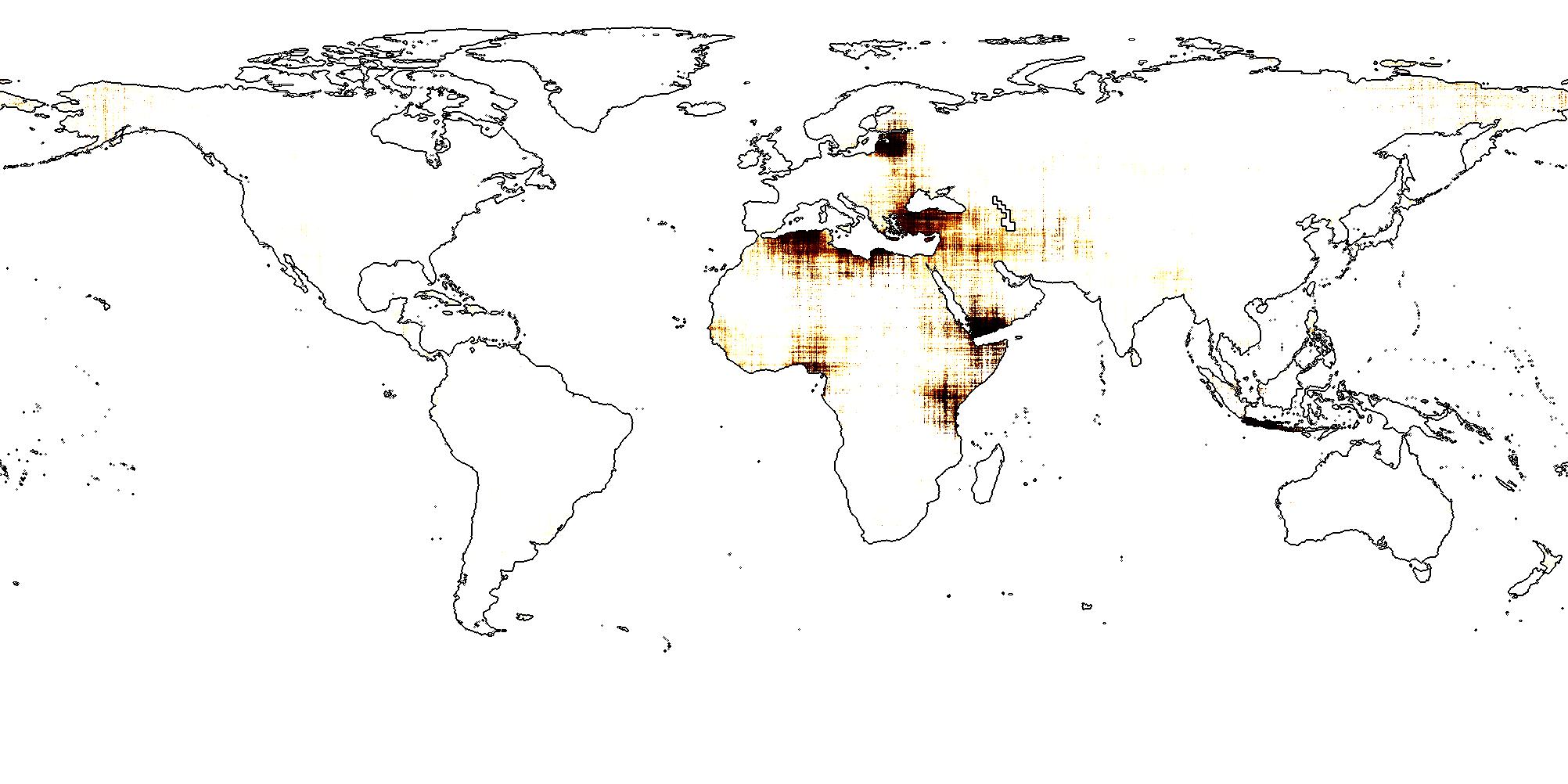}\vspacepred
		\caption[]{{\small 
		$\dft$
		}}    
		\label{fig:multi-unit_residential_dft_pred_intro}
	\end{subfigure}
	\caption{
	Applying location encoders to differentiate two visually similar species ((a)-(j)) or two visually similar land use types ((k)-(t)). Arctic fox and bat-eared fox might look very similar visually as shown in (a) and (f). However, they have different spatial distributions. 
(b) and (g) show their distinct patterns in species image locations. 
	(c)-(e): The predicted  distributions of Arctic fox from different location encoders (without images as input).
(h)-(j): The predicted  distributions of bat-eared fox. Similarly, it might be hard to differentiate factories/powerplants from multi-unit residential buildings only based on their overhead satellite imgeries as shown in (k) and (p). However, as shown in (l) and (q), they have very different global spatial distributions.
	(m)-(o) and (r)-(t) show the predicted spatial distributions of factories/powerplants and multi-unit residential buildings from different location encoders.  
    We can see that while $\aodha$ \citep{mac2019presence}  produces a over-generalized spatial distribution,
    $\spheregrid$ and $\dft$ (our model) produces more compact and fine-grained distributions on the polar region and in data sparse areas such as Africa (See Figure \ref{fig:4081_dist_intro}-\ref{fig:4081_spheregrid_intro}).
    $\grid$ \citep{mai2020multiscale} is between the two. For more examples, please see Figure \ref{fig:spesdist18} and \ref{fig:fmow_pred_dist}.
	} 
	\label{fig:spesdist_intro}
	\vspace*{-0.6cm}
\end{figure*}

Subsequently, this practice of ignoring the round Earth has been adopted by many recent geospatial artificial intelligence (GeoAI) \citep{hu2019geoai,janowicz2020geoai} research on problems such as climate extremes forecasting \citep{ham2019deep}, species distribution modeling \citep{berg2014birdsnap}, location representation learning \citep{mai2020multiscale}, and trajectory prediction \citep{rao2020lstm}. Due to the lack of interpretability of these deep neural network models, this issue has not attracted much attentions by the whole geospatial community.

It is acceptable that the projection errors might be neglectable in small-scale (e.g., neighborhood-level or city-level) geospatial studies. However, they become  non-negligible when we conduct research at a country scale or even global scale. Meanwhile, demand on representation and prediction learning at a global scale grows dramatically due to emerging global scale issues, such as the transition path of the latest pandemic \citep{Chinazzi2020}, long lasting issue for malaria \citep{Caminade2014}, under threaten global biodiversity \citep{DiMarcoetal2019, Ceballosetal2020}, and numerous ecosystem and social system responses for climate change \citep{Hansen&Cramer2015}. 
This trend urgently calls for GeoAI models that can avoid map projection errors and directly perform \textit{calculation on a round planet} \citep{chrisman2017calculating}. 
To achieve this goal, we need a representation learning model which can directly encode point coordinates on a spherical surface into the embedding space such that the resulting location embeddings preserve the spherical distances (e.g., great circle distance\footnote{\url{https://en.wikipedia.org/wiki/Great-circle_distance}}) between two points. With such \mai{a representation}, existing neural network architectures can operate on spherical-distance-kept location embeddings to enable the ability of \textit{calculating on a round planet}.

In fact, such location representation learning models are usually termed location encoders which were originally \mai{developed} to handle 2D or 3D Cartesian coordinates \mai{\citep{chu2019geo,mac2019presence,mai2020multiscale,zhong2020reconstructing,mai2022review,mildenhall2021nerf,schwarz2020graf,niemeyer2021giraffe,barron2021mipnerf,mari2022satnerf,xiangli2022bungeenerf}}.
Location encoders represent a point in a 2D \mai{or 3D Euclidean space \citep{zhong2020reconstructing,mildenhall2021nerf,schwarz2020graf,niemeyer2021giraffe}} into a high dimensional embedding such that the representations are more learning-friendly for downstream machine learning models. 
\mai{
For example, \spacevec~ \citep{mai2020multiscale,mai2020se} was developed for POI type classification, geo-aware image classification, and geographic question answering which can accurately model point distributions in a 2D Euclidean space. 
Recently, several popular location/position encoders widely used in the computer vision domain are also called neural implicit functions \citep{anokhin2021cips,he2021spatial,chen2021liif,niemeyer2021giraffe} which follow the idea of Neural Radiance Fields (\nerf) \citep{mildenhall2020nerf} to
map a 2D or 3D point coordinates to visual signals via a Fourier input mapping \citep{tancik2020fourier,anokhin2021cips,he2021spatial}, or so-called Fourier position encoding \citep{mildenhall2020nerf,schwarz2020graf,niemeyer2021giraffe}, followed by a Multi-Layer Perception (MLP). }
Until now, those 2D/3D \mai{Euclidean} location encoders have already shown promising performances on multiple tasks across different domains including geo-aware image classification \citep{chu2019geo,mac2019presence,mai2020multiscale}, POI classification \citep{mai2020multiscale}, trajectory prediction \citep{xu2018encoding}, geographic question answering \citep{mai2020se}, \mai{2D image superresolution\citep{anokhin2021cips,chen2021liif,he2021spatial}, 3D protein structure reconstruction \citep{zhong2020reconstructing}, 3D scenes representation for view synthesis \citep{mildenhall2020nerf,barron2021mipnerf,tancik2022blocknerf,mari2022satnerf,xiangli2022bungeenerf} and novel image/view generation \citep{schwarz2020graf,niemeyer2021giraffe}. } However, similarly to above mentioned France case, when applying the state-of-the-art (SOTA) 2D \mai{Euclidean} location encoders \citep{mac2019presence,mai2020multiscale} to large-scale real-world GPS coordinate datasets such as remote sensing images taken all over the world \mai{which require distance metric learning on the spherical surface}, a \textbf{map projection distortion problem} \citep{williamson1973,chrisman2017calculating} emerges, especially in the polar areas. 
\mai{On the other hand, the \nerf-style 3D Euclidean location encoders \citep{mildenhall2020nerf,schwarz2020graf,niemeyer2021giraffe} are commonly used to model point distances in the 3D Euclidean space, but not capable of accurately modeling the distances on a complex manifold such as spherical surfaces. Directly applying \nerf-style models on these datasets means these models have to approximate the spherical distances with 3D Euclidean distances which leads to a distance metric approximation error. }
This highlights the necessity of such \mai{a} spherical location encoder discussed above.

In this work, we propose a multi-scale spherical location encoder,
\emph{\modelname}, which can directly encode spherical coordinates while avoiding the map projection distortion \mai{and spherical-to-Euclidean distance approximation error}. 
The \mai{multi-scale} encoding method utilizes 2D Discrete Fourier Transform\footnote{\url{http://fourier.eng.hmc.edu/e101/lectures/Image_Processing/node6.html}} basis ($O(S^2)$ terms) or a subset ($O(S)$ terms) of it while still being able to correctly measure the spherical distance.
Following previous work we use location encoding to learn the geographic prior distribution of different image labels so that given an image and its associated location, we can combine the prediction of the location encoder and that from the state-of-the-art image classification models, e.g., inception V3 \citep{szegedy2016rethinking}, to improve the image classification accuracy. Figure \ref{fig:pos_enc} \mai{illustrates} the whole architecture. 
We demonstrate the effectiveness of \emph{\modelname} on
geo-aware image classification tasks including fine-grained species recognition \citep{chu2019geo,mac2019presence,mai2020multiscale}, Flickr image recognition \citep{tang2015improving,mac2019presence}, and remote sensing image classification  \citep{christie2018functional,ayush2020selfsup}. 
Figure \ref{fig:4084_aodha_intro}-\ref{fig:4084_spheregrid_intro} and \ref{fig:4081_aodha_intro}-\ref{fig:4081_spheregrid_intro} show the predicted species distributions of \textit{Arctic fox} and \textit{bat-eared fox} from three different models. 
Figure \ref{fig:factory_or_powerplant_aodha_pred_intro}-\ref{fig:factory_or_powerplant_dft_pred_intro} and \ref{fig:multi-unit_residential_aodha_pred_intro}-\ref{fig:multi-unit_residential_dft_pred_intro} show the predicted land use distributions of \textit{factory or powerplant} and \textit{multi-unit residential building} from three different models.
\textbf{In summary, the contributions of our work are:}
\begin{enumerate}
    \item We propose a spherical location encoder, \emph{\modelname}, which, as far as we know,  is the first inductive embedding encoding scheme which aims at preserving spherical distance.
\mai{    We also developed a unified view of distant reserving encoding methods on spheres based on Double Fourier Sphere (DFS) \citep{merilees1973,orszag1974}.
}
    \item We provide theoretical proof that 
\emph{\modelname} encodings can preserve spherical surface \mai{distances} between points. \mai{As a comparison, we also prove that the 2D location encoders \citep{gao2018learning,mai2020multiscale,mai2023csp} model latitude and longitude differences separately, and \nerf-style 3D location encoders \citep{mildenhall2020nerf,schwarz2020graf,niemeyer2021giraffe} model axis-wise differences between two points in 3D Euclidean space separately -- none of them can correctly model spherical distances. }

    \item We first conduct experiments on 20 synthetic datasets generated based on the mixture of von Mises–Fisher distribution (MvMF). We show that \emph{\modelname} is able to outperform all baselines including the state-of-the-art (SOTA) 2D location encoders \mai{and \nerf-style 3D location encoders} on all 20 synthetic datasets with an up to 30.8\% error rate reduction. Results show that \mai{2D location encoders are more powerful than \nerf-style 3D location encoders on all synthetic datasets. And
    compared with those 2D location encoders, } \emph{\modelname} is more effective when the dataset has a large data bias toward the polar area.
    
    \item We also conduct extensive experiments on seven real-world datasets for three geo-aware image classification tasks.
Results show that due to its \mai{spherical} distance preserving ability,  \emph{\modelname} outperforms \mai{both} the SOTA 2D location encoder models \mai{and \nerf-style 3D location encoders}.

    \item Further analysis shows that \mai{compared with 2D location encoders, }\emph{\modelname}~ is able to produce finer-grained and compact spatial distributions, and does significantly better in the polar regions and areas with sparse training samples.

\end{enumerate}

The rest of this paper is structured as follows. In Section \ref{sec:map_proj}, we motivate our work by highlighting the importance of the idea of calculating on the round planet.
\mai{Then, we provide a formal problem formulation of spherical location representation learning in Section \ref{sec:prob}. 
Next, }
we briefly summarize the related work in Section \ref{sec:relatedwork}.
The main contribution - \emph{\modelname} - is detailed discussed in Section \ref{sec:method}. 
\mai{Then,} Section \ref{sec:baselines} lists all baseline models we consider in this work.
\mai{The theoretical limitations of 2D location encoder $\grid$ as well as \nerf~style 3D location encoders are discussed in Section \ref{sec:thm_limit}. }
Section \ref{sec:synthetic} presents the \mai{experimental} results on the synthetic datasets.
Then, Section \ref{sec:exp} presents our experimental results on 7 real-world datasets for geo-aware image classification. Finally, we conclude this paper in Section \ref{sec:conclusion}. \maigch{Code and data of this work are available at \url{https://gengchenmai.github.io/sphere2vec-website/}.}

\begin{figure*}
	\centering \tiny
	\vspace*{-0.2cm}
\begin{subfigure}[b]{0.24\textwidth}  
		\centering 
		\includegraphics[width=\textwidth]{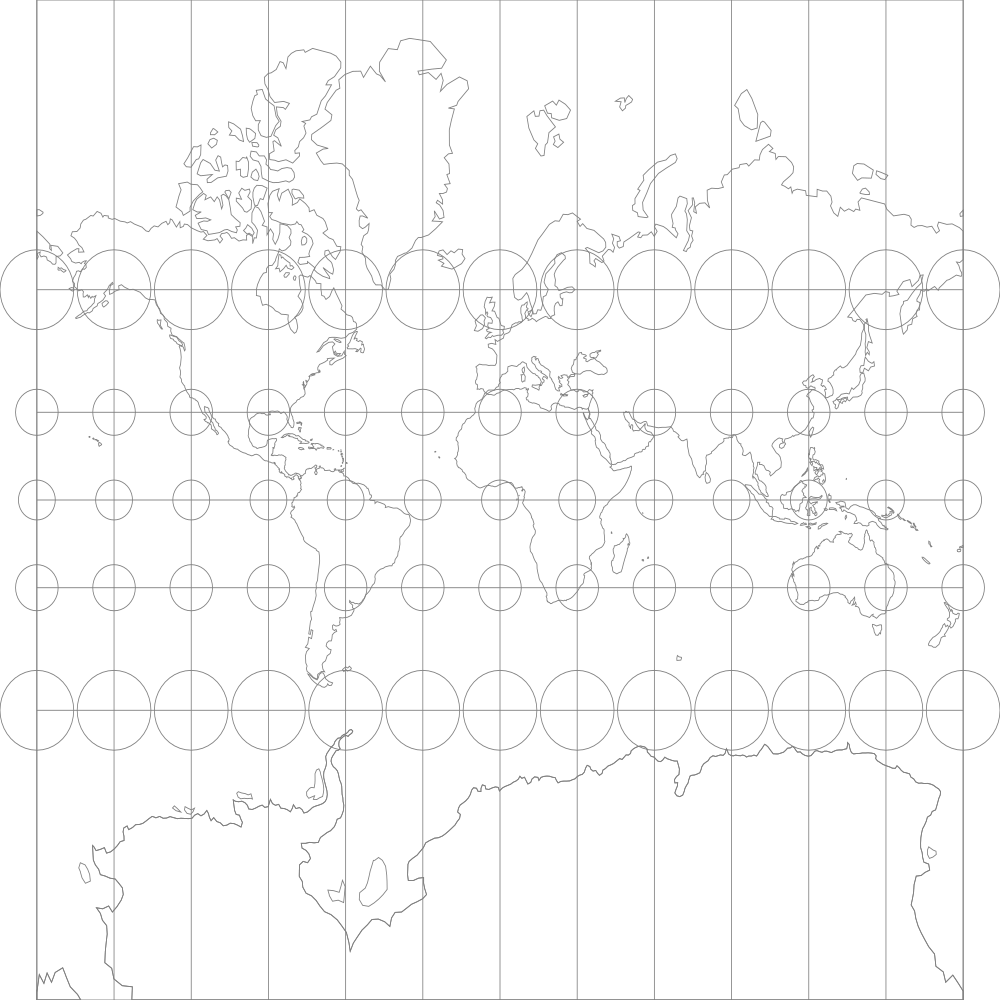}
		\vspace*{-0.3cm}
		\caption[]{{\small 
		Mercator
		}}    
		\label{fig:mercator}
	\end{subfigure}
	\hfill
	\begin{subfigure}[b]{0.24\textwidth}  
		\centering 
		\includegraphics[width=\textwidth]{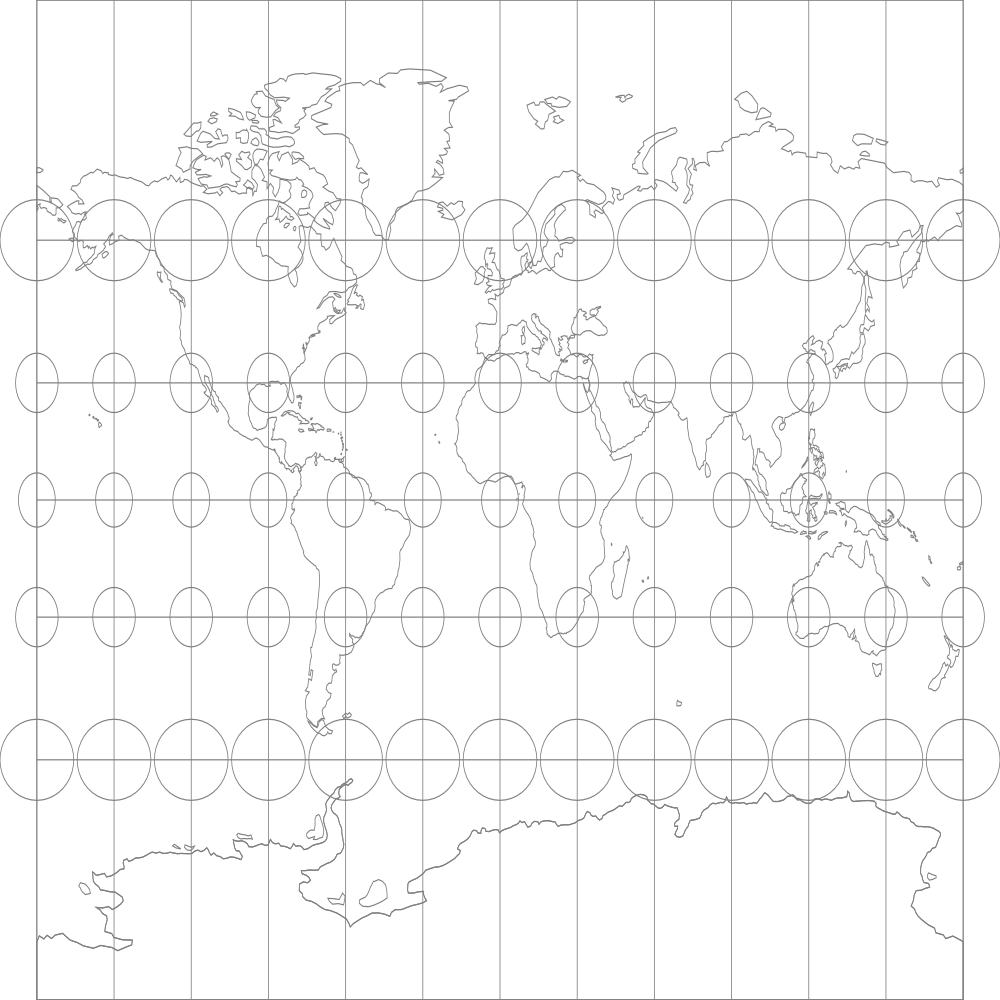}
		\vspace*{-0.3cm}
		\caption[]{{\small 
		Miller
		}}    
		\label{fig:miller}
	\end{subfigure}
	\hfill
	\begin{subfigure}[b]{0.24\textwidth}  
		\centering 
		\includegraphics[width=\textwidth]{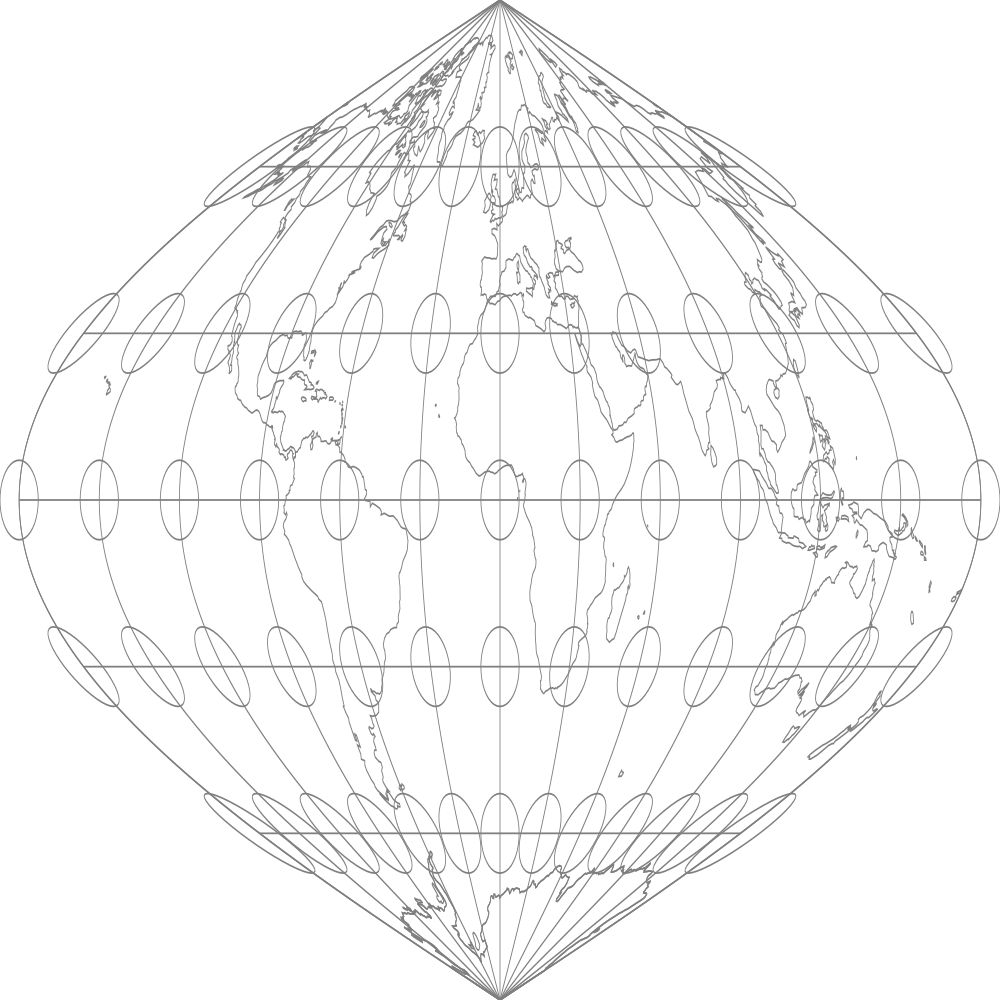}
\caption[]{{\small 
		Sinusoidal
		}}    
		\label{fig:sinusoidal}
	\end{subfigure}
	\hfill
	\begin{subfigure}[b]{0.24\textwidth}  
		\centering 
		\includegraphics[width=\textwidth]{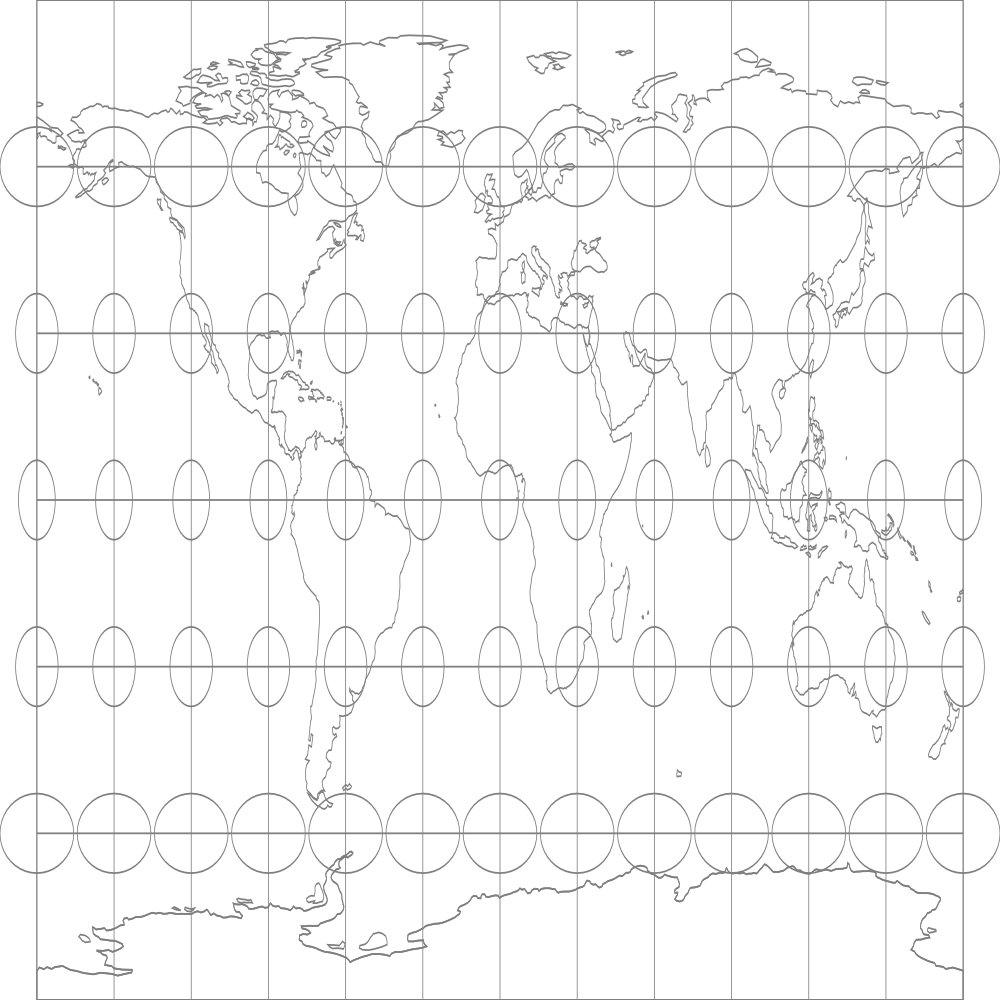}
		\vspace*{-0.3cm}
		\caption[]{{\small 
		Equirectangular
		}}    
		\label{fig:Equirectangle}
	\end{subfigure}
	\caption{An illustration for map projection distortion:
	(a)-(d): Tissot indicatrices for four projections.
The equal area circles are putted in different locations to show how the map distortion affect its shape.
	} 
	\label{fig:map_proj}
	\vspace*{-0.15cm}
\end{figure*}

\section{Calculating on a Round Planet} \label{sec:map_proj}

The blindness to the round Earth or the inappropriate usage of map projections can lead to tremendous and unexpected effects especially when we study a global scale problem since \textit{map projection distortion is unavoidable when projecting spherical coordinates into 2D space.} 

There are no map projection can preserve distances at all direction.
The so-called equidistant projection can only preserve distance on one direction, e.g., the longitude direction for the equirectangular projection (See Figure \ref{fig:Equirectangle}), while the conformal map projections (See Figure \ref{fig:mercator}) can preserve directions while resulting in a large distance distortion. 
For a comprehensive overview of map projections and their distortions, see \cite{mulcahy2001symbolization}. 

When we estimate probability distributions at a global scale (e.g., species distributions or land use types over the world) with a neural network architecture, using 2D Euclidean-based GeoAI models with projected spatial data instead of directly modeling these distributions on a spherical surface will lead to unavoidable map projection distortions and suboptimal results.
This highlights the importance of \textit{calculating on a round planet} \citep{chrisman2017calculating} and necessity of a spherical distance-kept location encoder.
 \section{\mai{Problem Formulation}}  \label{sec:prob}

\mai{
{\em Distributed representation of point-features on the spherical surface} can be formulated as follows. 
Given a set of points $\mP=\{\th_i\}$ on the surface of a sphere $\coordspasphere^{2}$, e.g., locations of remote sensing images taken all over the world, where $\th_i=(\lon_i,\lat_i) \in \coordspasphere^{2}$ indicates a point with  longitude $\lon_i \in  [-\pi , \pi)$ and latitude $\lat_i \in   [-{\pi}/{2} , {\pi}/{2}]$.
Define a function $\enc_{\mP,\params}(\th): \coordspasphere^{2} \to \Real^\embdim$, 
which is parameterized by $\params$ and maps any coordinate $\th$ in a spherical surface $\coordspasphere^{2}$ to a vector representation of $\embdim$ dimension. In the following, we use $\enc(\th)$ as an abbreviation for $\enc_{\mP,\params}(\th)$.
}

\mai{
Let $ \enc(\th) = \pemlp(PE_{\freq}(\th))$ 
where $\pemlp()$ is a learnable multi-layer perceptron with $\numresnet$ hidden layers and $\numneuron$ neurons per layer. We want to find a \textit{position encoding} function $PE_{\freq}(\th)$ which does a one-to-one mapping from each point $\th_i=(\lon_i,\lat_i) \in \coordspasphere^{2}$ to a multi-scale representation with $\freq$ be the total number of scales. 
}

\mai{
We expect to find a function $PE_{\freq}(\th)$ such that the resulting multi-scale representation of $\th$ preserves the spherical surface distance while it is more learning-friendly for the downstream neuron network model $\pemlp()$.
More concretely, we'd like to use position encoding functions which satisfy the following requirement:
\begin{align}
    \langle PE_{\freq}(\th_1), PE_{\freq}(\th_2) \rangle = \pefunc(\sd), \forall \th_1, \th_2 \in \coordspasphere^{2},
    \label{equ:prob_stat}
\end{align}
where $\langle \cdot, \cdot \rangle$ is the cosine similarity function between two embeddings. $\sd \in [0, \pi R]$ is the spherical surface distance between $\th_1, \th_2$, $R$ is the radius of this sphere, and $\pefunc(x)$ is a strictly monotonically decreasing function for $x \in [0, \pi R]$. 
}

 \section{Related Work}
\label{sec:relatedwork}

\mai{\subsection{Neural Implicit Functions and \nerf}} \label{subsec:nerf_related}

\mai{
As an increasingly popular family of models in the computer vision domain,  
neural implicit functions \citep{anokhin2021cips,he2021spatial,chen2021liif,niemeyer2021giraffe} refer to the neural network architectures that directly map a 2D or 3D coordinates into visual signals via a Fourier input mapping/position encoding \citep{tancik2020fourier,anokhin2021cips,he2021spatial,mildenhall2020nerf,schwarz2020graf,niemeyer2021giraffe}, followed by a Multi-Layer Perception (MLP).
}

\mai{
A good example is Neural Radiance Fields (\nerf) \citep{mildenhall2020nerf}, which combines neural implicit functions and volume rendering for novel view synthesis for 3D complex scenes. The idea of \nerf~ becomes very popular and many follow-up works have been done to revise the $\nerf$ model in order to achieve more accurate view synthesis. For example, NeRF in the Wild (NeRF-W) \citep{martin2021nerfw} was proposed to learn separate transient phenomena from each static scene to make the model robust to radiometric variation and transient objects. Shadow NeRF (S-NeRF) \citep{derksen2021snerf} was proposed to exploit the direction of solar rays to obtain a more realistic view synthesis on multi-view satellite photogrammetry. Similarly, Satellite NeRF (Sat-NeRF) \citep{mari2022satnerf} combines NeRF with native satellite camera models to achieve robustness to transient phenomena that cannot be explained by the position of the sun to solve the same task. 
A more noticeable example is GIRAFFE \citep{niemeyer2021giraffe} which is a NeRF-based deep generative model which achieves a more controllable image synthesis. All these NeRF variations mentioned above use the same NeRF Fourier position encoding. And they all use this position encoding in the same generative task -- novel image synthesis. Moreover, although S-NeRF and Sat-NeRF work on geospatial data, i.e., satellite images, they focus on rather small geospatial scales, e.g., city scales, in which map projection distortion can be ignored. 
In contrast, we investigate the advantages and drawbacks of various location encoders in large-scale (e.g., global-scale) geospatial prediction tasks which are discriminative tasks.
We use NeRF position encoding as one of our baselines.
}

\mai{
Several works also discussed the possibility to revise NeRF position encoding. 
The original encoding method takes a single 3D point as input which ignores both the relative footprint of the corresponding image pixel and the length of the interval along the ray which leads to aliasing artifacts when rendering novel camera trajectories \citep{tancik2022blocknerf}. To fix this issue, 
Mip-NeRF \citep{barron2021mipnerf} proposed a new Fourier position encoding called integrated positional encoding (IPE). Instead of encoding one single 3D point, IPE encodes 3D conical frustums approximated by multivariate Gaussian distributions which are sampled along the ray based on the projected pixel footprints.
Block-NeRF \citep{tancik2022blocknerf} adopted the IPE idea and showed how to scale NeRF to render city-scale scenes. Similarly, BungeeNeRF \citep{xiangli2022bungeenerf} also used the IPE model to develop a progressive NeRF that can do multi-scale rendering for satellite images in different spatial scales. 
In this work, we focus on encoding a single point on the spherical surface, not a 3D conical frustums. 
So IPE is not considered as one of the baselines.
}

\mai{
Neural implicit functions are also popular for other computer vision tasks such as image superresolution \citep{anokhin2021cips,chen2021liif,he2021spatial} and image compression \citep{dupont2021coin,strumpler2022implicit}.}

\subsection{Location Encoder}  \label{subsec:locenc_related}

Location encoders \citep{chu2019geo,mac2019presence,mai2020multiscale,zhong2020reconstructing,mai2023csp} are neural network architectures which encode points in low-dimensional (2D or 3D) spaces \citep{zhong2020reconstructing}) into high dimensional embeddings. 
There has been much research on developing inductive learning-based location encoders. Most of them directly apply Multi-Layer Perceptron (MLP) to 2D coordinates to get a high dimensional location embedding for downstream tasks such as pedestrian trajectory prediction \citep{xu2018encoding} and geo-aware image classification \citep{chu2019geo}.
Recently, Mac Adoha et al. \citep{mac2019presence} apply sinusoid functions to encode the latitude and longitude of each image before feeding into MLPs.
All of the above approaches deploy location encoding at a single-scale. 

Inspired by the position encoder in Transformer \citep{vaswani2017attention} and Neuroscience research on grid cells \citep{banino2018vector,cueva2018emergence} of mammals,  
\cite{mai2020multiscale} proposed to  apply multi-scale sinusoid functions to encode locations in 2D Euclidean space before feeding into MLPs.
The multi-scale representations have advantage of capturing spatial feature distributions with different characteristics. 
Similarly, \cite{zhong2020reconstructing} utilized a multi-scale location encoder for the position of proteins' atoms in 3D Euclidean space for protein structure reconstruction with great success. 
Location encoders can be incorporated into the state-of-art models for many tasks to make them spatially explicit \mai{\citep{yan2019spatial,janowicz2020geoai,mai2022symbolic,mai2023csp}}.

Compared with well-established \maigch{kernel}-based approaches \citep{scholkopf2001kernel,xu2018encoding} such as Radius Based Function (RBF) which requires \mai{memorizing} the training examples as the kernel centers for a robust prediction, inductive-learning-based location encoders \citep{chu2019geo,mac2019presence,mai2020multiscale,zhong2020reconstructing} have many advantages: 
1) They are more memory efficient since they do not need to memorize training samples; 
2) Unlike RBF, the performance on unseen locations does not depend on the number and distribution of kernels.
Moreover, \cite{gao2018learning} have shown that \mai{grid-like} periodic representation of locations can preserve absolute position information, relative distance\mai{,} and direction information in 2D Euclidean space. \cite{mai2020multiscale} further show that it benefits the generalizability of \mai{down-stream} models.
For a comprehensive survey of different location encoders, please refer to \cite{mai2022review}.

Despite all these \mai{successes} in location encoding research, none of them consider location representation learning on a spherical surface which is in fact critical for a global scale geospatial study. Our work aims at filling this gap.

\subsection{Machine Learning Models on Spheres}
\label{subsec:mapproj_related}
Recently, there has been \mai{an} increasing amount of work on designing machine learning models for prediction tasks on spherical surfaces. 
For \mai{the} omnidirectional image classification task, 
both \cite{cohen2018spherical} and \cite{coors2018spherenet} designed different spherical versions of the traditional convolutional neural network (CNN) models in which the CNN filters explicitly \mai{consider} map projection distortion. 
In terms of image geolocalization \citep{izbicki2019exploiting} and text geolocalization \citep{izbicki2019geolocating}, a loss function based on the mixture of von Mises-Fisher distributions (MvMF)-- a spherical \mai{analog} of the Gaussian mixture model (GMM)--  is used to replace the traditional cross-entropy loss for geolocalization models \citep{izbicki2019exploiting,izbicki2019geolocating}. 
All these works are closely related to geometric deep learning \citep{bronstein2017geometric}. 
They show the importance to consider the spherical geometry instead of projecting it back to a 2D plane, yet none of them considers representation learning of spherical coordinates in the embedding space.

\subsection{Spatially Explicit Artificial Intelligence}  \label{subsec:spex_related}
There has been much work in
\textit{improving the performance of 
current state-of-the-art 
artificial intelligence and machine learning models by
using spatial features or spatial inductive bias} -- so-called
spatially explicit artificial intelligence \mai{\citep{yan2017itdl,mai2019relaxing,yan2019spatial,yan2019graph,janowicz2020geoai,li2021tobler,zhu2021spatial,janowicz2022know,liu2022review,zhu2022reasoning,mai2022symbolic,mai2023towards,huang2023learning}}, or \textit{SpEx-AI}. 
 The spatial inductive bias 
in these models includes: 
 spatial dependency \citep{kejriwal2017neural,yan2019spatial}, 
 \mai{spatial heterogeneity \citep{berg2014birdsnap,chu2019geo,mac2019presence,mai2020multiscale,zhu2021spatial,gupta2021spatial,xie2021statistically}}, 
 map projection \citep{cohen2018spherical,coors2018spherenet,izbicki2019exploiting,izbicki2019geolocating}, 
 scale effect \citep{weyand2016planet,mai2020multiscale}, and so on.

 \subsection{Pseudospectral Methods on Spheres}  \label{subsec:pseudospectral}
Multiple studies have been focused on the numerical solutions on spheres, for example, in weather prediction 
\citep{orszag1972,orszag1974,merilees1973}.
The main idea is \mai{so-called} pseudospectral methods which leverage truncated discrete Fourier transformation on spheres to achieve computation efficiency while avoiding the error caused by map projection distortion. 
The particular set of basis functions to be used depends on the particular problem.
However, they do not aim at learning good representations in machine learning models. In this study, we try to make connections to these approaches and explore how their insights can be realized in a deep learning model. \section{Method} \label{sec:method}

Our main contribution - the design of spherical distance-kept location encoder $\enc(\th)$, \emph{\modelname} will be presented in Section \ref{subsec:encoder}. 
We developed a unified view of distance-reserving encoding on spheres based on Double Fourier Sphere (DFS) \citep{merilees1973,orszag1974}.
The resulting location embedding $\peemb  = \enc(\th)$ is a general-purpose embedding which can be utilized in different decoder architectures for various tasks. 
In Section \ref{subsec:img_cls_decoder}, we briefly show how to utilize the proposed $\enc(\th)$ in the geo-aware image classification task.

\subsection{\modelname}
\label{subsec:encoder}

The multi-scale location encoder defined in Section~\ref{sec:prob} is in the form of $ \enc(\th) = \pemlp(PE_{\nscale}(\th))$. 
$PE_{\nscale}(\th)$ is a concatenation of multi-scale spherical spatial features of $\nscale$ levels. In the following, we call $\enc(\th)$ {\it location encoder} and its component $PE_{\nscale}(\th)$ {\it position encoder}.

\begin{figure*}
	\centering \tiny
	\vspace*{-0.2cm}
	\begin{subfigure}[b]{0.162\textwidth}  
		\centering 
		\includegraphics[width=\textwidth]{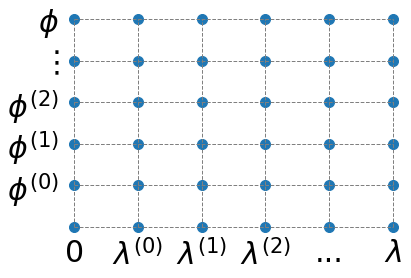}\vspace*{-0.2cm}
		\caption[]{{\small 
		$\dft$
		}}    
		\label{fig:dft}
	\end{subfigure}
	\begin{subfigure}[b]{0.162\textwidth}  
		\centering 
		\includegraphics[width=\textwidth]{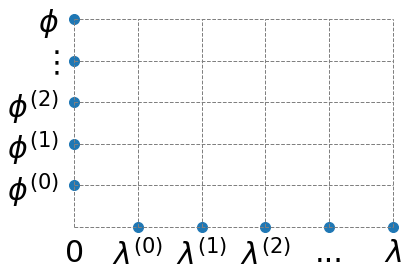}\vspace*{-0.2cm}
		\caption[]{{\small 
		$\grid$
		}}    
		\label{fig:grid}
	\end{subfigure}
	\begin{subfigure}[b]{0.162\textwidth}  
		\centering 
		\includegraphics[width=\textwidth]{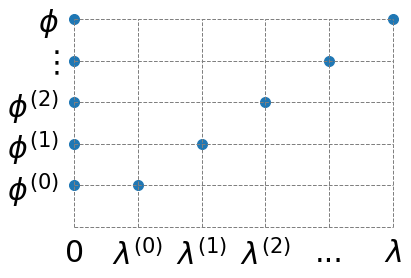}\vspace*{-0.2cm}
		\caption[]{{\small 
		$\sphere$
		}}    
		\label{fig:sphere}
	\end{subfigure}
	\begin{subfigure}[b]{0.162\textwidth}  
		\centering 
		\includegraphics[width=\textwidth]{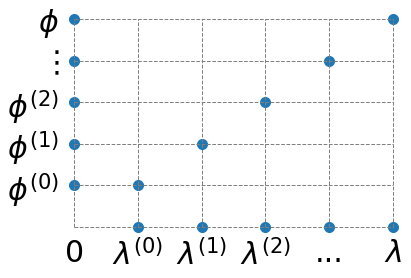}\vspace*{-0.2cm}
		\caption[]{{\small 
		$\spheregrid$
		}}    
		\label{fig:spheregrid}
	\end{subfigure}
	\begin{subfigure}[b]{0.162\textwidth}  
		\centering 
		\includegraphics[width=\textwidth]{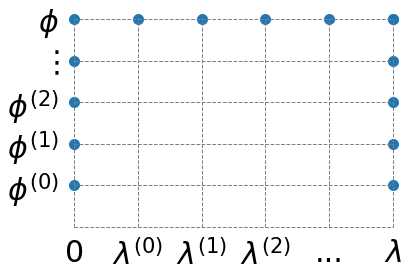}\vspace*{-0.2cm}
		\caption[]{{\small 
		$\spheremixscale$
		}}    
		\label{fig:spheremix}
		\end{subfigure}
		\begin{subfigure}[b]{0.162\textwidth}  
		\centering 
		\includegraphics[width=\textwidth]{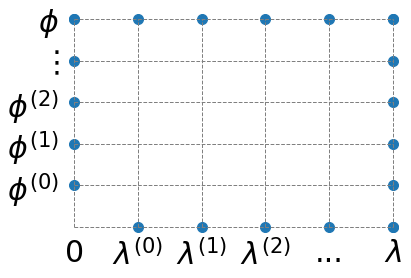}\vspace*{-0.2cm}
		\caption[]{{\small 
		$\spheregridmixscale$
		}}    
		\label{fig:spheremixgrid}
	\end{subfigure}

	\caption{Patterns of different encoders, blue points at $(\lon^{(m)},\lat^{(n)})$ mean interaction terms of trigonometric functions of $\lon^{(m)}$ and $\lat^{(n)}$ are included in the encoder, $\lon$ and $\lat$ axis correspond to single terms with no interactions.}
	\label{fig:allpatterns}
    \vspace*{-0.15cm}
\end{figure*}
 
\paragraph{$\dft$}
Double Fourier Sphere (DFS) \citep{merilees1973,orszag1974} is a simple yet successful pseudospectral method, which is computationally efficient and have been applied to analysis 
of large scale phenomenons such as weather \citep{sun2014} and blackholes \citep{bartnik2000}. Our first intuition is to use the base functions of DFS, which preserve periodicity in both the longitude and latitude directions, to help decompose $\th=(\lon,\lat)$ into a high dimensional vector:
\vspace{-0.1cm}
\begin{align} \begin{split}
   \spe{\dft}_{\nscale}(\th)=&\bigcup_{n=0}^{\nscale-1}[\sin\latz{n}, \cos\latz{n}] \cup  \bigcup_{m=0}^{\nscale-1} [\sin\lonz{m}, \cos\lonz{m}] \cup \\ 
    & \bigcup_{n=0}^{\nscale-1} \bigcup_{m=0}^{\nscale-1} [\cos\latz{n}\cos\lonz{m},  \cos\latz{n}\sin\lonz{m}, \\
& \sin\latz{n}\cos\lonz{m},\sin\latz{n}\sin\lonz{m}],
\label{equ:dfs}
\end{split}\end{align}
\vspace{-0.1cm}

where $\lonz{m}=\frac{\lon}{\fm}$, $\latz{n}=\frac{\lat}{\fn}$. $\fm$ and $\fn$ are scaling factors controlled by the current scale $m$ and $n$. 
Let $\minscale, \maxscale$ be the minimum and maximum  scaling factor, and $g = \frac{\maxscale}{\minscale}$.\footnote{In practice we fix $\maxscale=1$ meaning no scaling of $\lon,\lat$.}
$\fs=\minscale \cdot g^{\s/(\nscale-1)}$ where $\s$ is either $m$ or $n$.  
$\cup$ means vector concatenation and $\bigcup_{\s=0}^{\nscale-1}$ indicates vector concatenation through different scales.
It basically 
lets all the $\nscale$ scales of $\lat$ terms interact with all the $\nscale$ scales of $\lon$ terms in the encoder. 
This would introduce a position encoder with a $O(S^2)$ dimension output 
which increases the memory burden in training and hurts generalization. See Figure~\ref{fig:dft} for an illustration of the used $O(S^2)$ terms. An encoder might achieve better results by only using a subset of these terms.

In comparison, the state-of-the-art $\grid$ \citep{mai2020multiscale} encoder defines its position encoder as: \begin{align}
    \spe{\grid}_{\nscale}(\th)=\bigcup_{\s=0}^{\nscale-1}[\sin\lats,\cos\lats,\sin\lons,\cos\lons].
\label{equ:grid}
\end{align}
\vspace{-0.1cm}

Here, $\lons$ and $\lats$ have similar definitions as $\lonz{m}$ and $\latz{n}$ in Equation \ref{equ:dfs}. 
Figure~\ref{fig:grid} illustrates the used terms of $\grid$. We can see that $\grid$ employs a subset of terms from $\dft$. However, as we explained earlier, $\grid$ performs poorly at a global scale due to its inability to preserve spherical distances.

In the following we explore different subsets of DFS terms while achieving two goals: 1) efficient representation with $O(S)$ dimensions 2) preserving distance measures on a spherical surface. 

\paragraph{$\sphere$}
Inspired by the fact that any point  $(x,y,z)$ in 3D Cartesian coordinate can be expressed by $sin$ and $cos$ basis of spherical coordinates ($\lon$, $\lat$ plus radius)
\footnote{\url{https://en.wikipedia.org/wiki/Spherical_coordinate_system}},
we define the basic form of \modelname, namely $\sphere$ encoder: \begin{align}
   \spe{\sphere}_{\nscale}(\th)=\bigcup_{\s=0}^{\nscale-1}[\sin\lats,\cos\lats\cos\lons,\cos\lats\sin\lons].
    \label{equ:sphere}
\end{align}
\vspace{-0.1cm}

Figure~\ref{fig:sphere} illustrates the used terms of $\sphere$. 
To illustrate that $\sphere$ is good at capturing spherical distance, we take a close look at its basic case $\nscale=1$. When $\nscale=1$ and $\maxscale=1$, there is only one scale $\s=\nscale-1=0$ and we define $\fs=\minscale \cdot g^{\s/(\nscale-1)} = \maxscale = 1$. The multi-scale encoder degenerates to \begin{align}
        \spe{\sphere}_1(\th)=[\sin(\lat),\cos(\lat)\cos(\lon),\cos(\lat)\sin(\lon)].
    \label{equ:sphere1}
\end{align}
These three terms are included in the multi-scale version ($\nscale > 1$) and serve as the main terms at the largest scale and also the lowest frequency (when $\s=\nscale-1$). 
The high frequency terms are added to help the downstream neuron network to learn the point-feature more efficiently \citep{tancik2020fourier}. 
Interestingly, $\spe{\sphere}_1$ captures the spherical distance in a very explicit way:
\newtheorem{theorem}{Theorem}
\begin{theorem}\label{thm1}
Let $\th_1$, $\th_2$ be two points on the same sphere $ \coordspasphere^{2} $ with radius $\radius$, then 
\begin{align}
    \langle \spe{\sphere}_1(\th_1),  \spe{\sphere}_1(\th_2) \rangle = \cos(\frac{\sd}{\radius}),
\end{align}
where $\sd$ is the great circle distance between $\th_1$ and $\th_2$. Under this metric,
\begin{align}
    \|\spe{\sphere}_1(\th_1)- \spe{\sphere}_1(\th_2)\|
= 2\sin (\frac{\sd}{2\radius}).
\end{align}
Moreover, $\| \spe{\sphere}_1(\th_1)-\spe{\sphere}_1(\th_2) \|\approx \frac{\sd}{\radius}$,when $\sd$ is small w.r.t. $\radius$.
\end{theorem} 
\newtheorem{remark}{Remark}
See the proof in \mai{Appendix} \ref{sec:proof1}.

Since the central angle $\ca = \frac{\sd}{\radius} \in [0, \pi]$ and $\cos(x)$ is strictly monotonically decrease for $x \in [0, \pi], $ Theorem $\ref{thm1}$ 
shows that $\spe{\sphere}_1(\th)$ directly satisfies our expectation in Equation \ref{equ:prob_stat} where $\pefunc(x) = \cos(\frac{x}{\radius})$.

\paragraph{$\spheremixscale$}
Considering the fact that many geographical patterns are more sensitive to either latitude (e.g., temperature, sunshine duration) or longitude (e.g., timezones, geopolitical borderlines),  we might want to focus on increasing the resolution of either  $\lat$ or $\lon$ while holding the other relatively at a large scale.  
Therefore, we introduce a multi-scale position encoder  $\spheremixscale$, where interaction terms between $\lat$ and $\lon$ always have one of them fixed at the top scale:
\begin{align}
\begin{split}
    \spe{\spheremixscale}_{\nscale}(\th)=&\bigcup_{s=0}^{\nscale-1}[\sin\lats,\cos\lats\cos\lon,\cos\lat\cos\lons, \\
    &\cos\lats\sin\lon,\cos\lat\sin\lons].\\
\end{split}\end{align}
This new encoder ensures that the $\lat$ term interact with all the scales of $\lon$ terms (i.e., $\lons$ terms) and $\lon$ term interact with all the scales of $\lat$ terms (i.e., $\lats$ terms). See Figure \ref{fig:spheremix} for the used terms of $\spheremixscale$. 
Both $\spe{\sphere}_{\nscale}$ and $\spe{\spheremixscale}_{\nscale}$ are multi-scale versions of a spherical distance-kept encoder (See Equation \ref{equ:sphere1}) and keep that as the main term in their multi-scale representations.

\paragraph{$\spheregrid$ and $\spheregridmixscale$}

From the above analysis of the two proposed position encoders and the SOTA $\grid$ encoders, we know that $\grid$ pays more attention to the sum of $\cos$ difference of latitudes and longitudes, while our proposed encoders pay more attention to the spherical distances. 
In order to capture both information, we consider merging $\grid$ with each proposed encoders to get more powerful models that encode geographical information from different angles. 
\begin{align}
\spe{\spheregrid}_{\nscale}(\th) &= \spe{\sphere}_{\nscale}(\th) \cup \spe{\grid}_{\nscale}(\th), \label{eq:spheregrid} \\ 
\spe{\spheregridmixscale}_{\nscale}(\th)&= \spe{\spheremixscale}_{\nscale}(\th) \cup \spe{\grid}_{\nscale}(\th).
\label{eq:spheregridmixscale} \end{align}
We hypothesize that encoding these terms in the multi-scale representation would make the training of the encoder easier and the order of output dimension is still $O(S)$.
See Figure \ref{fig:spheregrid} and \ref{fig:spheremixgrid} for the used terms of $\spheregrid$ and $\spheregridmixscale$.

In location encoding, the uniqueness of the encoding results (i.e., no two different points on a sphere having the same position encoding) is very important. $PE_{\nscale}(\th)$ in the five proposed methods are by design one-to-one mapping. 
\begin{theorem}
$\forall * \in \{
\dft, \sphere,\spheregrid,\spheremixscale,$ \\ $\spheregridmixscale\}$, $PE^*_{\nscale}(\th)$ is an injective function.
\label{the:injective}
\end{theorem}
See the proof in \mai{Appendix} \ref{sec:proof2}.

\subsection{Applying \modelname~to Geo-Aware Image Classification}  \label{subsec:img_cls_decoder}

Follow the practice of \cite{mac2019presence} and \cite{mai2020multiscale}, we formulate the \textit{geo-aware image classification task} \citep{chu2019geo,mac2019presence} as follow:
Given an image $\image$ taken from location/point $\th$, we estimate which category $\classy$ it belongs to.
If we assume that $\image$ and $\th$ are independent given $\classy$
and an even-prior $P(\classy)$, 
then we have
\begin{align}
    &P(\classy|\image,\th) 
    = \dfrac{P(\image, \th | \classy)P(\classy)}{P(\image, \th)} 
    =P(\image|\classy)
    P(\th | \classy )
    \dfrac{P(\classy)}{P(\image, \th)} \\
    &= \dfrac{P(\classy|\image)P(\image)}{P(\classy)} 
    \dfrac{P( \classy|\th )P(\th)}{P(\classy)} 
    \dfrac{P(\classy)}{P(\image, \th)} \\
    &= P(\classy|\th)P(\classy|\image)
    \dfrac{P(\image)P(\th)}{P(\classy)P(\image, \th)} 
    \; \propto \; P(\classy|\th)P(\classy|\image)
    \label{eq:img_pred}
\end{align}
$P(\classy|\image)$ can be obtained by fine-tuning \mai{the state-of-the-art image classification model }
for a specific task, such as a pretrained InceptionV3 network \citep{mac2019presence} for species recognition, or a pretrained MoCo-V2+TP \citep{ayush2020selfsup} for RS image classification. 
\mai{To be more specific, we use a pretrained image encoder $\imgenc()$ to extract the embedding for each input image, i.e., $\imgenc(\image)$.
Then in order to compute $P(\classy|\image)$, we can either 1) fine-tune an image classifier $\imgcls$ based on these frozen image embeddings, or 2) fine-tune the whole image encoder architecture $\imgcls(\imgenc(\image))$. Here, $\imgcls$ is a multilayer perceptron (MLP) followed by a softmax activation function. Both \cite{mac2019presence} and \cite{mai2020multiscale} adopted the second approach which fine-tunes the whole image classification architecture. We also adopt the second approach to have a fair comparison with all these previous methods. Please refer to Section \ref{sec:ablation_img_loc_comb} for an ablation study on this.
}
The idea is illustrated in the orange box in Figure \ref{fig:pos_enc}.

In this work, we focus on the second component -- estimating the geographic prior distribution of image label $\classy$ over the spherical surface $P(\classy|\th)$ (the blue box in Figure \ref{fig:pos_enc}). This probability distribution can be estimated by using a location encoder $\enc()$. We can use either our proposed \emph{\modelname} or some existing 2D \mai{\citep{mai2020multiscale,mac2019presence,chu2019geo} or 3D \citep{mari2022satnerf,martin2021nerfw}} Euclidean location encoders. 
More concretely, we have $P(\classy|\th) \; \propto \; \act(\enc(\th)\classemb_{:,\classy})$ where $\act()$ is a sigmoid activation function. 
$\classemb \in \Real^{\embdim \times \numclass}$ is a class embedding matrix (the location classifier in Figure \ref{fig:pos_enc}) where the $\classy_{th}$ column $\classemb_{:,\classy} \in \Real^{\embdim}$ indicates the class embedding for class $\classy$. $\embdim$ indicates the dimension of location embedding $\mathbf{p}[\th] = \enc(\th)$ and $\numclass$ is the total number of image classes.

The major objective is to learn $P(\classy|\th) \; \propto \; \act(\enc(\th)\classemb_{:,\classy})$ such that all observed species occurrences (all image locations $\th$ as well as their associated species class $\classy$) have maximum probabilities. \cite{mac2019presence} used a loss function which is based on maximum likelihood estimation (MLE). Given a set of training samples - data points and their associated class labels $\sampleset = \{(\th, \classy)\}$, the loss function $\imgclsloss(\sampleset)$ is defined as:
\begin{align}
\begin{split}
    \imgclsloss(\sampleset)  = 
    & \sum_{(\th, \classy) \in \sampleset} \sum_{\th^{-} \in \negsamp(\th)} \Big ( 
    \lossweight\log(\act(\enc(\th)\classemb_{:,\classy})) \\ 
    & + \sum_{i=1,i \neq \classy}^{\numclass} \log(1 - \act(\enc(\th)\classemb_{:,i}) )  \\ 
    & + \sum_{i=1}^{\numclass} \log(1 - \act(\enc(\th^{-})\classemb_{:,i}) ) \Big )
\end{split}
\label{equ:imgloss}
\end{align}

Here, $\lossweight$ is a hyperparameter to increase the weight of positive samples. $\negsamp(\th)$ represents the negative sample set of point $\th$ in which $\th^{-} \in \negsamp(\th)$ is a negative sample uniformly generated from the spherical surface given each data point $\th$. 
Equation \ref{equ:imgloss} can be seen as a modified version of the cross-entropy loss used in binary classification. The first term is the positive sample term weighted by $\lossweight$. The second term is the normal negative term used in cross-entropy loss. The third term is added to consider uniformly sampled locations as negative samples.

Figure \ref{fig:pos_enc} illustrates the whole workflow. During training time, the image classification module (the orange box) and location classification module (the blue box) are supervised trained separately. During the inference time, the probabilities $P(\classy|\image)$ and $P(\classy|\th)$ computed from these two modules are multiplied to yield the final prediction.

\section{ Baselines }\label{sec:baselines}

In order to understand the advantage of \mai{spherical-distance-kept} location encoders, we compare different versions of \emph{\modelname} with multiple baselines:
 \begin{itemize}
\item $\tile$
divides the study area $A$ (e.g., the \mai{earth's} surface) into grids with equal intervals along the latitude and longitude direction. Each grid has an embedding to be used as the encoding for every location $\th$ fall into this grid. 
This is a common practice \mai{adopted} by many previous \mai{works} when dealing with coordinate data \citep{berg2014birdsnap,adams2015frankenplace,tang2015improving}.

\item $\aodha$
is a location encoder model introduced by \citet{mac2019presence}. Given a location $\th = (\lon, \lat)$, it uses a coordinate wrap mechanism to convert each dimension of $\th$ into 2 numbers \mai{:  
\begin{align}
	\spe{\aodha}_1(\th) = [\sin(\lon), \cos(\lon), \sin(2\lat), \cos(2\lat)].
	\label{equ:wrap}
\end{align}}
Then the results are passed through a multi-layered fully connected neural network $\pemlp^{\aodha}()$ which consists of
an initial fully connected layer, followed by a series of $\numresnet$ residual blocks, each consisting of two fully connected layers ($\numneuron$ hidden neurons) with a dropout layer in between. We adopt the official code of \citet{mac2019presence}\footnote{\url{http://www.vision.caltech.edu/~macaodha/projects/geopriors/}} for this implementation. We can see that $\aodha$ still follows our general definition of location encoders $ \enc(\th) = \pemlp(PE_{\freq}(\th))$ where $\nscale  = 1$.

\item $\aodhaffn$
is similar to $\aodha$ except that it replaces $\pemlp^{\aodha}()$ with $\peffn()$, a simple learnable multi-layer perceptron with $\numresnet$ hidden layers and $\numneuron$ neurons per layer as that \emph{\modelname} has. $\aodhaffn$ is used to exclude the effect of different $\pemlp()$ on the performance of location encoders. In the following, all location encoder baselines use $\peffn()$ as the learnable neural network component so that we can directly compare the effect of different position encoding $\spe{*}_{\nscale}$ on the model performance.

\item $\xyz$
first converts $\th_i=(\lon_i,\lat_i) \in \coordspasphere^{2}$ into 3D Cartesian coordinates $(x, y, z)$ centered at the sphere center by following Equation \ref{equ:xyz} before feeding into a multilayer perceptron $\pemlp()$. Here, we let $(x, y, z)$ to locate on a unit sphere with radius $\radius=1$. As we can see, $\xyz$ is just a special case of $\sphere$ when $\nscale = 1$, i.e., $\spe{\sphere}_{1}$.
\begin{align}
\begin{split}
\spe{\xyz}_{\nscale}(\th) &= [z, x, y] =  \spe{\sphere}_{1}\\
&= [\sin\lat,\cos\lat\cos\lon,\cos\lat\sin\lon] 
\end{split}
\label{equ:xyz}
\end{align}

\item ${\rbf}$ 
randomly samples $M$ points from the training dataset as RBF anchor points \{$\th^{anchor}_{m}, m=1...M$\}, and use gaussian kernels $\exp{\big(-\dfrac{\parallel \th_i  - \th^{anchor}_{m} \parallel^2}{2\kernelsize^2}\big)}$ on each anchor points, where $\kernelsize$ is the kernel size. 
Each input point $\th_i$ is encoded as a $M$-dimension RBF feature vector, i.e., $\spe{\rbf}_{M}$, which is fed into $\peffn()$ to obtain the location embedding. This is a strong baseline for representing floating number features in machine learning models \mai{used by \citet{mai2020multiscale}}.

\item $\rff$, i.e., \textit{Random Fourier Features} \citep{rahimi2007rf, nguyen2017rf}, first encodes location $\th$ into a $\rffdim$ dimension vector - 
$
\spe{\rff}_{\rffdim}(\th) = \rffenc(\th) = \frac{\sqrt{2}}{\sqrt{\rffdim}} \bigcup_{i=1}^{\rffdim}[
\cos{(\dirvec_i^T \th + \shift_i)}]$
where 
$
\dirvec_i \iidsim
\mathcal{N}(\mathbf{0}, \rffcov^2 I)$ is a direction vector whose each dimension is independently sampled from a normal distribution. 
$\shift_i$ is uniformly sampled from $[0, 2\pi]$. $I$ is an identity matrix. 
Each component of $\rffenc(\bx)$ first projects $\th$ into a random direction $\dirvec_i$ and makes a shift by $\shift_i$. Then it wraps this line onto the unit cirle in $\Real^2$ with the cosine function. \citet{rahimi2007rf} show that $\rffenc(\th) ^T \rffenc(\th')$ is an unbiased estimator of the Gaussian kernal $K(\th,\th')$. $\rffenc(\th)$ is consist of $\rffdim$ different estimates to produce \mai{an approximation with a further lower variance. }
To make $\rff$ comparable to other baselines, we feed $\rffenc(\th)$ into $\peffn()$ to produce the final location embedding.

\item $\grid$ is a multi-scale location encoder on 2D Euclidean space proposed by \cite{mai2020multiscale}. Here, we simply treat $\th = (\lon, \lat)$ as 2D coordinate. It first use $\spe{\grid}_{\nscale}(\th)$ shown in Equation \ref{equ:grid} to encode location $\th$ into a multi-scale representation and then feed it into $\peffn()$ to produce the final location embedding. 

\item  $\theory$ is another multi-scale location encoder on 2D Euclidean space proposed by \cite{mai2020multiscale}. It use a position encoder $\spe{\theory}_{\nscale}(\th)$ shown in Equation \ref{equ:theory}. Here, $\ths = [\lons, \lats] = [\frac{\lon}{\fs}, \frac{\lat}{\fs}]$ and $\unit_1 = [1,0]^T, \unit_2 = [-1/2,\sqrt{3}/2]^T, \unit_3 = [-1/2,-\sqrt{3}/2]^T \in \Real^2$ are three unit vectors which orient $2\pi/3$ apart from each other. The encoding results are feed into $\peffn()$ to produce the final location embedding. \begin{align}
    \spe{\theory}_{\nscale}(\th)=\bigcup_{s=0}^{\nscale-1}\bigcup_{j=1}^{3}[\sin(\langle \ths, \unit_{j} \rangle), \cos(\langle \ths, \unit_{j} \rangle)].
    \label{equ:theory}
\end{align}

\item \mai{$\nerf$ indicates a multiscale location encoder adapted from the positional encoder $\spe{\nerf}_{\nscale}(\th)$ used by Neural Radiance Fields (NeRF) \citep{mildenhall2020nerf} and many NeRF variations such as NeRF-W \citep{martin2021nerfw}, S-NeRF \citep{derksen2021snerf}, Sat-NeRF \citep{mari2022satnerf}, GIRAFFE \citep{niemeyer2021giraffe}, etc., which was proposed for novel view synthesis for 3D scenes. Here, $\nerf$ can be treated as a multiscale version of $\xyz$. It first converts $\th=(\lon,\lat) \in \coordspasphere^{2}$ into 3D Cartesian coordinates $(x, y, z)$ centered at the unit sphere center. Here, $(x, y, z)$ are normalized to lie in $[-1,1]$, i.e., $\radius=1$. Different from $\xyz$, it uses NeRF-style positional encoder $\spe{\nerf}_{\nscale}(\th)$ in Equation \ref{equ:nerf} to process $(x, y, z)$ into a multiscale representation. To make it comparable with other location encoders, we further feed $\spe{\nerf}_{\nscale}(\th)$ into $\peffn()$ to get the final location embedding.
}
\mai{
\begin{align}
\begin{split}
    & \spe{\nerf}_{\nscale}(\th) =\bigcup_{s=0}^{\nscale-1}\bigcup_{p\in \{z, x, y\}}[\sin(2^{s}\pi p), \cos(2^{s}\pi p)], \\
    & where \; [z, x, y] = [\sin\lat,\cos\lat\cos\lon,\cos\lat\sin\lon] .
\end{split}
\label{equ:nerf}
\end{align}
}

\end{itemize}

All types of \emph{\modelname} as well as all baseline models we compared \mai{except $\tile$ }
share the same model set up - $ \enc(\th) = \pemlp(PE_{\nscale}(\th))$. 
The main difference is the position encoder $PE_{\nscale}(\th)$ used in different models.
$PE_{\nscale}(\th)$ used by $\grid$, $\theory$\mai{, $\nerf$,} and different types of \emph{\modelname} encode the input coordinates in a multi-scale fashion by using different sinusoidal functions with different frequencies. Many previous work call this practice ``Fourier input mapping'' \citep{rahaman2019spectral,tancik2020fourier,basri2020frequency,anokhin2021image}. The difference is that $\grid$ and $\theory$ use the Fourier features from 2D Euclidean space\mai{, $\nerf$ uses the predefined Fourier scales to directly encode the points in 3D Euclidean space,} while our \emph{\modelname} uses all or the subset of Double Fourier Sphere Features to take into account the spherical geometry and the distance distortion it brings.

All models are implemented in PyTorch. We use the original implementation of $\aodha$ from \citet{mac2019presence} and the implementation of $\grid$ and $\theory$ from \citet{mai2020multiscale}. 
\mai{Since the original implementation of NeRF\footnote{\url{https://github.com/bmild/nerf}} \citep{mildenhall2020nerf} is in TensorFlow, we reimplement $\nerf$ in PyTorch Framework by following their codes. }
We train and evaluate each model on a Ubuntu machine with 2 GeForce GTX Nvidia GPU cores, each of which has 10GB memory. 
\mai{\section{Theoretical Limitations of $\grid$ and $\nerf$} \label{sec:thm_limit}} 

\mai{\subsection{Theoretical Limitations of $\grid$} \label{sec:grid_thm}} 

\mai{
We first provide mathematic proofs to demonstrate why $\grid$ is not suitable to model spherical distances.
\begin{theorem}\label{thm:grid}
Let $\th_1$, $\th_2$ be two points on the same sphere $ \coordspasphere^{2} $ with radius $\radius$, then we have
\begin{align}
\begin{split}
    &\langle \spe{\grid}_\nscale(\th_1),  \spe{\grid}_\nscale(\th_2) \rangle \\
    &=\sum_{s=0}^{\nscale-1} \Big(\cos(\latz{s}_1-\latz{s}_2) +\cos(\lonz{s}_1-\lonz{s}_2) \Big) \\
    &=\sum_{s=0}^{\nscale-1} \Big(\cos(\frac{\lat_1-\lat_2}{\fs}) +\cos(\frac{\lon_1-\lon_2}{\fs}) \Big),
\end{split}
\label{eq:grid_cos_multi}
\end{align}
When $\nscale=1$, we have
\begin{align}
\begin{split}
    \langle \spe{\grid}_1(\th_1),  \spe{\grid}_1(\th_2) \rangle =\cos(\lat_1-\lat_2) +\cos(\lon_1-\lon_2),
\end{split}
\label{eq:grid_cos}
\end{align}
\end{theorem} 
}

\mai{Theorem \ref{thm:grid} is very easy to prove based on the angle difference formula, so we skip its proof. This result indicates that $\grid$ models the latitude  and longitude differences of $\th_1$ and $\th_2$ independently rather than spherical distance.
This introduces problems when encoding locations in the polar area.
Let's consider data pairs $\th_1=(\lon_1,\lat)$ and $\th_2=(\lon_2,\lat)$, the distance between them in output space of $\spe{\grid}_\nscale$ is:
\begin{align}
\begin{split}
    &\|\spe{\grid}_\nscale(\th_1)-\spe{\grid}_\nscale(\th_2)\|^2 \\
&=\|\spe{\grid}_\nscale(\th_1)\|^2 + \|\spe{\grid}_\nscale(\th_2)\|^2 \\
&-2\langle \spe{\grid}_\nscale(\th_1),  \spe{\grid}_\nscale(\th_2) \rangle \\
&= 2-2\sum_{s=0}^{\nscale-1} \cos(\frac{\lon_1-\lon_2}{\fs}) 
\end{split}
\label{eq:grid_diff_multi}
\end{align}
This distance stays as a constant for any values of $\lat$. However, when $\lat$ varies from $-\frac{\pi}{2}$ to $\frac{\pi}{2}$, the actual spherical distance changes in a wide range, e.g., the actual distance between the data pair at $\lat=-\frac{\pi}{2}$ (South Pole) is 0 while the distance between the data pair at $\lat=0$ (Equator), gets the maximum value. This problem in measuring distances also has a negative impact on $\grid$'s ability to model distributions
in areas with sparse sample points because it is hard to learn the true spherical distances.
}

\mai{
In fact, in our experiments ($\nscale > 1$), we observe that $\grid$ reaches peak performance at much smaller $\minscale$ than that of \emph{\modelname} encodings. Moreover, $\sphere$ outperforms $\grid$ near polar regions where $\grid$ produces large distances though the spherical distances are small (A, B in Figure \ref{fig:pos_enc}).
}

\mai{\subsection{Theoretical Limitations of $\nerf$} \label{sec:nerf_thm}}

\mai{
Since $\nerf$ is widely used for 3D representation learning \citep{mildenhall2020nerf,niemeyer2021giraffe}, a natural question is why not just use $\nerf$ for the geographic prediction tasks on the spherical surface, which can be embedded in the 3D space.
In this section, we discuss the theoretical limitations of $\nerf$ 3D multiscale encoding in the scenario of spherical encoding. }

\mai{
\begin{theorem}\label{thm3}
Let $\th_1, \th_2 \in \coordspasphere^{2}$ be two points on the spherical surface. 
Given their 3D Euclidean representations, i.e.,  
$\th_1=(z_1,x_1,y_1)$, $\th_2=(z_2,x_2,y_2)$, we define $\Delta\th = \th_1 - \th_2 = [z_1-z_2, x_1-x_2, y_1-y_2] = [\Delta\th_{z},\Delta\th_{x},\Delta\th_{y}]$ as the difference between them in the 3D Euclidean space.
Under $\nerf$ encoding  (Equation \ref{equ:nerf}), the distance between them satisfies   
\begin{align}
    \begin{split}
    &\|\spe{\nerf}_{\nscale}(\th_1)-\spe{\nerf}_{\nscale}(\th_2)\|^2  \\
    &= \sum_{s=0}^{\nscale-1} \Big(4\sin^2(2^{s-1}\pi \Delta\th_z) + 4\sin^2(2^{s-1}\pi \Delta\th_x)  \\
    &+ 4\sin^2(2^{s-1}\pi \Delta\th_y)\Big)  \\
    &= \sum_{s=0}^{\nscale-1} 4\|\vec{Y}_{s}\|^2,\\
\end{split}
    \label{equ:nerf_proof}
\end{align}
where $\vec{Y}_{s}=[\sin(2^{s-1}\pi \Delta\th_z), \sin(2^{s-1}\pi \Delta\th_x), \sin(2^{s-1}\pi \Delta\th_y)]$.
\end{theorem}
}

\mai{
See the proof in Appendix \ref{proof3}.
}

\mai{
\begin{theorem}\label{thm4}
$\nerf$ is not an injective function.
\end{theorem}
}

\mai{
Theorem \ref{thm4} is very easy to prove based on Theorem \ref{thm3}.
Since $\nerf$ requires $\radius=1$, when $\th_1=(1,0,0)$ and $\th_2=(-1,0,0)$, i.e., they are the north and south pole, we have $\Delta\th = [2, 0, 0]$. The distance between their multiscale $\nerf$ encoding is,
\begin{align}
    \|\spe{\nerf}_{\nscale}(\th_1)-\spe{\nerf}_{\nscale}(\th_2)\|^2=\sum_{s=0}^{\nscale-1} 4 \sin^2(2^{s}\pi)=0,
\end{align}
Since Equation \ref{equ:nerf_proof} is symmetrical for the x,y, and z axis, we will have the same problems when $\th_1=(0,1,0)$, $\th_2=(0,-1,0)$ or $\th_1=(0,0,1)$, $\th_2=(0,0,-1)$. This indicates that even though these three pairs of points have the largest spherical distances, they have identical $\nerf$ multiscale representations. This illustrates that $\nerf$ is \textit{not} an injective function.
}

\mai{
Theorem \ref{thm3} shows that, unlike \modelname, the distance between two $\nerf$ location embedding is not a monotonic increasing function of $\sd$, but a non-monotonic function of the coordinates of $\Delta\th$, the axis-wise differences between two points in 3D Euclidean space. 
So $\nerf$ does not preserve spherical distance for spherical points, but rather models $\Delta\th_{z},\Delta\th_{x},\Delta\th_{y}$ separately. }

\begin{figure*}
	\centering \small \vspace*{-0.2cm}
	\begin{subfigure}[b]{0.48\textwidth}  
		\centering 
		\includegraphics[width=\textwidth]{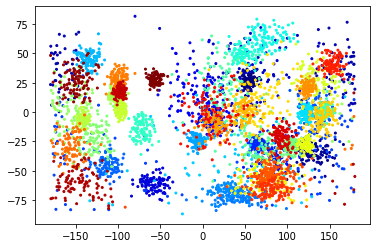}
		\vspacecomp
		\caption[]{{\small 
		U1 dataset in 2D degree Space ($\vmfkappamax=16$)
		}}    
		\label{fig:vmfC50S100L1H16_2d}
	\end{subfigure}
	\hfill
	\begin{subfigure}[b]{0.48\textwidth}  
		\centering 
		\includegraphics[width=\textwidth]{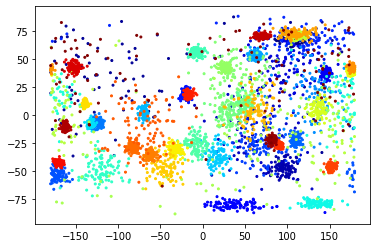}
		\vspacecomp
		\caption[]{{\small 
		U2 dataset in 2D degree Space ($\vmfkappamax=32$)
		}}    
		\label{fig:vmfC50S100L1H32_2d}
	\end{subfigure}
	\hfill
	\begin{subfigure}[b]{0.48\textwidth}  
		\centering 
		\includegraphics[width=\textwidth]{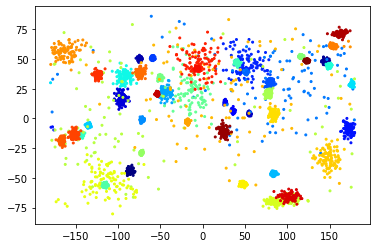}
		\vspacecomp
		\caption[]{{\small 
		U3 dataset in 2D degree Space ($\vmfkappamax=64$)
		}}    
		\label{fig:vmfC50S100L1H64IDX1_2d}
	\end{subfigure}
	\hfill
	\begin{subfigure}[b]{0.48\textwidth}  
		\centering 
		\includegraphics[width=\textwidth]{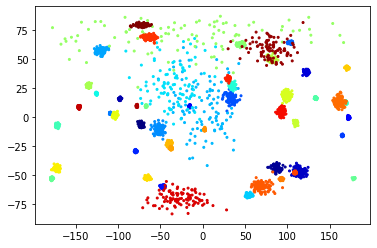}
		\vspacecomp
		\caption[]{{\small 
		U4 dataset in 2D degree Space ($\vmfkappamax=128$)
		}}    
		\label{fig:vmfC50S100L1H128IDX1_2d}
	\end{subfigure}
	\hfill
	\begin{subfigure}[b]{0.68\textwidth}  
		\centering 
		\includegraphics[width=\textwidth]{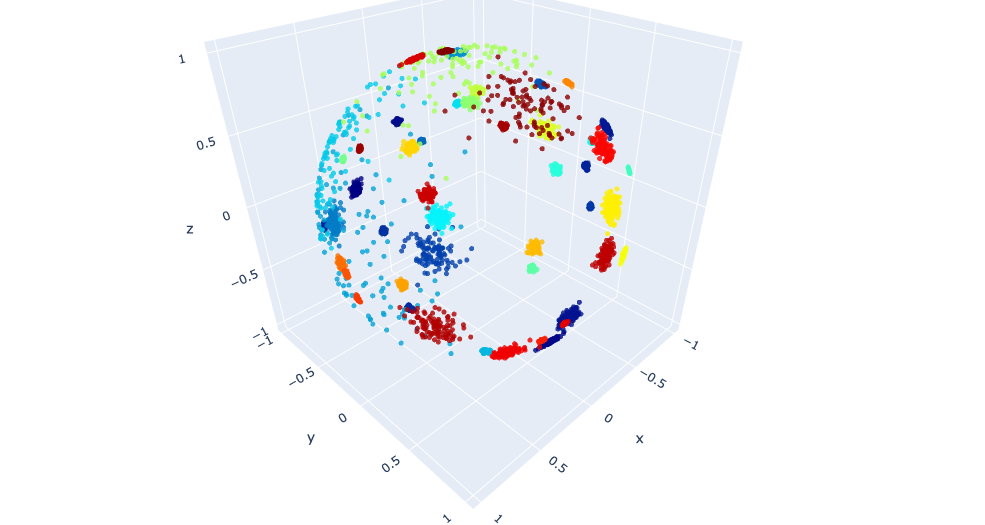}
		\vspacecomp
		\caption[]{{\small 
		U4 dataset in 3D space ($\vmfkappamax=128$)
		}}    
		\label{fig:vmfC50S100L1H128IDX1_3d}
	\end{subfigure}
	\caption{The data distributions of four synthetic datasets (U1, U2, U3, and U4) generated from the uniform sampling method. (e) shows the U4 dataset in a 3D Euclidean space. We can see that if we treat these datasets as 2D data points as $\grid$ and $\theory$, the $\vmf$ distributions in the polar areas will be stretched and look like 2D aniostropic multivariate Gaussian distributions. However, this kind of systematic bias can be avoided if we use a spherical location encoder as $\modelname$.
	} 
	\label{fig:synthetic_uniform_2d_3d}
	\vspace*{-0.15cm}
\end{figure*}

\begin{figure*}
	\centering \small \vspace*{-0.2cm}
	\begin{subfigure}[b]{0.48\textwidth}  
		\centering 
		\includegraphics[width=\textwidth]{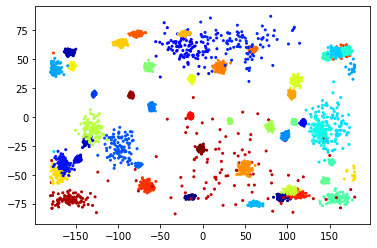}
		\vspacecomp
		\caption[]{{\small 
		S1.3 dataset in 2D degree Space ($\numvmfMUinterval=5$)
		}}    
		\label{fig:vmfIN5CL10MrandomS100L1H64_2d}
	\end{subfigure}
	\hfill
	\begin{subfigure}[b]{0.48\textwidth}  
		\centering 
		\includegraphics[width=\textwidth]{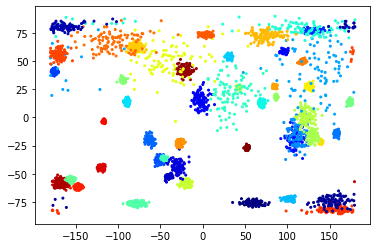}
		\vspacecomp
		\caption[]{{\small 
		S2.3 dataset in 2D degree Space ($\numvmfMUinterval=10$)
		}}    
		\label{fig:vmfIN10CL5MrandomS100L1H64_2d}
	\end{subfigure}
	\hfill
	\begin{subfigure}[b]{0.48\textwidth}  
		\centering 
		\includegraphics[width=\textwidth]{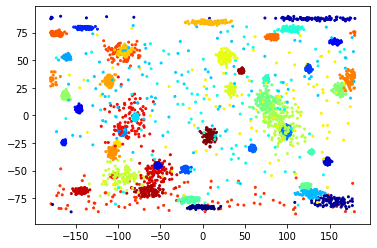}
		\vspacecomp
		\caption[]{{\small 
		S3.3 dataset in 2D degree Space ($\numvmfMUinterval=25$)
		}}    
		\label{fig:vmfIN25CL2MrandomS100L1H64_2d}
	\end{subfigure}
	\hfill
	\begin{subfigure}[b]{0.48\textwidth}  
		\centering 
		\includegraphics[width=\textwidth]{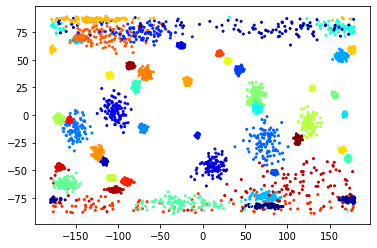}
		\vspacecomp
		\caption[]{{\small 
		S4.3 dataset in 2D degree Space ($\numvmfMUinterval=50$)
		}}    
		\label{fig:vmfIN50CL1MrandomS100L1H64_2d}
	\end{subfigure}
	\caption{
	The data distributions of four synthetic datasets (S1.3, S2.3, S3.3, and S4.3) generated from the stratified sampling method with $\vmfkappamax=64$. We can see that when $\numvmfMUinterval$ increases, a more fine-grain stratified sampling is carried out. The resulting dataset has a larger data bias toward the polar areas.
	} 
	\label{fig:synthetic_stratified_2d_3d}
	\vspace*{-0.15cm}
\end{figure*}
 
\section{Experiments with Synthetic Datasets}  \label{sec:synthetic}

\mai{Theorem \ref{thm1} and \ref{the:injective} provide theoretical guarantees}
 of \emph{\modelname} for spherical distance preservation. 
To empirically verify the effectiveness of 
\emph{\modelname} in a controlled setting, we construct a set of synthetic datasets and evaluate our \emph{\modelname} and all baseline models on these datasets. To make a simpler task, different from the setting shown in Figure \ref{fig:pos_enc}, 
we skip the image encoder component and only focus on the location encoder training and evaluation. For each synthetic dataset, we simulate a set of spherical coordinates as the geo-locations of images to train different location encoders. And in the evaluation step, the performances of different models are  computed directly based on $P(\classy|\th)$ only, but not $P(\classy|\th)P(\classy|\image)$.

\subsection{Synthetic Dataset Generation} \label{sec:synthetic_generation}

We utilize the von Mises–Fisher distribution ($\vmf$) \citep{izbicki2019exploiting}, an analogy of the 2D Gaussian distribution on the spherical surface $\coordspasphere^{2}$ to generate synthetic data points\footnote{\url{https://www.tensorflow.org/probability/api\_docs/python/tfp/distributions/VonMisesFisher\#sample}}. 
The probability density function of $\vmf$ is defined as
\begin{align}
    \begin{split}
    \vmf(\th; \mu, \vmfkappa) & = \dfrac{\vmfkappa}{2\pi\sinh(\vmfkappa)} \exp(\vmfkappa\vmfmu^{T}\latlontoxyz(\th))    \end{split}
    \label{equ:vmf_fun}
\end{align}
where $\latlontoxyz(\th) = [x, y, z] = [\cos\lat\cos\lon,\cos\lat\sin\lon,\sin\lat]$,
which converts $\th$ into a coordinates in the 3D Euclidean space on the surface of a unit sphere. 
A $\vmf$ distribution is controlled by two parameters -- the mean direction $\vmfmu \in \Real^{3}$ and concentration parameter $\vmfkappa \in \Real^{+}$. $\vmfmu$ indicates the center of a $\vmf$ distribution which is a 3D unit vector. $\vmfkappa$ is a positive real number which controls the concentration of $\vmf$. A higher $\vmfkappa$ indicates more compact $\vmf$ distribution, while $\vmfkappa = 1$ correspond to a $\vmf$ distribution with standard deviation covering half of the unit sphere.

To simulate multi-modal distributions, we generate spherical coordinates based on a mixture of von Mises–Fisher distributions (MvMF). 
We assume  $\numvmfcla$ classes with even prior, and each classes follows a $\vmf$ distribution.
To create a dataset we first sample $\numvmfcla$ sets of parameters $\{(\mu_i, \vmfkappa_i)\}$ ($\numvmfcla=50$).
Then we draw $\numvmfsamplepercla$ samples, i.e., spherical coordinates, for each class ($\numvmfsamplepercla=100$).
So in total, each generated synthetic dataset has 5000 data points for 50 balanced classes. 

The concentration parameter $\vmfkappa_i$ is sampled by first drawing  $r$ from an uniform distribtuion $U(\vmfkappamin, \vmfkappamax)$, and then take the square $r^2$. The square helps to avoid sampling many large $\vmfkappa_i$ which yield very concentrated $\vmf$ distributions that are rather easy to be classified. 
We fix $\vmfkappamin=1$ and vary $\vmfkappamax$ in $[16, 32, 64, 128]$.

For the center parameter $\mu_i$  we adopt two sampling approaches:
\begin{enumerate}
    \item \textbf{Uniform Sampling}: We uniformly sample $\numvmfcla$ centers ($\mu_i$) from the surface of a unit sphere. 
We generate four synthetic datasets (for different values of $\vmfkappamax$) and indicate them as U1, U2, U3, U4. 
    See Table \ref{tab:synthetic_tab} for the parameters we use to generate these datasets. 
    \item \textbf{Stratified Sampling}: We first evenly divide the latitude range $[-\pi/2, \pi/2]$ into $\numvmfMUinterval$ intervals. 
    Then we uniformly sample $\numvmfMUclaperinterval$ centers ($\mu_i$) from the spherical surface defined by each latitude interval. 
    Since the latitude intervals in polar regions have smaller spherical surface area, this stratified sampling method has higher density in the polar regions. 
    We keep $\numvmfMUinterval \times \numvmfMUclaperinterval = \numvmfcla = 50$ fixed and varies $\numvmfMUinterval$ in $[5, 10, 25, 50]$. 
Combined with the 4 $\vmfkappamax$ choices, this procedure yields  16 different synthetic datasets. We denote them as $Si.j$. See Table \ref{tab:synthetic_tab} for the parameters we use to generate these datasets. 
\end{enumerate}

Figure \ref{fig:vmfC50S100L1H16_2d}-\ref{fig:vmfC50S100L1H128IDX1_2d} visualize the data point distributions of U1, U2, U3, U4 which derived from the uniform sampling method in 2D space. Figure \ref{fig:vmfC50S100L1H128IDX1_3d} visualized the U4 dataset in a 3D Euclidean space. We can see that when $\vmfkappamax$ is larger, the variation of point density among different $\vmf$ distributions becomes larger. Some $\vmf$ are very concentrated and the resulting data points are easier to be classified. Moreover, if we treat these datasets as 2D data points as $\grid$ and $\theory$ do, $\vmf$ distributions in the polar areas will be stretched 
to very extended shapes making model learning more difficult. 
However, this kind of systematic bias can be avoided if we use a spherical location encoder as \emph{\modelname}.

Figure \ref{fig:synthetic_stratified_2d_3d} visualizes the data distributions of four synthetic datasets with stratified sampling method. They have different $\numvmfMUinterval$ but the same $\vmfkappamax$. We can see that when $\numvmfMUinterval$ increases, a more fine-grain stratified sampling is carried out. The resulting dataset has a larger data bias toward the polar areas.

\begin{table*}[]
\caption{
Compare $\spheregridmixscale$ to baselines on synthetic datasets. We use Top1 as the evaluation metric. 
U1 - U4 indicate 4 synthetic datasets generated based on the uniform sampling approach (see Section \ref{sec:synthetic_generation}). S1.1 - S4.4 indicate 16 synthetic datasets generated based on the stratified sampling apprach. 
For all datasets have $\numvmfcla=50$ and $\numvmfsamplepercla=100$. 
For each model, we perform grid search on its hyperparameters for each dataset and report the best Top1 accuracy.
The $\Delta Top1$ column shows the absolute  performance improvement of $\spheregridmixscale$ over the best baseline model (bolded) for each dataset.
The $ER$ column shows the relative reduction of error compared to the best baseline model (bolded). 
We can see that $\spheregridmixscale$ can outperform all other baseline models on all of these 20 synthetic datasets. 
The absolute Top1 accuracy improvement can be as much as 2.0\% for datasets with lower precisions, and the error rate deduction can be as much as $30.8\%$ for datasets with high precisions.
	}
	\label{tab:synthetic_tab}
	\centering
	\setlength{\tabcolsep}{3pt}
	
{
\footnotesize
\begin{tabular}{
c|c|cc|cc|
|cccccccc|c||c|c}
\toprule
ID & Method                       & $\numvmfMUinterval$       & $\numvmfMUclaperinterval$ & $\vmfkappamin$         & $\vmfkappamax$ & $\xyz$  & $\aodha$      & $\aodha$+\textit{ffn} & $\rff$  & $\rbf$           & $\grid$      & $\theory$        & \mai{$\nerf$} & $\spheregridmixscale$ & $\Delta Top1$ & $ER$ \\ \hline
U1 &
\multirow{4}{*}{uniform}     & \multirow{4}{*}{-}  & \multirow{4}{*}{-}          & \multirow{4}{*}{1} & 16          & 67.2 & 67.0          & 66.9          & 66.8 & 46.6          & \textbf{68.6} & 67.8         & \mai{62.7} & \textbf{69.2}      & 0.6        & -1.9       \\
U2 &                                           &                     &                             &                    & 32          & 73.1 & 75.1          & 73.9          & 72.3 & 58.4          & 76.2          & \textbf{76.5} &  \mai{72.5} & \textbf{77.4}      & 0.9        & -3.8       \\
U3 &                                                 &                     &                             &                    & 64          & 86.1 & 90.1          & 88.3          & 89.0 & 91.7          & 92.3          & \textbf{92.7} &  \mai{90.1} & \textbf{93.3}      & 0.6        & -8.2       \\
U4 &                                               &                     &                             &                    & 128         & 91.8 & 94.9    & 92.3          & 92.5 & 97.4          & 97.5          & \textbf{97.7} & \mai{95.7} & \textbf{98.0}      & 0.3        & -13.0      \\ \hline S1.1 &                  \multirow{16}{*}{stratified} & \multirow{4}{*}{5}  & \multirow{4}{*}{10}         & \multirow{4}{*}{1} & 16          & 68.7 & 69.7          & 68.8          & 68.6 & \textbf{70.5} & 69.5          & 69.4          & \mai{66.5} & \textbf{72.3}      & 1.8        & -6.1       \\
S1.2 &                                             &                     &                             &                    & 32          & 76.7 & 79.1          & 78.1          & 78.4 & 81.1          & \textbf{81.2} & 79.2       & \mai{76.1}   & \textbf{82.9}      & 1.7        & -9.0       \\
S1.3 &                                             &                     &                             &                    & 64          & 91.2 & 92.5          & 92.9          & 92.6 & 94.7          & 94.8          & \textbf{94.9} & \mai{92.1} & \textbf{95.4}      & 0.5        & -9.8       \\
S1.4 &                                           &                     &                             &                    & 128         & 86.5 & 91.6          & 88.3          & 92.4 & 93.5          & \textbf{95.2} & 94.9          & \mai{92.4} & \textbf{96.1}      & 0.9        & -18.7      \\ \cline{1-1} \cline{3-16} 
S2.1 &                                         & \multirow{4}{*}{10} & \multirow{4}{*}{5}          & \multirow{4}{*}{1} & 16          & 70.5 & 71.3          & 70.7          & 70.4 & 46.6          & \textbf{72.0} & 70.7          & \mai{67.0} & \textbf{74.0}      & 2.0        & -7.1       \\
S2.2 &                                             &                     &                             &                    & 32          & 76.1 & 79.7          & 78.2          & 78.6 & 61.2          & \textbf{80.9} & 80.5      & \mai{77.6}    & \textbf{82.3}      & 1.4        & -7.3       \\
S2.3 &                                             &                     &                             &                    & 64          & 88.0 & 89.9          & 88.2          & 88.5 & 80.0          & \textbf{92.5} & 91.9      & \mai{89.0}    & \textbf{93.3}      & 0.8        & -10.7      \\
S2.4 &                                        &                     &                             &                    & 128         & 94.4 & 96.6          & 96.7          & 95.5 & 94.0          & \textbf{97.6} & 97.6          &  \mai{96.2} & \textbf{98.1}      & 0.5        & -20.8      \\ \cline{1-1} \cline{3-16} 
S3.1 &                                             & \multirow{4}{*}{25} & \multirow{4}{*}{2}          & \multirow{4}{*}{1} & 16          & 66.2 & 66.3          & 64.7          & 65.6 & \textbf{67.1} & 66.7          & 66.7          & \mai{61.3} & \textbf{68.3}      & 1.2        & -3.6       \\
S3.2 &                                           &                     &                             &                    & 32          & 80   & 82.5          & 80.7          & 81.6 & 83.4          & \textbf{84.5} & 82.1          & \mai{80.3} & \textbf{85.9}      & 1.4        & -9.0       \\
S3.3 &                                         &                     &                             &                    & 64          & 85.4 & 86.0          & 85.7          & 86.2 & 89.1          & \textbf{89.6} & 88.6          & \mai{86.1} & \textbf{91.0}      & 1.4        & -13.5      \\
S3.4 &                                           &                     &                             &                    & 128         & 93.2 & 96.0          & 94.8          & 95.7 & 97.2          & 97.3          & \textbf{97.4} & \mai{96.7} & \textbf{98.0}      & 0.6        & -23.1      \\ \cline{1-1} \cline{3-16} 
S4.1 &                                             & \multirow{4}{*}{50} & \multirow{4}{*}{1}          & \multirow{4}{*}{1} & 16          & 64.8 & \textbf{67.4} & 66.0          & 66.3 & 66.9          & 67.1          & 64.5          & \mai{62.9} & \textbf{68.4}      & 1          & -3.1       \\
S4.2 &                                             &                     &                             &                    & 32          & 75.6 & 78.2          & 77.4          & 77.4 & 78.4          & \textbf{80.1} & 78.3    & \mai{75.7}      & \textbf{81.0}      & 0.9        & -4.5       \\
S4.3 &                                           &                     &                             &                    & 64          & 91.3 & 93.9          & 93.7          & 93.8 & 95.0          & \textbf{95.2} & 94.0          & \mai{92.5} & \textbf{96.1}      & 0.9        & -18.7      \\
S4.4 &                                             &                     &                             &                    & 128         & 94.3 & 95.5          & 94.4          & 94.7 & 95.4          & \textbf{97.4} & 96.5      & \mai{95.2}    & \textbf{98.2}      & 0.8        & -30.8     \\ \bottomrule
\end{tabular}
}
\end{table*}
 
\subsection{Synthetic Dataset Evaluation Results} \label{sec:synthetic_eval}

We evaluate all baseline models as well as $\spheregridmixscale$ on those generated 20 syhthetic datasets as described above. 
For each model, we do grid search on their hyperparameters for each dataset including supervised learning rate $\lr$, the number of scales $\nscale$, the minimum scaling factor $\minscale$, the number of hidden layers and number of neurons used in $\peffn(\cdot)$ -- $\numresnet$ and $\numneuron $. The best performance of each model is reported in Table \ref{tab:synthetic_tab}. 
We use Top1 as the evaluation metric. The Topk classification accuracy is defined as follow
\begin{align}
    TOP_k = \dfrac{1}{|\dataset|}\sum_{i=1}^{|\dataset|} \mathbf{1}(Rank(\th_i, \classy_i) \geqslant k)
    \label{equ:topk_loc}
\end{align}
where $\dataset = \{(\th_i, \classy_i)\}$ is a set of location $\th_i$ and label $\classy_i$ tuples which indicates the whole validation or testing set. $|\dataset|$ denotes the total number of samples in $\dataset$. $Rank(\th_i, \classy_i)$ indicates the rank of the ground truth label $\classy_i$ in the ranked listed of all classes based on the probability score $P(\classy_i|\th_i)$ given by a specific location encoder. A lower rank indicates a better model prediction. $\mathbf{1}(*)$ is a function return 1 when the condition $*$ is true and 0 otherwise. A higher Topk indicates a better performance.

Some observations can be made from Table \ref{tab:synthetic_tab}:
\begin{enumerate}
    \item $\spheregridmixscale$ is able to outperform all baselines on all 20 synthetic datasets. The absolute Top1 improvement can go up to 2\% and the error rate deduction can go up to $30.8\%$. This shows the robustness of $\spheregridmixscale$.
    \item When the dataset is fairly easy to  classify (i.e., all baseline models can produce 95+\% Top1 accuracy), $\spheregridmixscale$ is still able to further improve the performance and gives a very large error rate reduction (up to 30.8\%). 
    This indicates that $\spheregridmixscale$ is very robust and reliable for datasets with different distribution characteristics.
    \item Comparing the error rate of different stratified sampling generated datasets (S1.j - S4.j) we can see that when we keep $\vmfkappamax$ fixed and increase $\numvmfMUinterval$, the relative error reduction $ER$ 
become larger. Increasing $\numvmfMUinterval$ means we do a more \textit{fine-grain} stratified sampling. The resulting datasets should sample more $\vmf$ distributions in the polar regions. This phenomenon shows that \textbf{when the dataset has a larger data bias towards the polar area, we expect $\spheregridmixscale$ to be more effective.}
    \item \mai{From Table \ref{tab:synthetic_tab}, we can also see that among all the baseline methods, $\grid$ achieves the best performances on most datasets (12 out of 20), followed by $\theory$ (5 out of 20). This observation aligns the experiment results from \citet{mai2020multiscale} which shows the advantages of multiscale location representation versus single-scale representations. }
    \item \mai{It is interesting to see that although $\nerf$ is also a multiscale location encoding approach, it underperforms $\grid$ and $\theory$ on all synthetic datasets. We guess the reasons are 1) $\nerf$ treats geo-coordinates as 3D Euclidean coordinates and ignores the fact that they are all on the spherical surface which yields more modeling freedom and makes it more difficult for $\peffn$ to learn; 
2) $\nerf$ uses predefined Fourier scales, i.e., $\{2^0, 2^1,...,2^s,...,2^{\nscale-1}\}$, while $\grid$, $\theory$, and $\modelname$ are more flexible in terms of Fourier scale choices which are controlled by $\maxscale$ and $\minscale$.  }
\end{enumerate}

\section{Experiment with Geo-Aware Image Classification}
\label{sec:exp}

Next, we empirically evaluate the performances of our \emph{\modelname} as well as \mai{all 9 baseline methods}
on 7 real-world datasets for the geo-aware image classification task.

\subsection{Dataset} \label{sec:dataset}

\begin{figure*}
	\centering \tiny
	\begin{subfigure}[b]{0.195\textwidth}  
		\centering 
		\includegraphics[width=\textwidth]{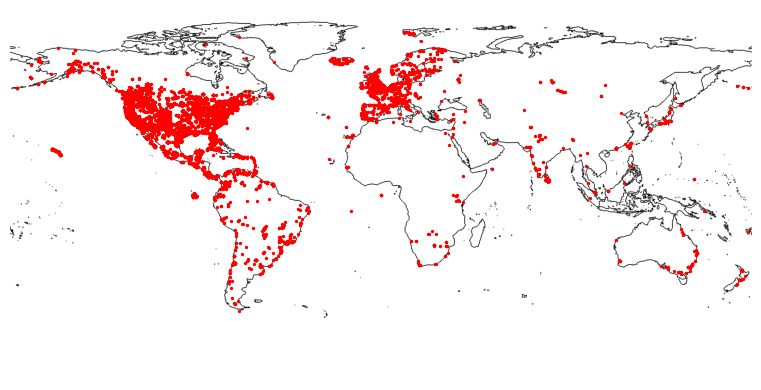}\vspacepred
		\caption[]{{\small 
		BirdSnap Train
		}}    
		\label{fig:birdsnap_orig_train_locs}
	\end{subfigure}
	\hfill
	\begin{subfigure}[b]{0.195\textwidth}  
		\centering 
		\includegraphics[width=\textwidth]{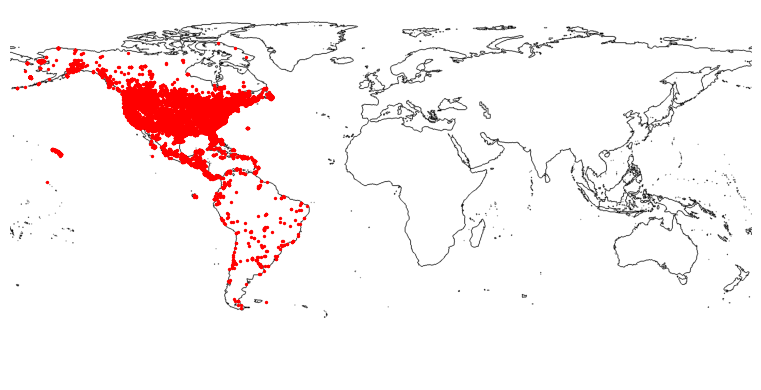}\vspacepred
		\caption[]{{\small 
		BirdSnap† Train
		}}    
		\label{fig:birdsnap_ebird_train_locs}
	\end{subfigure}
	\hfill
	\begin{subfigure}[b]{0.195\textwidth}  
		\centering 
		\includegraphics[width=\textwidth]{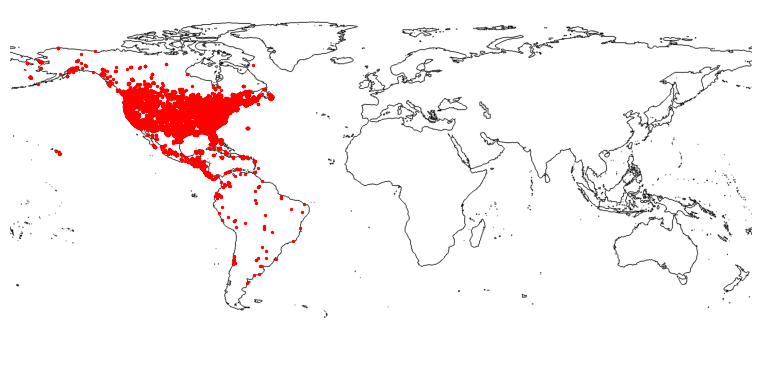}\vspacepred
		\caption[]{{\small 
		NABirds† Train
		}}    
		\label{fig:nabirds_train_locs}
	\end{subfigure}
	\hfill
	\begin{subfigure}[b]{0.195\textwidth}  
		\centering 
		\includegraphics[width=\textwidth]{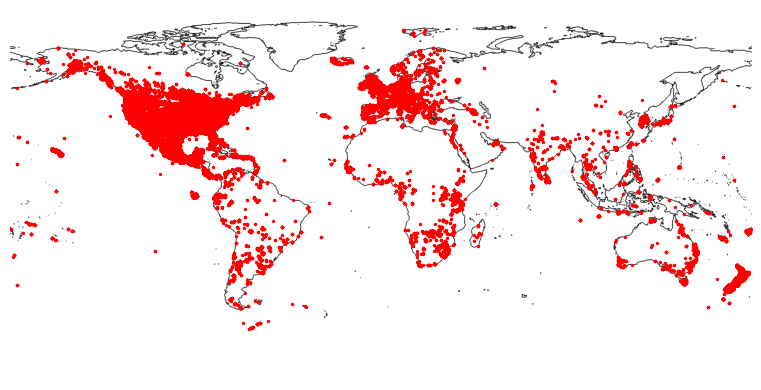}\vspacepred
		\caption[]{{\small 
		iNat2017 Train
		}}    
		\label{fig:inat_2017_train_locs}
	\end{subfigure}
	\hfill
	\begin{subfigure}[b]{0.195\textwidth}  
		\centering 
		\includegraphics[width=\textwidth]{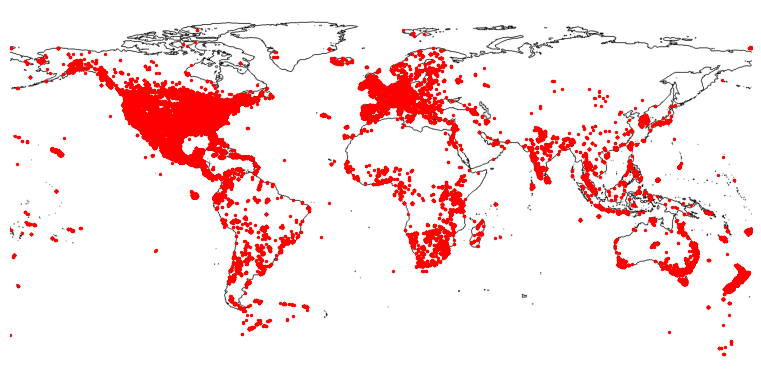}\vspacepred
		\caption[]{{\small 
		iNat2018 Train
		}}    
		\label{fig:inat_2018_train_locs}
	\end{subfigure}
	\hfill
	\begin{subfigure}[b]{0.195\textwidth}  
		\centering 
		\includegraphics[width=\textwidth]{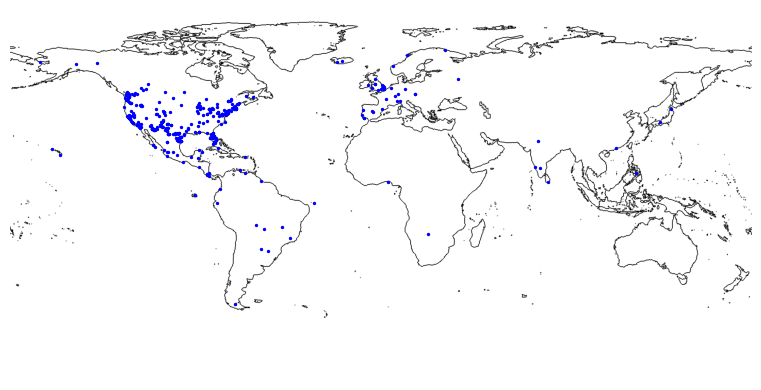}\vspacepred
		\caption[]{{\small 
		BirdSnap Test
		}}    
		\label{fig:birdsnap_orig_test_locs}
	\end{subfigure}
	\hfill
	\begin{subfigure}[b]{0.195\textwidth}  
		\centering 
		\includegraphics[width=\textwidth]{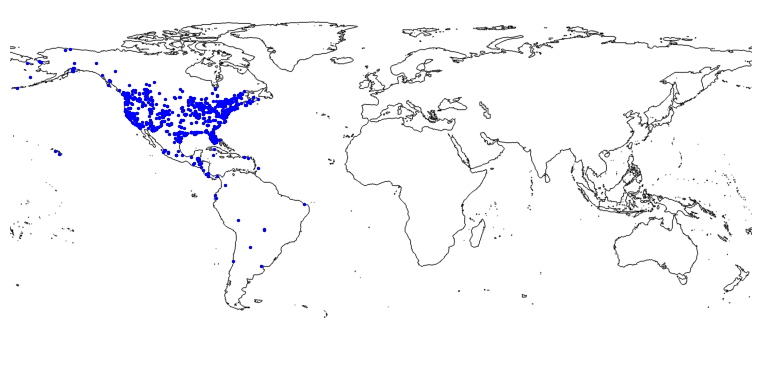}\vspacepred
		\caption[]{{\small 
		BirdSnap† Test
		}}    
		\label{fig:birdsnap_ebird_test_locs}
	\end{subfigure}
	\hfill
	\begin{subfigure}[b]{0.195\textwidth}  
		\centering 
		\includegraphics[width=\textwidth]{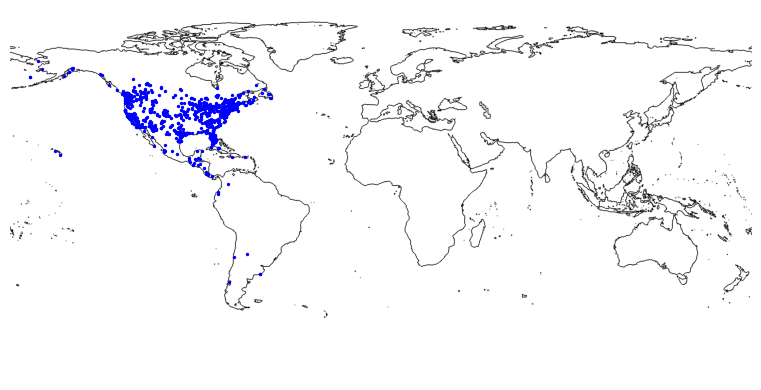}\vspacepred
		\caption[]{{\small 
		NABirds† Test
		}}    
		\label{fig:nabirds_test_locs}
	\end{subfigure}
	\hfill
	\begin{subfigure}[b]{0.195\textwidth}  
		\centering 
		\includegraphics[width=\textwidth]{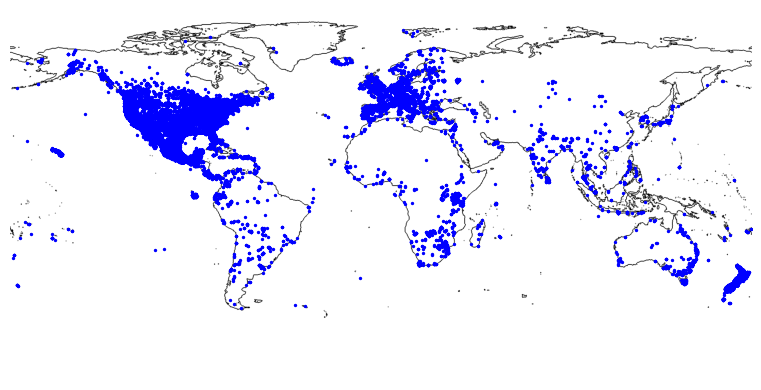}\vspacepred
		\caption[]{{\small 
		iNat2017 Val
		}}    
		\label{fig:inat_2017_val_locs}
	\end{subfigure}
	\hfill
	\begin{subfigure}[b]{0.195\textwidth}  
		\centering 
		\includegraphics[width=\textwidth]{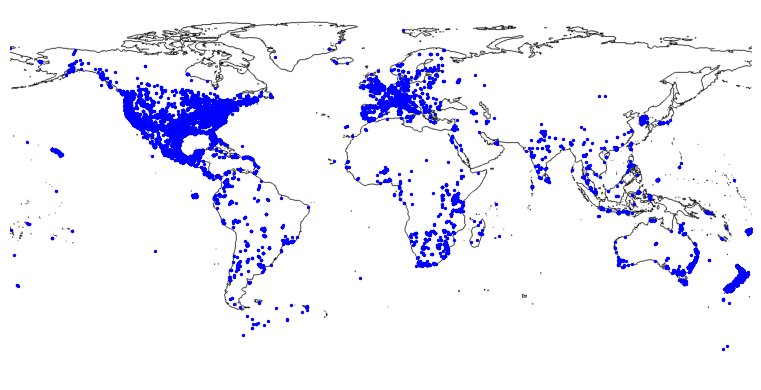}\vspacepred
		\caption[]{{\small 
		iNat2018 Val
		}}    
		\label{fig:inat_2018_val_locs}
	\end{subfigure}
	\caption{
	Training, validation/testing locations of different fine-grained species recognition datasets. Different datasets use either validation or testing dataset to evaluate model performance. So we plot their corresponding image geographic distributions.
	} 
	\label{fig:species_loc_dist}
\end{figure*}

\begin{figure*}
	\centering \tiny
	\begin{subfigure}[b]{0.24\textwidth}  
		\centering 
		\includegraphics[width=\textwidth]{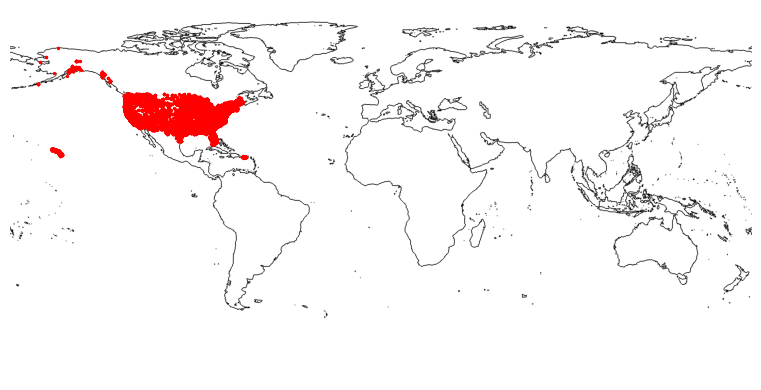}\vspacepred
		\caption[]{{\small 
		YFCC Train
		}}    
		\label{fig:yfcc_train_locs}
	\end{subfigure}
\begin{subfigure}[b]{0.24\textwidth}  
		\centering 
		\includegraphics[width=\textwidth]{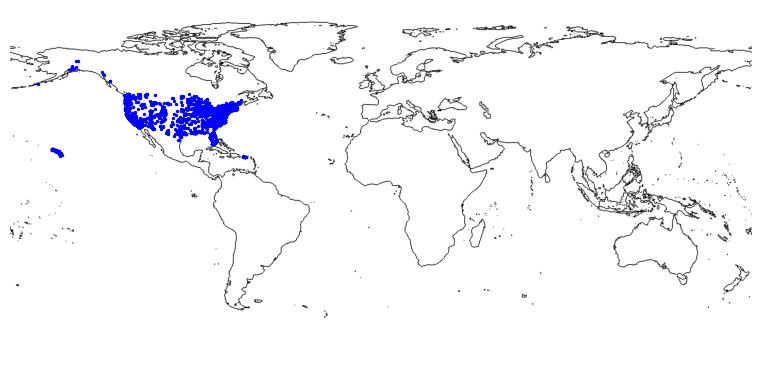}\vspacepred
		\caption[]{{\small 
		YFCC Test
		}}    
		\label{fig:yfcc_test_locs}
	\end{subfigure}
\begin{subfigure}[b]{0.24\textwidth}  
		\centering 
		\includegraphics[width=\textwidth]{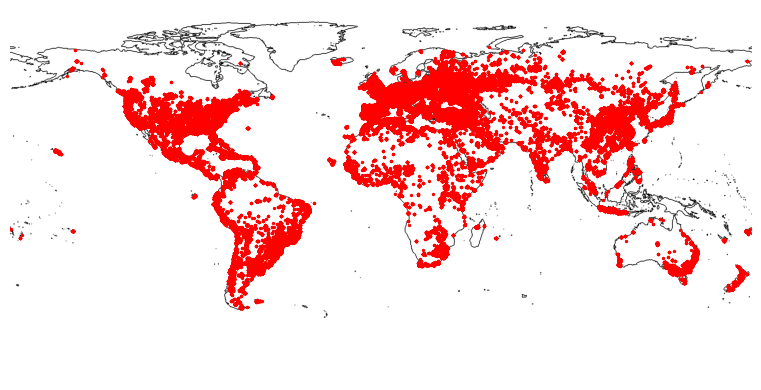}\vspacepred
		\caption[]{{\small 
		fMoW Train
		}}    
		\label{fig:fmow_train_locs}
	\end{subfigure}
\begin{subfigure}[b]{0.24\textwidth}  
		\centering 
		\includegraphics[width=\textwidth]{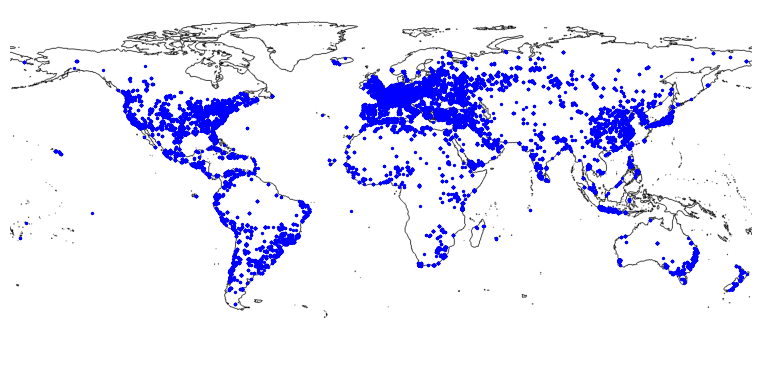}\vspacepred
		\caption[]{{\small 
		fMoW Val
		}}    
		\label{fig:fmow_test_locs}
	\end{subfigure}
	\caption{
	Training and validation/testing locations of Flickr image recognition (YFCC) and RS image classification (fMoW).
	} 
	\label{fig:rs_loc_dist}
\end{figure*}

More specifically, we test the performances of different location encoders on seven datasets from three different problems: fine-grained species recognition, Flickr image recognition, and remote sensing image classification.
The statistics of these seven datasets are shown in Table \ref{tab:data_stat}. Figure \ref{fig:species_loc_dist} and \ref{fig:rs_loc_dist} show the spatial distributions of the training, validation/testing data of these datasets.

\begin{table}[]
\caption{Dataset statistics on different geo-aware image classification datasets. "Train", "Val", and "Test" column indicates the number of data samples in each dataset. "\#Class" column indicates the total number of classes for each dataset.
	}
	\label{tab:data_stat}
	\centering
	\setlength{\tabcolsep}{3pt}
\begin{tabular}{c|c|ccc|c}
\toprule
     Task                            & Dataset   & Train  & Val   & Test & \#Class \\ \hline
\multirow{5}{*}{Species Recog.} & BirdSnap  & 19133  & 443   & 443  & 500     \\
                                     & BirdSnap† & 42490  & 980   & 980  & 500     \\
                                     & NABirds†  & 22599  & 1100  & 1100 & 555     \\
                                     & iNat2017  & 569465 & 93622 & -    & 5089    \\
                                     & iNat2018  & 436063 & 24343 & -    & 8142    \\ \hline
Flickr                                  & YFCC      & 66739  & 4449  & 4449 & 100     \\ \hline
RS                                   & fMoW      & 363570 & 53040 & -    & 62  \\
\bottomrule
\end{tabular}
\end{table}

\paragraph{Fine-Grained Species Recognition}
We use five widely used fine-grained species recognition image datasets in which each data sample is a tuple of an image $\image$, a location $\th$, and its ground truth class $\classy$: 
\begin{enumerate}
    \item \textbf{BirdSnap}: An image dataset about bird species based on BirdSnap dataset \citep{berg2014birdsnap} which consists of 500 bird species that are commonly found in the North America.
The original BirdSnap dataset \citep{berg2014birdsnap} did not provided the location metadata. \cite{mac2019presence} recollected the images and location data based on the original image URLs.
    \item \textbf{BirdSnap†}: An enriched \mai{BirdSnap} dataset constructed by \cite{mac2019presence} by simulating locations, dates, and photographers from the eBrid dataset \citep{sullivan2009ebird}.
    \item \textbf{NABirds†}: Another image dataset about North American bird species constructed by \cite{mac2019presence} based on the NABirds dataset \citep{van2015building} in which the location metadata were also simulated from the eBrid dataset \citep{sullivan2009ebird}. 
    \item \textbf{iNat2017}: The species recognition dataset used in the iNaturalist 2017 challenges\footnote{\url{https://github.com/visipedia/inat_comp/tree/master/2017}} \citep{van2018inaturalist} with 5089 unique categories.
    \item \textbf{iNat2018}: The species recognition dataset used in the iNaturalist 2018 challenges\footnote{\url{https://github.com/visipedia/inat_comp/tree/master/2018}} \citep{van2018inaturalist} with 8142 unique categories.
\end{enumerate}

\paragraph{Flickr Image Classification}
We use the Yahoo Flickr Creative Commons 100M
dataset\footnote{\url{https://yahooresearch.tumblr.com/post/89783581601/one-hundred-million-creative-commons-flickr-images}} (YFCC100M-GEO100 dataset) which is a set of geo-tagged Flickr photos collected by Yahoo! Research. Here, we denote this dataset as YFCC. YFCC has been used in \cite{tang2015improving,mac2019presence} for geo-aware image classification. See Figure \ref{fig:yfcc_train_locs} and \ref{fig:yfcc_test_locs} for the spatial distributions of the training and test dataset of YFCC.

\paragraph{Remote Sensing Image Classification}
We use the Functional Map of the World dataset (denoted as fMoW) \citep{klocek2019functional} as one representative remote sensing (RS) image classification dataset. The fMoW dataset contains about 363K training and 53K validation remote sensing images which are classfied into 62 different land use types. They are 4-band or 8-band multispectral remote sensing images. 4-band images are collected from the QuickBird-2 or GeoEye-1 satellite systems while 8-band images are from WorldView-2 or WorldView-3. We use the fMoW-rgb version of fMoW dataset which are JPEG compressed version of these remote sensing images with only the RGB bands. 
The reason we pick fMoM is that 1) the fMoW dataset contains RS images with diverse land use types collected all over the world (see Figure \ref{fig:fmow_train_locs} and \ref{fig:fmow_test_locs}); 2) it is a large RS image dataset with location metadata available. In contrast, the UC Merced dataset \citep{yang2010ucmerced} consist of RS images collected from only 20 US cities. The EuroSAT dataset \citep{helber2019eurosat} contained RS images collected from 30 European countries. And the location metadata of the RS images from these two datasets are not publicly available. 
\mai{Global coverage} of the RS images is important in our experiment since we focus on studying how the map projection distortion problem \mai{and spherical-to-Euclidean distance approximation error} can be solved by \emph{\modelname} on a global scale geospatial problem.
The reason we use the RGB version is that this dataset version has an existing pretrained image encoder -- MoCo-V2+TP \citep{ayush2020selfsup} available to use. We do not need to train our own remote sensing image encoder.

\subsection{Geo-Aware Image Classification}  \label{sucsec:img_cls_exp}

\begin{table*}[ht!]
\begin{minipage}{\linewidth} 
    \caption{The Top1 classification accuracy of different  geo-aware image classification models
over three tasks: species recognition, 
Flickr image classification (YFCC), and remote sensing (RS) image classification (fMOW \citep{christie2018functional}).
See Section \ref{sec:baselines} for the description of each baseline. 
$\tile$ indicates the results reported by \cite{mac2019presence}.
$\aodha*$ indicates the original results reported by \cite{mac2019presence} while $\aodha$ is the best results we obtained by rerunning their code. 
Since the test sets for iNat2017, iNat2018, and fMoW are not open-sourced, we report results on validation sets.
The best performance of the baseline models and \emph{\modelname} are highlighted as bold.
All compared models use location only while ignoring time.
The original result reported by \cite{ayush2020selfsup} for No Prior on fMOW is 69.05. 
We obtain 69.84 by  retraining their implementation.  "Avg" column indicates the average performance of each model on all five species recognition datasets. 
See Section \ref{sec:hyperpara_tune}
for hyperparameter tuning details. 
}
	\label{tab:imgcls_eval}
	\centering
{
\setlength{\tabcolsep}{2pt}
\begin{tabular}{l|ccccc|c||c|c}
\toprule
Task                      & \multicolumn{6}{c||}{Species Recognition}                                                             & Flickr            & RS             \\ \hline
Dataset                    & BirdSnap       & BirdSnap†      & NABirds†       & iNat2017       & iNat2018       & Avg            & YFCC           & fMOW           \\ \hline
P(y|x) - Prior Type        & Test           & Test           & Test           & Val            & Val            & -              & Test           & Val            \\ \hline
No Prior (i.e. image model)      & 70.07                                 & 70.07                                 & 76.08                                 & 63.27                                 & 60.20                                  & 67.94                                 & 50.15                                 & 69.84                                 \\
$\tile$ \citep{tang2015improving}                           & 70.16                                 & 72.33                                 & 77.34                                 & 66.15                                 & 65.61                                 & 70.32                                 & 50.43                                 & -                                     \\
$\xyz$                            & 71.85          & 78.97          &  81.20           & 69.39          & 71.75          &  74.63          &  50.75          &  70.18          \\
$\aodha*$ \citep{mac2019presence}  & 71.66                                 & 78.65                                 & 81.15                                 & 69.34                                 & 72.41                                 & 74.64                                 & 50.70                                 & -                                     \\
$\aodha$ &  71.87          &  79.06          &  \textbf{81.62} &  69.22          &  72.92          &  74.94          &  50.90          &  70.29          \\
$\aodhaffn$                    &  \textbf{71.99} &  79.21          &  81.36          &  69.40           &  71.95          &  74.78          &  50.76          &  70.28          \\
$\rbf$\citep{mai2020multiscale}                             &  71.78          &  79.40           &  81.32          &  68.52          &  71.35          &  74.47         &  51.09          &  70.65          \\
$\rff$\citep{rahimi2007random}                            &  71.92          &  79.16          &  81.30           &  69.36          &  71.80           &  74.71         &  50.67          &  70.27          \\
\emph{\spacevec}-$\grid$ \citep{mai2020multiscale}                           &  \mai{71.70}           &  \mai{79.72} &  \mai{81.24}          &  \mai{69.46}          &  \mai{73.02}          &  \mai{75.03}          &  \mai{\textbf{51.18}} & \mai{\textbf{70.80}} \\
\emph{\spacevec}-$\theory$ \citep{mai2020multiscale}                         &  \mai{71.88}          &  \mai{\textbf{79.75}}          &  \mai{81.30}          &  \mai{\textbf{69.47}} &  \mai{\textbf{73.03}} &  \mai{\textbf{75.09}} &  \mai{\textbf{51.16}} &  \mai{\textbf{70.81}} \\ 
\mai{$\nerf$ \citep{mildenhall2020nerf}}  &  \mai{71.66}          &  \mai{79.66}           &  \mai{81.32}          &  \mai{69.45}          &  \mai{73.00}          &  \mai{75.02}          &  \mai{50.97}          &  \mai{70.64 }         \\
\hline
\emph{\modelname}-$\sphere$                        &  \mai{72.11}          &  \mai{79.80}           &  \mai{81.88}          &  \mai{69.68}          &  \mai{73.29}          &  \mai{75.35}          &  \mai{\textbf{51.34}}          &  \mai{71.00}          \\
\emph{\modelname}-$\spheregrid$                      & \mai{\textbf{72.41}}                        & \mai{80.11}                                 & \mai{\textbf{81.97}}                                 & \mai{\textbf{69.75}}                        & \mai{73.31}                                 & \mai{75.51}                                 & \mai{51.28}                                 & \mai{71.03}          \\
\emph{\modelname}-$\spheremixscale$                        & \mai{72.06}                                 & \mai{79.84}                                 & \mai{81.94}                                 & \mai{69.72}                        & \mai{73.25}                                 & \mai{75.36}                                 & \mai{\textbf{51.35}}                        & \mai{70.99}          \\
\emph{\modelname}-$\spheregridmixscale$                   &  \mai{72.24}          &  \mai{\textbf{80.57}} &  \mai{81.94}  &  \mai{69.67} &  \mai{\textbf{73.80}} &  \mai{\textbf{75.64}} &  \mai{51.24}          &  \mai{71.10}          \\
\emph{\modelname}-$\dft$                      & 71.75                                 & 79.18                                 & 81.39                                 & 69.65                                 & 73.24                                 & 75.04                                 & 51.15                                 & \textbf{71.46}     
\\
\bottomrule
\end{tabular} }
\end{minipage}
\end{table*} 

To test the effectiveness of \emph{\modelname}, 
we conduct geo-aware image classification experiments on seven large-scale real-world datasets as we described in Section \ref{sec:dataset}. 

Beside the baselines described in Section \ref{sec:baselines}, we also consider $No \; Prior$, which represents 
\mai{an full supervised trained image classifier }
without using any location information, i.e., predicting image labels purely based on image information $P(\classy|\image)$.

Table \ref{tab:imgcls_eval} compares the Top1 classification accuracy of five variants of \emph{\modelname} models against those of nine baseline models as we discussed in Section \ref{sec:baselines}. 

Similar to Equation \ref{equ:topk_loc}, the Topk classification accuracy on geo-aware image classification task is defined as follow
\begin{align}
    TOP_k = \dfrac{1}{|\dataset|}\sum_{i=1}^{|\dataset|} \mathbf{1}(Rank(\th_i, \image_i, \classy_i) \geqslant k)
    \label{equ:topk}
\end{align}
where $\dataset = \{(\th_i, \image_i, \classy_i)\}$ is a set of location $\th_i$, image $\image_i$, and label $\classy_i$ tuples which indicates the whole validation or testing set. $|\dataset|$ denotes the total number of samples in $\dataset$. $Rank(\th_i, \image_i, \classy_i)$ indicates the rank of the ground truth label $\classy_i$ in the ranked listed of all classes based on the probability score $P(\classy_i|\th_i)P(\classy_i|\image_i)$ given by a specific geo-aware image classification model. $\mathbf{1}(*)$ is defined the same as that in Equation \ref{equ:topk_loc}. 

From Table \ref{tab:imgcls_eval}, we can see that the \emph{\modelname} models outperform  baselines on all seven datasets, 
and the variants with linear number of DFS terms ($\sphere$, $\spheregrid$, $\spheremixscale$, and $\spheregridmixscale$) works as well as or even better than $\dft$.
This clearly show the advantages of \emph{\modelname} to handle large-scale geographic datasets. On the five species recognition datasets, $\spheregridmixscale$ achieves the best performance while $\spheremixscale$ and $\dft$ achieve the best performance on YFCC and fMoW correspondingly.
\mai{Similar to our findings in the synthetic dataset experiments, $\grid$ and $\theory$ also outperform or are comaprable to $\nerf$ on all 7 real-world datasets.
}

\subsection{Hyperparameter Analysis} \label{sec:hyperpara_tune}

In order to find the best hyperparameter combinations for each model on each dataset, we use grid search to do hyperparameter tuning including 
supervised training learning rate $\lr = [0.01, 0.005, 0.002, 0.001, 0.0005, 0.00005]$, 
the number of scales $\nscale = [16, 32, 64]$, 
the minimum scaling factor $\minscale = [0.1  0.05  0.02  0.01  0.005 0.001 0.0001]$, 
the number of hidden layers and number of neurons used in $\peffn(\cdot)$ -- $\numresnet = [1,2,3,4]$ and $\numneuron = [256, 512, 1024]$, 
the dropout rate in $\peffn(\cdot)$ -- $dropout = [0.1, 0.2, 0.3, 0.4, 0.5, 0.6, 0.7]$. 
We also test multiple options for the nonlinear function used for $\peffn(\cdot)$ including ReLU, LeakyReLU, and Sigmoid.
The maximum scaling factor $\maxscale$ can be determined based on the range of latitude $\lat$ and longitude $\lon$. 
For $\grid$ and $\theory$, we use $\maxscale = 360$ and for all \emph{\modelname}, we use $\maxscale = 1$. 
As for $\rbf$ and $\rff$, we also tune their hyperparamaters including kernel size $\kernelsize = [0.5, 1, 2, 10]$ as well as the number of kernels $M = [100, 200, 500]$.

Based on hyperparameter tuning, we find out using 0.5 as the dropout rate and ReLU as the nonlinear activation function for $\peffn(\cdot)$ works best for every location encoder. 
Moreover, we find out $\lr$ and $\minscale$ are the most important hyperparameters. 
Table \ref{tab:imgcls_eval_param} shows the best hyperparameter combinations of different \emph{\modelname} models on different geo-aware image classification datasets. We use a smaller $\nscale$ for $\dft$ since it has $O(\nscale^{2})$ terms while the other models have $O(\nscale)$ terms. $\dft$ with $\nscale=8$ yield a similar number of terms to the other models with $\nscale=32$ (see Table \ref{tab:dim}). 
Interestingly, all five \emph{\modelname} models ($\sphere$, $\spheregrid$, $\spheremixscale$, $\spheregridmixscale$, and $\dft$) show the best performance on the first six datasets with the same hyperparamter combinations. On the fMoW dataset, five \emph{\modelname} achieve the best performances with different $\minscale$ but sharing other hyperparameters.
This indicates that the proposed \emph{\modelname} models show similar performance over different hyperparameter combinations.

\begin{table}[!]
\caption{
The best hyperparameter combinations of \emph{\modelname} models on different geo-aware image classification datasets. The best $\nscale$ is 8 for $\dft$ and 32 for all others;
and we fix the maximum scale $\maxscale$ as 1.
Here, $\minscale$ indicates the minimum scale.
$\numresnet$ and $\numneuron$ are the number of hidden layers and the number of neurons in  $\pemlp()$ respectively.
}
\label{tab:imgcls_eval_param}
\centering
\begin{tabular}{l|cccc}
\toprule
Dataset   & Model & $\lr$  & $\minscale$ & $\numneuron$ \\ \hline
{BirdSnap} & All & 0.001  & $10^{-6}$& 512  \\ {BirdSnap$\dagger$} &  All & 0.001  & $10^{-4}$ & 1024 \\ {NABirds$\dagger$ } & All & 0.001  & $10^{-4}$   & 1024  \\ {iNat2017} &  All & 0.0001 & $10^{-2}$  & 1024 \\ {iNat2018} & All & 0.0005  & $10^{-3}$  & 1024 \\  \hline
{YFCC} & All & 0.001  & $5\times10^{-3}$  & 512 \\ \hline
\multirow{5}{*}{fMoW} & \sphere             & \multirow{5}{*}{0.01} & $10^{-3}$  & \multirow{5}{*}{512} \\
                      & {\spheregrid }         &                       & $10^{-4}$ &                      \\
                      & \spheremixscale     &                       & $10^{-3}$  &                      \\
                      & {\spheregridmixscale } &                       & $5\times10^{-4}$ &                      \\
                      & \dft                &                       & $10^{-4}$ &         \\            
\bottomrule
\end{tabular}
\end{table}

\begin{table}[]
\centering
\caption{Dimension of position encoding for different models in terms of total scales $S$}
\label{tab:dim}
\setlength{\tabcolsep}{2pt}
\begin{tabular}{cccccc}
\hline
Model     & $\sphere$ & $\spheregrid$ & $\spheremixscale$ & $\spheregridmixscale$ & $\dft$   \\ \hline
Dim. & $3S$      & $6S$          & $5S$              & $8S$                  & $4S^2+4S$ \\ \hline
\end{tabular}
\label{modeldim}
\end{table}

We also find out that using a deeper MLP as $\peffn(\cdot)$, i.e., a larger $\numresnet$ does not necessarily lead to better classification accuracy. In many cases, one hidden layer -- $\numresnet = 1$ achieves the best performance for many kinds of location encoders. We discuss this in detail in Section \ref{sec:perform_mlp_depth}.

Based on the hyperparameter tuning, the best hyperparameter combinations are selected for different models on different datasets. The best results are reported in Table \ref{tab:imgcls_eval}. Note that each model has been running for 5 times and its mean Top1 score is reported. Due to the limit of space, the standard deviation of each model's performance on each dataset is not included in Table \ref{tab:imgcls_eval}. However, we report the standard deviations of all models' performance on three datasets in Section \ref{sec:perform_mlp_depth}. 
\subsection{Model Performance Sensitivity Analysis}
\label{sec:eval_mlp}

\subsubsection{Model Performance Distribution Comparison} \label{sec:perform_hist_compare}
\begin{figure*}[ht!]
	\centering \small \begin{subfigure}[b]{0.49\textwidth}  
		\centering 
        \includegraphics[width=\textwidth]{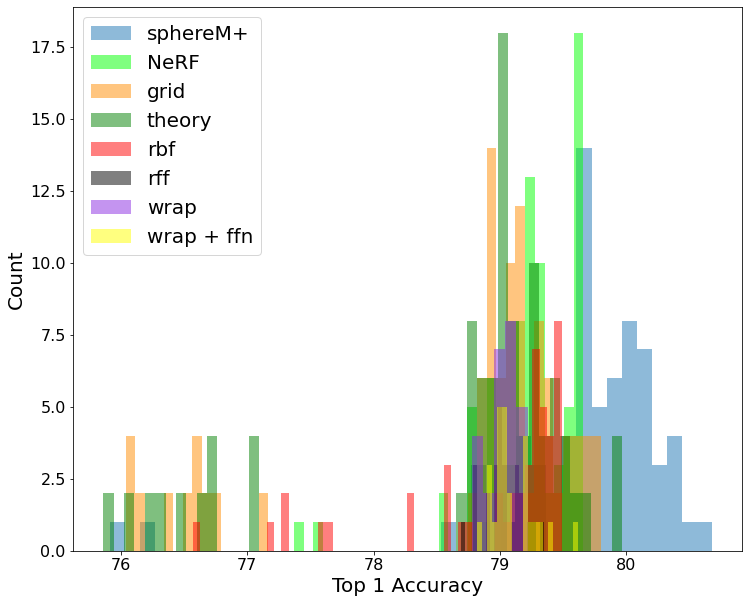}
		\vspacecomp
		\caption[]{{\small 
		 \mai{BirdSnap† dataset}
		}}    
		\label{fig:bridsnap_ebird_hist}
	\end{subfigure}
	\hfill
	\begin{subfigure}[b]{0.49\textwidth}  
		\centering 
        \includegraphics[width=\textwidth]{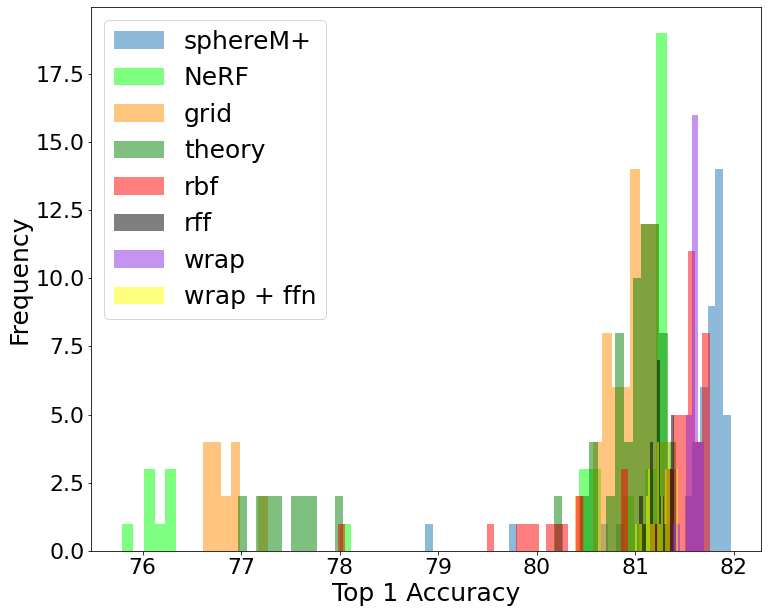}
		\vspacecomp
		\caption[]{{\small 
		 \mai{NABirds† dataset}
		}}    
		\label{fig:nabirds_hist}
	\end{subfigure}

	\hfill
	\begin{subfigure}[b]{0.49\textwidth}  
		\centering 
		\includegraphics[width=\textwidth]{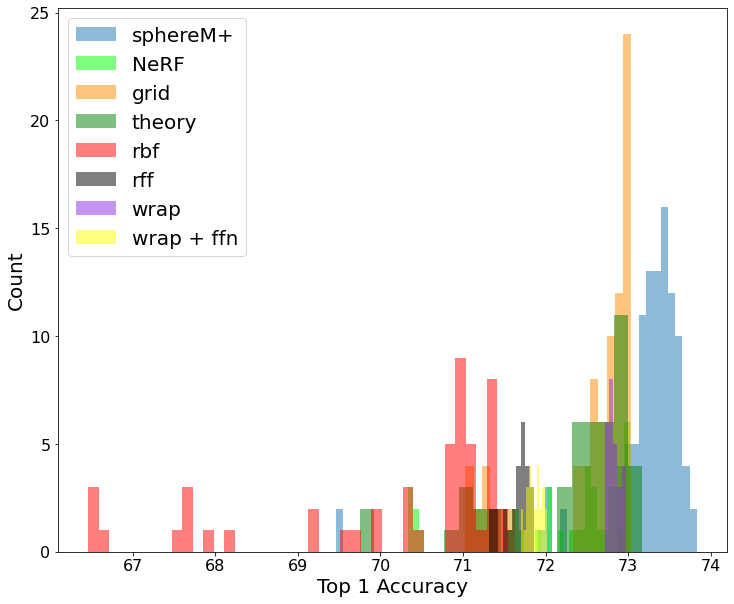}
		\vspacecomp
		\caption[]{{\small 
		\mai{iNat2018 dataset}
		}}    
		\label{fig:inat2018_hist}
	\end{subfigure}
	\hfill
	\begin{subfigure}[b]{0.49\textwidth}  
		\centering 
		\includegraphics[width=\textwidth]{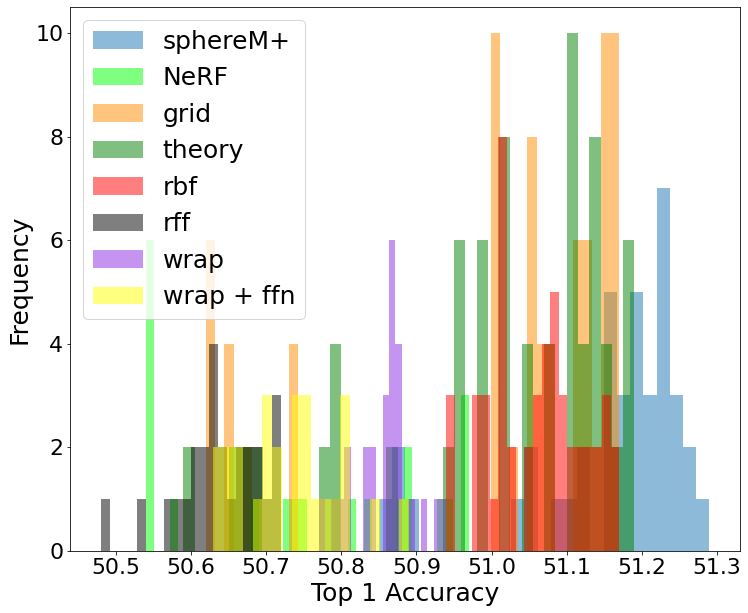}
		\vspacecomp
		\caption[]{{\small 
		\mai{YFCC dataset}
		}}    
		\label{fig:yfcc_hist}
	\end{subfigure}
	\caption{
 \mai{The model performance (Top1 accuracy) distributions/histograms of different models under different hyperparameter combinations on (a) the BirdSnap† dataset, (b) the NABirds† dataset, (c) the iNat2018 dataset, and (d) the YFCC dataset.
	X axis: the Top1 accuracy scores of the respective model; 
	Y axis: the frequency of different hyperparameter combinations of the same model falling in the same Top1 Accuracy bin.
	In all four plots, each color indicate a Top1 accuracy histogram of one specific model on a specific dataset. This histogram shows the model's sensitivity towards different hyperparameter combinations. We can see that in all four plots, the histogram of $\spheregridmixscale$ (the blue histogram) are clearly different from all baseline models' histograms. This shows the clear advantage of $\spheregridmixscale$ over all baselines.
}
	} 
	\label{fig:perform_dist}
	\vspace*{-0.15cm}
\end{figure*}

To have a better understanding of the performance difference between \emph{\modelname} and all baseline models, we visualize the distributions/histograms of Top1 accuracy scores of different models on the BirdSnap†, NABirds†, iNat2018, and YFCC dataset under different hyperparameter combinations.
More specifically, after the hyperparameter tuning process described in Section \ref{sec:hyperpara_tune}, for each location encoder and each dataset we get a collection of trained models with different hyperparameter combinations. They correspond to a distribution/histograms of Top1 accuracy scores for this model on the respective dataset. Figure \ref{fig:perform_dist} compares the histogram of $\spheregridmixscale$ and all baseline models on four datasets. We can see that the histogram of $\spheregridmixscale$ is clearly above those of all baselines. This further demonstrates the superiority of \emph{\modelname} over all baselines.

\subsubsection{Performance Sensitivity to the Depth of MLP} \label{sec:perform_mlp_depth}
To further understand how the performances of different location encoders vary according to the depth of the multi-layer perceptron $\peffn()$, we conduct a performance sensitivity analysis. 
Table \ref{tab:imgcls_eval_mlp} is a complementary of Table \ref{tab:imgcls_eval} which compares the performance of $\spheregridmixscale$ with all baseline models on the geo-aware image classification task. The results on three datasets are shown here including BirdSnap†, NABirds†, and iNat2018. For each model, we vary the depth of its $\peffn()$, i.e., $\numresnet = [1,2,3,4]$. The best evaluation results with each $\numresnet$ are reported. Moreover, we run each model with one specific $\numresnet$ 5 times and report the standard deviation of the Top1 accuray, indicated in ``()''. Several observations can be made based on Table \ref{tab:imgcls_eval_mlp}:
\begin{enumerate}
    \item Although the absolute performance improvement between $\spheregridmixscale$ and the best baseline model is not very large -- \mai{0.91\%, 0.62\%, and 0.77\%} for three datasets respectively, \textbf{these performance improvements are statistically significant given the standard deviations of these Top1 scores.} 
    \item \textbf{These performance improvements are comparable to those from the previous studies on the same tasks.} In other words, the small margin is due to the nature of these datasets. For example, \citet{mai2020multiscale} showed that $\grid$ or $\theory$ has 0.79\%, 0.44\% absolute Top1 accuracy improvement on BirdSnap† and NABirds† dataset respectively. \citet{mac2019presence} showed that $\aodha$ has 0.09\%, 0\%, 0.04\% absolute Top1 accuracy improvement on BirdSnap, BirdSnap† and NABirds† dataset. Here, we only consider the results of $\aodha$ that only uses location information, but not temporal information. Although \citet{mac2019presence} showed that compared with $\tile$ and nearest neighbor methods, $\aodha$ has 3.19\% and 3.71\% performance improvement on iNat2017 and iNat2018 datatset, these large margins are mainly because the baselines they used are rather weak. When we consider the typical $\rbf$ and $\rff$ \citep{rahimi2007random} used in our study, their performance \mai{improvements} are down to -0.02\% and 0.61\%.  
    \item By comparing the performances of the same model with different \mai{depths} of its $\peffn()$, i.e., $\numresnet$, we can see that the model performance is not sensitive to $\numresnet$. In fact, in most cases, one layer $\peffn()$ achieves the best result. This indicates that \textbf{the depth of the MLP does not significantly affect the model performance and a deeper MLP does not necessarily lead to a better performance}. 
    In other words, \textbf{
the systematic bias (i.e., distance distortion) introduced by \mai{$\grid$, $\theory$, and $\nerf$} can not later be compensated by a deep MLP.
} It shows the importance of designing a spherical-distance-aware location encoder.
\end{enumerate}

\begin{table}[ht!]
\caption{
The impact of the depth $\numresnet$ of multi-layer perceptrons $\peffn()$ on Top1 accuracy for various models. The numbers in ``()'' indicates the standard deviations 
estimated from 5 independent train/test runs.
We find that 
the model performances are not very sensitive to  $\peffn()$, and,
in most cases, one layer $\peffn()$ achieve the best result.
In other words, 
the larger performance gaps in fact \mai{come} from different $PE_{\nscale}(\cdot)$ we use.
Moreover, given the \mai{performance} variance of each model, we can see that $\spheregridmixscale$ outperforms other baseline models on all these three datasets and the margins are statistically significant. The same conclusion can be drawn based on our experiments on other datasets. Here, we only show results on three datasets as an illustrative example. }
	\label{tab:imgcls_eval_mlp}
	\centering
\setlength{\tabcolsep}{3pt}
\begin{tabular}{l|c|ccc}
\toprule
Dataset                             &    & BirdSnap†    & NABirds†     & iNat2018     \\ \hline
&   $\numresnet$                                                          & Test         & Test         & Val          \\ \hline
\multirow{4}{*}{$\xyz$}                & 1                                                              & 78.81 (0.10) & 81.08 (0.05) & 71.60 (0.08)  \\
                                    & 2                                                              & 78.83 (0.10) & 81.20 (0.09)  & 71.70 (0.02)  \\
                                    & 3                                                              & 78.97 (0.06) & 81.11 (0.06) & 71.75 (0.04) \\
                                    & 4                                                              & 78.84 (0.09) & 81.02 (0.03) & 71.71 (0.03) \\ \hline
\multirow{4}{*}{$\aodha$}           & 1                                                              & 79.04 (0.13) & 81.60 (0.04) & 72.89 (0.08) \\
                                    & 2                                                              & 78.94 (0.13) & \textbf{81.62 (0.04)} & 72.84 (0.07) \\
                                    & 3                                                              & 79.08 (0.15) & 81.53 (0.02) & 72.92 (0.05) \\
                                    & 4                                                              & 79.06 (0.11) & 81.51 (0.09) & 72.77 (0.06) \\ \hline
\multirow{4}{*}{$\aodhaffn$}      & 1                                                              & 78.97 (0.09) & 81.23 (0.06) & 71.90 (0.05) \\
                                    & 2                                                              & 79.02 (0.15) & 81.36 (0.04) & 71.95 (0.05) \\
                                    & 3                                                              & 79.21 (0.14) & 81.35 (0.05) & 71.94 (0.04) \\
                                    & 4                                                              & 79.06 (0.09) & 81.27 (0.13) & 71.93 (0.04) \\ \hline
\multirow{4}{*}{$\rbf$}                & 1                                                              & 79.40 (0.13) & 81.32 (0.08) & 71.02 (0.18) \\
                                    & 2                                                              & 79.38 (0.12) & 81.22 (0.11) & 71.29 (0.20) \\
                                    & 3                                                              & 79.40 (0.04) & 81.31 (0.07) & 71.35 (0.21) \\
                                    & 4                                                              & 79.25 (0.05) & 81.30 (0.07) & 71.21 (0.19) \\  \hline
\multirow{4}{*}{$\rff$ }                & 1                                                              & 78.96 (0.18) & 81.27 (0.07) & 71.76 (0.06) \\
                                    & 2                                                              & 78.97 (0.04) & 81.28 (0.05) & 71.71 (0.09) \\
                                    & 3                                                              & 79.07 (0.12) & 81.30 (0.11) & 71.80 (0.04) \\
                                    & 4                                                              & 79.16 (0.13) & 81.22 (0.11) & 71.46 (0.05) \\  \hline
\multirow{4}{*}{\mai{$\grid$} }           & \mai{1}                                                              & \mai{79.72 (0.07)} & \mai{81.24 (0.06)} & \mai{73.02 (0.02)} \\
                                    & \mai{2}                                                              & \mai{79.05 (0.06)} & \mai{81.09 (0.07)} & \mai{72.87 (0.05)}  \\
                                    & \mai{3}                                                              & \mai{79.23 (0.12)} & \mai{80.95 (0.14)} & \mai{72.69 (0.05)} \\
                                    & \mai{4}                                                              & \mai{78.97 (0.10)} & \mai{80.71 (0.10)} & \mai{72.51 (0.07)} \\ \hline
\multirow{4}{*}{\mai{$\theory$} }             & \mai{1}                                                              & \mai{79.75 (0.17)} & \mai{81.23 (0.02)} & \mai{\textbf{73.03 (0.09)}} \\
                                    & \mai{2}                                                              & \mai{79.08 (0.20)} & \mai{81.30 (0.11)} & \mai{72.70 (0.02)} \\
                                    & \mai{3}                                                              & \mai{78.94 (0.19)} & \mai{81.00 (0.09)} & \mai{72.49 (0.08)} \\
                                    & \mai{4}                                                              & \mai{79.07 (0.14)} & \mai{80.64 (0.14)} & \mai{72.35 (0.07)} \\ \hline
\multirow{4}{*}{\mai{$\nerf$} }             & \mai{1}                                                              & \mai{\textbf{79.66 (0.00)}} & \mai{81.27 (0.00)} & \mai{73.00 (0.01)} \\
                                    & \mai{2}                                                              & \mai{79.65 (0.02)} & \mai{81.29 (0.00)} & \mai{72.97 (0.03)} \\
                                    & \mai{3}                                                              & \mai{79.40 (0.05)} & \mai{\textbf{81.32 (0.01)}} & \mai{72.88 (0.02)} \\
                                    & \mai{4}                                                              & \mai{79.24 (0.04)} & \mai{81.23 (0.00)} & \mai{72.80 (0.02)} \\
                                    \noalign{\hrule height 2pt}
\multirow{4}{*}{\mai{$\spheregridmixscale$}} & \mai{1}                                                              & \mai{\textbf{80.57 (0.08)}} & \mai{81.87 (0.02)} & \mai{\textbf{73.80 (0.05)}} \\
                                    & \mai{2}                                                              & \mai{79.82 (0.14)} & \mai{81.83 (0.04)} & \mai{73.42 (0.06)} \\
                                    & \mai{3}                                                              & \mai{80.03 (0.08)} & \mai{\textbf{81.94 (0.04)}} & \mai{73.40 (0.05)} \\
                                    & \mai{4}                                                              & \mai{79.90 (0.15)} & \mai{81.84 (0.09)} & \mai{73.20 (0.04)} \\
                                    \bottomrule
\end{tabular}
\end{table}

\subsubsection{\mai{Ablation Studies on Approaches for Image and Location Fusion}} \label{sec:ablation_img_loc_comb}

\begin{table}
\caption{
\mai{Ablation Studies on different ways to combine image and location information on the iNat2018 dataset.
``Fusion'' column indicates different methods to fuse image and location information.
``$\imgenc(\cdot)$'' and ``$\enc(\cdot)$'' indicates the type of image encoder and location encoder used for each model. 
``$\imgenc(\cdot)$ Train'' denotes different ways to train the image encoder. ``Frozen'' means we use an InceptionV3 network pre-trained on ImageNet as an image feature extractor and freeze its learnable parameters while only finetuning the last softmax layer.
``Finetune'' means we finetune the whole image encoder $\imgenc(\cdot)$.
}
}
	\label{tab:ablation_img_loc_comb}
	\centering
\setlength{\tabcolsep}{1pt}
\begin{tabular}{>{\color{black}}l|>{\color{black}}l|>{\color{black}}l|>{\color{black}}l}
\toprule
Model           & Concat (Frozen) & Concat (Finetune) & Post Fusion    \\ \hline
$\imgenc(\cdot)$       & InceptionV3             & InceptionV3               & InceptionV3       \\ \hline
$\imgenc(\cdot)$ Train & Frozen               & Finetune               & Finetune       \\ \hline
$\enc(\cdot)$       & $\spheregridmixscale$             & $\spheregridmixscale$              & $\spheregridmixscale$       \\ \hline
Top1            & 48.74                & 73.35                  & \textbf{73.72} \\ \bottomrule
\end{tabular}
\end{table}

\mai{
In Section \ref{subsec:img_cls_decoder}, we discuss how we fusion the predictions from the image encoder and location encoder together for the final model prediction. However, there are other ways to fuse the image and location information. In this section, we conduct ablation studies on different image and location fusion approaches:
\begin{itemize}
    \item \textbf{Post Fusion} is the method we adopt from \citet{mac2019presence} which is illustrated in Figure \ref{fig:pos_enc}. The image encoder $\imgenc(\cdot)$ and location encoder $\enc(\cdot)$ are trained separately and their final predictions are combined.
    \item \textbf{Concat (Img. Finetune)} indicates a method in which the image embedding $\imgenc(\image)$ and the location embedding $\enc(\th)$ are concatenated together and fed into a softmax layer for the final prediction. The whole architecture is trained end-to-end.
    \item \textbf{Concat (Img. Frozen)} indicates the same model architecture as Concat (Img. Finetune). The only difference is that $\imgenc(\cdot)$ is initialized by a pretrained weight and its learnable parameters are frozen during the image and location join training.
\end{itemize}
}

\mai{
We conduct experiments on iNat2018 dataset and the results are shown in Table \ref{tab:ablation_img_loc_comb}. We can see that:
\begin{itemize}
    \item \textit{Post Fusion}, the method we adopt in our study, achieves the best Top1 score and outperforms both \textit{Concat} approaches. This result is aligned with the results of \citet{chu2019geo}.
    \item \textit{Concat (Img. Frozen)} shows a significantly lower performance than \textit{Concat (Img. Finetune)}. This is understandable and consistent with the existing literature \citep{ayush2020selfsup} since the linear probing method, \textit{Concat (Img. Frozen)}, usually underperforms a fully fine-tuning method, \textit{Concat (Img. Finetune)}. 
    \item Although \textit{Post Fusion} only shows a small margin over \textit{Concat (Img. Finetune)}, the training process of \textit{Post Fusion} is much easier since we can separate the training process of the image encoder $\imgenc(\cdot)$ and location encoder $\enc(\cdot)$. In contrast, \textit{Concat (Img. Finetune)} has to train a large network which is hard to do hyperparameter tuning.
\end{itemize}
}

\begin{figure*}
	\centering \small \begin{subfigure}[b]{0.59\textwidth}  
		\centering 
\includegraphics[width=\textwidth]{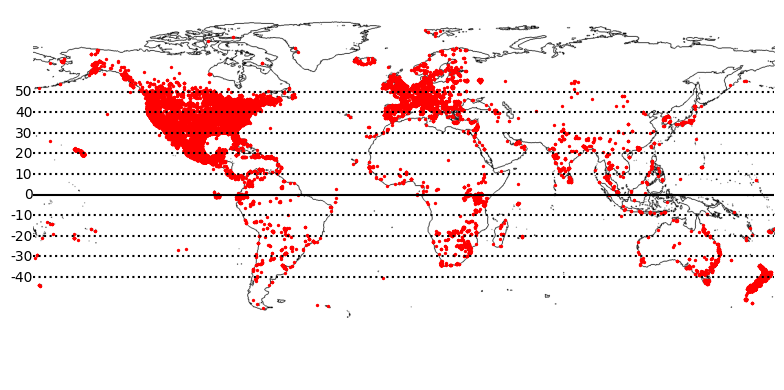}
		\vspacecomp
		\caption[]{{\small 
		Validation Locations
		}}    
		\label{fig:inat17_locs}
	\end{subfigure}
	\hfill
	\begin{subfigure}[b]{0.383\textwidth}  
		\centering 
		\includegraphics[width=\textwidth]{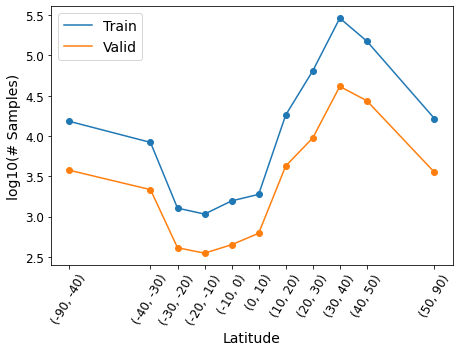}
		\vspacecomp
		\caption[]{{\small 
Samples per $\lat$ band
		}}    
		\label{fig:inat17_num_sample}
	\end{subfigure}
	\hfill
	\begin{subfigure}[b]{0.59\textwidth}  
		\centering \includegraphics[width=\textwidth]{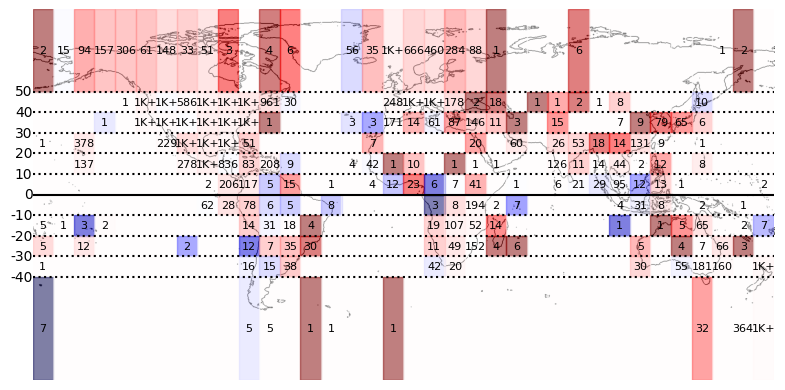}
		\vspacecomp
		\caption[]{{\small 
		$\Delta MRR$ per cell
		}}    
		\label{fig:inat17_cell_mrr}
	\end{subfigure}
	\hfill
	\begin{subfigure}[b]{0.4\textwidth}  
		\centering 
		\includegraphics[width=\textwidth]{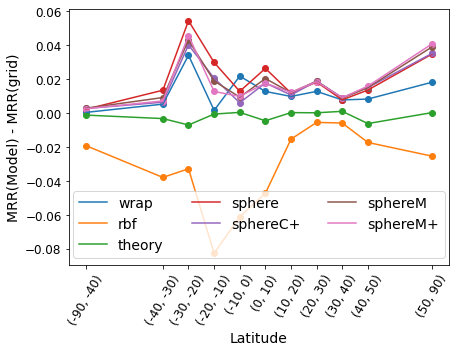}
		\vspacecomp
		\caption[]{{\small 
		$\Delta MRR$ per $\lat$ band
		}}    
		\label{fig:inat17_mrr}
	\end{subfigure}

\caption{The data distribution of the iNat2017 dataset and model performance comparison on it:
(a) Sample locations for validation set of the iNat2017 dataset where the dashed and solid lines indicates latitudes; 
	(b) The number of training and validation samples in different latitude intervals.
(c) $\Delta MRR = MRR(\spheregrid) - MRR(\grid)$ 
for each latitude-longitude cell. Red and blue color indicates positive and negative $\Delta MRR$ while darker color means high absolute value. The number on each cell indicates the number of validation data points  while "1K+" means there are more than 1K points in a cell.
	(d) $\Delta MRR$ between a model and baseline $\grid$ on the validation dataset in different latitude bands.
} 
	\label{fig:inat17_analysis}
	\vspace*{-0.15cm}
\end{figure*}

\begin{figure*}[ht!]
	\centering \small \begin{subfigure}[b]{0.4\textwidth}  
		\centering 
		\includegraphics[width=\textwidth]{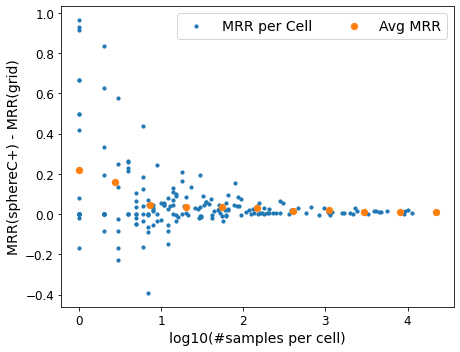}
		\vspacecomp
		\caption[]{{\small 
$\Delta MRR$ per cell
		}}    
		\label{fig:inat17_cell_scatter}
	\end{subfigure}
	\hfill
	\begin{subfigure}[b]{0.4\textwidth}  
		\centering 
		\includegraphics[width=\textwidth]{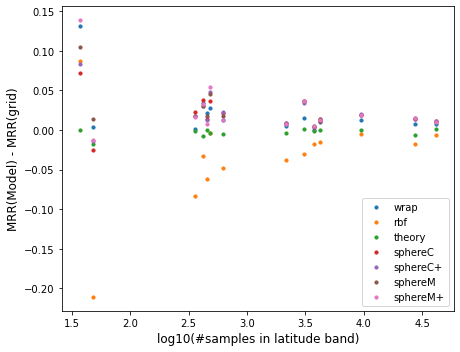}
		\vspacecomp
		\caption[]{{\small 
$\Delta MRR$ per $\lat$ band
		}}    
		\label{fig:inat17_band_scatter}
	\end{subfigure}
	\caption{The number of sample v.s. the model performance improvements on the iNat2017 dataset:
	(a) The number of validation samples v.s. $\Delta MRR = MRR($\spheregrid) - MRR($\grid)$ per latitude-longitude cell defined in Figure \ref{fig:inat17_cell_mrr}. The orange dots represent moving averages.
	(b) The number of validation samples v.s. $\Delta MRR$ per latitude band defined in Figure \ref{fig:inat17_mrr}.
} 
	\label{fig:inat17_analysis_2}
	\vspace*{-0.15cm}
\end{figure*}

\begin{figure*}
	\centering \small \begin{subfigure}[b]{0.59\textwidth}  
		\centering 
\includegraphics[width=\textwidth]{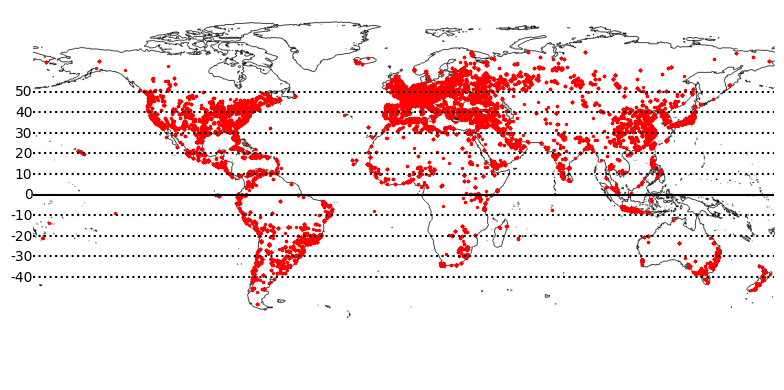}
		\vspacecomp
		\caption[]{{\small 
		Validation Locations
		}}    
		\label{fig:fmow_locs}
	\end{subfigure}
	\hfill
	\begin{subfigure}[b]{0.383\textwidth}  
		\centering 
		\includegraphics[width=\textwidth]{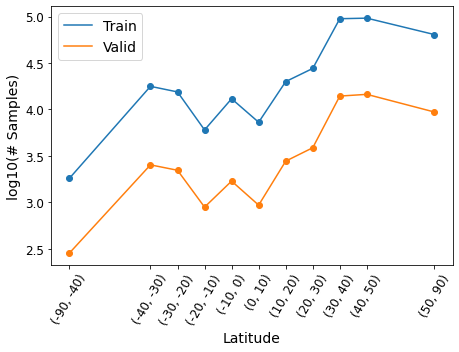}
		\vspacecomp
		\caption[]{{\small 
Samples per $\lat$ band
		}}    
		\label{fig:fmow_num_sample}
	\end{subfigure}
	\hfill
	\begin{subfigure}[b]{0.59\textwidth}  
		\centering \includegraphics[width=\textwidth]{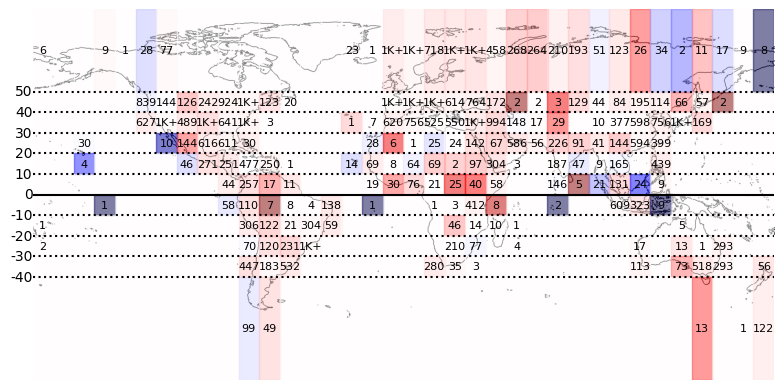}
		\vspacecomp
		\caption[]{{\small 
		$\Delta MRR$ per cell
		}}    
		\label{fig:fmow_cell_mrr}
	\end{subfigure}
	\hfill
	\begin{subfigure}[b]{0.4\textwidth}  
		\centering 
		\includegraphics[width=\textwidth]{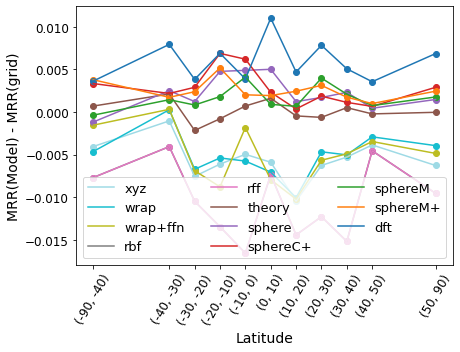}
		\vspacecomp
		\caption[]{{\small 
		$\Delta MRR$ per $\lat$ band
		}}    
		\label{fig:fmow_mrr}
	\end{subfigure}

\caption{The data distribution of the fMoW dataset and model performance comparison on it:
(a) Sample locations for validation set of the fMoW dataset; 
(b) The number of training and validation samples in different latitude intervals.
(c) $\Delta MRR = MRR(\dft) - MRR(\grid)$ 
for each latitude-longitude cell. Red and blue color indicates positive and negative $\Delta MRR$ while darker color means high absolute value. The number on each cell indicates the number of validation data points  while "1K+" means there are more than 1K points in a cell.
	(d) $\Delta MRR$ between a model and baseline $\grid$ on the validation dataset in different latitude bands.
} 
	\label{fig:fmow_analysis}
	\vspace*{-0.15cm}
\end{figure*} \subsection{Understand the Superiority of \modelname} \label{sec:superiority_analysis}

Based on the theoretical analysis of \emph{\modelname} in Section \ref{sec:proof}, we make two hypotheses to explain the superiority of \emph{\modelname} over 2D Euclidean location encoders such as $\theory$, $\grid$:

\begin{itemize}
    \item [A:] Our spherical-distance-kept \emph{\modelname} have a significant advantage over 2D location encoders in the polar area where we expect a large map projection distortion.
    \item [B:] \emph{\modelname} outperforms 2D location encoders in areas with sparse sample points because it is difficult for $\grid$ and $\theory$ to learn spherical distances in these areas with less samples but \emph{\modelname} can handle it due to its theoretical guarantee for spherical distance preservation.
\end{itemize}

To validate these two hypotheses, 
we use iNat2017 and fMoW 
to conduct multiple \mai{empirical analyses}. Table \ref{tab:imgcls_eval} uses Top1 classification accuracy as the evaluation metric to be aligned with several previous \mai{works} \citep{mac2019presence,mai2020multiscale,ayush2020selfsup}. However, Top1 only considers the samples whose ground truth labels are \mai{top-ranked while} ignoring all the other samples' ranks. In contrast, mean reciprocal rank (MRR) considers the ranks of all samples. Equation \ref{equ:mrr} shows the definition of MRR:
\begin{align}
    MRR = \dfrac{1}{|\dataset|}\sum_{i=1}^{|\dataset|} \dfrac{1}{Rank(\th_i, \image_i, \classy_i)}.
    \label{equ:mrr}
\end{align}
where $|\dataset|$ and $Rank(\th_i, \image_i, \classy_i)$ have the same definition as those in Equation \ref{equ:topk}. A higher MRR indicates better model performance. \mai{Because of} the advantage of MRR, we use MRR as the evaluation metric to compare different models.

\subsubsection{Analysis on the iNat2017 Dataset}
Figure \ref{fig:inat17_analysis} and \ref{fig:inat17_analysis_2} show the analysis results on the iNat2017 dataset. Figure \ref{fig:inat17_locs} shows the image locations in the iNat2017 validation dataset. We split this dataset into different latitude bands as indicated by the black lines in Figure \ref{fig:inat17_locs}. The \mai{numbers} of samples in each latitude \mai{band} for the training and validation dataset of iNat2017 are visualized in Figure \ref{fig:inat17_num_sample}. We can see that more samples are available in the North hemisphere, especially when $\lat > 10^{\circ}$.

We compare the MRR scores of different models in different geographic regions to see how the differences \mai{in} MRR change across space.
We compute MRR difference between $\spheregrid$ to $\grid$, i.e., $\Delta MRR = MRR(\spheregrid) - MRR(\grid)$, in different latitude-longitude cell and visualize them in Figure \ref{fig:inat17_cell_mrr}. Here, the color of cells is proportional to $\Delta MRR$. Red and blue color indicates positive and negative $\Delta MRR$ and white color \mai{indicates} nearly zero MRR.
Darker color corresponds to a high absolute $\Delta MRR$ value. Numbers in cells indicate the total number of validation samples in this cell. We can see that $\spheregrid$ outperforms $\grid$ in almost all cells near the North Pole since all these cells are in red color. This observation confirms our Hypothesis A. However, we also see two blue cells \mai{at} the South Pole. But given the fact that these cells only \mai{contain} 5 and 7 samples, we assume these two blue cells attributed to the stochasticity \mai{involved} during the neural network training.

To further validate Hypothesis A, we compute MRR scores of different models in different latitude bands. The $\Delta MRR$ between each model to $\grid$ in different latitude bands are visualized in Figure \ref{fig:inat17_mrr}. We can clearly see that 4 \emph{\modelname}~ models have larger $\Delta MRR$ near the North Pole which validates Hypothesis A. Moreover, \emph{\modelname}~ has advantages on bands with less data samples, e.g. $\lat \in [-30^{\circ}, -20^{\circ})$. This observation also confirms Hypothesis B.

To further understand the relation between the model performance and the number of data samples in different geographic regions, we contrast the number of samples with $\Delta MRR$.
Figure \ref{fig:inat17_cell_scatter} contrasts the number of samples per cell with the $\Delta MRR = MRR(\spheregrid) - MRR(\grid)$ per cell (denoted as blues dots). 
We classify latitude-longitude cells into different groups based on the number of samples and an average MRR is computed for each group (denoted as the yellow dots). We can see $\spheregrid$ has more advantages over $\grid$ on cells with \mai{fewer data samples}. This shows the robustness of $\spheregrid$ on data sparse area. 
Similarly, Figure \ref{fig:inat17_band_scatter} contrasts the number of samples in each latitude band with $\Delta MRR$ between different models and $\grid$ per band. We can see that 4 \emph{\modelname} show advantages over $\grid$ in bands with \mai{fewer samples}. $\rbf$ is particularly bad in data sparse bands which is a typical \mai{drawback} for kernel-based methods. The observations from Figure \ref{fig:inat17_cell_scatter} and \ref{fig:inat17_band_scatter} confirm our Hypothesis B.

\subsubsection{Analysis on the fMoW Dataset}
\mai{Following} the same practice of Figure \ref{fig:inat17_analysis}, Figure \ref{fig:fmow_analysis} shows similar analysis results on the fMoW dataset. Figure \ref{fig:fmow_locs} visualizes the \mai{sample} locations in the fMoW validation dataset and Figure \ref{fig:fmow_num_sample} shows the numbers of training and validation samples in each latitude band. Similar to the iNat2017 dataset, we can see that for the fMoW dataset more samples are available in the North hemisphere, especially when $\lat > 20^{\circ}$.

Similar to Figure \ref{fig:inat17_cell_mrr}, Figure \ref{fig:fmow_cell_mrr} shows the $\Delta MRR = MRR(\dft) - MRR(\grid)$ for each latitude-longitude cell. Red and blue color indicates positive and negative $\Delta MRR$. Similar observations can be seen from Figure \ref{fig:inat17_cell_mrr}. $\dft$ has advantages over $\grid$ in most cells near the North pole and South Pole. $\grid$ only wins in a few pole cells with small numbers of samples. This observation confirms our Hypothesis A.

Similar to Figure \ref{fig:inat17_mrr}, Figure \ref{fig:fmow_mrr} visualizes the $\Delta MRR$ between each model to $\grid$ in different latitude bands on the fMoW dataset. We can see that all \emph{\modelname} models can outperform $\grid$ on all latitude bands.  $\dft$ has a clear advantage over all the other models on all bands. Moreover, all \emph{\modelname} models have clear advantages over $\grid$ near the North pole and South pole which further confirms our Hypothesis A. In latitude band $\lat \in [0^{\circ}, 10^{\circ})$ where we have \mai{fewer} training samples (see Figure \ref{fig:fmow_num_sample}), $\dft$ has clear advantages over other models which confirms our Hypothesis B.

\subsection{Visualize Estimated Spatial Distributions} \label{sec:locenc_viz_distri}

\begin{figure*}[!htbp]
	\centering \tiny
	\vspace*{-0.2cm}
	\begin{subfigure}[b]{0.10\textwidth}  
		\centering 
		\includegraphics[width=\textwidth]{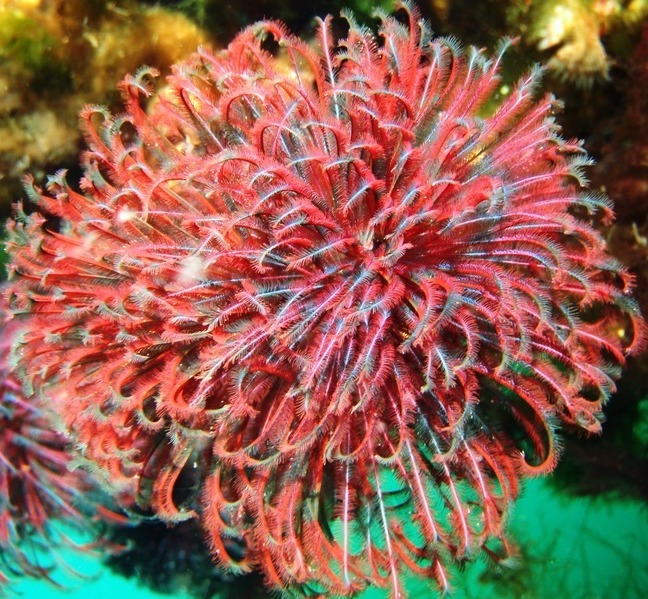}\vspace*{-0.2cm}
		\caption[]{{\small 
		Image
		}}    
		\label{fig:Eudistylia_vancouveri_img}
	\end{subfigure}
	\hfill
	\begin{subfigure}[b]{0.22\textwidth}  
		\centering 
		\includegraphics[width=\textwidth]{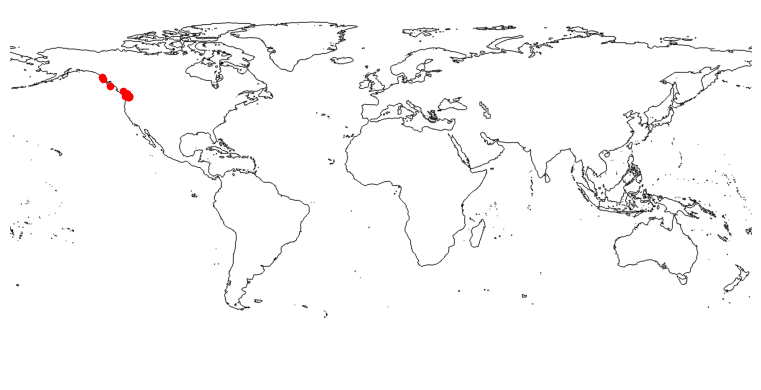}\vspacepred
		\caption[]{{\small 
		Feather duster worm
		}}    
		\label{fig:0002_dist}
	\end{subfigure}
	\hfill
	\begin{subfigure}[b]{0.22\textwidth}  
		\centering 
		\includegraphics[width=\textwidth]{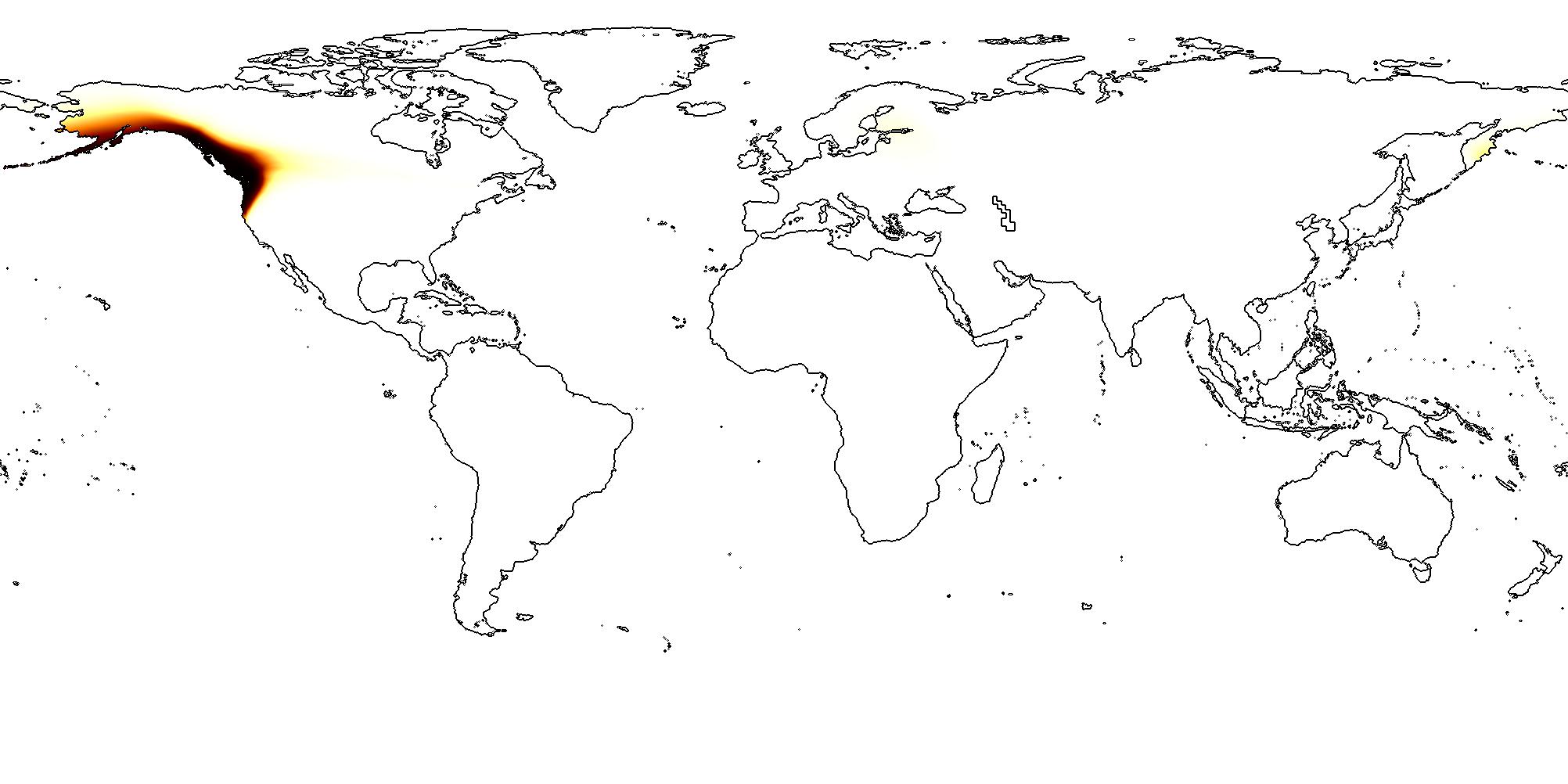}\vspacepred
		\caption[]{{\small 
		$\aodha$
		}}    
		\label{fig:0002_aodha}
	\end{subfigure}
	\hfill
	\begin{subfigure}[b]{0.22\textwidth}  
		\centering 
		\includegraphics[width=\textwidth]{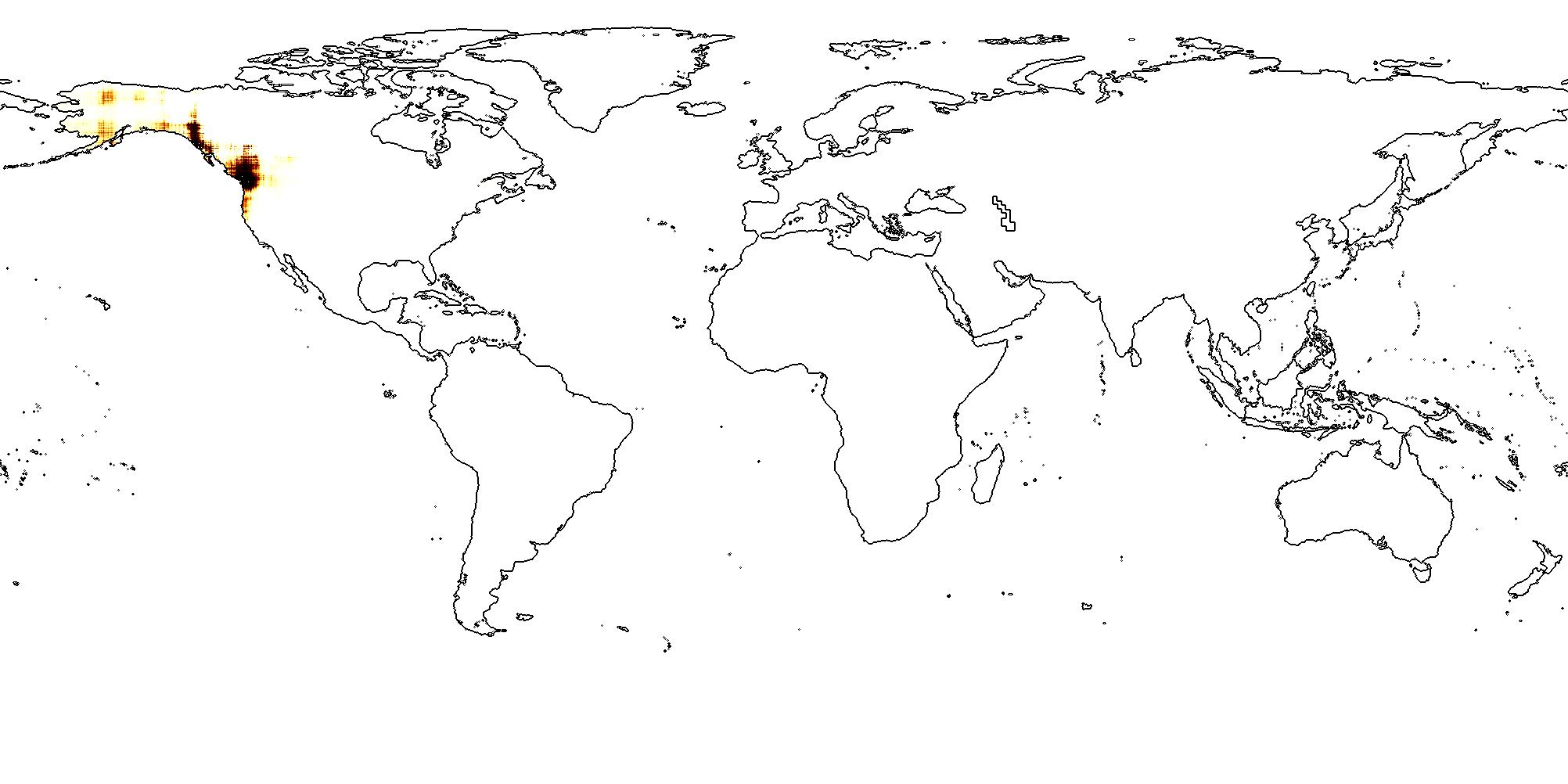}\vspacepred
		\caption[]{{\small 
		$\grid$
		}}    
		\label{fig:0002_grid}
	\end{subfigure}
	\hfill
	\begin{subfigure}[b]{0.22\textwidth}  
		\centering 
		\includegraphics[width=\textwidth]{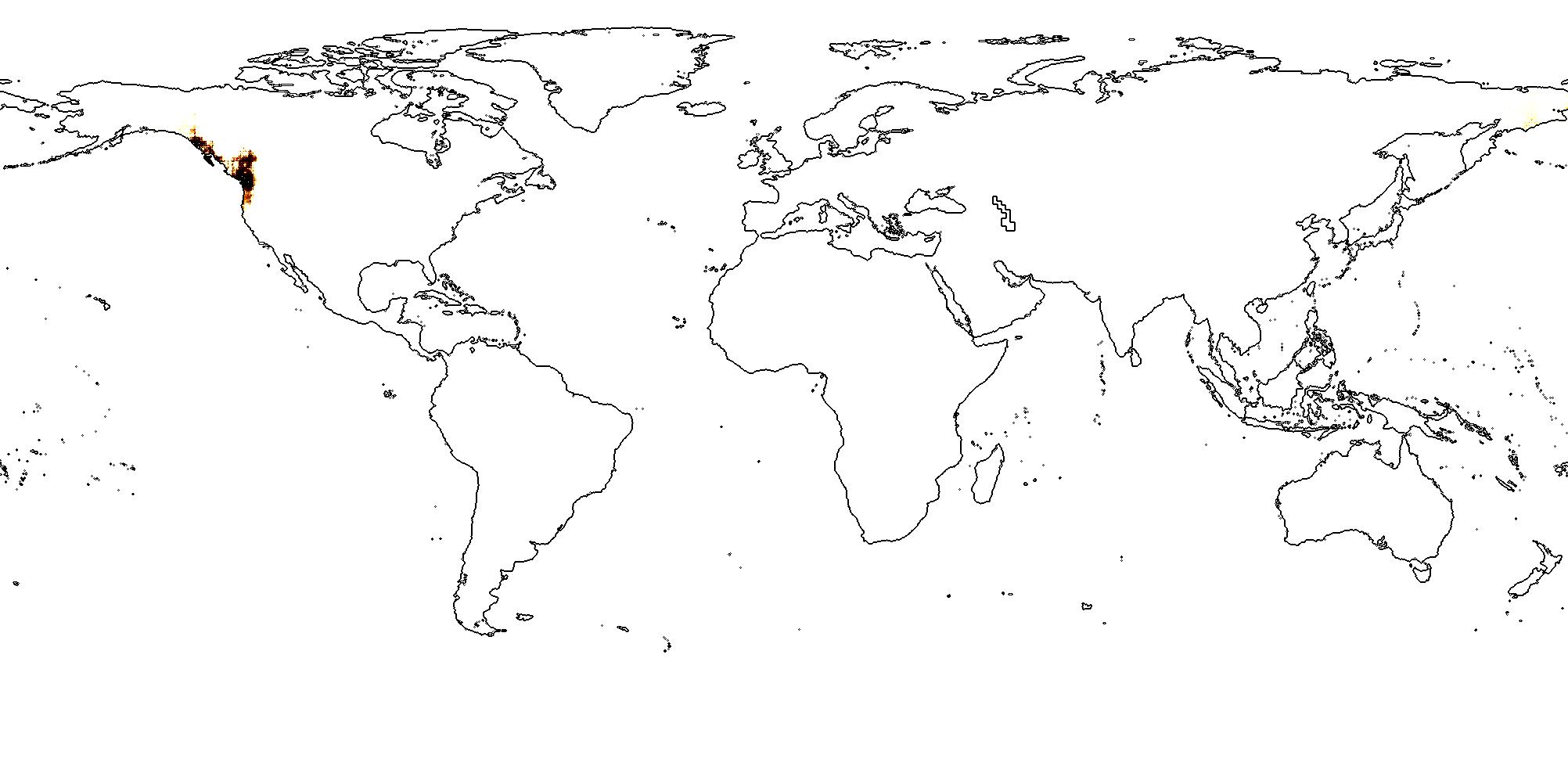}\vspacepred
		\caption[]{{\small 
		$\spheregrid$
		}}    
		\label{fig:0002_spheregrid}
	\end{subfigure}
	
	\hfill
	\begin{subfigure}[b]{0.10\textwidth}  
		\centering 
		\includegraphics[width=\textwidth]{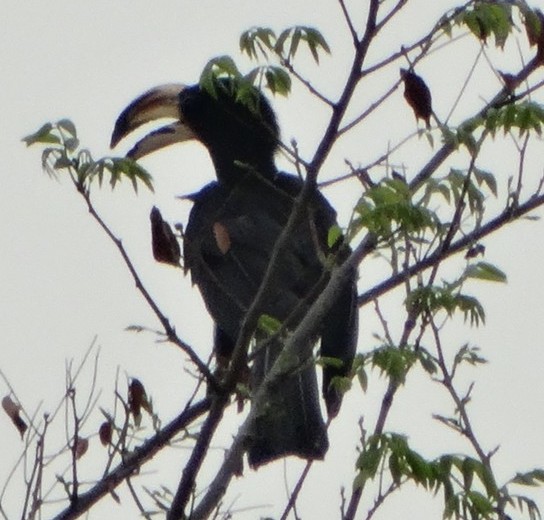}\vspace*{-0.2cm}
		\caption[]{{\small 
		Image
		}}    
		\label{fig:Lophoceros_fasciatus_img}
	\end{subfigure}
	\hfill
	\begin{subfigure}[b]{0.22\textwidth}  
		\centering 
		\includegraphics[width=\textwidth]{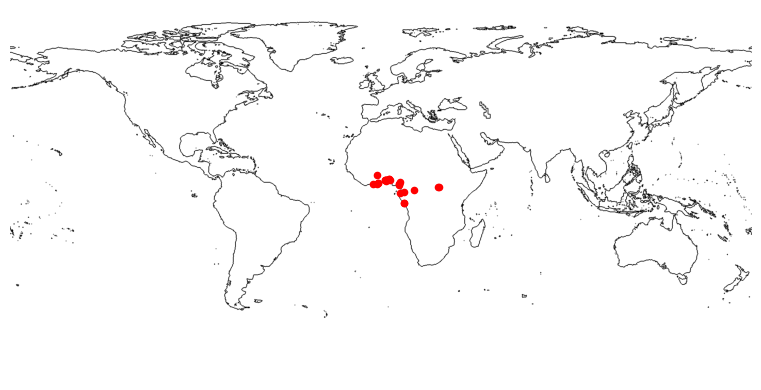}\vspacepred
		\caption[]{{\small 
African pied hornbill
		}}    
		\label{fig:2891_dist}
	\end{subfigure}
	\hfill
	\begin{subfigure}[b]{0.22\textwidth}  
		\centering 
		\includegraphics[width=\textwidth]{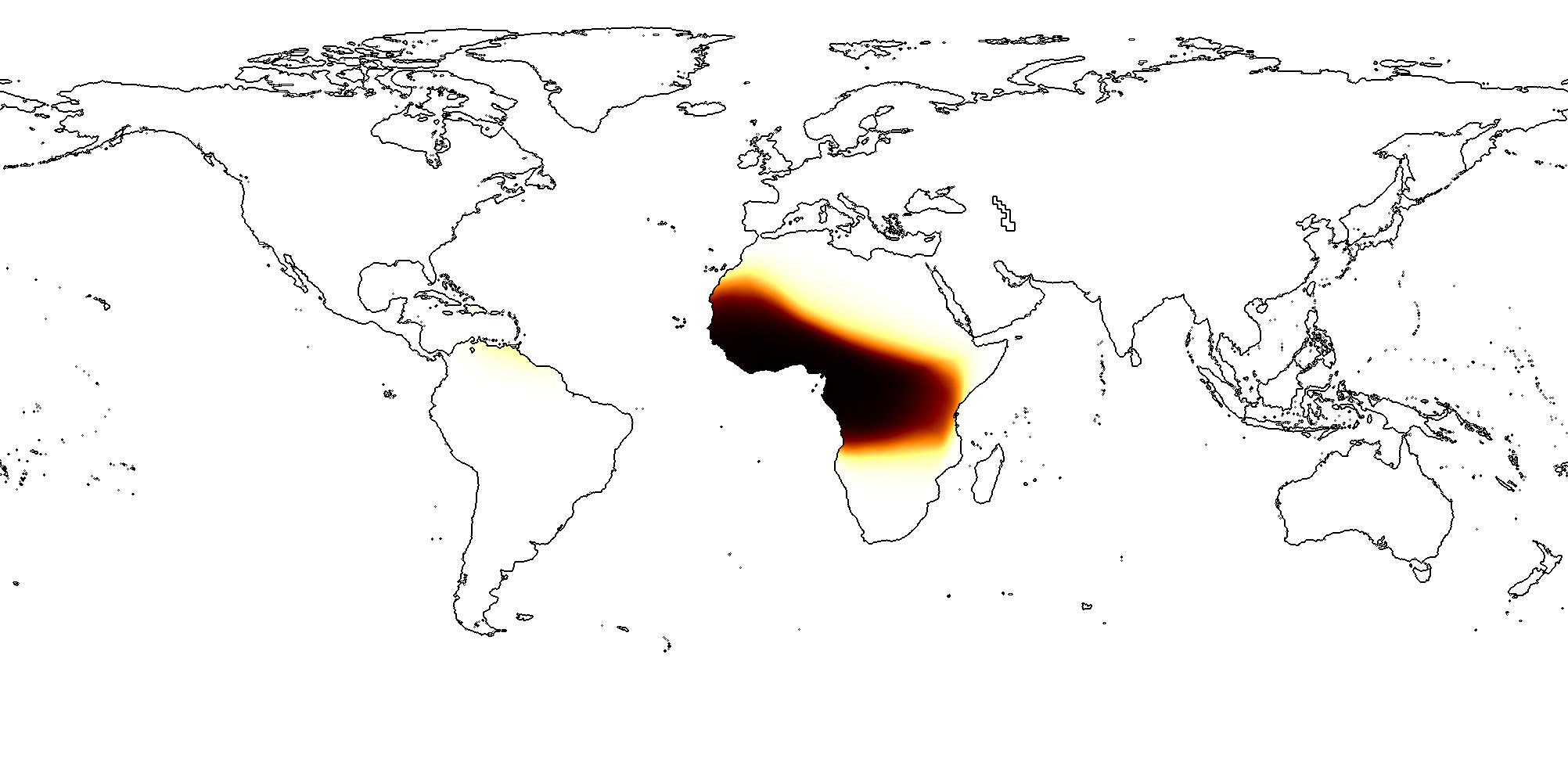}\vspacepred
		\caption[]{{\small 
		$\aodha$
		}}    
		\label{fig:2891_aodha}
	\end{subfigure}
	\hfill
	\begin{subfigure}[b]{0.22\textwidth}  
		\centering 
		\includegraphics[width=\textwidth]{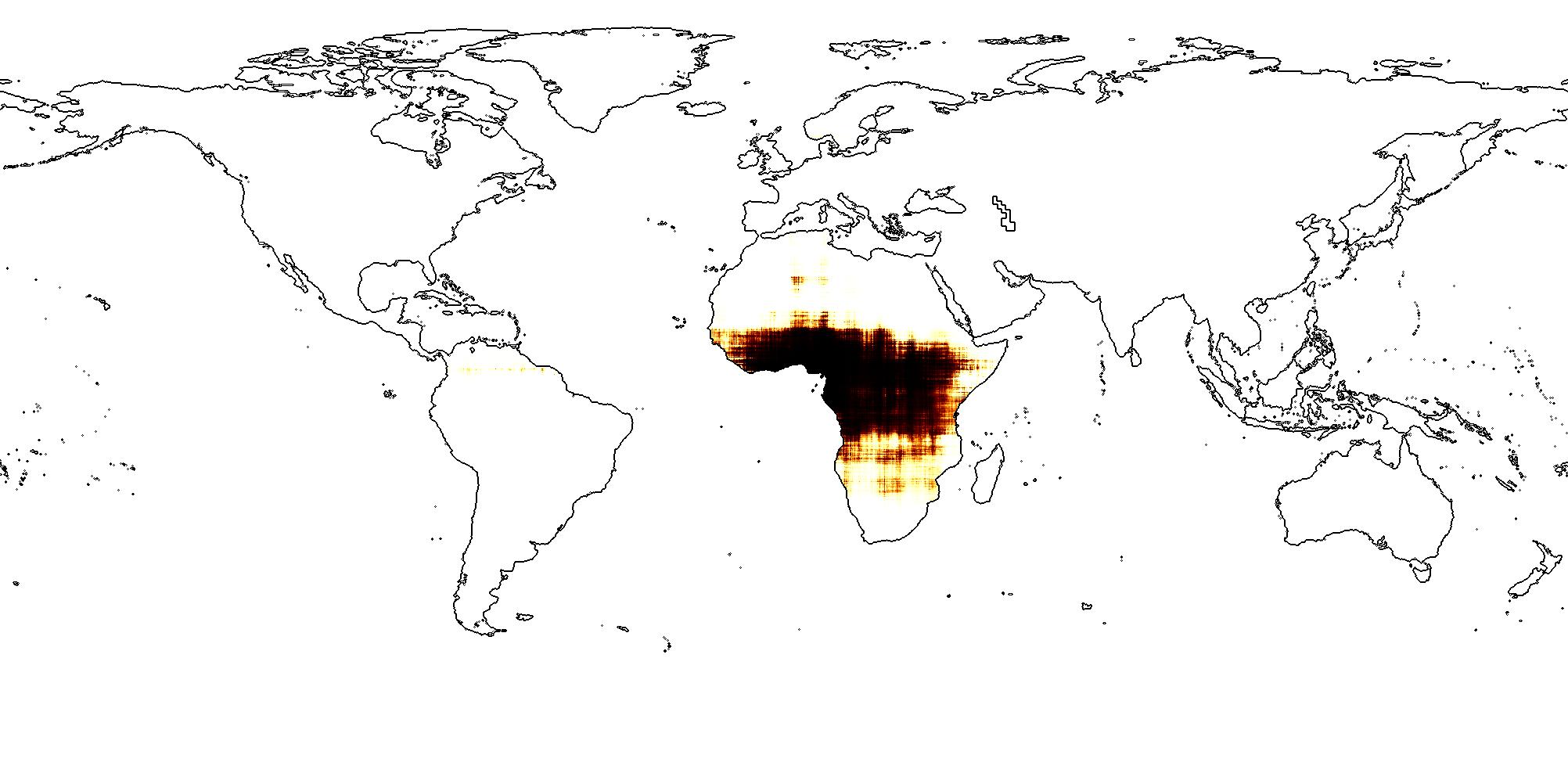}\vspacepred
		\caption[]{{\small 
		$\grid$
		}}    
		\label{fig:2891_grid}
	\end{subfigure}
	\hfill
	\begin{subfigure}[b]{0.22\textwidth}  
		\centering 
		\includegraphics[width=\textwidth]{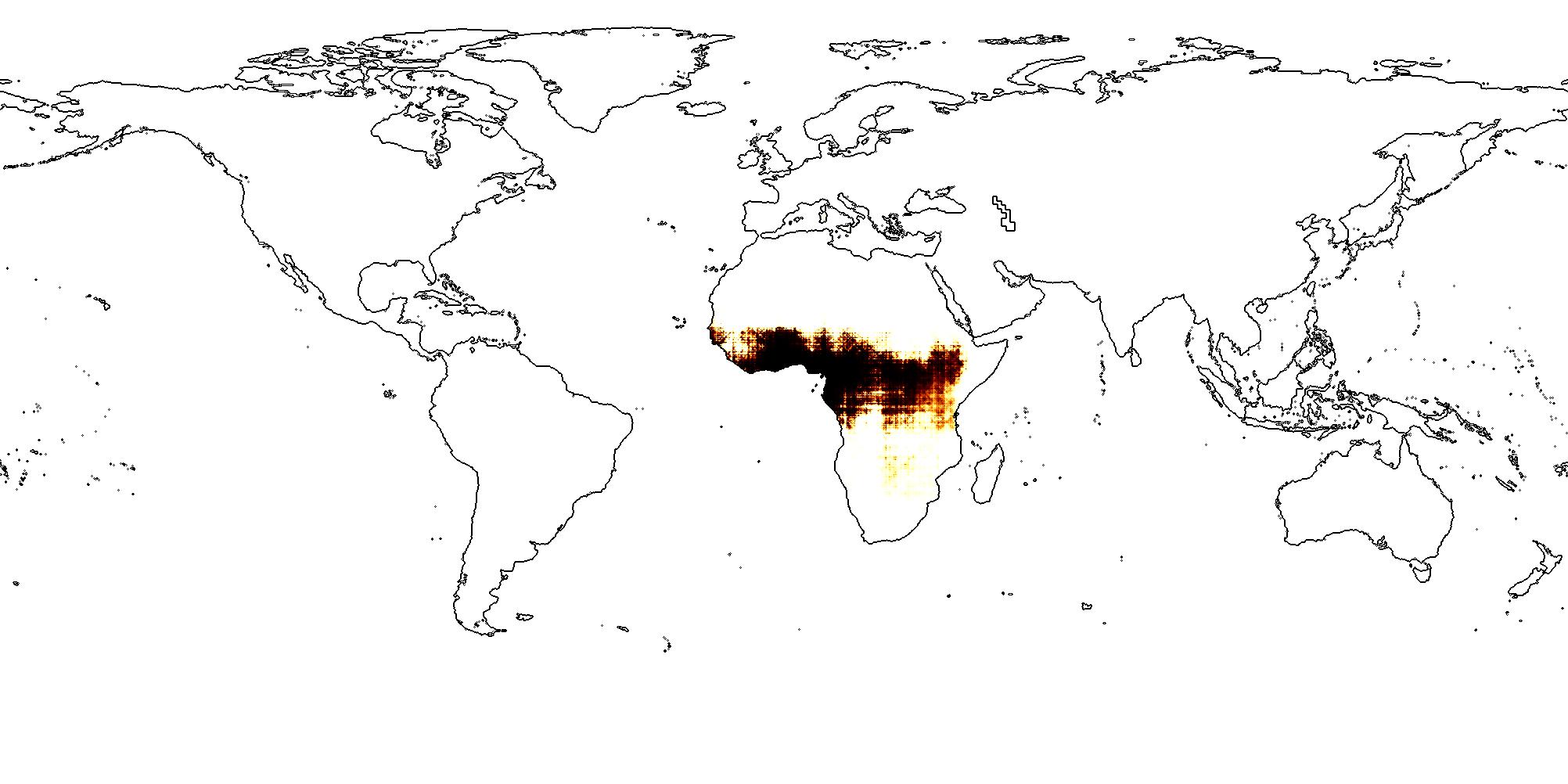}\vspacepred
		\caption[]{{\small 
		$\spheregrid$
		}}    
		\label{fig:2891_spheregrid}
	\end{subfigure}

	\hfill
	\begin{subfigure}[b]{0.10\textwidth}  
		\centering 
		\includegraphics[width=\textwidth]{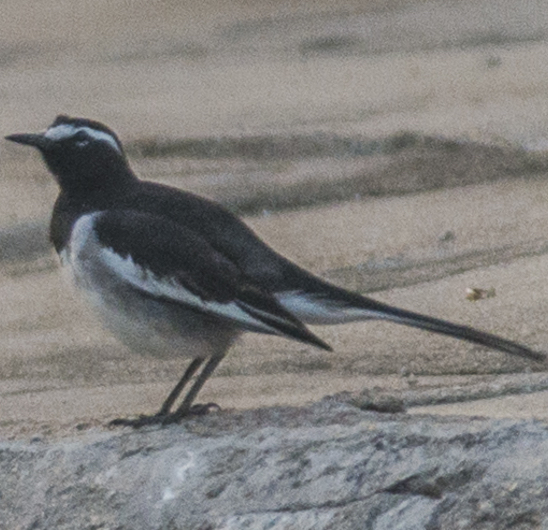}\vspace*{-0.2cm}
		\caption[]{{\small 
		Image
		}}    
		\label{fig:Motacilla_maderaspatensis_img}
	\end{subfigure}
	\hfill
	\begin{subfigure}[b]{0.22\textwidth}  
		\centering 
		\includegraphics[width=\textwidth]{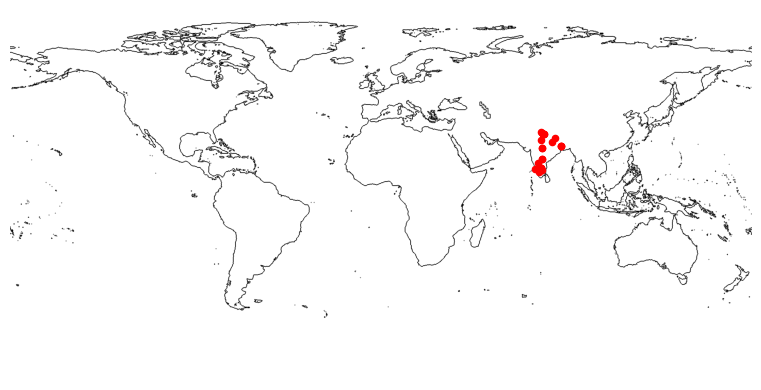}\vspacepred
		\caption[]{{\small 
White-browed wagtail
		}}    
		\label{fig:3475_dist}
	\end{subfigure}
	\hfill
	\begin{subfigure}[b]{0.22\textwidth}  
		\centering 
		\includegraphics[width=\textwidth]{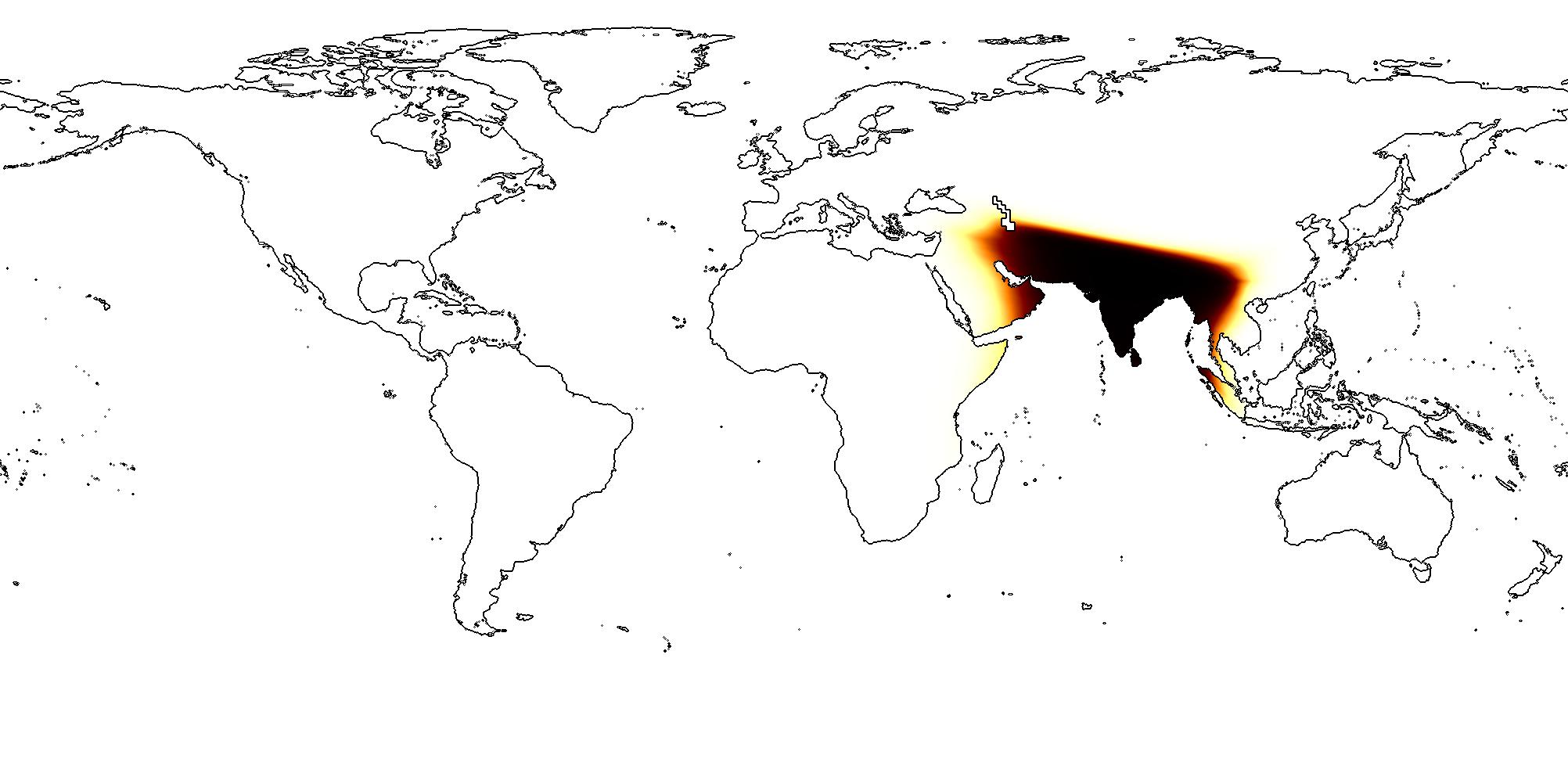}\vspacepred
		\caption[]{{\small 
		$\aodha$
		}}    
		\label{fig:3475_aodha}
	\end{subfigure}
	\hfill
	\begin{subfigure}[b]{0.22\textwidth}  
		\centering 
		\includegraphics[width=\textwidth]{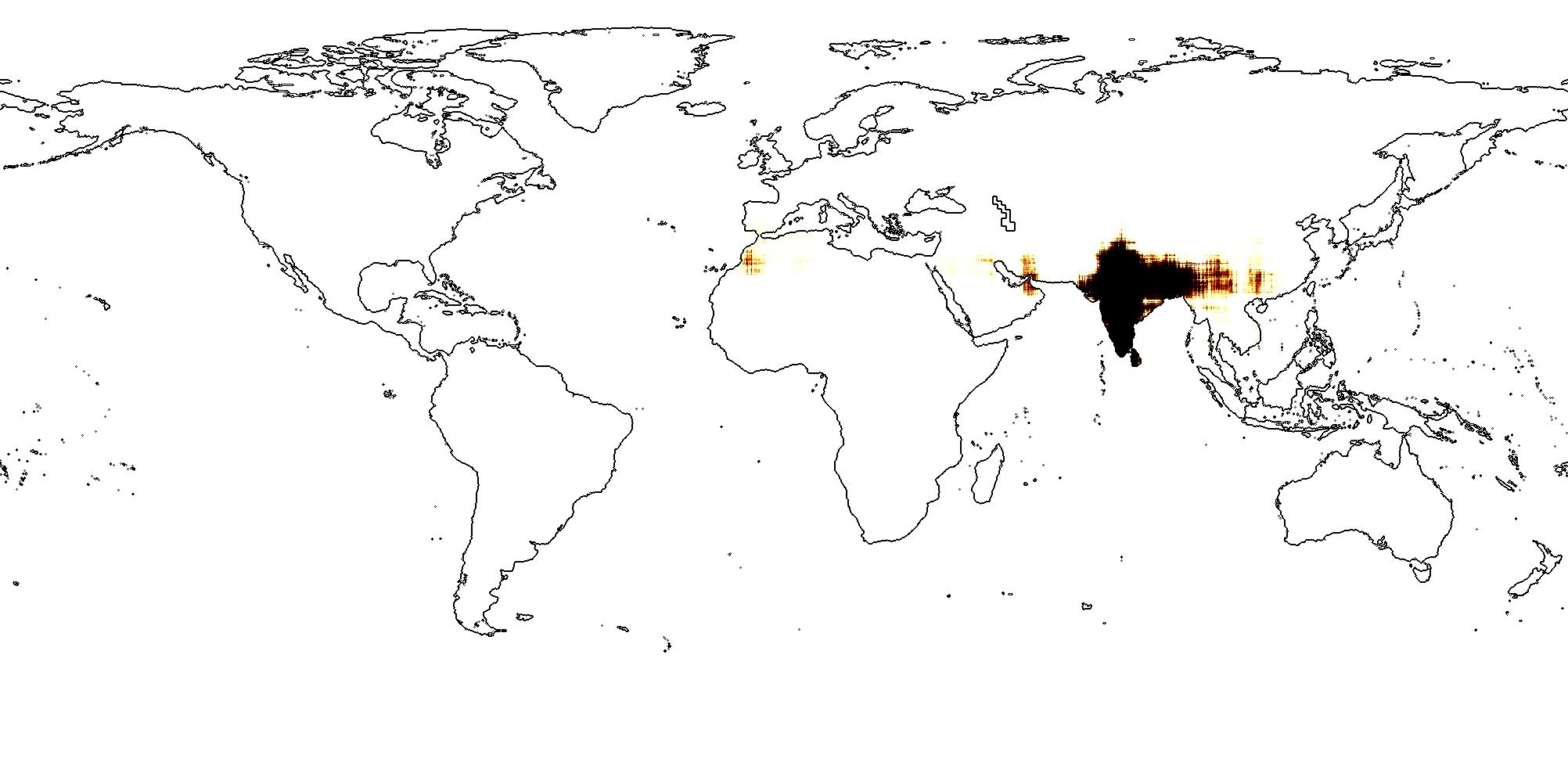}\vspacepred
		\caption[]{{\small 
		$\grid$
		}}    
		\label{fig:3475_grid}
	\end{subfigure}
	\hfill
	\begin{subfigure}[b]{0.22\textwidth}  
		\centering 
		\includegraphics[width=\textwidth]{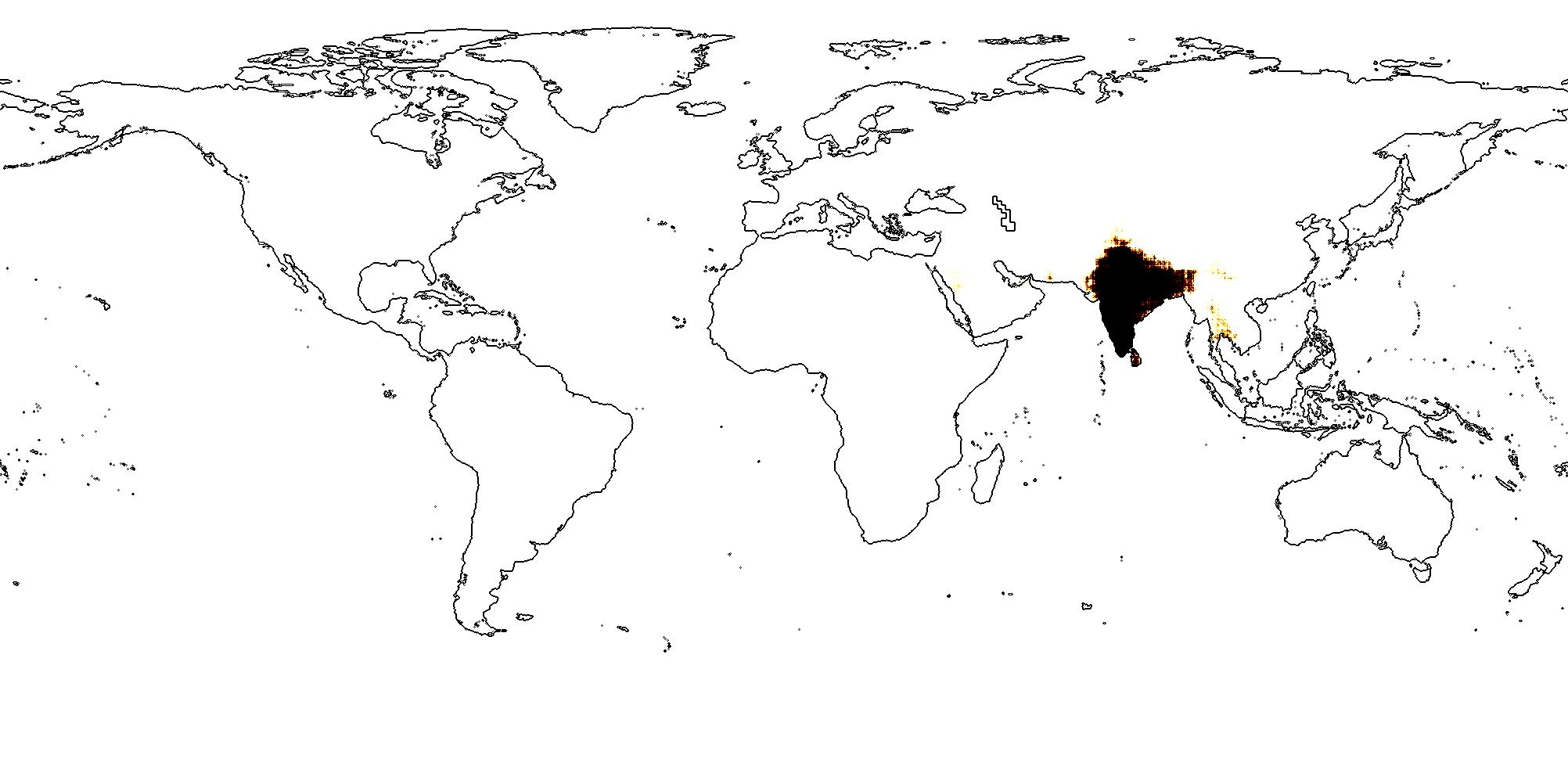}\vspacepred
		\caption[]{{\small 
		$\spheregrid$
		}}    
		\label{fig:3475_spheregrid}
	\end{subfigure}

    \begin{subfigure}[b]{0.10\textwidth}  
		\centering 
		\includegraphics[width=\textwidth]{./Vulpes_lagopus-21026-final.png}\vspace*{-0.2cm}
		\caption[]{{\small 
		Image
		}}    
		\label{fig:fox1_full}
	\end{subfigure}
	\hfill
	\begin{subfigure}[b]{0.22\textwidth}  
		\centering 
		\includegraphics[width=\textwidth]{./4084_cnt_25.png}\vspacepred
		\caption[]{{\small 
		Arctic Fox }}    
		\label{fig:4084_dist}
	\end{subfigure}
	\hfill
	\begin{subfigure}[b]{0.22\textwidth}  
		\centering 
		\includegraphics[width=\textwidth]{./gt_4084_Vulpes-lagopus_geo_net_predict.jpg}\vspacepred
		\caption[]{{\small 
		$\aodha$
		}}    
		\label{fig:4084_aodha}
	\end{subfigure}
	\hfill
	\begin{subfigure}[b]{0.22\textwidth}  
		\centering 
		\includegraphics[width=\textwidth]{./gt_4084_Vulpes-lagopus_gridcell_predict.jpg}\vspacepred
		\caption[]{{\small 
		$\grid$
		}}    
		\label{fig:4084_grid}
	\end{subfigure}
	\hfill
	\begin{subfigure}[b]{0.22\textwidth}  
		\centering 
		\includegraphics[width=\textwidth]{./gt_4084_Vulpes-lagopus_spheregrid_predict.jpg}\vspacepred
		\caption[]{{\small 
		$\spheregrid$
		}}    
		\label{fig:4084_spheregrid}
	\end{subfigure}
	
	\begin{subfigure}[b]{0.10\textwidth}  
		\centering 
		\includegraphics[width=\textwidth]{./Otocyon_megalotis-12189-final.png}\vspace*{-0.2cm}
		\caption[]{{\small 
		Image
		}}    
		\label{fig:fox2}
	\end{subfigure}
	\hfill
	\begin{subfigure}[b]{0.22\textwidth}  
		\centering 
		\includegraphics[width=\textwidth]{./4081_cnt_22.png}\vspacepred
		\caption[]{{\small 
		Bat-Eared Fox }}    
		\label{fig:4081_dist}
	\end{subfigure}
	\hfill
	\begin{subfigure}[b]{0.22\textwidth}  
		\centering 
		\includegraphics[width=\textwidth]{./gt_inat_2018_4081_Otocyon-megalotis_geo_net_predict.jpeg}\vspacepred
		\caption[]{{\small 
		$\aodha$
		}}    
		\label{fig:4081_aodha}
	\end{subfigure}
	\hfill
	\begin{subfigure}[b]{0.22\textwidth}  
		\centering 
		\includegraphics[width=\textwidth]{./gt_inat_2018_4081_Otocyon-megalotis_gridcell_predict.jpeg}\vspacepred
		\caption[]{{\small 
		$\grid$
		}}    
		\label{fig:4081_grid}
	\end{subfigure}
	\hfill
	\begin{subfigure}[b]{0.22\textwidth}  
		\centering 
		\includegraphics[width=\textwidth]{./gt_inat_2018_4081_Otocyon-megalotis_spheregridmixscale_predict.jpeg}\vspacepred
		\caption[]{{\small 
		$\spheregrid$
		}}    
		\label{fig:4081_spheregrid}
	\end{subfigure}
	
	\hfill
	\begin{subfigure}[b]{0.10\textwidth}  
		\centering 
		\includegraphics[width=\textwidth]{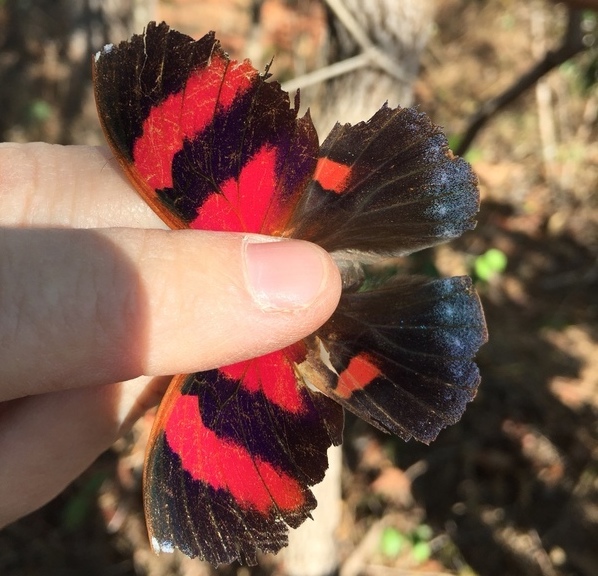}\vspace*{-0.2cm}
		\caption[]{{\small 
		Image
		}}    
		\label{fig:Siderone_galanthis_img}
	\end{subfigure}
	\hfill
	\begin{subfigure}[b]{0.22\textwidth}  
		\centering 
		\includegraphics[width=\textwidth]{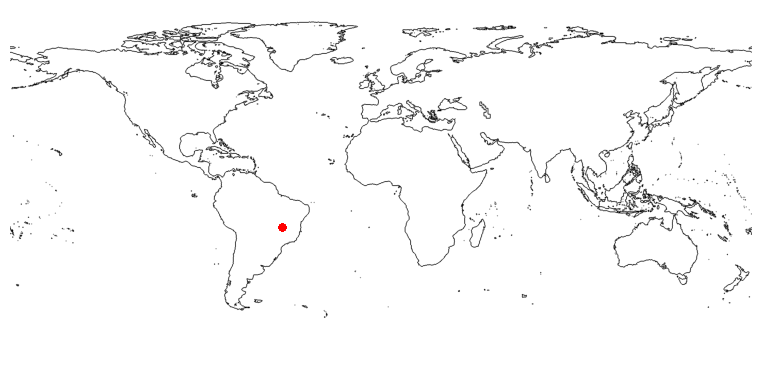}\vspacepred
		\caption[]{{\small 
Red-striped leafwing
		}}    
		\label{fig:1555_dist}
	\end{subfigure}
	\hfill
	\begin{subfigure}[b]{0.22\textwidth}  
		\centering 
		\includegraphics[width=\textwidth]{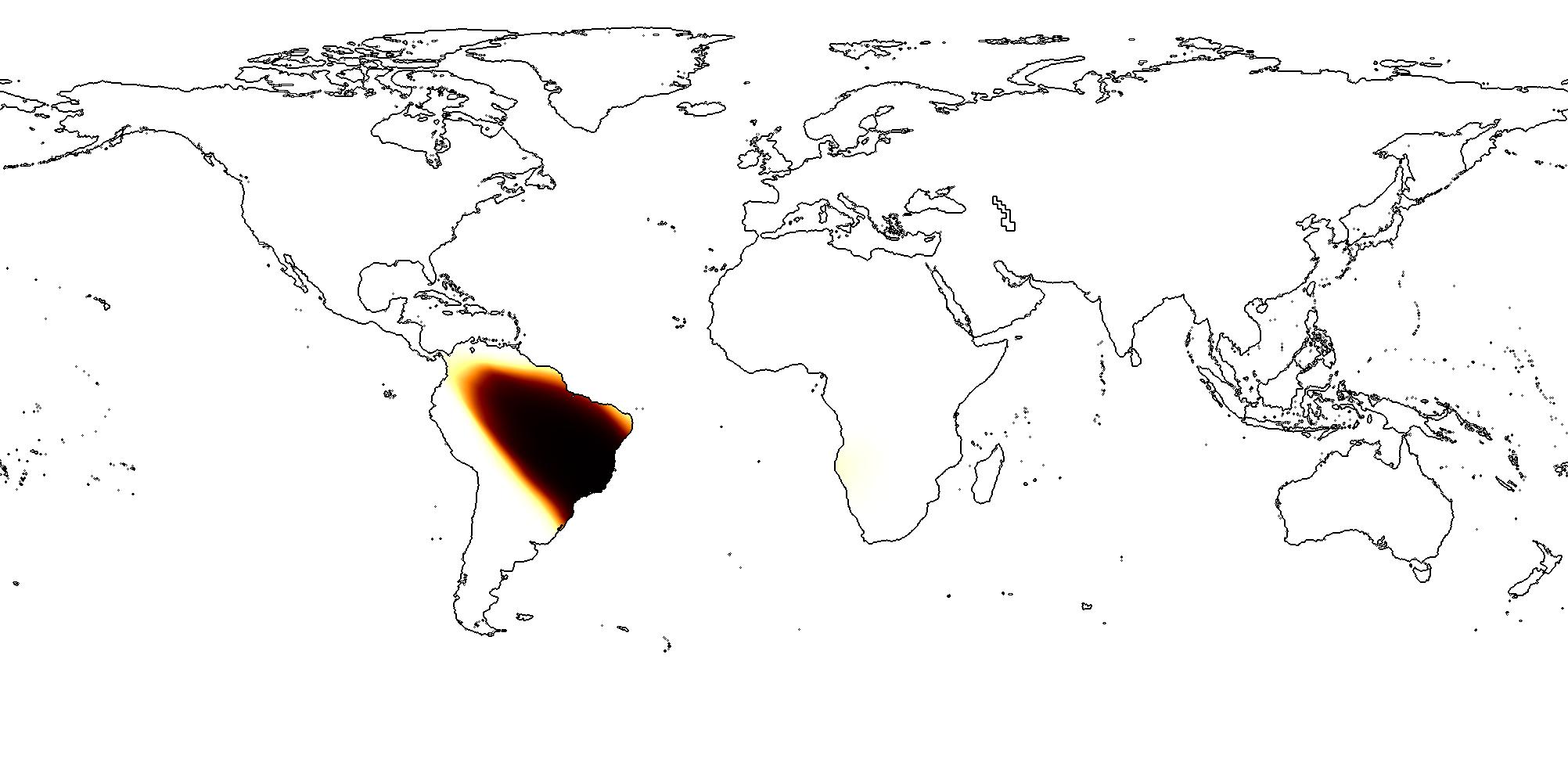}\vspacepred
		\caption[]{{\small 
		$\aodha$
		}}    
		\label{fig:1555_aodha}
	\end{subfigure}
	\hfill
	\begin{subfigure}[b]{0.22\textwidth}  
		\centering 
		\includegraphics[width=\textwidth]{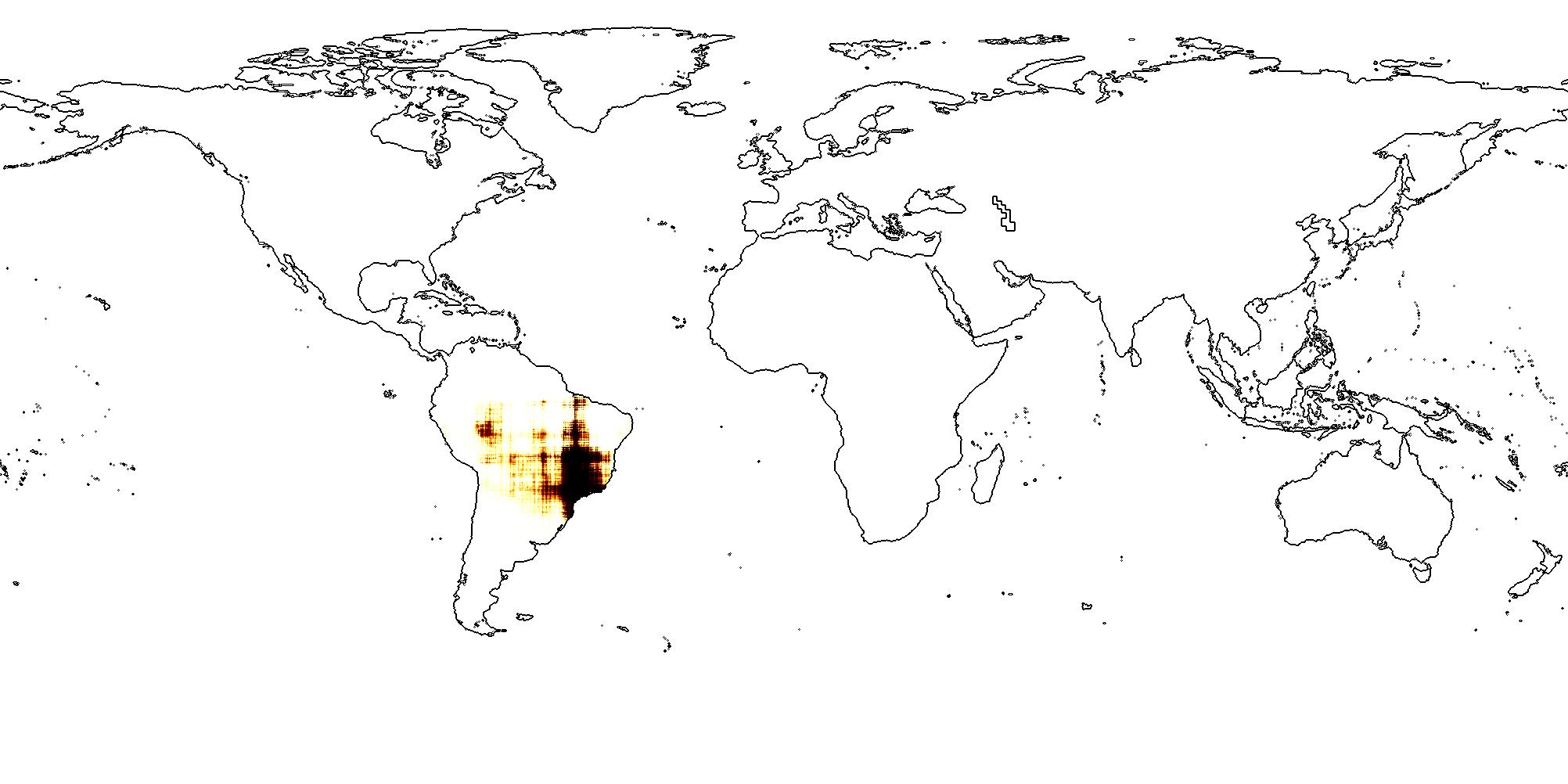}\vspacepred
		\caption[]{{\small 
		$\grid$
		}}    
		\label{fig:1555_grid}
	\end{subfigure}
	\hfill
	\begin{subfigure}[b]{0.22\textwidth}  
		\centering 
		\includegraphics[width=\textwidth]{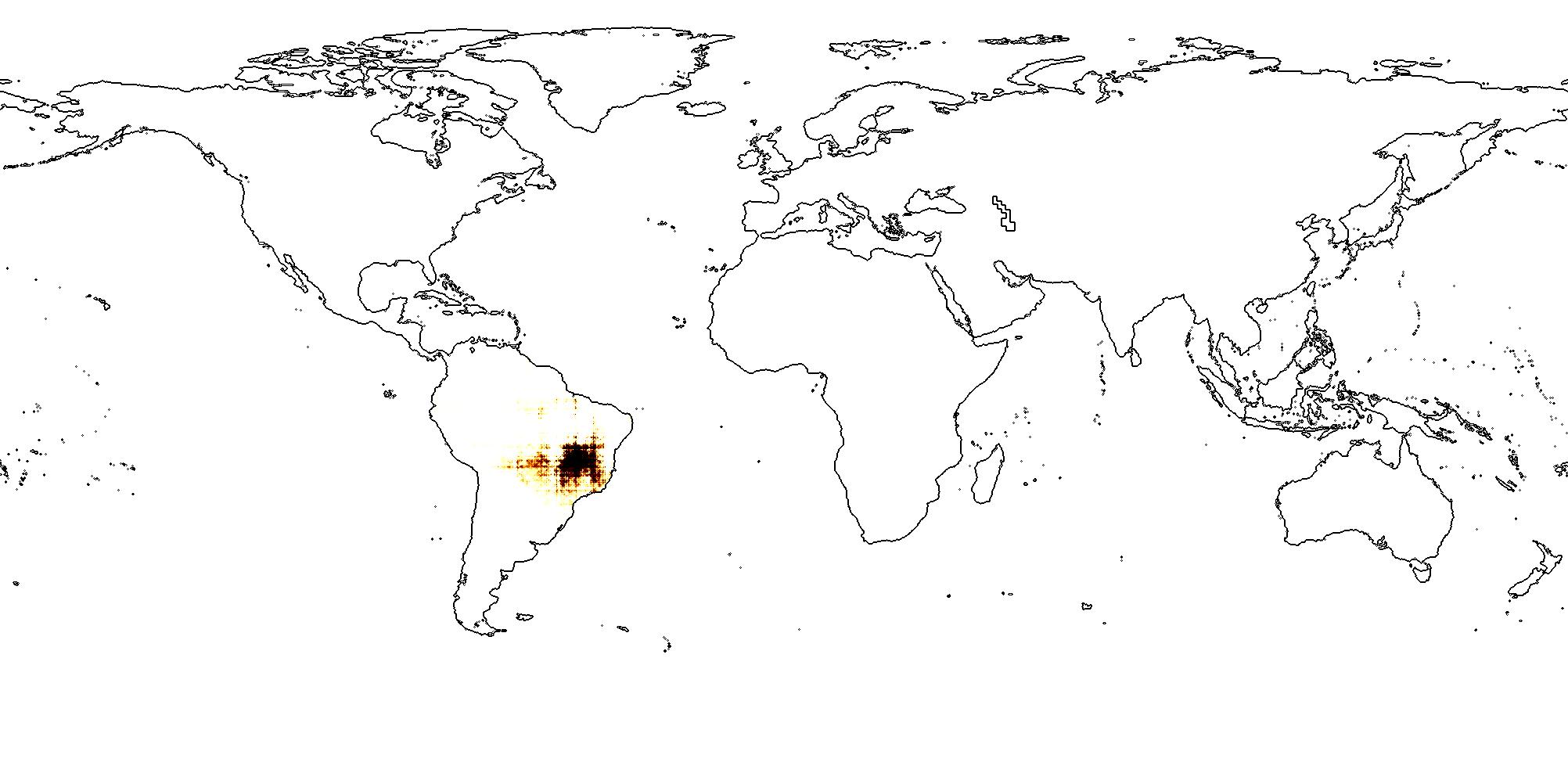}\vspacepred
		\caption[]{{\small 
		$\spheregrid$
		}}    
		\label{fig:1555_spheregrid}
	\end{subfigure}
	
	\hfill
	\begin{subfigure}[b]{0.10\textwidth}  
		\centering 
		\includegraphics[width=\textwidth]{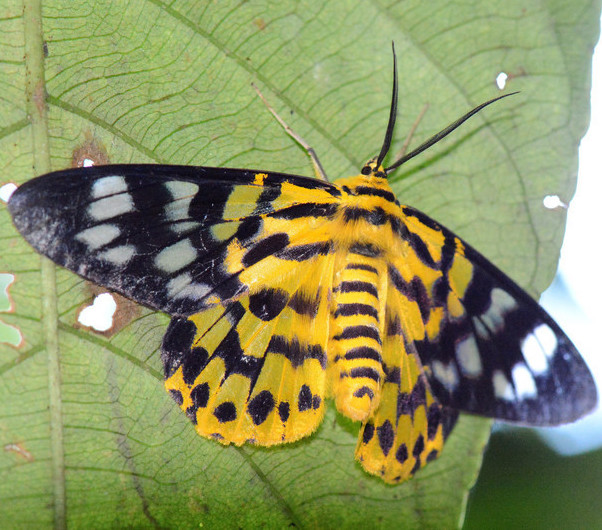}\vspace*{-0.2cm}
		\caption[]{{\small 
		Image
		}}    
		\label{fig:Dysphania_militaris_img}
	\end{subfigure}
	\hfill
	\begin{subfigure}[b]{0.22\textwidth}  
		\centering 
		\includegraphics[width=\textwidth]{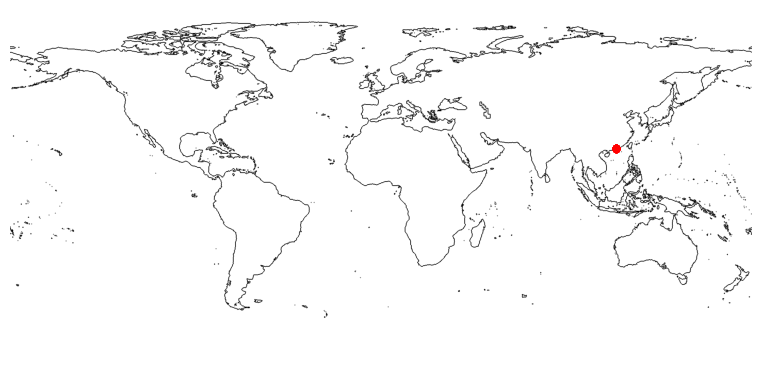}\vspacepred
		\caption[]{{\small 
		False Tiger Moth
		}}    
		\label{fig:0904_dist}
	\end{subfigure}
	\hfill
	\begin{subfigure}[b]{0.22\textwidth}  
		\centering 
		\includegraphics[width=\textwidth]{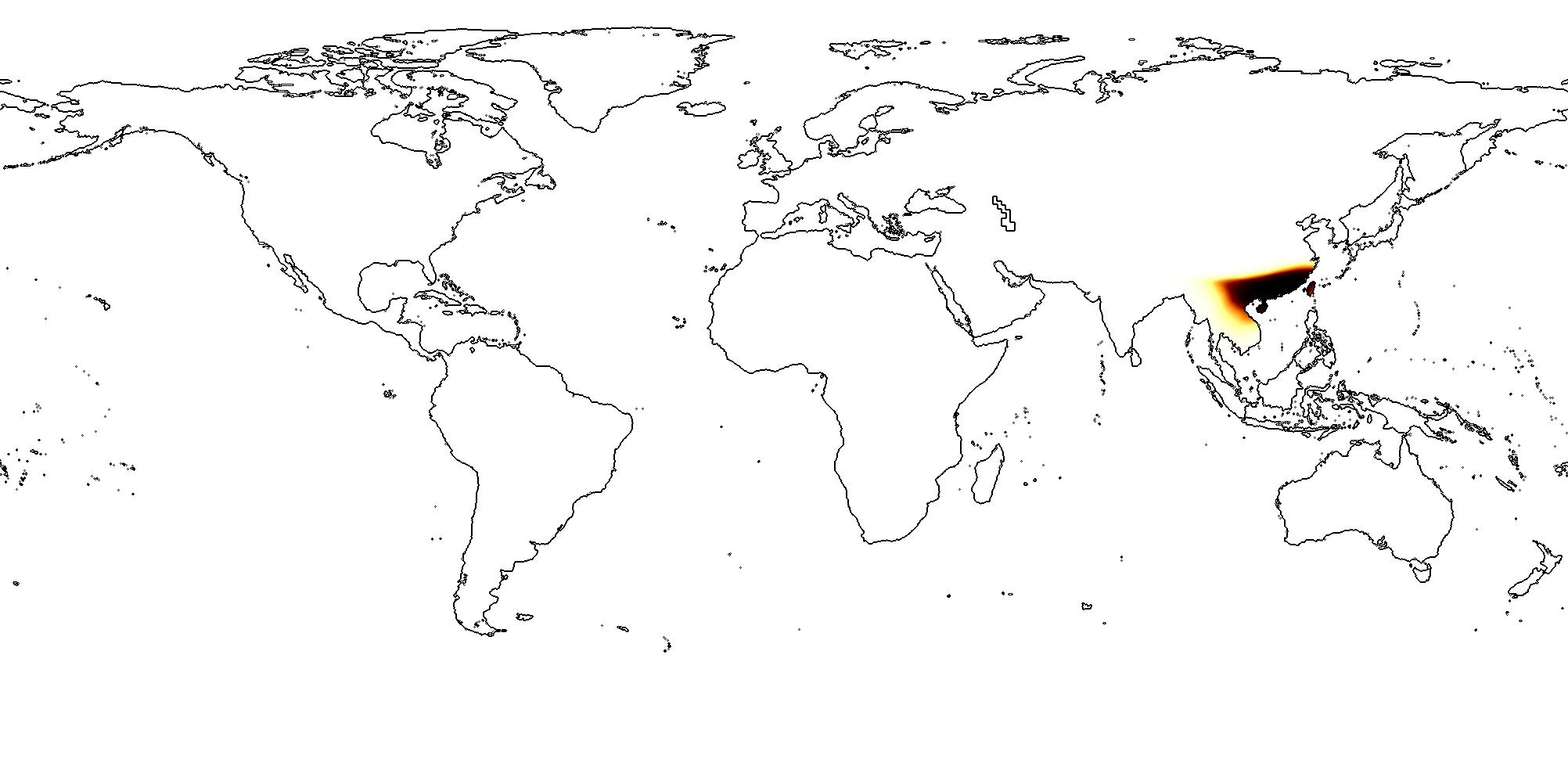}\vspacepred
		\caption[]{{\small 
		$\aodha$
		}}    
		\label{fig:0904_aodha}
	\end{subfigure}
	\hfill
	\begin{subfigure}[b]{0.22\textwidth}  
		\centering 
		\includegraphics[width=\textwidth]{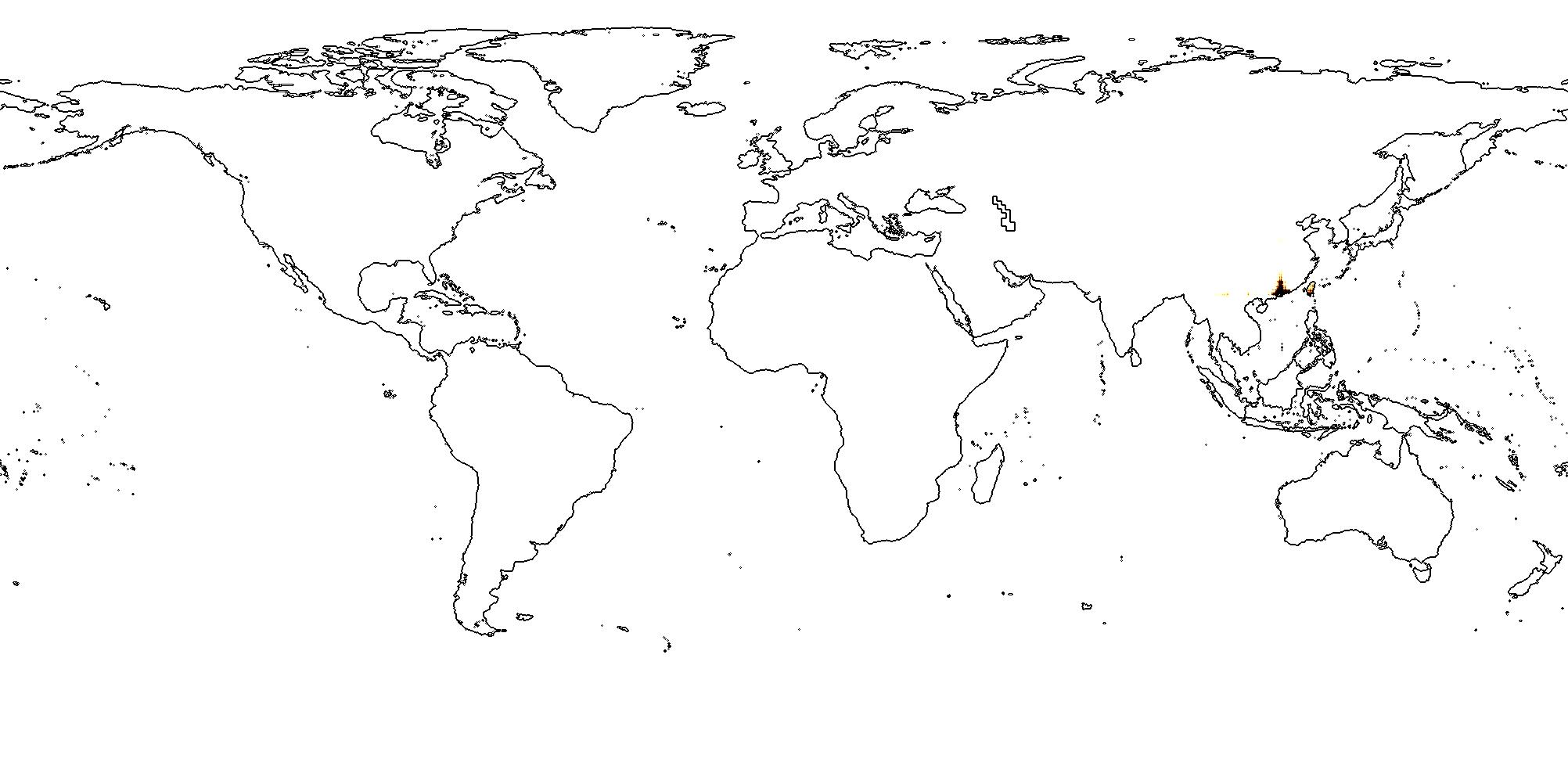}\vspacepred
		\caption[]{{\small 
		$\grid$
		}}    
		\label{fig:0904_grid}
	\end{subfigure}
	\hfill
	\begin{subfigure}[b]{0.22\textwidth}  
		\centering 
		\includegraphics[width=\textwidth]{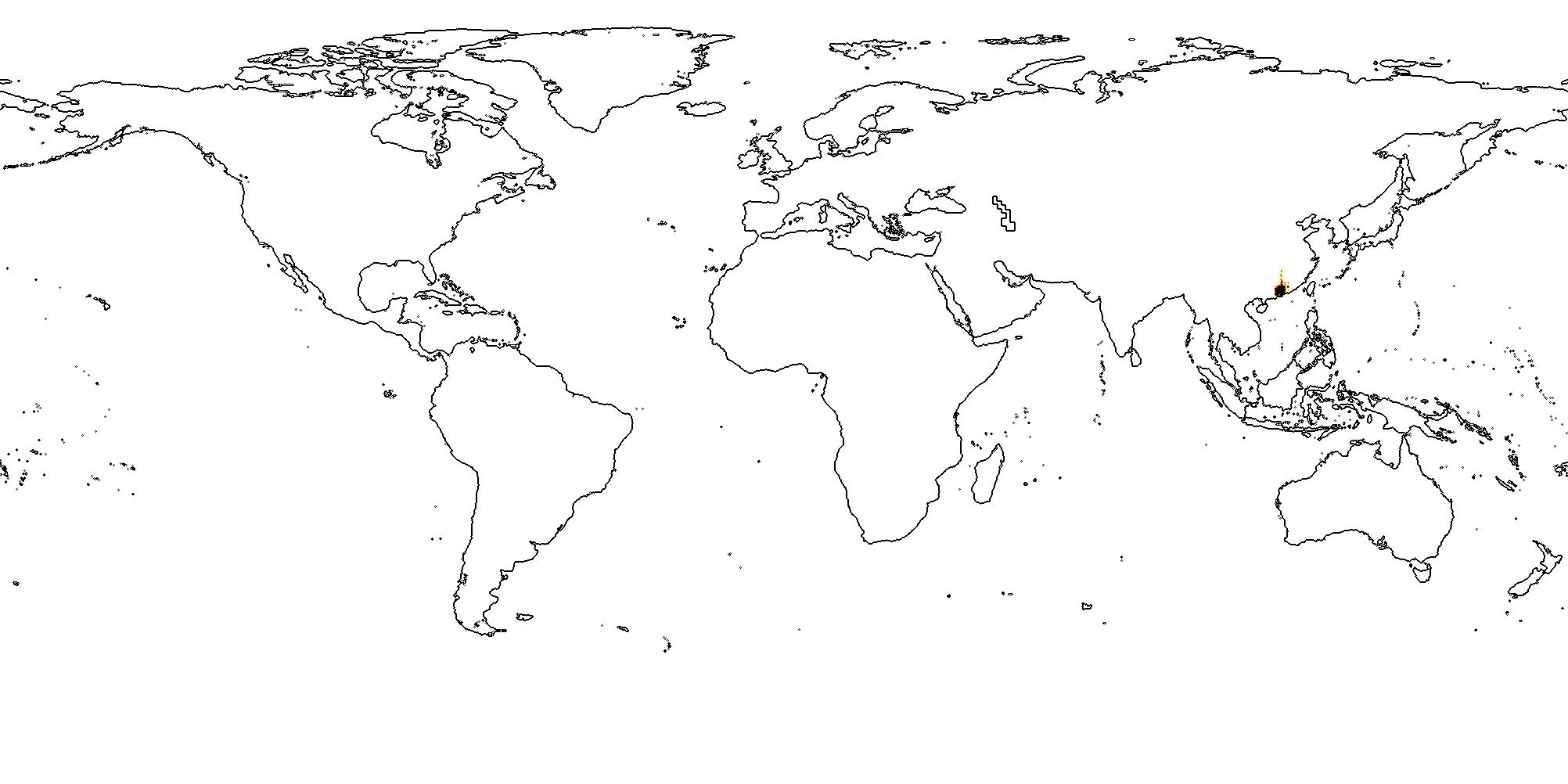}\vspacepred
		\caption[]{{\small 
		$\spheregrid$
		}}    
		\label{fig:0904_spheregrid}
	\end{subfigure}
	
	\caption{Comparison of the predicted spatial distributions of example species in the iNat2018 dataset from different location encoders. 
Each row indicates one specific species. We show one marine polychaete worm species, two bird species, two fox species, and two butterfly species. 
	The first and second figure of each row show an example figure as well as the data points of this species from iNat2018 training data.
} 
	\label{fig:spesdist18}
	\vspace*{-0.15cm}
\end{figure*}

\begin{figure*}[!htbp]
	\centering \tiny
	\vspace*{-0.2cm}
	\begin{subfigure}[b]{0.10\textwidth}  
		\centering 
		\includegraphics[width=\textwidth]{./fmow_16_factory_or_powerplant_img.png}\vspace*{-0.2cm}
		\caption[]{{\small 
		Image
		}}    
		\label{fig:factory_or_powerplant_img}
	\end{subfigure}
	\hfill
	\begin{subfigure}[b]{0.22\textwidth}  
		\centering 
		\includegraphics[width=\textwidth]{./fmow_16_factory_or_powerplant_locs.png}\vspacepred
		\caption[]{{\small 
		Factory or powerplant
		}}    
		\label{fig:factory_or_powerplant_loc}
	\end{subfigure}
	\hfill
	\begin{subfigure}[b]{0.22\textwidth}  
		\centering 
		\includegraphics[width=\textwidth]{./gt_0016_factory_or_powerplant_geo_net_predict.jpeg}\vspacepred
		\caption[]{{\small 
		$\aodha$
		}}    
		\label{fig:factory_or_powerplant_aodha_pred}
	\end{subfigure}
	\hfill
	\begin{subfigure}[b]{0.22\textwidth}  
		\centering 
		\includegraphics[width=\textwidth]{./gt_0016_factory_or_powerplant_gridcell_predict.jpeg}\vspacepred
		\caption[]{{\small 
		$\grid$
		}}    
		\label{fig:factory_or_powerplant_grid_pred}
	\end{subfigure}
	\hfill
	\begin{subfigure}[b]{0.22\textwidth}  
		\centering 
		\includegraphics[width=\textwidth]{./gt_0016_factory_or_powerplant_dft_predict.jpeg}\vspacepred
		\caption[]{{\small 
		$\dft$
		}}    
		\label{fig:factory_or_powerplant_dft_pred}
	\end{subfigure}

	\hfill
	\begin{subfigure}[b]{0.10\textwidth}  
		\centering 
		\includegraphics[width=\textwidth]{./fmow_30_multi-unit_residential_img.png}\vspace*{-0.2cm}
		\caption[]{{\small 
		Image
		}}    
		\label{fig:multi-unit_residential_img}
	\end{subfigure}
	\hfill
	\begin{subfigure}[b]{0.22\textwidth}  
		\centering 
		\includegraphics[width=\textwidth]{./fmow_30_multi-unit_residential_locs.png}\vspacepred
		\caption[]{{\small 
		Multi-unit residential
		}}    
		\label{fig:multi-unit_residential_loc}
	\end{subfigure}
	\hfill
	\begin{subfigure}[b]{0.22\textwidth}  
		\centering 
		\includegraphics[width=\textwidth]{./gt_0030_multi-unit_residential_geo_net_predict.jpeg}\vspacepred
		\caption[]{{\small 
		$\aodha$
		}}    
		\label{fig:multi-unit_residential_aodha_pred}
	\end{subfigure}
	\hfill
	\begin{subfigure}[b]{0.22\textwidth}  
		\centering 
		\includegraphics[width=\textwidth]{./gt_0030_multi-unit_residential_gridcell_predict.jpeg}\vspacepred
		\caption[]{{\small 
		$\grid$
		}}    
		\label{fig:multi-unit_residential_grid_pred}
	\end{subfigure}
	\hfill
	\begin{subfigure}[b]{0.22\textwidth}  
		\centering 
		\includegraphics[width=\textwidth]{./gt_0030_multi-unit_residential_dft_predict.jpeg}\vspacepred
		\caption[]{{\small 
		$\dft$
		}}    
		\label{fig:multi-unit_residential_dft_pred}
	\end{subfigure}

	\hfill
	\begin{subfigure}[b]{0.10\textwidth}  
		\centering 
		\includegraphics[width=\textwidth]{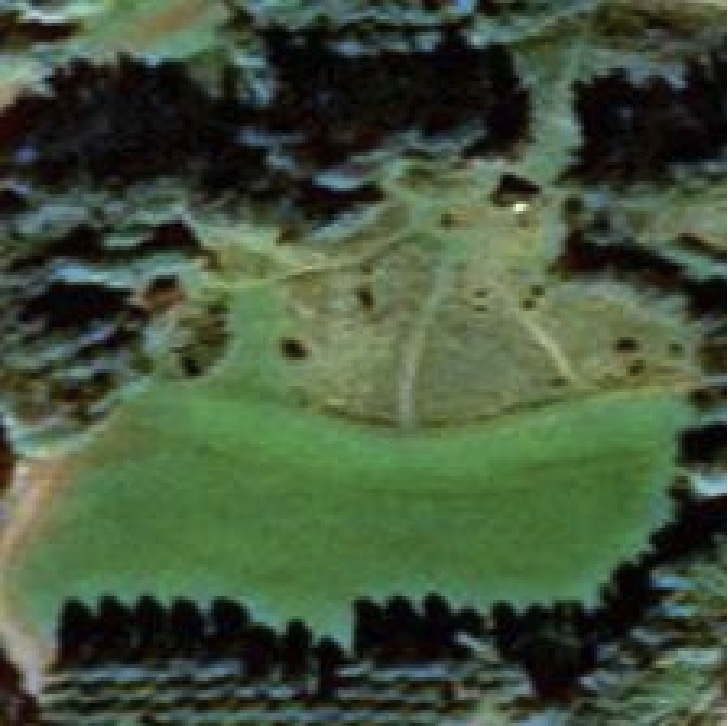}\vspace*{-0.2cm}
		\caption[]{{\small 
		Image
		}}    
		\label{fig:park_img}
	\end{subfigure}
	\hfill
	\begin{subfigure}[b]{0.22\textwidth}  
		\centering 
		\includegraphics[width=\textwidth]{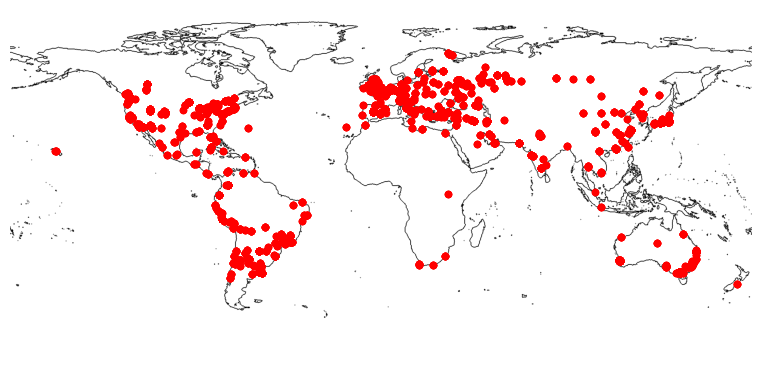}\vspacepred
		\caption[]{{\small 
		Park
		}}    
		\label{fig:park_loc}
	\end{subfigure}
	\hfill
	\begin{subfigure}[b]{0.22\textwidth}  
		\centering 
		\includegraphics[width=\textwidth]{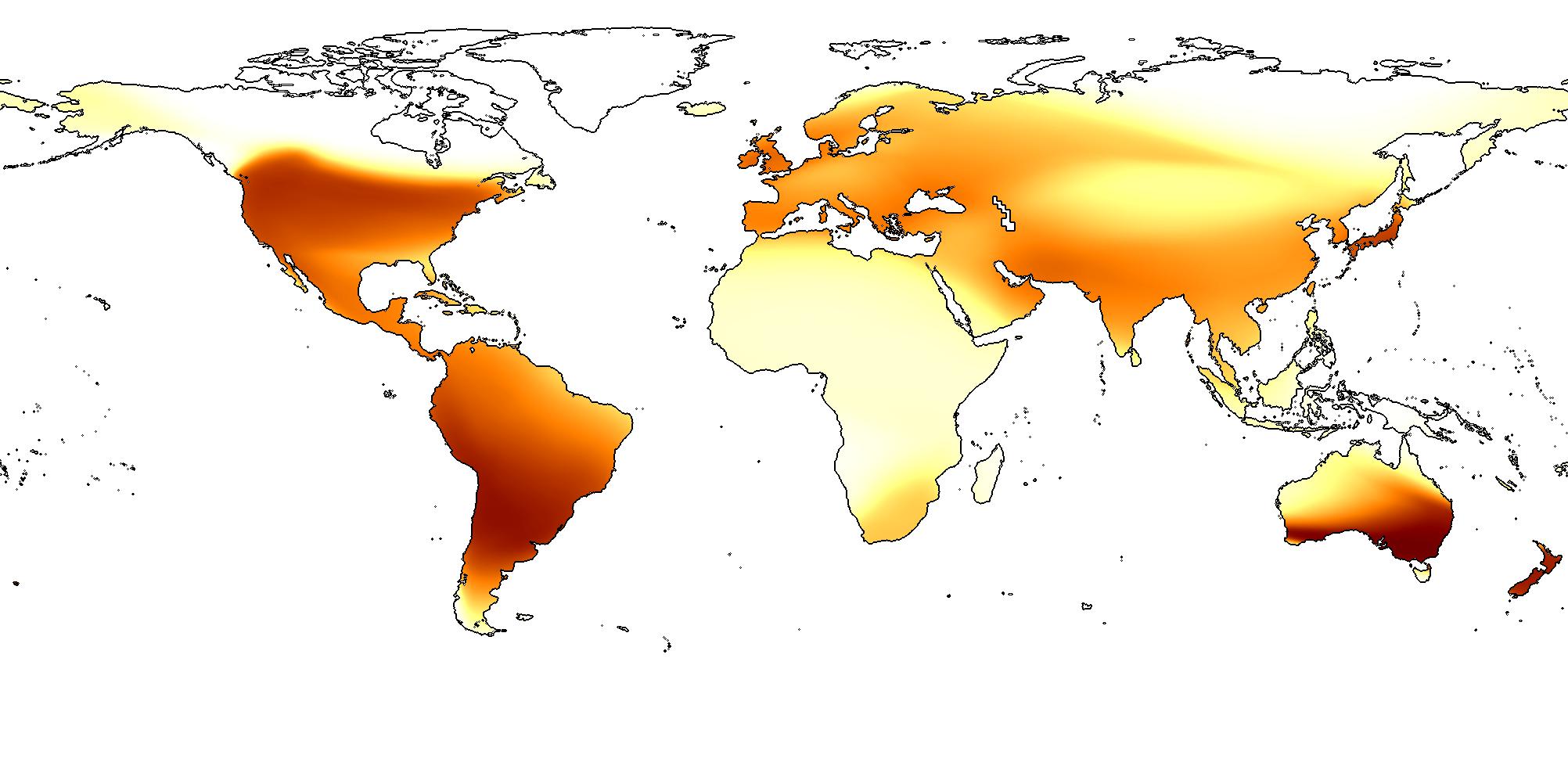}\vspacepred
		\caption[]{{\small 
		$\aodha$
		}}    
		\label{fig:park_aodha_pred}
	\end{subfigure}
	\hfill
	\begin{subfigure}[b]{0.22\textwidth}  
		\centering 
		\includegraphics[width=\textwidth]{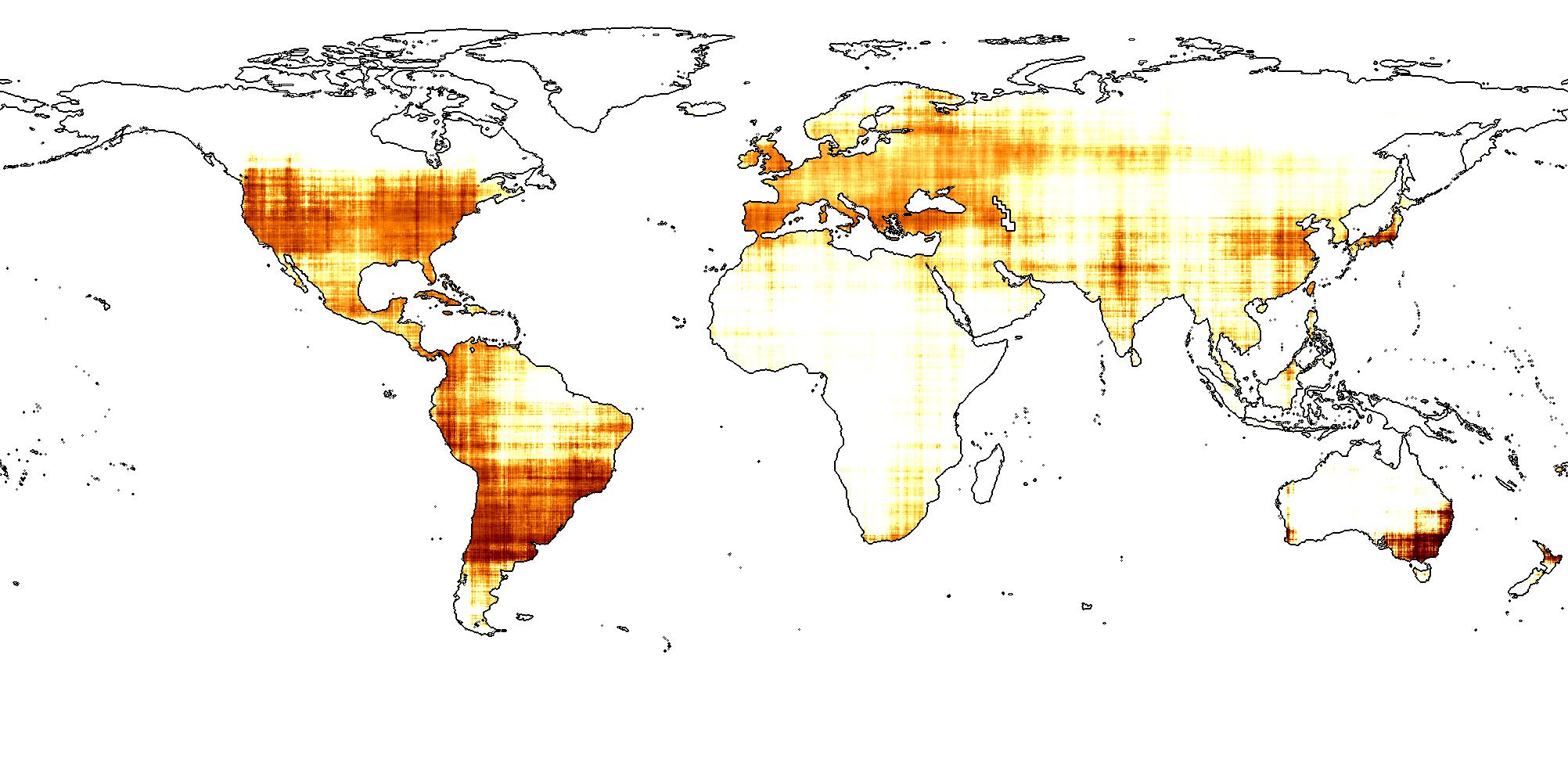}\vspacepred
		\caption[]{{\small 
		$\grid$
		}}    
		\label{fig:park_grid_pred}
	\end{subfigure}
	\hfill
	\begin{subfigure}[b]{0.22\textwidth}  
		\centering 
		\includegraphics[width=\textwidth]{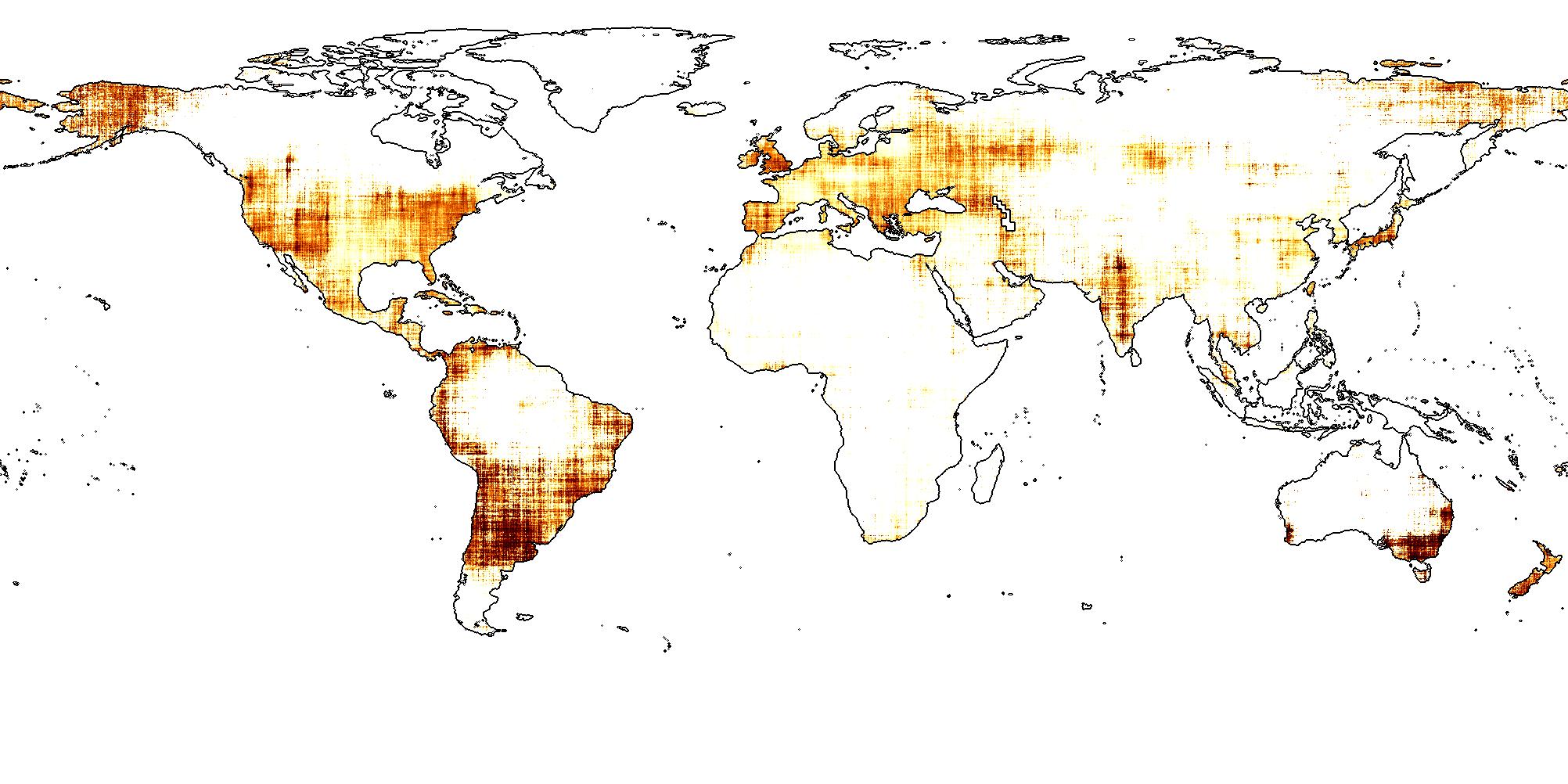}\vspacepred
		\caption[]{{\small 
		$\dft$
		}}    
		\label{fig:park_dft_pred}
	\end{subfigure}
	
	\hfill
	\begin{subfigure}[b]{0.10\textwidth}  
		\centering 
		\includegraphics[width=\textwidth]{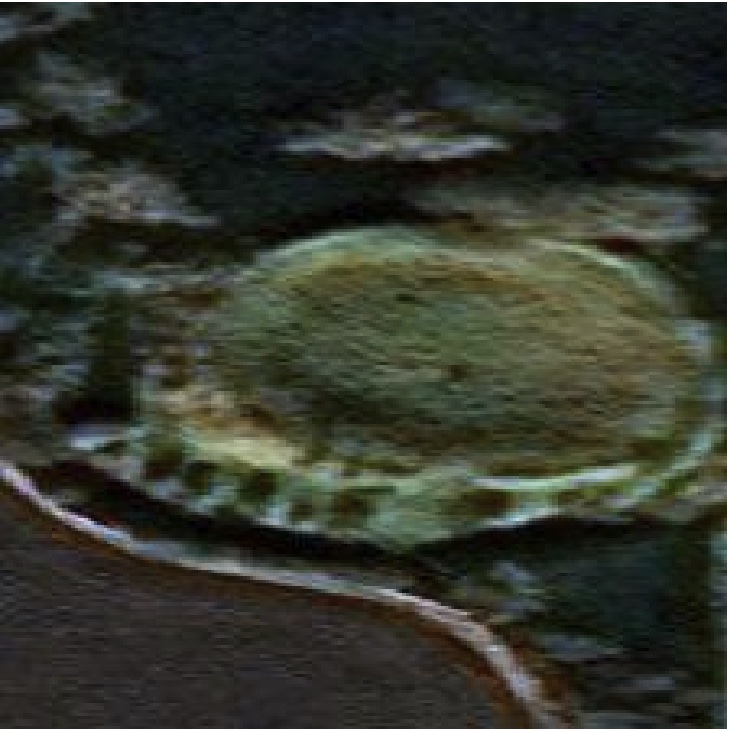}\vspace*{-0.2cm}
		\caption[]{{\small 
		Image
		}}    
		\label{fig:archaeological_site_img}
	\end{subfigure}
	\hfill
	\begin{subfigure}[b]{0.22\textwidth}  
		\centering 
		\includegraphics[width=\textwidth]{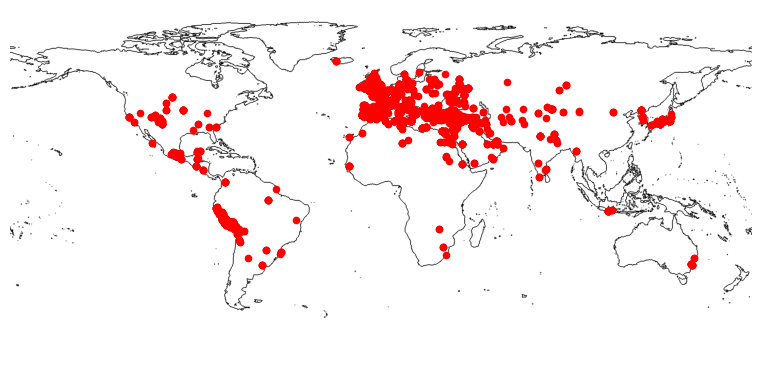}\vspacepred
		\caption[]{{\small 
		Archaeological site
		}}    
		\label{fig:archaeological_site_loc}
	\end{subfigure}
	\hfill
	\begin{subfigure}[b]{0.22\textwidth}  
		\centering 
		\includegraphics[width=\textwidth]{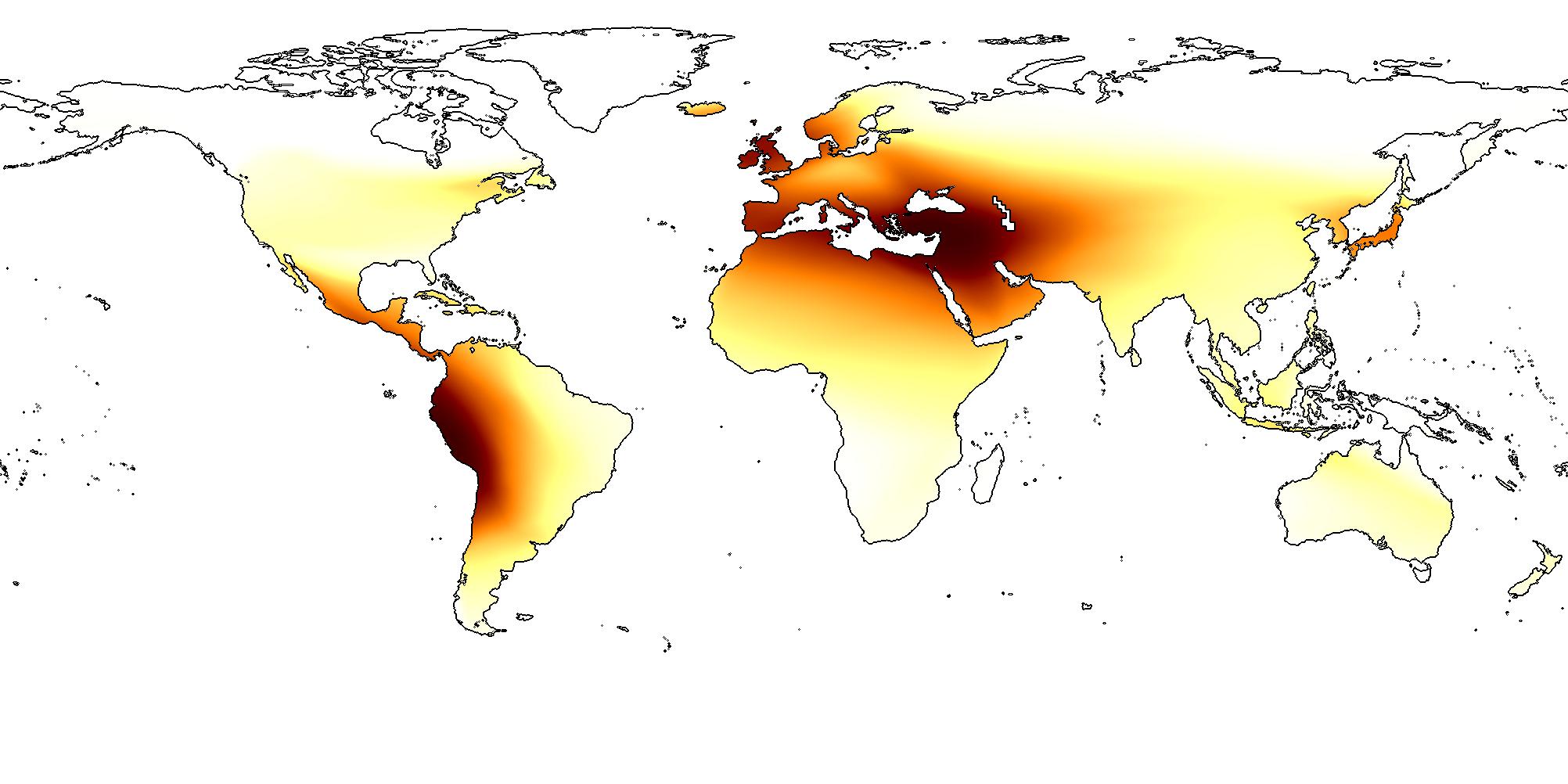}\vspacepred
		\caption[]{{\small 
		$\aodha$
		}}    
		\label{fig:archaeological_site_aodha_pred}
	\end{subfigure}
	\hfill
	\begin{subfigure}[b]{0.22\textwidth}  
		\centering 
		\includegraphics[width=\textwidth]{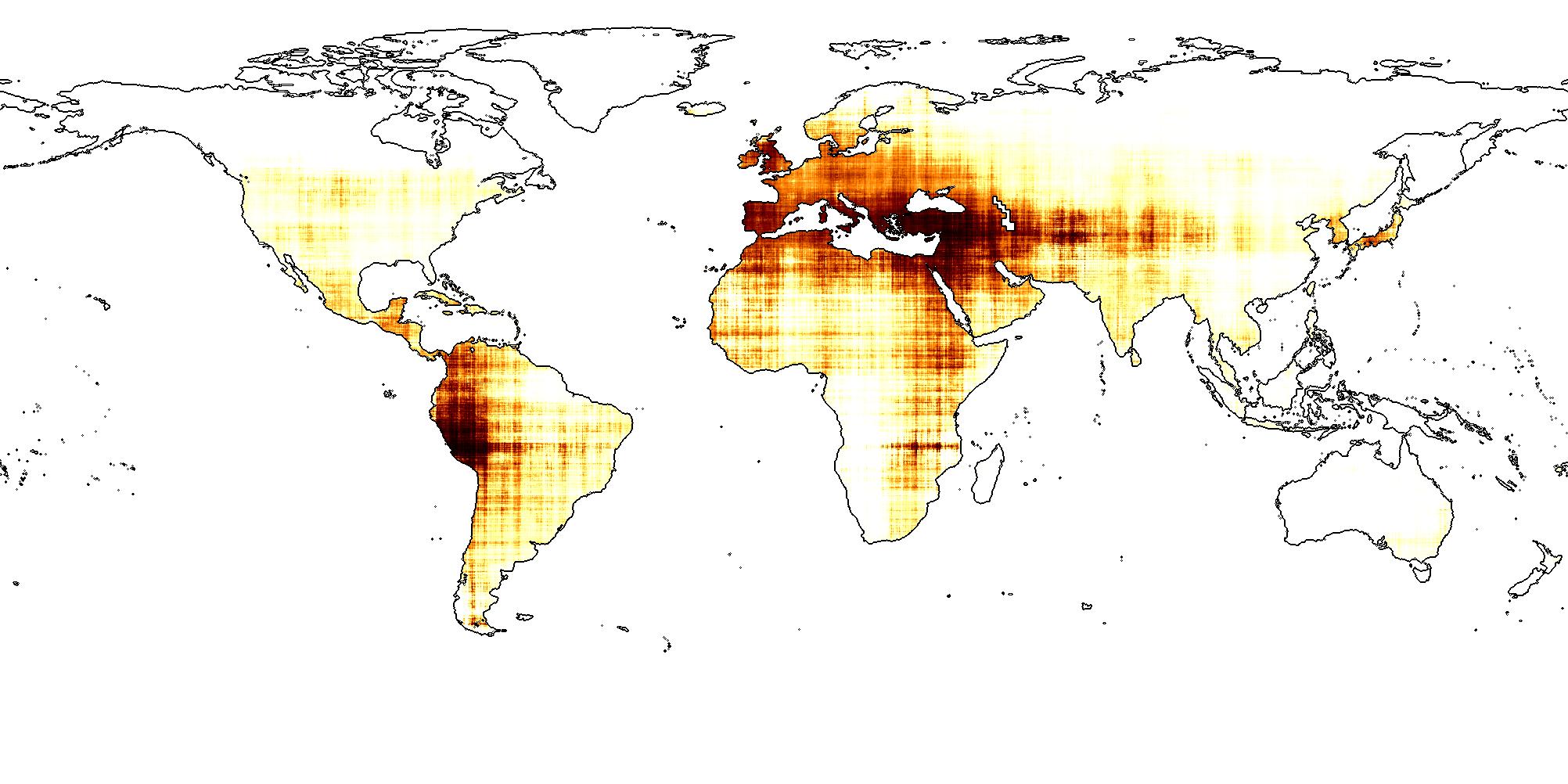}\vspacepred
		\caption[]{{\small 
		$\grid$
		}}    
		\label{fig:archaeological_site_grid_pred}
	\end{subfigure}
	\hfill
	\begin{subfigure}[b]{0.22\textwidth}  
		\centering 
		\includegraphics[width=\textwidth]{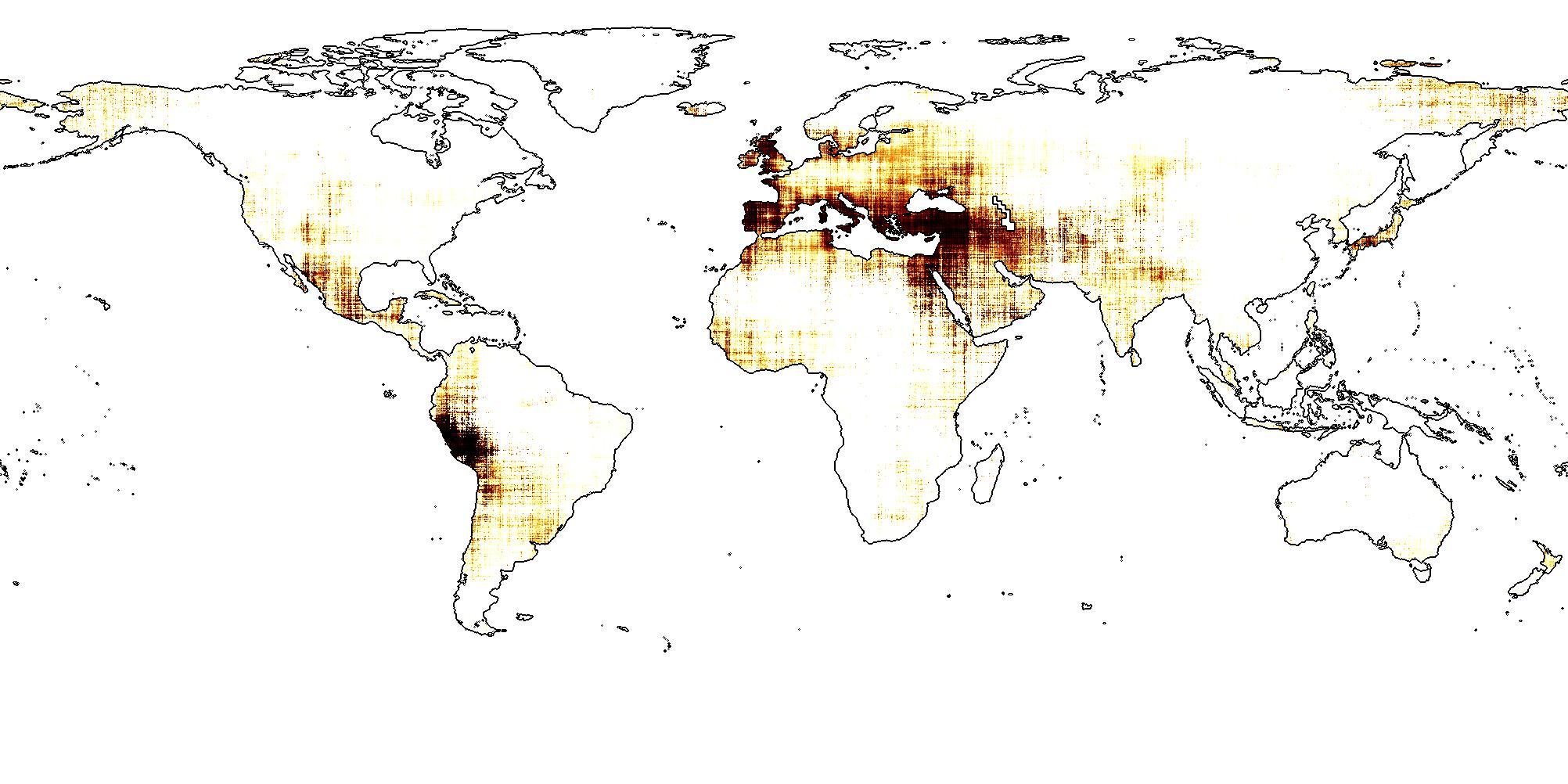}\vspacepred
		\caption[]{{\small 
		$\dft$
		}}    
		\label{fig:archaeological_site_dft_pred}
	\end{subfigure}
	\caption{Comparison of the predicted spatial distributions of example land use types in the fMoW dataset from different location encoders. 
Each row indicates one specific land use type.
	The first and second figure of each row show an example figure as well as the data points of this land use types from the fMoW training data.
	As shown in Figure (a) and (f), although factories or powerplants and multi-unit residential type look very similar from overhead satellite imageries, they have very distinct spatial distribution (Figure (b) and (g)). Similarly, parks and archaeological sites look similar from satellites imageries (Figure (k) and (p)) which are usually covered by vegetation. However, they have very distinct spatial distribution (Figure (l) and (q)).
	We compare the predicted spatial distribution of each land use type from three different location encoders: $\aodha$, $\grid$, and $\dft$. 
} 
	\label{fig:fmow_pred_dist}
	\vspace*{-0.15cm}
\end{figure*}

To have a better understanding of how well different location encoders model the geographic prior distributions of different image labels, 
we use iNat2018 and fMoW data as examples and plot the predicted spatial distributions of different example species/land use types from different location encoders, and compare them with the training sample locations of the corresponding species or land use types (see Figure \ref{fig:spesdist18} and \ref{fig:fmow_pred_dist}).

\subsubsection{Predicted Species Distribution for iNat2018}
From Figure \ref{fig:spesdist18}, we can see that $\aodha$ \citep{mac2019presence} produces rather over-generalized species distributions due to the fact that it is a single-scale location encoder. 
$\spheregrid$ (our model) produces a more compact and fine-grained distribution in each geographic region, especially in the polar region and in \mai{data-sparse} areas such as Africa and Asia.
The distributions produced by $\grid$ \citep{mai2020multiscale} \mai{are} between these two. However, $\grid$ has limited spatial distribution modeling ability in the polar area (e.g., Figure \ref{fig:0002_grid} and \ref{fig:4084_grid}) as well as \mai{data-sparse} regions.

For example, in the white-browed wagtail example, $\aodha$ produces \mai{an over-generalized} spatial distribution which covers India, East Saudi Arabia, and the Southwest of China (See Figure \ref{fig:3475_aodha}). However, according to the training sample locations (Figure \ref{fig:3475_dist}), white-browed wagtails only \mai{occur} in India. $\grid$ is better than $\aodha$ but still \mai{produces} a distribution covering the Southwest of China. $\spheregrid$ produces the best compact distribution estimation. Similarly, for the red-striped leafwing, the sample locations are clustered in a small region \mai{in West Africa} while $\aodha$ produces \mai{an over-generalized distribution} (see Figure \ref{fig:1555_aodha}). $\grid$ produces a better distribution estimation (see Figure \ref{fig:1555_grid}) but it still has a over-generalized issue. 
Our $\spheregrid$ produces the best estimation among these three models -- a compact distribution estimation covering the exact West Africa region (See Figure \ref{fig:1555_spheregrid}).

\subsubsection{Predicted Land Use Distribution for fMoW}
Similar visualizations are made for some example land use types in the fMoW dataset, i.e., Figure \ref{fig:fmow_pred_dist}. 
Factories/powerplants (Figure \ref{fig:factory_or_powerplant_loc}) might look similar to multi-unit residential buildings (Figure \ref{fig:multi-unit_residential_img}) from overhead satellite imageries. But they have very different geographic distributions (Figure \ref{fig:factory_or_powerplant_loc} and \ref{fig:multi-unit_residential_loc}). 
A similarly situation can be seen for parks (Figure \ref{fig:park_img} and \ref{fig:park_loc}) and archaeological sites (Figure \ref{fig:archaeological_site_img} and \ref{fig:archaeological_site_loc}).

The estimated spatial distributions of these four land use types from three location encoders, i.e., $\aodha$, $\grid$, and $\dft$ are visualized. Just like what we see from Figure \ref{fig:spesdist18}, similar observations can be made.
$\aodha$ usually produces over-generalized distributions. $\dft$ generates more compact and accurate distributions while $\grid$ is between these two. We also find out that $\grid$ will generate some grid-like patterns due to the \mai{use of sinusoidal functions}. $\dft$ suffers less from it and produces more \mai{accurate} distributions.

\begin{figure*}[!htbp]
	\centering \tiny

	\vspace*{-0.2cm}
	\begin{subfigure}[b]{0.245\textwidth}  
		\centering 
		\includegraphics[width=\textwidth]{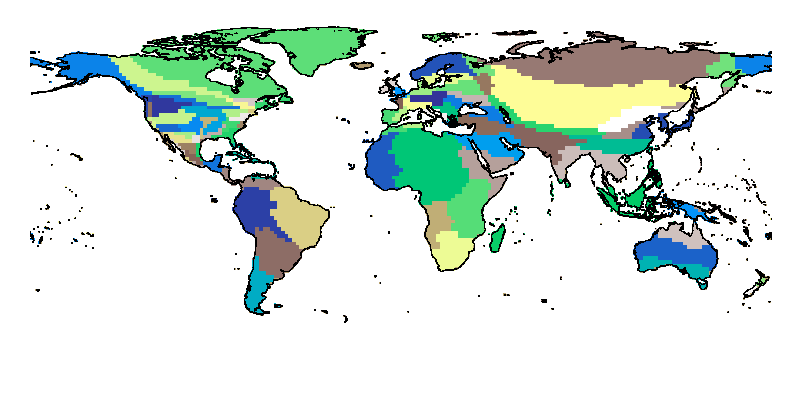}\vspaceclustering
		\caption[]{{\small 
		$\aodha$ }}    
		\label{fig:inat17_aodha}
	\end{subfigure}
	\hfill
	\begin{subfigure}[b]{0.245\textwidth}  
		\centering 
		\includegraphics[width=\textwidth]{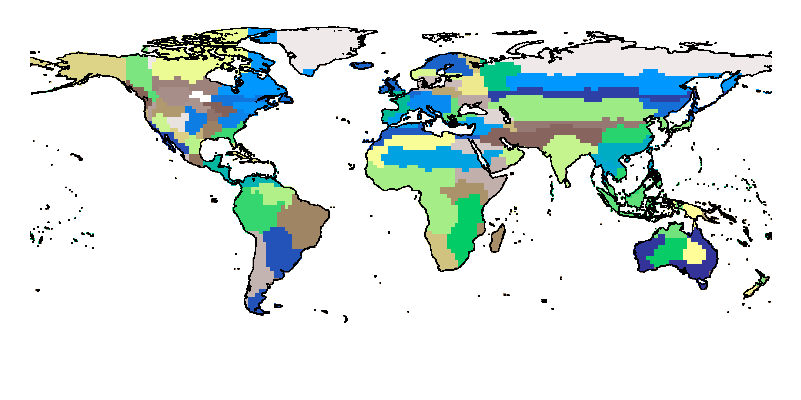}\vspaceclustering
		\caption[]{{\small 
		$\grid$ $(\minscale = 10^{-2})$
		}}    
		\label{fig:inat17_grid_2}
	\end{subfigure}
	\hfill
	\begin{subfigure}[b]{0.245\textwidth}  
		\centering 
		\includegraphics[width=\textwidth]{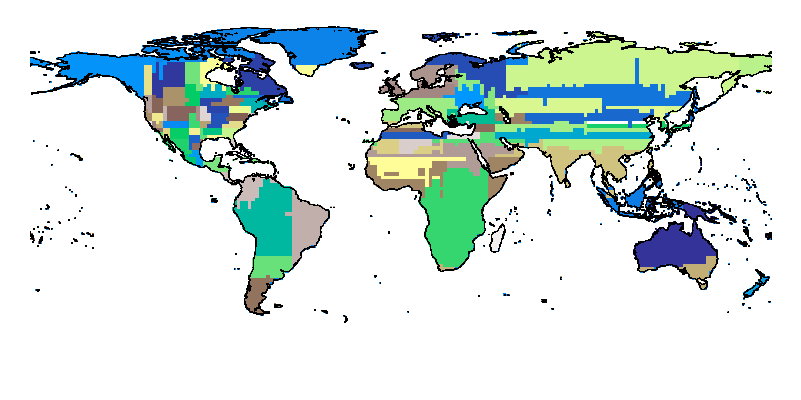}\vspaceclustering
		\caption[]{{\small 
		$\grid$ $(\minscale = 10^{-6})$
		}}    
		\label{fig:inat17_grid}
	\end{subfigure}
	\hfill
	\begin{subfigure}[b]{0.245\textwidth}  
		\centering 
		\includegraphics[width=\textwidth]{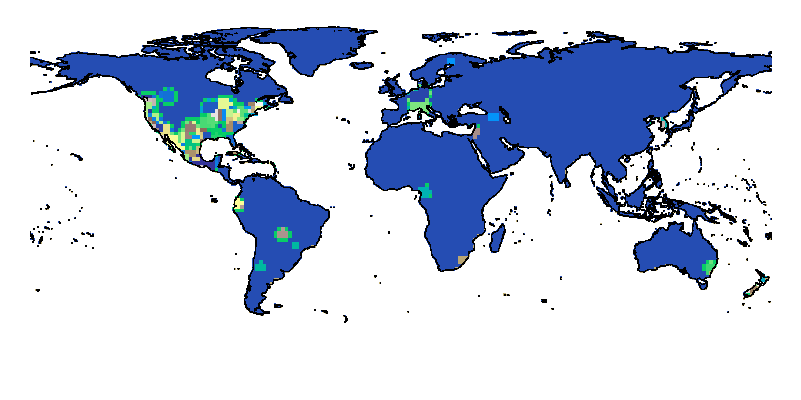}\vspaceclustering
		\caption[]{{\small 
		$\rbf$ $(\sigma = 1, \numkernel = 200)$
		}}    
		\label{fig:inat17_rbf}
	\end{subfigure}
	\hfill
	\begin{subfigure}[b]{0.245\textwidth}  
		\centering 
		\includegraphics[width=\textwidth]{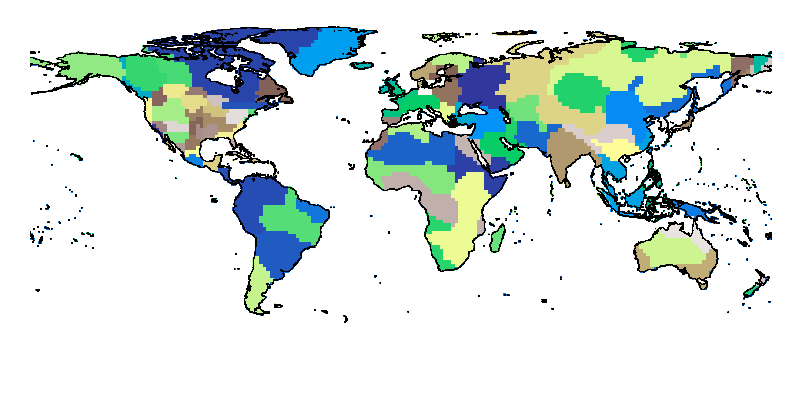}\vspaceclustering
		\caption[]{{\small 
		$\theory$ $ (\minscale = 10^{-2})$
		}}    
		\label{fig:inat17_theory_2}
	\end{subfigure}
	\hfill
	\begin{subfigure}[b]{0.245\textwidth}  
		\centering 
		\includegraphics[width=\textwidth]{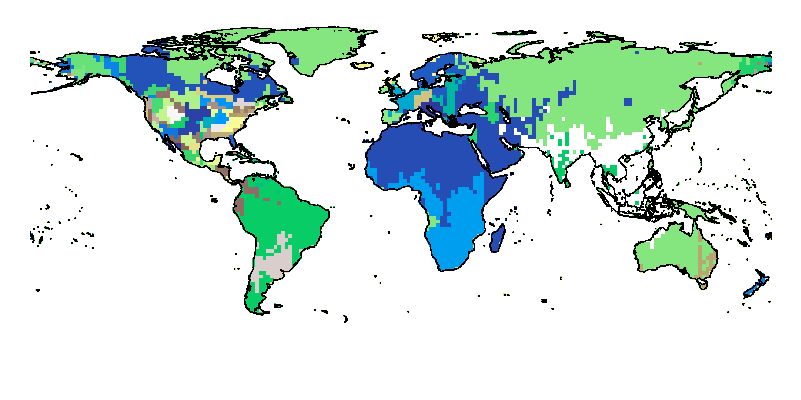}\vspaceclustering
		\caption[]{{\small 
		$\theory$ $ (\minscale = 10^{-6})$
		}}    
		\label{fig:inat17_theory}
	\end{subfigure}
        \hfill
	\begin{subfigure}[b]{0.245\textwidth}  
		\centering 
		\includegraphics[width=\textwidth]{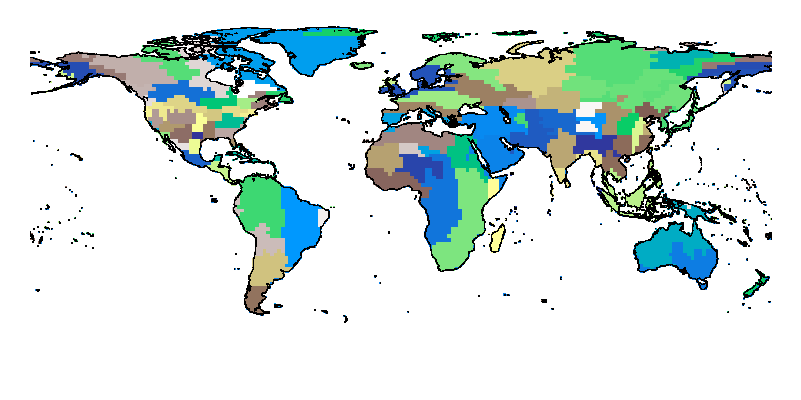}\vspaceclustering
		\caption[]{{\small 
		\mai{$\nerf$ $ (\nscale = 32)$}
		}}    
		\label{fig:inat17_nerf}
	\end{subfigure}
	\hfill
	\begin{subfigure}[b]{0.245\textwidth}  
		\centering 
		\includegraphics[width=\textwidth]{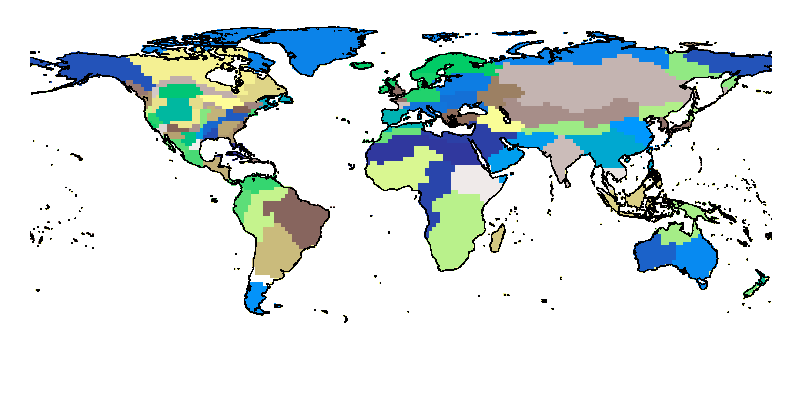}\vspaceclustering
		\caption[]{{\small 
		$\spheremixscale$ $ (\minscale = 10^{-1})$
		}}    
		\label{fig:inat17_spheremixscale_1}
	\end{subfigure}
	\hfill
	\begin{subfigure}[b]{0.245\textwidth}  
		\centering 
		\includegraphics[width=\textwidth]{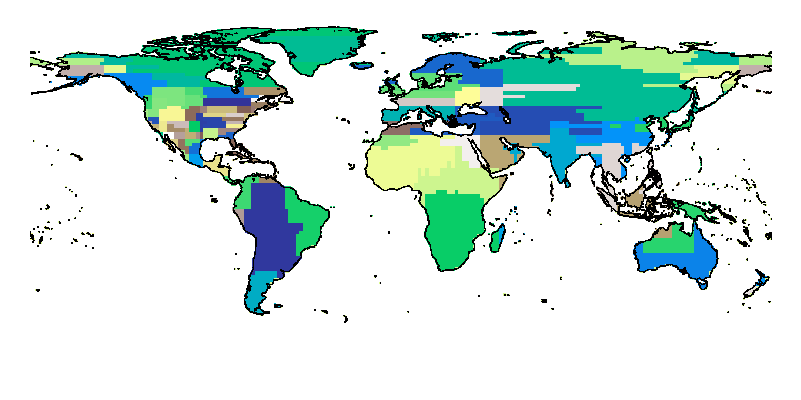}\vspaceclustering
		\caption[]{{\small 
		$\spheremixscale$ $ (\minscale = 10^{-2})$
		}}    
		\label{fig:inat17_spheremixscale_2}
	\end{subfigure}
\hfill
	\begin{subfigure}[b]{0.245\textwidth}  
		\centering 
		\includegraphics[width=\textwidth]{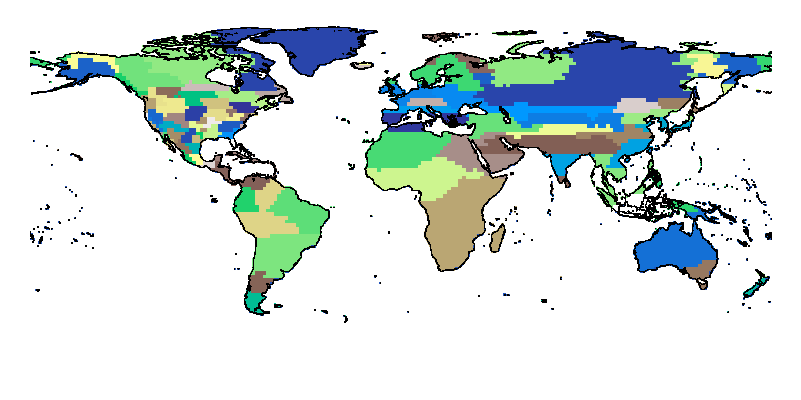}\vspaceclustering
		\caption[]{{\small 
		$\sphere$ $ (\minscale = 10^{-2})$
		}}    
		\label{fig:inat17_sphere}
	\end{subfigure}
	\hfill
	\begin{subfigure}[b]{0.245\textwidth}  
		\centering 
		\includegraphics[width=\textwidth]{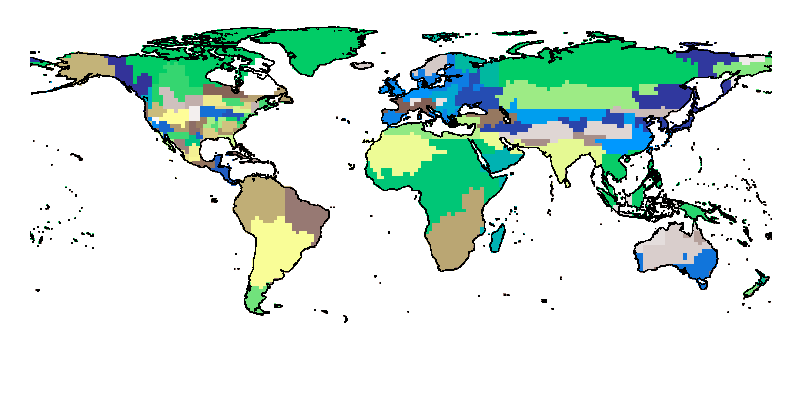}\vspaceclustering
		\caption[]{{\small 
		$\spheregrid$ $ (\minscale = 10^{-2})$
		}}    
		\label{fig:inat17_spheregrid}
	\end{subfigure}
	\hfill
	\begin{subfigure}[b]{0.245\textwidth}  
		\centering 
		\includegraphics[width=\textwidth]{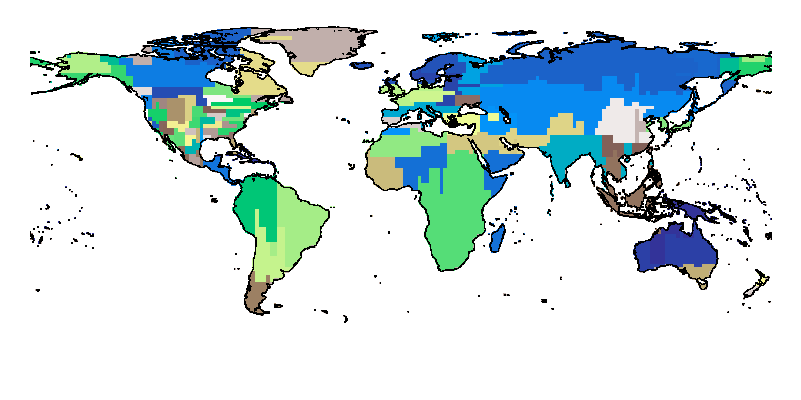}\vspaceclustering
		\caption[]{{\small 
		$\spheregridmixscale$ $ (\minscale = 10^{-2})$
		}}    
		\label{fig:inat17_spheregridmixscale}
	\end{subfigure}
	\hfill
	\begin{subfigure}[b]{0.245\textwidth}  
		\centering 
		\includegraphics[width=\textwidth]{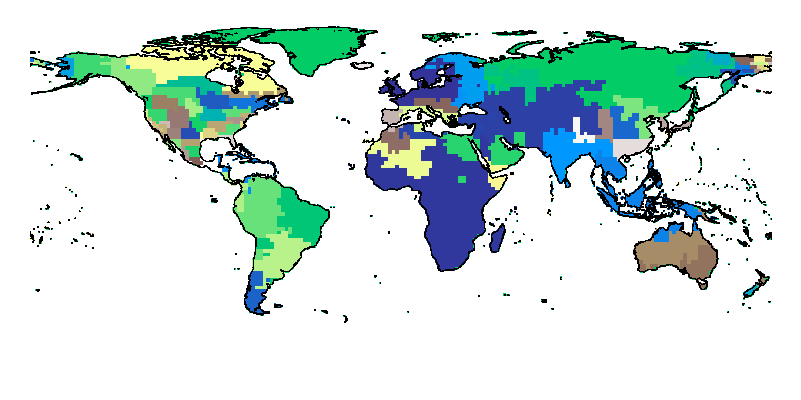}\vspaceclustering
		\caption[]{{\small 
		$\dft$ $ (\minscale = 10^{-2})$
		}}    
		\label{fig:inat17_dft}
	\end{subfigure}

	\caption[]{Embedding clusterings of different location encoders trained on the iNat2017 dataset. 
	(a) $\aodha*$ with 4 hidden ReLU layers of 256 neurons; 
	(d) $\rbf$ with the best kernel size $\kernelsize=1$ and number of anchor points $\numkernel=200$; 
(b)(c)(e)(f) are $\emph{\spacevec}$ models \citep{mai2020multiscale} with different min scale $\minscale = \{10^{-6}, 10^{-2}\}$.\textsuperscript{a}
    \mai{(g) is $\nerf$ with $\minscale=32$, and 1 hidden ReLU layer of 512 neurons. }
\mai{(h)-(m)} are different \emph{\modelname} models.\textsuperscript{b}
} 
	\scriptsize
	\textsuperscript{a} {They share the same best hyperparameters:  $\freq = 64$,  $\maxscale = 1$, and 1 hidden ReLU layers of 512 neurons.} \\
	\textsuperscript{b}{They share the same best hyperparameters: $\freq = 32$, $\maxscale = 1$, and 1 hidden ReLU layers of 1024 neurons.}
	\label{fig:inat17}
\end{figure*}

\begin{figure*}[!htbp]
	\centering \tiny
	\vspace*{-0.2cm}
	\begin{subfigure}[b]{0.24\textwidth}  
		\centering 
		\includegraphics[width=\textwidth]{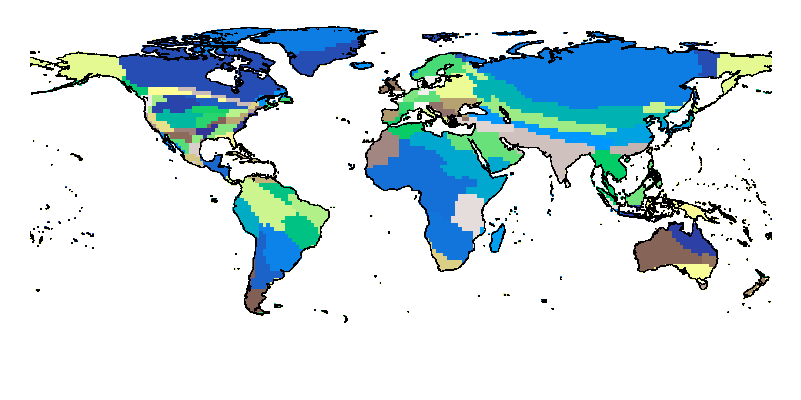}\vspaceclustering
		\caption[]{{\small 
		$\aodha$ }}    
		\label{fig:inat18_aodha}
	\end{subfigure}
	\hfill
	\begin{subfigure}[b]{0.24\textwidth}  
		\centering 
		\includegraphics[width=\textwidth]{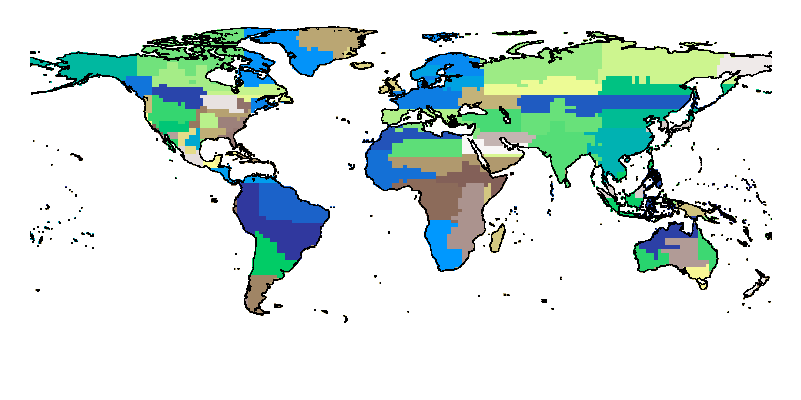}\vspaceclustering
		\caption[]{{\small 
		$\grid$ $(\minscale = 10^{-3})$
		}}    
		\label{fig:inat18_grid_3}
	\end{subfigure}
	\hfill
	\begin{subfigure}[b]{0.24\textwidth}  
		\centering 
		\includegraphics[width=\textwidth]{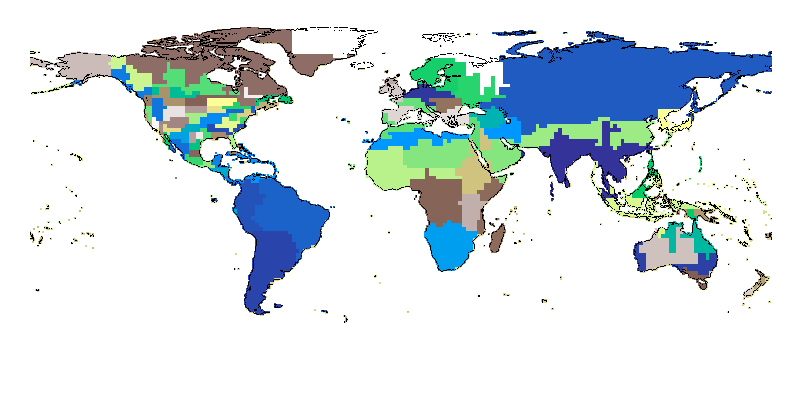}\vspaceclustering
		\caption[]{{\small 
		$\grid$ $(\minscale = 10^{-6})$
		}}    
		\label{fig:inat18_grid}
	\end{subfigure}
	\hfill
	\begin{subfigure}[b]{0.24\textwidth}  
		\centering 
		\includegraphics[width=\textwidth]{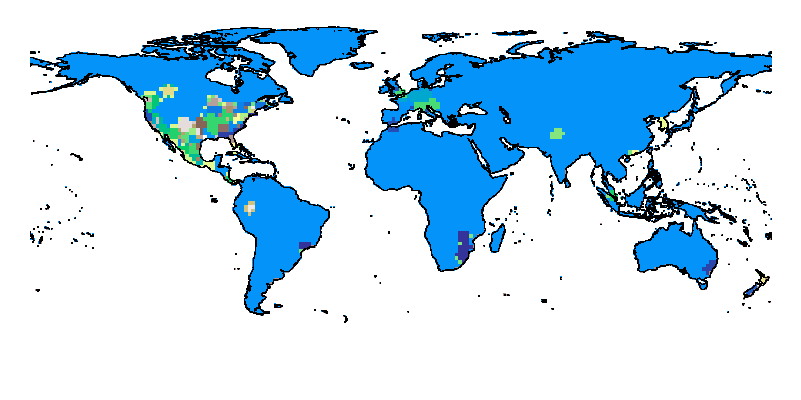}\vspaceclustering
		\caption[]{{\small 
		$\rbf$ $(\sigma = 1, \numkernel = 200)$
		}}    
		\label{fig:inat18_rbf}
	\end{subfigure}
	\hfill
	\begin{subfigure}[b]{0.24\textwidth}  
		\centering 
		\includegraphics[width=\textwidth]{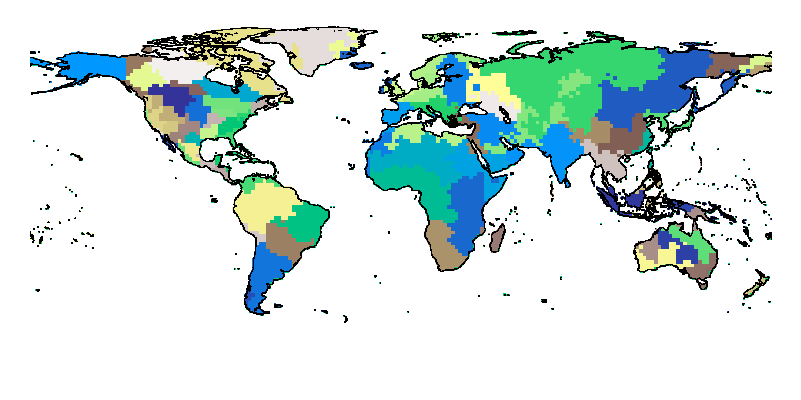}\vspaceclustering
		\caption[]{{\small 
		$\theory$ $ (\minscale = 10^{-3})$
		}}    
		\label{fig:inat18_theory_3}
	\end{subfigure}
	\hfill
	\begin{subfigure}[b]{0.24\textwidth}  
		\centering 
		\includegraphics[width=\textwidth]{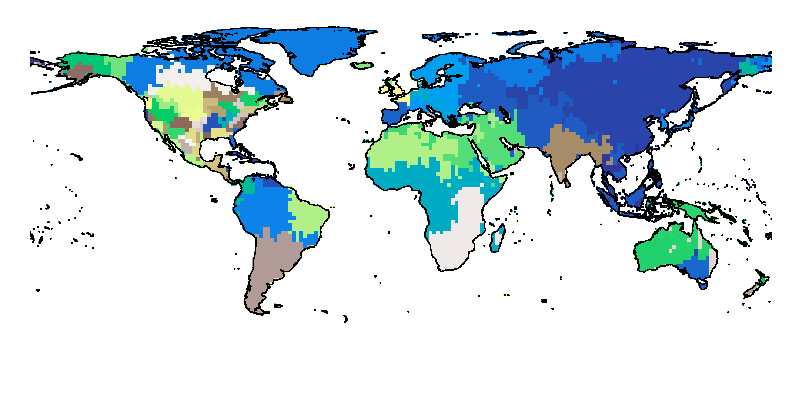}\vspaceclustering
		\caption[]{{\small 
		$\theory$ $ (\minscale = 10^{-6})$
		}}    
		\label{fig:inat18_theory}
	\end{subfigure}
        \hfill
	\begin{subfigure}[b]{0.24\textwidth}  
		\centering 
		\includegraphics[width=\textwidth]{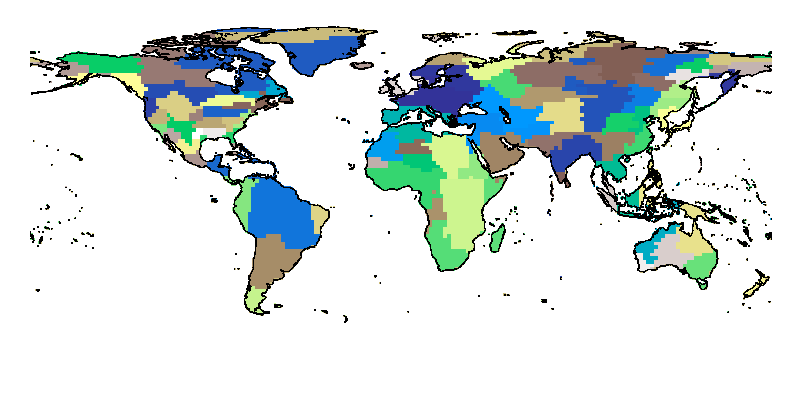}\vspaceclustering
		\caption[]{{\small 
		\nerf{$\nerf$ $ (\nscale = 32)$}
		}}    
		\label{fig:inat18_nerf}
	\end{subfigure}
	\hfill
	\begin{subfigure}[b]{0.24\textwidth}  
		\centering 
		\includegraphics[width=\textwidth]{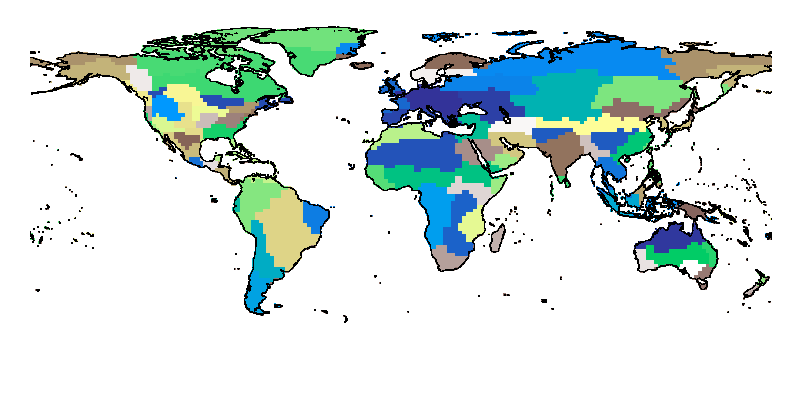}\vspaceclustering
		\caption[]{{\small 
		$\spheremixscale$ $ (\minscale = 10^{-1})$
		}}    
		\label{fig:inat18_spheremixscale_1}
	\end{subfigure}
	\hfill
	\begin{subfigure}[b]{0.24\textwidth}  
		\centering 
		\includegraphics[width=\textwidth]{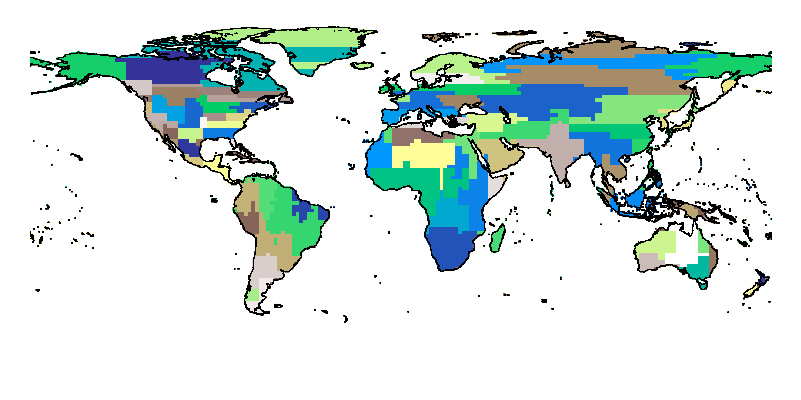}\vspaceclustering
		\caption[]{{\small 
		$\spheremixscale$ $ (\minscale = 10^{-3})$
		}}    
		\label{fig:inat18_spheremixscale_3}
	\end{subfigure}
\hfill
	\begin{subfigure}[b]{0.24\textwidth}  
		\centering 
		\includegraphics[width=\textwidth]{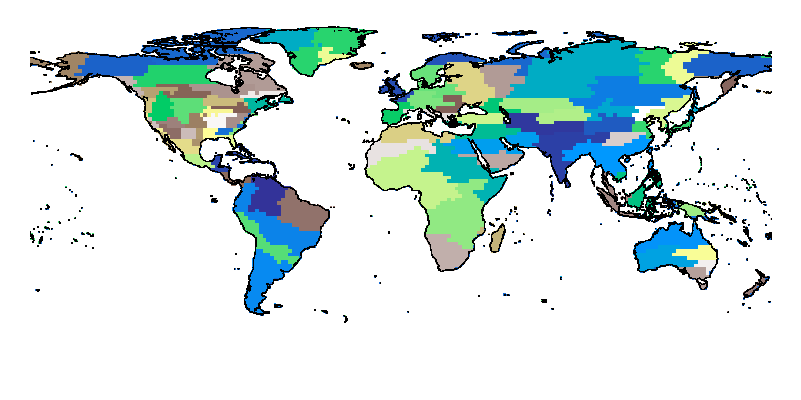}\vspaceclustering
		\caption[]{{\small 
		$\sphere$ $ (\minscale = 10^{-3})$
		}}    
		\label{fig:inat18_sphere}
	\end{subfigure}
	\hfill
	\begin{subfigure}[b]{0.24\textwidth}  
		\centering 
		\includegraphics[width=\textwidth]{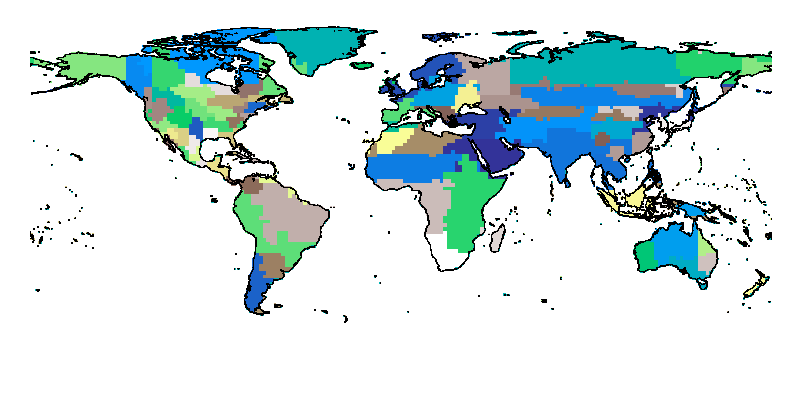}\vspaceclustering
		\caption[]{{\small 
		$\spheregrid$ $ (\minscale = 10^{-3})$
		}}    
		\label{fig:inat18_spheregrid}
	\end{subfigure}
	\hfill
	\begin{subfigure}[b]{0.24\textwidth}  
		\centering 
		\includegraphics[width=\textwidth]{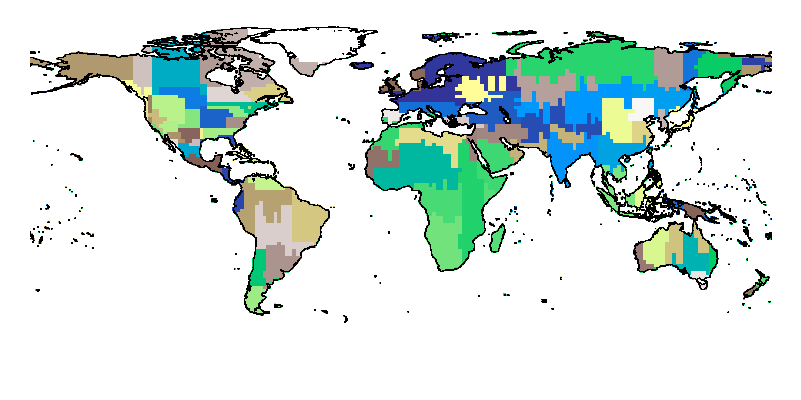}\vspaceclustering
		\caption[]{{\small 
		$\spheregridmixscale$ $ (\minscale = 10^{-3})$
		}}    
		\label{fig:inat18_spheregridmixscale}
	\end{subfigure}
	\hfill
	\begin{subfigure}[b]{0.24\textwidth}  
		\centering 
		\includegraphics[width=\textwidth]{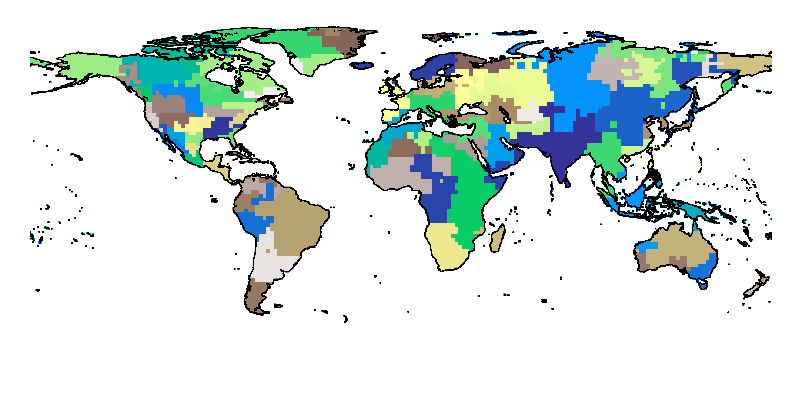}\vspaceclustering
		\caption[]{{\small 
		$\dft$ $ (\minscale = 10^{-3})$
		}}    
		\label{fig:inat18_dft}
	\end{subfigure}

	\caption[]{Embedding clusterings of different location encoders trained on the iNat2018 dataset.
(a) $\aodha$ with 4 hidden ReLU layers of 256 neurons; 
	(d) $\rbf$ with the best kernel size $\kernelsize=1$ and number of anchor points $\numkernel=200$; 
(b)(c)(e)(f) are \emph{\emph{\spacevec}} models \citep{mai2020multiscale} with different min scale $\minscale = \{10^{-6}, 10^{-3}\}$.\textsuperscript{a}
    \mai{(g) is $\nerf$ with $\minscale=32$, and 1 hidden ReLU layer of 512 neurons. }
\mai{(h)-(m)} are \emph{\modelname} models with different min scale $\minscale$.\textsuperscript{b}
} 
	\scriptsize
	\textsuperscript{a} {They share the same best hyperparameters:  $\freq = 64$,  $\maxscale = 1$, and 1 hidden ReLU layer of 512 neurons.} \\
	\textsuperscript{b}{They share the same best hyperparameters: $\freq = 32$, $\maxscale = 1$, and 1 hidden ReLU layers of 1024 neurons.}
	\label{fig:inat18}
\end{figure*}

\subsection{Location Embedding Clustering} \label{sec:locenc_clustering}
To show how the trained location encoders learn the image label distributions, we divide the globe into small latitude-longitude cells and use a location encoder (e.g., \emph{\modelname} or other baseline location encoders) trained on the iNat2017 or iNat2018 dataset to produce a location embedding for the center of each cell. Then we do agglomerative clustering\footnote{\url{https://scikit-learn.org/stable/modules/generated/sklearn.cluster.AgglomerativeClustering.html}} on all these embeddings to produce a clustering map. Figure \ref{fig:inat17} and \ref{fig:inat18} show the clustering results for different models with different hyperparameters on the iNat2017 and iNat2018 \mai{datasets}.

From Figure \ref{fig:inat17}, we can see that:
\begin{enumerate}
    \item In all these clustering maps, nearby locations are clustered together which indicates their location embeddings are similar to each other. This confirms that the learned location encoder can preserve distance information. \item In the $\rbf$ clustering map shown in Figure \ref{fig:inat17_rbf}, except \mai{North America}, almost all the other regions are in the same cluster. This is \mai{because compared with North America, all other regions have fewer training samples.} This indicates that $\rbf$ can not generate a reliable spatial distribution estimation in \mai{data-sparse} regions.
    \item The clustering maps of $\grid$ (Figure \ref{fig:inat17_grid_2} and \ref{fig:inat17_grid}) show horizontal strip-like clusters. More specifically, in Figure \ref{fig:inat17_grid}, the boundaries of many clusters are parallel to the longitude and latitude lines. We \mai{hypothesize} that these kinds of artifacts \mai{are} created because $\grid$ measures the latitude and longitude differences separately (see \mai{Theorem \ref{thm:grid}}) which cannot measure the spherical distance correctly.
    \item $\aodha$ (Figure \ref{fig:inat17_aodha}), $\spheremixscale$ (Figure \ref{fig:inat17_spheremixscale_1}, $\sphere$ (Figure \ref{fig:inat17_sphere}), $\spheregrid$ (Figure \ref{fig:inat17_spheregrid}), $\spheregridmixscale$ (Figure \ref{fig:inat17_spheregridmixscale}), and $\dft$ (Figure \ref{fig:inat17_dft}) show reasonable geographic clustering maps. Each cluster has rather \mai{naturally} looked curvilinear boundaries rather than linear boundaries. We think this reflects the true mixture of different species distributions. However, as we \mai{showed} in Section \ref{sec:locenc_viz_distri}, the single-scale $\aodha$ produces over-generalized distribution while \emph{\modelname} can produce more compact distribution estimation.
\end{enumerate}

Similar conclusions can be \mai{drawn} from Figure \ref{fig:inat18}. We believe those figures visually demonstrate the superiority of \emph{\modelname}.

 \section{Conclusion}  \label{sec:conclusion}
In this work, we propose a \mai{general-purpose} multi-scale spherical location encoder - \emph{\modelname} which can encode any location on the spherical surface into a high \mai{dimensional} vector which is learning-friendly for downstream neuron network models. 
We provide theoretical proof that \emph{\modelname} is able to preserve the spherical surface distance between points. 
\mai{
As a comparison, we also prove that the 2D location encoders such as $\grid$ \citep{gao2018learning,mai2020multiscale} model the latitude and longitude difference of two points separately.
And \nerf-style 3D location encoders \citep{mildenhall2020nerf,schwarz2020graf,niemeyer2021giraffe} 
model the axis-wise differences between two points in 3D Euclidean space separately. Both of them cannot model the true spherical distance.
}
To verify the superiority of \emph{\modelname} in a controlled setting, we generate 20 synthetic datasets and evaluate \emph{\modelname} and all baselines on them. Results show that \emph{\modelname} can outperform all baselines on all 20 sythetic datasets and the error rate reduction can go up to 30.8\%. The results indicate that when the underlying dataset has a larger data bias towards the polar area, we expect a bigger performance improvement of \emph{\modelname}.
We further conduct experiments on three geo-aware image classification tasks with 7 large-scale real-world datasets.
Results shows that \emph{\modelname} can outperform the state-of-the-art 2D location encoders on all 7 datasets. Further analysis shows that \emph{\modelname} is especially excel at polar regions as well data-sparse areas.

Encoding point-features on a spherical surface is a fundamental problem, especially in geoinformatics, geography, meteorology, oceanography, geoscience, and environmental science. Our proposed \emph{\modelname} is a general-purpose spherical-distance-reserving encoding which realizes our idea of directly calculating on the round planet. It can be utilized in a wide range of geospatial prediction tasks. In this work, we only conduct experiments on geo-aware image classification and spatial distribution estimation. Except for the tasks we discussed above, the potential applications include areas like public health, epidemiology, agriculture, economy, ecology, and environmental engineering, and researches like large-scale human mobility and trajectory prediction \citep{xu2018encoding}, geographic question answering \citep{mai2020se}, global biodiversity hotspot prediction \citep{myers2000biodiversity,DiMarcoetal2019,Ceballosetal2020}, weather forecasting and climate change \citep{dupont2021generative,ham2019deep}, global pandemic study and its relation to air pollution \citep{wu2020exposure}, and so on. In general, we expect our proposed \emph{\modelname} will benefit various \textit{AI for social goods}\footnote{\url{https://ai.google/social-good/}} applications which \maigch{involve} predictive modeling at global scales. Moreover, \emph{\modelname} can also contribute \maigch{to} the idea of developing a foundation model for geospatial artificial intelligence \maigch{\citep{mai2022foundation,mai2023opportunities}} in general.

\section*{Declaration of Competing Interest}
The authors declare that they have no known competing financial interests or personal relationships that could have appeared to influence the work reported in this paper.

\section*{Acknowledgement}

We would like to thank Prof. Keith Clarke for his suggestions on different map projection distortion errors and his help on generating Figure \ref{fig:map_proj}.

This work is mainly funded by the National Science Foundation under Grant No. 2033521 A1 – KnowWhereGraph: Enriching and Linking Cross-Domain Knowledge Graphs using Spatially-Explicit AI Technologies. Gengchen Mai acknowledges the support of UCSB Schmidt Summer Research Accelerator Award, Microsoft AI for Earth Grant, the Office of the Director of National Intelligence (ODNI), Intelligence Advanced Research Projects Activity (IARPA) via 2021-2011000004, \maigch{and the Office of Research Internal Research Support Co-funding Grant at University of Georgia}. 
\maigch{Stefano Ermon acknowledges support from NSF (\#1651565), AFOSR (FA95501910024), ARO (W911NF-21-1-0125), Sloan Fellowship, and CZ Biohub.}
Any opinions, findings, conclusions, or recommendations expressed in this material are those of the authors and do not necessarily reflect the views of the National Science Foundation.

\printcredits

\bibliographystyle{cas-model2-names}


\begin{thebibliography}{94}
\expandafter\ifx\csname natexlab\endcsname\relax\def\natexlab#1{#1}\fi
\providecommand{\url}[1]{\texttt{#1}}
\providecommand{\href}[2]{#2}
\providecommand{\path}[1]{#1}
\providecommand{\DOIprefix}{doi:}
\providecommand{\ArXivprefix}{arXiv:}
\providecommand{\URLprefix}{URL: }
\providecommand{\Pubmedprefix}{pmid:}
\providecommand{\doi}[1]{\href{http://dx.doi.org/#1}{\path{#1}}}
\providecommand{\Pubmed}[1]{\href{pmid:#1}{\path{#1}}}
\providecommand{\bibinfo}[2]{#2}
\ifx\xfnm\relax \def\xfnm[#1]{\unskip,\space#1}\fi
\bibitem[{Adams et~al.(2015)Adams, McKenzie and
  Gahegan}]{adams2015frankenplace}
\bibinfo{author}{Adams, B.}, \bibinfo{author}{McKenzie, G.},
  \bibinfo{author}{Gahegan, M.}, \bibinfo{year}{2015}.
\newblock \bibinfo{title}{Frankenplace: interactive thematic mapping for ad hoc
  exploratory search}, in: \bibinfo{booktitle}{Proceedings of the 24th
  international conference on world wide web},
  \bibinfo{organization}{International World Wide Web Conferences Steering
  Committee}. pp. \bibinfo{pages}{12--22}.
\bibitem[{Anokhin et~al.(2021a)Anokhin, Demochkin, Khakhulin, Sterkin,
  Lempitsky and Korzhenkov}]{anokhin2021cips}
\bibinfo{author}{Anokhin, I.}, \bibinfo{author}{Demochkin, K.},
  \bibinfo{author}{Khakhulin, T.}, \bibinfo{author}{Sterkin, G.},
  \bibinfo{author}{Lempitsky, V.}, \bibinfo{author}{Korzhenkov, D.},
  \bibinfo{year}{2021}a.
\newblock \bibinfo{title}{Image generators with conditionally-independent pixel
  synthesis}, in: \bibinfo{booktitle}{Proceedings of the IEEE/CVF Conference on
  Computer Vision and Pattern Recognition}, pp. \bibinfo{pages}{14278--14287}.
\bibitem[{Anokhin et~al.(2021b)Anokhin, Demochkin, Khakhulin, Sterkin,
  Lempitsky and Korzhenkov}]{anokhin2021image}
\bibinfo{author}{Anokhin, I.}, \bibinfo{author}{Demochkin, K.},
  \bibinfo{author}{Khakhulin, T.}, \bibinfo{author}{Sterkin, G.},
  \bibinfo{author}{Lempitsky, V.}, \bibinfo{author}{Korzhenkov, D.},
  \bibinfo{year}{2021}b.
\newblock \bibinfo{title}{Image generators with conditionally-independent pixel
  synthesis}, in: \bibinfo{booktitle}{Proceedings of the IEEE/CVF Conference on
  Computer Vision and Pattern Recognition}, pp. \bibinfo{pages}{14278--14287}.
\bibitem[{Ayush et~al.(2020)Ayush, Uzkent, Meng, Tanmay, Burke, Lobell and
  Ermon}]{ayush2020selfsup}
\bibinfo{author}{Ayush, K.}, \bibinfo{author}{Uzkent, B.},
  \bibinfo{author}{Meng, C.}, \bibinfo{author}{Tanmay, K.},
  \bibinfo{author}{Burke, M.}, \bibinfo{author}{Lobell, D.},
  \bibinfo{author}{Ermon, S.}, \bibinfo{year}{2020}.
\newblock \bibinfo{title}{Geography-aware self-supervised learning}.
\newblock \bibinfo{journal}{arXiv preprint arXiv:2011.09980} .
\bibitem[{Banino et~al.(2018)Banino, Barry, Uria, Blundell, Lillicrap,
  Mirowski, Pritzel, Chadwick, Degris, Modayil et~al.}]{banino2018vector}
\bibinfo{author}{Banino, A.}, \bibinfo{author}{Barry, C.},
  \bibinfo{author}{Uria, B.}, \bibinfo{author}{Blundell, C.},
  \bibinfo{author}{Lillicrap, T.}, \bibinfo{author}{Mirowski, P.},
  \bibinfo{author}{Pritzel, A.}, \bibinfo{author}{Chadwick, M.J.},
  \bibinfo{author}{Degris, T.}, \bibinfo{author}{Modayil, J.}, et~al.,
  \bibinfo{year}{2018}.
\newblock \bibinfo{title}{Vector-based navigation using grid-like
  representations in artificial agents}.
\newblock \bibinfo{journal}{Nature} \bibinfo{volume}{557},
  \bibinfo{pages}{429}.
\bibitem[{Barron et~al.(2021)Barron, Mildenhall, Tancik, Hedman, Martin-Brualla
  and Srinivasan}]{barron2021mipnerf}
\bibinfo{author}{Barron, J.T.}, \bibinfo{author}{Mildenhall, B.},
  \bibinfo{author}{Tancik, M.}, \bibinfo{author}{Hedman, P.},
  \bibinfo{author}{Martin-Brualla, R.}, \bibinfo{author}{Srinivasan, P.P.},
  \bibinfo{year}{2021}.
\newblock \bibinfo{title}{Mip-nerf: A multiscale representation for
  anti-aliasing neural radiance fields}, in: \bibinfo{booktitle}{Proceedings of
  the IEEE/CVF International Conference on Computer Vision}, pp.
  \bibinfo{pages}{5855--5864}.
\bibitem[{Bartnik and Norton(2000)}]{bartnik2000}
\bibinfo{author}{Bartnik, R.}, \bibinfo{author}{Norton, A.},
  \bibinfo{year}{2000}.
\newblock \bibinfo{title}{Numerical methods for the einstein equations in null
  quasi-spherical coordinates}.
\newblock \bibinfo{journal}{SIAM Journal on Scientific Computing}
  \bibinfo{volume}{22}, \bibinfo{pages}{917--950}.
\bibitem[{Basri et~al.(2020)Basri, Galun, Geifman, Jacobs, Kasten and
  Kritchman}]{basri2020frequency}
\bibinfo{author}{Basri, R.}, \bibinfo{author}{Galun, M.},
  \bibinfo{author}{Geifman, A.}, \bibinfo{author}{Jacobs, D.},
  \bibinfo{author}{Kasten, Y.}, \bibinfo{author}{Kritchman, S.},
  \bibinfo{year}{2020}.
\newblock \bibinfo{title}{Frequency bias in neural networks for input of
  non-uniform density}, in: \bibinfo{booktitle}{International Conference on
  Machine Learning}, \bibinfo{organization}{PMLR}. pp.
  \bibinfo{pages}{685--694}.
\bibitem[{Berg et~al.(2014)Berg, Liu, Woo~Lee, Alexander, Jacobs and
  Belhumeur}]{berg2014birdsnap}
\bibinfo{author}{Berg, T.}, \bibinfo{author}{Liu, J.},
  \bibinfo{author}{Woo~Lee, S.}, \bibinfo{author}{Alexander, M.L.},
  \bibinfo{author}{Jacobs, D.W.}, \bibinfo{author}{Belhumeur, P.N.},
  \bibinfo{year}{2014}.
\newblock \bibinfo{title}{{BirdSnap}: Large-scale fine-grained visual
  categorization of birds}, in: \bibinfo{booktitle}{Proceedings of the IEEE
  Conference on Computer Vision and Pattern Recognition}, pp.
  \bibinfo{pages}{2011--2018}.
\bibitem[{Boyer(2012)}]{boyer2012history}
\bibinfo{author}{Boyer, C.B.}, \bibinfo{year}{2012}.
\newblock \bibinfo{title}{History of analytic geometry}.
\newblock \bibinfo{publisher}{Courier Corporation}.
\bibitem[{Bronstein et~al.(2017)Bronstein, Bruna, LeCun, Szlam and
  Vandergheynst}]{bronstein2017geometric}
\bibinfo{author}{Bronstein, M.M.}, \bibinfo{author}{Bruna, J.},
  \bibinfo{author}{LeCun, Y.}, \bibinfo{author}{Szlam, A.},
  \bibinfo{author}{Vandergheynst, P.}, \bibinfo{year}{2017}.
\newblock \bibinfo{title}{Geometric deep learning: going beyond euclidean
  data}.
\newblock \bibinfo{journal}{IEEE Signal Processing Magazine}
  \bibinfo{volume}{34}, \bibinfo{pages}{18--42}.
\bibitem[{Caminade et~al.(2014)Caminade, Kovats, Rocklov, Tompkins, Morse,
  Col{\'o}n-Gonz{\'a}lez, Stenlund, Martens and Lloyd}]{Caminade2014}
\bibinfo{author}{Caminade, C.}, \bibinfo{author}{Kovats, S.},
  \bibinfo{author}{Rocklov, J.}, \bibinfo{author}{Tompkins, A.M.},
  \bibinfo{author}{Morse, A.P.}, \bibinfo{author}{Col{\'o}n-Gonz{\'a}lez,
  F.J.}, \bibinfo{author}{Stenlund, H.}, \bibinfo{author}{Martens, P.},
  \bibinfo{author}{Lloyd, S.J.}, \bibinfo{year}{2014}.
\newblock \bibinfo{title}{Impact of climate change on global malaria
  distribution}.
\newblock \bibinfo{journal}{Proceedings of the National Academy of Sciences}
  \bibinfo{volume}{111}, \bibinfo{pages}{3286--3291}.
\newblock \URLprefix \url{https://www.pnas.org/content/111/9/3286},
  \DOIprefix\doi{10.1073/pnas.1302089111},
  \href{http://arxiv.org/abs/https://www.pnas.org/content/111/9/3286.full.pdf}{\tt
  arXiv:https://www.pnas.org/content/111/9/3286.full.pdf}.
\bibitem[{Ceballos et~al.(2020)Ceballos, Ehrlich and Raven}]{Ceballosetal2020}
\bibinfo{author}{Ceballos, G.}, \bibinfo{author}{Ehrlich, P.R.},
  \bibinfo{author}{Raven, P.H.}, \bibinfo{year}{2020}.
\newblock \bibinfo{title}{Vertebrates on the brink as indicators of biological
  annihilation and the sixth mass extinction}.
\newblock \bibinfo{journal}{Proceedings of the National Academy of Sciences}
  \URLprefix \url{https://www.pnas.org/content/early/2020/05/27/1922686117},
  \DOIprefix\doi{10.1073/pnas.1922686117},
  \href{http://arxiv.org/abs/https://www.pnas.org/content/early/2020/05/27/1922686117.full.pdf}{\tt
  arXiv:https://www.pnas.org/content/early/2020/05/27/1922686117.full.pdf}.
\bibitem[{Chen et~al.(2021)Chen, Liu and Wang}]{chen2021liif}
\bibinfo{author}{Chen, Y.}, \bibinfo{author}{Liu, S.}, \bibinfo{author}{Wang,
  X.}, \bibinfo{year}{2021}.
\newblock \bibinfo{title}{Learning continuous image representation with local
  implicit image function}, in: \bibinfo{booktitle}{Proceedings of the IEEE/CVF
  conference on computer vision and pattern recognition}, pp.
  \bibinfo{pages}{8628--8638}.
\bibitem[{Chinazzi et~al.(2020)Chinazzi, Davis, Ajelli, Gioannini, Litvinova,
  Merler, Pastore~y Piontti, Mu, Rossi, Sun, Viboud, Xiong, Yu, Halloran,
  Longini and Vespignani}]{Chinazzi2020}
\bibinfo{author}{Chinazzi, M.}, \bibinfo{author}{Davis, J.T.},
  \bibinfo{author}{Ajelli, M.}, \bibinfo{author}{Gioannini, C.},
  \bibinfo{author}{Litvinova, M.}, \bibinfo{author}{Merler, S.},
  \bibinfo{author}{Pastore~y Piontti, A.}, \bibinfo{author}{Mu, K.},
  \bibinfo{author}{Rossi, L.}, \bibinfo{author}{Sun, K.},
  \bibinfo{author}{Viboud, C.}, \bibinfo{author}{Xiong, X.},
  \bibinfo{author}{Yu, H.}, \bibinfo{author}{Halloran, M.E.},
  \bibinfo{author}{Longini, I.M.}, \bibinfo{author}{Vespignani, A.},
  \bibinfo{year}{2020}.
\newblock \bibinfo{title}{The effect of travel restrictions on the spread of
  the 2019 novel coronavirus (covid-19) outbreak}.
\newblock \bibinfo{journal}{Science} \bibinfo{volume}{368},
  \bibinfo{pages}{395--400}.
\newblock \URLprefix \url{https://science.sciencemag.org/content/368/6489/395},
  \DOIprefix\doi{10.1126/science.aba9757},
  \href{http://arxiv.org/abs/https://science.sciencemag.org/content/368/6489/395.full.pdf}{\tt
  arXiv:https://science.sciencemag.org/content/368/6489/395.full.pdf}.
\bibitem[{Chrisman(2017)}]{chrisman2017calculating}
\bibinfo{author}{Chrisman, N.R.}, \bibinfo{year}{2017}.
\newblock \bibinfo{title}{Calculating on a round planet}.
\newblock \bibinfo{journal}{International Journal of Geographical Information
  Science} \bibinfo{volume}{31}, \bibinfo{pages}{637--657}.
\bibitem[{Christie et~al.(2018)Christie, Fendley, Wilson and
  Mukherjee}]{christie2018functional}
\bibinfo{author}{Christie, G.}, \bibinfo{author}{Fendley, N.},
  \bibinfo{author}{Wilson, J.}, \bibinfo{author}{Mukherjee, R.},
  \bibinfo{year}{2018}.
\newblock \bibinfo{title}{Functional map of the world}, in:
  \bibinfo{booktitle}{Proceedings of the IEEE Conference on Computer Vision and
  Pattern Recognition}, pp. \bibinfo{pages}{6172--6180}.
\bibitem[{Chu et~al.(2019)Chu, Potetz, Wang, Howard, Song, Brucher, Leung and
  Adam}]{chu2019geo}
\bibinfo{author}{Chu, G.}, \bibinfo{author}{Potetz, B.}, \bibinfo{author}{Wang,
  W.}, \bibinfo{author}{Howard, A.}, \bibinfo{author}{Song, Y.},
  \bibinfo{author}{Brucher, F.}, \bibinfo{author}{Leung, T.},
  \bibinfo{author}{Adam, H.}, \bibinfo{year}{2019}.
\newblock \bibinfo{title}{Geo-aware networks for fine grained recognition}, in:
  \bibinfo{booktitle}{Proceedings of the IEEE International Conference on
  Computer Vision Workshops}, pp. \bibinfo{pages}{0--0}.
\bibitem[{Cohen et~al.(2018)Cohen, Geiger, K{\"o}hler and
  Welling}]{cohen2018spherical}
\bibinfo{author}{Cohen, T.S.}, \bibinfo{author}{Geiger, M.},
  \bibinfo{author}{K{\"o}hler, J.}, \bibinfo{author}{Welling, M.},
  \bibinfo{year}{2018}.
\newblock \bibinfo{title}{Spherical {CNN}s}, in:
  \bibinfo{booktitle}{Proceedings of ICLR 2018}.
\bibitem[{Coors et~al.(2018)Coors, Paul~Condurache and
  Geiger}]{coors2018spherenet}
\bibinfo{author}{Coors, B.}, \bibinfo{author}{Paul~Condurache, A.},
  \bibinfo{author}{Geiger, A.}, \bibinfo{year}{2018}.
\newblock \bibinfo{title}{Sphere{N}et: Learning spherical representations for
  detection and classification in omnidirectional images}, in:
  \bibinfo{booktitle}{Proceedings of the European Conference on Computer Vision
  (ECCV)}, pp. \bibinfo{pages}{518--533}.
\bibitem[{Cueva and Wei(2018)}]{cueva2018emergence}
\bibinfo{author}{Cueva, C.J.}, \bibinfo{author}{Wei, X.X.},
  \bibinfo{year}{2018}.
\newblock \bibinfo{title}{Emergence of grid-like representations by training
  recurrent neural networks to perform spatial localization}, in:
  \bibinfo{booktitle}{International Conference on Learning Representations}.
\bibitem[{Derksen and Izzo(2021)}]{derksen2021snerf}
\bibinfo{author}{Derksen, D.}, \bibinfo{author}{Izzo, D.},
  \bibinfo{year}{2021}.
\newblock \bibinfo{title}{Shadow neural radiance fields for multi-view
  satellite photogrammetry}, in: \bibinfo{booktitle}{Proceedings of the
  IEEE/CVF Conference on Computer Vision and Pattern Recognition}, pp.
  \bibinfo{pages}{1152--1161}.
\bibitem[{Di~Marco et~al.(2019)Di~Marco, Ferrier, Harwood, Hoskins and
  Watson}]{DiMarcoetal2019}
\bibinfo{author}{Di~Marco, M.}, \bibinfo{author}{Ferrier, S.},
  \bibinfo{author}{Harwood, T.D.}, \bibinfo{author}{Hoskins, A.J.},
  \bibinfo{author}{Watson, J.E.M.}, \bibinfo{year}{2019}.
\newblock \bibinfo{title}{Wilderness areas halve the extinction risk of
  terrestrial biodiversity}.
\newblock \bibinfo{journal}{Nature} \bibinfo{volume}{573},
  \bibinfo{pages}{582--585}.
\newblock \URLprefix \url{https://doi.org/10.1038/s41586-019-1567-7},
  \DOIprefix\doi{10.1038/s41586-019-1567-7}.
\bibitem[{Dupont et~al.()Dupont, Golinski, Alizadeh, Teh and
  Doucet}]{dupont2021coin}
\bibinfo{author}{Dupont, E.}, \bibinfo{author}{Golinski, A.},
  \bibinfo{author}{Alizadeh, M.}, \bibinfo{author}{Teh, Y.W.},
  \bibinfo{author}{Doucet, A.}, .
\newblock \bibinfo{title}{Coin: Compression with implicit neural
  representations}, in: \bibinfo{booktitle}{Neural Compression: From
  Information Theory to Applications--Workshop@ ICLR 2021}.
\bibitem[{Dupont et~al.(2021)Dupont, Teh and Doucet}]{dupont2021generative}
\bibinfo{author}{Dupont, E.}, \bibinfo{author}{Teh, Y.W.},
  \bibinfo{author}{Doucet, A.}, \bibinfo{year}{2021}.
\newblock \bibinfo{title}{Generative models as distributions of functions}.
\newblock \bibinfo{journal}{arXiv preprint arXiv:2102.04776} .
\bibitem[{Gao et~al.(2019)Gao, Xie, Zhu and Wu}]{gao2018learning}
\bibinfo{author}{Gao, R.}, \bibinfo{author}{Xie, J.}, \bibinfo{author}{Zhu,
  S.C.}, \bibinfo{author}{Wu, Y.N.}, \bibinfo{year}{2019}.
\newblock \bibinfo{title}{Learning grid cells as vector representation of
  self-position coupled with matrix representation of self-motion}, in:
  \bibinfo{booktitle}{International Conference on Learning Representations}.
\bibitem[{Gupta et~al.(2021)Gupta, Molnar, Xie, Knight and
  Shekhar}]{gupta2021spatial}
\bibinfo{author}{Gupta, J.}, \bibinfo{author}{Molnar, C.},
  \bibinfo{author}{Xie, Y.}, \bibinfo{author}{Knight, J.},
  \bibinfo{author}{Shekhar, S.}, \bibinfo{year}{2021}.
\newblock \bibinfo{title}{Spatial variability aware deep neural networks
  (svann): A general approach}.
\newblock \bibinfo{journal}{ACM Transactions on Intelligent Systems and
  Technology (TIST)} \bibinfo{volume}{12}, \bibinfo{pages}{1--21}.
\bibitem[{Ham et~al.(2019)Ham, Kim and Luo}]{ham2019deep}
\bibinfo{author}{Ham, Y.G.}, \bibinfo{author}{Kim, J.H.}, \bibinfo{author}{Luo,
  J.J.}, \bibinfo{year}{2019}.
\newblock \bibinfo{title}{Deep learning for multi-year enso forecasts}.
\newblock \bibinfo{journal}{Nature} \bibinfo{volume}{573},
  \bibinfo{pages}{568--572}.
\bibitem[{Hansen and Cramer(2015)}]{Hansen&Cramer2015}
\bibinfo{author}{Hansen, G.}, \bibinfo{author}{Cramer, W.},
  \bibinfo{year}{2015}.
\newblock \bibinfo{title}{Global distribution of observed climate change
  impacts}.
\newblock \bibinfo{journal}{Nature Climate Change} \bibinfo{volume}{5},
  \bibinfo{pages}{182--185}.
\newblock \URLprefix \url{https://doi.org/10.1038/nclimate2529},
  \DOIprefix\doi{10.1038/nclimate2529}.
\bibitem[{Harmel(2009)}]{harmel2009nouveau}
\bibinfo{author}{Harmel, A.}, \bibinfo{year}{2009}.
\newblock \bibinfo{title}{Le nouveau syst{\`e}me r{\'e}glementaire lambert 93}.
\newblock \bibinfo{journal}{G{\'e}omatique Expert} \bibinfo{volume}{68},
  \bibinfo{pages}{26--30}.
\bibitem[{He et~al.(2021)He, Wang, Lai, Zhang, Meng, Burke, Lobell and
  Ermon}]{he2021spatial}
\bibinfo{author}{He, Y.}, \bibinfo{author}{Wang, D.}, \bibinfo{author}{Lai,
  N.}, \bibinfo{author}{Zhang, W.}, \bibinfo{author}{Meng, C.},
  \bibinfo{author}{Burke, M.}, \bibinfo{author}{Lobell, D.},
  \bibinfo{author}{Ermon, S.}, \bibinfo{year}{2021}.
\newblock \bibinfo{title}{Spatial-temporal super-resolution of satellite
  imagery via conditional pixel synthesis}.
\newblock \bibinfo{journal}{Advances in Neural Information Processing Systems}
  \bibinfo{volume}{34}, \bibinfo{pages}{27903--27915}.
\bibitem[{Helber et~al.(2019)Helber, Bischke, Dengel and
  Borth}]{helber2019eurosat}
\bibinfo{author}{Helber, P.}, \bibinfo{author}{Bischke, B.},
  \bibinfo{author}{Dengel, A.}, \bibinfo{author}{Borth, D.},
  \bibinfo{year}{2019}.
\newblock \bibinfo{title}{Eurosat: A novel dataset and deep learning benchmark
  for land use and land cover classification}.
\newblock \bibinfo{journal}{IEEE Journal of Selected Topics in Applied Earth
  Observations and Remote Sensing} \bibinfo{volume}{12},
  \bibinfo{pages}{2217--2226}.
\bibitem[{Hu et~al.(2019)Hu, Gao, Lunga, Li, Newsam and Bhaduri}]{hu2019geoai}
\bibinfo{author}{Hu, Y.}, \bibinfo{author}{Gao, S.}, \bibinfo{author}{Lunga,
  D.}, \bibinfo{author}{Li, W.}, \bibinfo{author}{Newsam, S.},
  \bibinfo{author}{Bhaduri, B.}, \bibinfo{year}{2019}.
\newblock \bibinfo{title}{Geoai at acm sigspatial: progress, challenges, and
  future directions}.
\newblock \bibinfo{journal}{Sigspatial Special} \bibinfo{volume}{11},
  \bibinfo{pages}{5--15}.
\bibitem[{Huang et~al.(2023)Huang, Zhang, Mai, Guo and Cui}]{huang2023learning}
\bibinfo{author}{Huang, W.}, \bibinfo{author}{Zhang, D.}, \bibinfo{author}{Mai,
  G.}, \bibinfo{author}{Guo, X.}, \bibinfo{author}{Cui, L.},
  \bibinfo{year}{2023}.
\newblock \bibinfo{title}{Learning urban region representations with pois and
  hierarchical graph infomax}.
\newblock \bibinfo{journal}{ISPRS Journal of Photogrammetry and Remote Sensing}
  \bibinfo{volume}{196}, \bibinfo{pages}{134--145}.
\bibitem[{Izbicki et~al.(2019a)Izbicki, Papalexakis and
  Tsotras}]{izbicki2019exploiting}
\bibinfo{author}{Izbicki, M.}, \bibinfo{author}{Papalexakis, E.E.},
  \bibinfo{author}{Tsotras, V.J.}, \bibinfo{year}{2019}a.
\newblock \bibinfo{title}{Exploiting the earth’s spherical geometry to
  geolocate images}, in: \bibinfo{booktitle}{Joint European Conference on
  Machine Learning and Knowledge Discovery in Databases},
  \bibinfo{organization}{Springer}. pp. \bibinfo{pages}{3--19}.
\bibitem[{Izbicki et~al.(2019b)Izbicki, Papalexakis and
  Tsotras}]{izbicki2019geolocating}
\bibinfo{author}{Izbicki, M.}, \bibinfo{author}{Papalexakis, V.},
  \bibinfo{author}{Tsotras, V.}, \bibinfo{year}{2019}b.
\newblock \bibinfo{title}{Geolocating tweets in any language at any location},
  in: \bibinfo{booktitle}{Proceedings of the 28th ACM International Conference
  on Information and Knowledge Management}, pp. \bibinfo{pages}{89--98}.
\bibitem[{Janowicz et~al.(2020)Janowicz, Gao, McKenzie, Hu and
  Bhaduri}]{janowicz2020geoai}
\bibinfo{author}{Janowicz, K.}, \bibinfo{author}{Gao, S.},
  \bibinfo{author}{McKenzie, G.}, \bibinfo{author}{Hu, Y.},
  \bibinfo{author}{Bhaduri, B.}, \bibinfo{year}{2020}.
\newblock \bibinfo{title}{Geo{AI}: Spatially explicit artificial intelligence
  techniques for geographic knowledge discovery and beyond}.
\bibitem[{Janowicz et~al.(2022)Janowicz, Hitzler, Li, Rehberger, Schildhauer,
  Zhu, Shimizu, Fisher, Cai, Mai et~al.}]{janowicz2022know}
\bibinfo{author}{Janowicz, K.}, \bibinfo{author}{Hitzler, P.},
  \bibinfo{author}{Li, W.}, \bibinfo{author}{Rehberger, D.},
  \bibinfo{author}{Schildhauer, M.}, \bibinfo{author}{Zhu, R.},
  \bibinfo{author}{Shimizu, C.}, \bibinfo{author}{Fisher, C.K.},
  \bibinfo{author}{Cai, L.}, \bibinfo{author}{Mai, G.}, et~al.,
  \bibinfo{year}{2022}.
\newblock \bibinfo{title}{Know, know where, knowwheregraph: A densely
  connected, cross-domain knowledge graph and geo-enrichment service stack for
  applications in environmental intelligence}.
\newblock \bibinfo{journal}{AI Magazine} \bibinfo{volume}{43},
  \bibinfo{pages}{30--39}.
\bibitem[{Kejriwal and Szekely(2017)}]{kejriwal2017neural}
\bibinfo{author}{Kejriwal, M.}, \bibinfo{author}{Szekely, P.},
  \bibinfo{year}{2017}.
\newblock \bibinfo{title}{Neural embeddings for populated geonames locations},
  in: \bibinfo{booktitle}{International Semantic Web Conference},
  \bibinfo{organization}{Springer}. pp. \bibinfo{pages}{139--146}.
\bibitem[{Klocek et~al.(2019)Klocek, Maziarka, Wolczyk, Tabor, Nowak and
  Smieja}]{klocek2019functional}
\bibinfo{author}{Klocek, S.}, \bibinfo{author}{Maziarka, L.},
  \bibinfo{author}{Wolczyk, M.}, \bibinfo{author}{Tabor, J.},
  \bibinfo{author}{Nowak, J.}, \bibinfo{author}{Smieja, M.},
  \bibinfo{year}{2019}.
\newblock \bibinfo{title}{Hypernetwork functional image representation}, in:
  \bibinfo{editor}{Tetko, I.V.}, \bibinfo{editor}{Kurkov{\'{a}}, V.},
  \bibinfo{editor}{Karpov, P.}, \bibinfo{editor}{Theis, F.J.} (Eds.),
  \bibinfo{booktitle}{Artificial Neural Networks and Machine Learning - {ICANN}
  2019 - 28th International Conference on Artificial Neural Networks, Munich,
  Germany, September 17-19, 2019, Proceedings - Workshop and Special Sessions},
  \bibinfo{publisher}{Springer}. pp. \bibinfo{pages}{496--510}.
\bibitem[{Li et~al.(2021)Li, Hsu and Hu}]{li2021tobler}
\bibinfo{author}{Li, W.}, \bibinfo{author}{Hsu, C.Y.}, \bibinfo{author}{Hu,
  M.}, \bibinfo{year}{2021}.
\newblock \bibinfo{title}{Tobler’s first law in geoai: A spatially explicit
  deep learning model for terrain feature detection under weak supervision}.
\newblock \bibinfo{journal}{Annals of the American Association of Geographers}
  \bibinfo{volume}{111}, \bibinfo{pages}{1887--1905}.
\bibitem[{Liu and Biljecki(2022)}]{liu2022review}
\bibinfo{author}{Liu, P.}, \bibinfo{author}{Biljecki, F.},
  \bibinfo{year}{2022}.
\newblock \bibinfo{title}{A review of spatially-explicit geoai applications in
  urban geography}.
\newblock \bibinfo{journal}{International Journal of Applied Earth Observation
  and Geoinformation} \bibinfo{volume}{112}, \bibinfo{pages}{102936}.
\bibitem[{Mac~Aodha et~al.(2019)Mac~Aodha, Cole and Perona}]{mac2019presence}
\bibinfo{author}{Mac~Aodha, O.}, \bibinfo{author}{Cole, E.},
  \bibinfo{author}{Perona, P.}, \bibinfo{year}{2019}.
\newblock \bibinfo{title}{Presence-only geographical priors for fine-grained
  image classification}, in: \bibinfo{booktitle}{Proceedings of the IEEE
  International Conference on Computer Vision}, pp.
  \bibinfo{pages}{9596--9606}.
\bibitem[{Mai et~al.(2022a)Mai, Hu, Gao, Cai, Martins, Scholz, Gao and
  Janowicz}]{mai2022symbolic}
\bibinfo{author}{Mai, G.}, \bibinfo{author}{Hu, Y.}, \bibinfo{author}{Gao, S.},
  \bibinfo{author}{Cai, L.}, \bibinfo{author}{Martins, B.},
  \bibinfo{author}{Scholz, J.}, \bibinfo{author}{Gao, J.},
  \bibinfo{author}{Janowicz, K.}, \bibinfo{year}{2022}a.
\newblock \bibinfo{title}{Symbolic and subsymbolic geoai: Geospatial knowledge
  graphs and spatially explicit machine learning}.
\newblock \bibinfo{journal}{Trans GIS} \bibinfo{volume}{26},
  \bibinfo{pages}{3118--3124}.
\bibitem[{Mai et~al.(2023a)Mai, Huang, Sun, Song, Mishra, Liu, Gao, Liu, Cong,
  Hu et~al.}]{mai2023opportunities}
\bibinfo{author}{Mai, G.}, \bibinfo{author}{Huang, W.}, \bibinfo{author}{Sun,
  J.}, \bibinfo{author}{Song, S.}, \bibinfo{author}{Mishra, D.},
  \bibinfo{author}{Liu, N.}, \bibinfo{author}{Gao, S.}, \bibinfo{author}{Liu,
  T.}, \bibinfo{author}{Cong, G.}, \bibinfo{author}{Hu, Y.}, et~al.,
  \bibinfo{year}{2023}a.
\newblock \bibinfo{title}{On the opportunities and challenges of foundation
  models for geospatial artificial intelligence}.
\newblock \bibinfo{journal}{arXiv preprint arXiv:2304.06798} .
\bibitem[{Mai et~al.(2020a)Mai, Janowicz, Cai, Zhu, Regalia, Yan, Shi and
  Lao}]{mai2020se}
\bibinfo{author}{Mai, G.}, \bibinfo{author}{Janowicz, K.},
  \bibinfo{author}{Cai, L.}, \bibinfo{author}{Zhu, R.},
  \bibinfo{author}{Regalia, B.}, \bibinfo{author}{Yan, B.},
  \bibinfo{author}{Shi, M.}, \bibinfo{author}{Lao, N.}, \bibinfo{year}{2020}a.
\newblock \bibinfo{title}{{SE}-{KGE}: A location-aware knowledge graph
  embedding model for geographic question answering and spatial semantic
  lifting}.
\newblock \bibinfo{journal}{Transactions in GIS}
  \DOIprefix\doi{10.1111/tgis.12629}.
\bibitem[{Mai et~al.(2022b)Mai, Janowicz, Hu, Gao, Yan, Zhu, Cai and
  Lao}]{mai2022review}
\bibinfo{author}{Mai, G.}, \bibinfo{author}{Janowicz, K.}, \bibinfo{author}{Hu,
  Y.}, \bibinfo{author}{Gao, S.}, \bibinfo{author}{Yan, B.},
  \bibinfo{author}{Zhu, R.}, \bibinfo{author}{Cai, L.}, \bibinfo{author}{Lao,
  N.}, \bibinfo{year}{2022}b.
\newblock \bibinfo{title}{A review of location encoding for geoai: methods and
  applications}.
\newblock \bibinfo{journal}{International Journal of Geographical Information
  Science} \bibinfo{volume}{36}, \bibinfo{pages}{639--673}.
\bibitem[{Mai et~al.(2020b)Mai, Janowicz, Yan, Zhu, Cai and
  Lao}]{mai2020multiscale}
\bibinfo{author}{Mai, G.}, \bibinfo{author}{Janowicz, K.},
  \bibinfo{author}{Yan, B.}, \bibinfo{author}{Zhu, R.}, \bibinfo{author}{Cai,
  L.}, \bibinfo{author}{Lao, N.}, \bibinfo{year}{2020}b.
\newblock \bibinfo{title}{Multi-scale representation learning for spatial
  feature distributions using grid cells}, in: \bibinfo{booktitle}{The Eighth
  International Conference on Learning Representations},
  \bibinfo{organization}{openreview}.
\bibitem[{Mai et~al.(2023b)Mai, Jiang, Sun, Zhu, Xuan, Cai, Janowicz, Ermon and
  Lao}]{mai2023towards}
\bibinfo{author}{Mai, G.}, \bibinfo{author}{Jiang, C.}, \bibinfo{author}{Sun,
  W.}, \bibinfo{author}{Zhu, R.}, \bibinfo{author}{Xuan, Y.},
  \bibinfo{author}{Cai, L.}, \bibinfo{author}{Janowicz, K.},
  \bibinfo{author}{Ermon, S.}, \bibinfo{author}{Lao, N.},
  \bibinfo{year}{2023}b.
\newblock \bibinfo{title}{Towards general-purpose representation learning of
  polygonal geometries}.
\newblock \bibinfo{journal}{GeoInformatica} \bibinfo{volume}{27},
  \bibinfo{pages}{289--340}.
\bibitem[{Mai et~al.(2023c)Mai, Lao, He, Song and Ermon}]{mai2023csp}
\bibinfo{author}{Mai, G.}, \bibinfo{author}{Lao, N.}, \bibinfo{author}{He, Y.},
  \bibinfo{author}{Song, J.}, \bibinfo{author}{Ermon, S.},
  \bibinfo{year}{2023}c.
\newblock \bibinfo{title}{Csp: Self-supervised contrastive spatial pre-training
  for geospatial-visual representations}, in: \bibinfo{booktitle}{International
  Conference on Machine Learning}, \bibinfo{organization}{PMLR}.
\bibitem[{Mai et~al.(2019)Mai, Yan, Janowicz and Zhu}]{mai2019relaxing}
\bibinfo{author}{Mai, G.}, \bibinfo{author}{Yan, B.},
  \bibinfo{author}{Janowicz, K.}, \bibinfo{author}{Zhu, R.},
  \bibinfo{year}{2019}.
\newblock \bibinfo{title}{Relaxing unanswerable geographic questions using a
  spatially explicit knowledge graph embedding model}, in:
  \bibinfo{booktitle}{AGILE: The 22nd Annual International Conference on
  Geographic Information Science}, \bibinfo{organization}{Springer}. pp.
  \bibinfo{pages}{21--39}.
\bibitem[{Mai et~al.(2022c)Mai, Cundy, Choi, Hu, Lao and
  Ermon}]{mai2022foundation}
\bibinfo{author}{Mai, G.M.}, \bibinfo{author}{Cundy, C.},
  \bibinfo{author}{Choi, K.}, \bibinfo{author}{Hu, Y.}, \bibinfo{author}{Lao,
  N.}, \bibinfo{author}{Ermon, S.}, \bibinfo{year}{2022}c.
\newblock \bibinfo{title}{Towards a foundation model for geospatial artificial
  intelligence}, in: \bibinfo{booktitle}{Proceedings of the 30th SIGSPATIAL
  international conference on advances in geographic information systems}.
\newblock \DOIprefix\doi{10.1145/3557915.3561043}.
\bibitem[{Mar{\'\i} et~al.(2022)Mar{\'\i}, Facciolo and
  Ehret}]{mari2022satnerf}
\bibinfo{author}{Mar{\'\i}, R.}, \bibinfo{author}{Facciolo, G.},
  \bibinfo{author}{Ehret, T.}, \bibinfo{year}{2022}.
\newblock \bibinfo{title}{Sat-nerf: Learning multi-view satellite
  photogrammetry with transient objects and shadow modeling using rpc cameras},
  in: \bibinfo{booktitle}{Proceedings of the IEEE/CVF Conference on Computer
  Vision and Pattern Recognition}, pp. \bibinfo{pages}{1311--1321}.
\bibitem[{Martin-Brualla et~al.(2021)Martin-Brualla, Radwan, Sajjadi, Barron,
  Dosovitskiy and Duckworth}]{martin2021nerfw}
\bibinfo{author}{Martin-Brualla, R.}, \bibinfo{author}{Radwan, N.},
  \bibinfo{author}{Sajjadi, M.S.}, \bibinfo{author}{Barron, J.T.},
  \bibinfo{author}{Dosovitskiy, A.}, \bibinfo{author}{Duckworth, D.},
  \bibinfo{year}{2021}.
\newblock \bibinfo{title}{Nerf in the wild: Neural radiance fields for
  unconstrained photo collections}, in: \bibinfo{booktitle}{Proceedings of the
  IEEE/CVF Conference on Computer Vision and Pattern Recognition}, pp.
  \bibinfo{pages}{7210--7219}.
\bibitem[{Merilees(1973)}]{merilees1973}
\bibinfo{author}{Merilees, P.E.}, \bibinfo{year}{1973}.
\newblock \bibinfo{title}{The pseudospectral approximation applied to the
  shallow water equations on a sphere}.
\newblock \bibinfo{journal}{Atmosphere} \bibinfo{volume}{11},
  \bibinfo{pages}{13--20}.
\newblock \DOIprefix\doi{10.1080/00046973.1973.9648342}.
\bibitem[{Mildenhall et~al.(2020)Mildenhall, Srinivasan, Tancik, Barron,
  Ramamoorthi and Ng}]{mildenhall2020nerf}
\bibinfo{author}{Mildenhall, B.}, \bibinfo{author}{Srinivasan, P.P.},
  \bibinfo{author}{Tancik, M.}, \bibinfo{author}{Barron, J.T.},
  \bibinfo{author}{Ramamoorthi, R.}, \bibinfo{author}{Ng, R.},
  \bibinfo{year}{2020}.
\newblock \bibinfo{title}{Nerf: Representing scenes as neural radiance fields
  for view synthesis}, in: \bibinfo{booktitle}{ECCV}.
\bibitem[{Mildenhall et~al.(2021)Mildenhall, Srinivasan, Tancik, Barron,
  Ramamoorthi and Ng}]{mildenhall2021nerf}
\bibinfo{author}{Mildenhall, B.}, \bibinfo{author}{Srinivasan, P.P.},
  \bibinfo{author}{Tancik, M.}, \bibinfo{author}{Barron, J.T.},
  \bibinfo{author}{Ramamoorthi, R.}, \bibinfo{author}{Ng, R.},
  \bibinfo{year}{2021}.
\newblock \bibinfo{title}{Nerf: Representing scenes as neural radiance fields
  for view synthesis}.
\newblock \bibinfo{journal}{Communications of the ACM} \bibinfo{volume}{65},
  \bibinfo{pages}{99--106}.
\bibitem[{Morlin-Yron(2017)}]{cnn2017map}
\bibinfo{author}{Morlin-Yron, S.}, \bibinfo{year}{2017}.
\newblock \bibinfo{title}{What’s the real size of africa? how western states
  used maps to downplay size of continent}.
\newblock \bibinfo{journal}{CNN} \URLprefix
  \url{https://www.cnn.com/2016/08/18/africa/real-size-of-africa}.
\bibitem[{Mulcahy and Clarke(2001)}]{mulcahy2001symbolization}
\bibinfo{author}{Mulcahy, K.A.}, \bibinfo{author}{Clarke, K.C.},
  \bibinfo{year}{2001}.
\newblock \bibinfo{title}{Symbolization of map projection distortion: a
  review}.
\newblock \bibinfo{journal}{Cartography and geographic information science}
  \bibinfo{volume}{28}, \bibinfo{pages}{167--182}.
\bibitem[{Myers et~al.(2000)Myers, Mittermeier, Mittermeier, Da~Fonseca and
  Kent}]{myers2000biodiversity}
\bibinfo{author}{Myers, N.}, \bibinfo{author}{Mittermeier, R.A.},
  \bibinfo{author}{Mittermeier, C.G.}, \bibinfo{author}{Da~Fonseca, G.A.},
  \bibinfo{author}{Kent, J.}, \bibinfo{year}{2000}.
\newblock \bibinfo{title}{Biodiversity hotspots for conservation priorities}.
\newblock \bibinfo{journal}{Nature} \bibinfo{volume}{403},
  \bibinfo{pages}{853}.
\bibitem[{Nguyen et~al.(2017)Nguyen, Le, Bui and Phung}]{nguyen2017rf}
\bibinfo{author}{Nguyen, T.D.}, \bibinfo{author}{Le, T.}, \bibinfo{author}{Bui,
  H.}, \bibinfo{author}{Phung, D.}, \bibinfo{year}{2017}.
\newblock \bibinfo{title}{Large-scale online kernel learning with random
  feature reparameterization}, in: \bibinfo{booktitle}{Proceedings of the
  Twenty-Sixth International Joint Conference on Artificial Intelligence,
  {IJCAI-17}}, pp. \bibinfo{pages}{2543--2549}.
\newblock \DOIprefix\doi{10.24963/ijcai.2017/354}.
\bibitem[{Niemeyer and Geiger(2021)}]{niemeyer2021giraffe}
\bibinfo{author}{Niemeyer, M.}, \bibinfo{author}{Geiger, A.},
  \bibinfo{year}{2021}.
\newblock \bibinfo{title}{Giraffe: Representing scenes as compositional
  generative neural feature fields}, in: \bibinfo{booktitle}{Proceedings of the
  IEEE/CVF Conference on Computer Vision and Pattern Recognition}, pp.
  \bibinfo{pages}{11453--11464}.
\bibitem[{Orszag(1972)}]{orszag1972}
\bibinfo{author}{Orszag, S.A.}, \bibinfo{year}{1972}.
\newblock \bibinfo{title}{Comparison of pseudospectral and spectral
  approximation}.
\newblock \bibinfo{journal}{Appl. Math.} \bibinfo{volume}{51},
  \bibinfo{pages}{253--259}.
\bibitem[{Orszag(1974)}]{orszag1974}
\bibinfo{author}{Orszag, S.A.}, \bibinfo{year}{1974}.
\newblock \bibinfo{title}{Fourier series on spheres}.
\newblock \bibinfo{journal}{Mon. Wea. Rev.} \bibinfo{volume}{102},
  \bibinfo{pages}{56–75}.
\bibitem[{Rahaman et~al.(2019)Rahaman, Baratin, Arpit, Draxler, Lin, Hamprecht,
  Bengio and Courville}]{rahaman2019spectral}
\bibinfo{author}{Rahaman, N.}, \bibinfo{author}{Baratin, A.},
  \bibinfo{author}{Arpit, D.}, \bibinfo{author}{Draxler, F.},
  \bibinfo{author}{Lin, M.}, \bibinfo{author}{Hamprecht, F.},
  \bibinfo{author}{Bengio, Y.}, \bibinfo{author}{Courville, A.},
  \bibinfo{year}{2019}.
\newblock \bibinfo{title}{On the spectral bias of neural networks}, in:
  \bibinfo{booktitle}{International Conference on Machine Learning},
  \bibinfo{organization}{PMLR}. pp. \bibinfo{pages}{5301--5310}.
\bibitem[{Rahimi and Recht(2008)}]{rahimi2007rf}
\bibinfo{author}{Rahimi, A.}, \bibinfo{author}{Recht, B.},
  \bibinfo{year}{2008}.
\newblock \bibinfo{title}{Random features for large-scale kernel machines}, in:
  \bibinfo{editor}{Platt, J.C.}, \bibinfo{editor}{Koller, D.},
  \bibinfo{editor}{Singer, Y.}, \bibinfo{editor}{Roweis, S.T.} (Eds.),
  \bibinfo{booktitle}{Advances in Neural Information Processing Systems},
  \bibinfo{publisher}{Curran Associates, Inc.}. pp.
  \bibinfo{pages}{1177--1184}.
\bibitem[{Rahimi et~al.(2007)Rahimi, Recht et~al.}]{rahimi2007random}
\bibinfo{author}{Rahimi, A.}, \bibinfo{author}{Recht, B.}, et~al.,
  \bibinfo{year}{2007}.
\newblock \bibinfo{title}{Random features for large-scale kernel machines.},
  in: \bibinfo{booktitle}{NIPS}, \bibinfo{organization}{Citeseer}.
  p.~\bibinfo{pages}{5}.
\bibitem[{Rao et~al.(2020)Rao, Gao, Kang and Huang}]{rao2020lstm}
\bibinfo{author}{Rao, J.}, \bibinfo{author}{Gao, S.}, \bibinfo{author}{Kang,
  Y.}, \bibinfo{author}{Huang, Q.}, \bibinfo{year}{2020}.
\newblock \bibinfo{title}{Lstm-trajgan: A deep learning approach to trajectory
  privacy protection}, in: \bibinfo{booktitle}{11th International Conference on
  Geographic Information Science (GIScience 2021)-Part I},
  \bibinfo{organization}{Schloss Dagstuhl-Leibniz-Zentrum f{\"u}r Informatik}.
\bibitem[{Sch{\"o}lkopf(2001)}]{scholkopf2001kernel}
\bibinfo{author}{Sch{\"o}lkopf, B.}, \bibinfo{year}{2001}.
\newblock \bibinfo{title}{The kernel trick for distances}, in:
  \bibinfo{booktitle}{Advances in Neural Information Processing Systems}, pp.
  \bibinfo{pages}{301--307}.
\bibitem[{Schwarz et~al.(2020)Schwarz, Liao, Niemeyer and
  Geiger}]{schwarz2020graf}
\bibinfo{author}{Schwarz, K.}, \bibinfo{author}{Liao, Y.},
  \bibinfo{author}{Niemeyer, M.}, \bibinfo{author}{Geiger, A.},
  \bibinfo{year}{2020}.
\newblock \bibinfo{title}{Graf: Generative radiance fields for 3d-aware image
  synthesis}.
\newblock \bibinfo{journal}{Advances in Neural Information Processing Systems}
  \bibinfo{volume}{33}, \bibinfo{pages}{20154--20166}.
\bibitem[{Sokol(2021)}]{nyt2021map}
\bibinfo{author}{Sokol, J.}, \bibinfo{year}{2021}.
\newblock \bibinfo{title}{Can this new map fix our distorted views of the
  world?}
\newblock \bibinfo{journal}{New York Times} \URLprefix
  \url{https://www.nytimes.com/2021/02/24/science/new-world-map.html}.
\bibitem[{Str{\"u}mpler et~al.(2022)Str{\"u}mpler, Postels, Yang, Gool and
  Tombari}]{strumpler2022implicit}
\bibinfo{author}{Str{\"u}mpler, Y.}, \bibinfo{author}{Postels, J.},
  \bibinfo{author}{Yang, R.}, \bibinfo{author}{Gool, L.V.},
  \bibinfo{author}{Tombari, F.}, \bibinfo{year}{2022}.
\newblock \bibinfo{title}{Implicit neural representations for image
  compression}, in: \bibinfo{booktitle}{Computer Vision--ECCV 2022: 17th
  European Conference, Tel Aviv, Israel, October 23--27, 2022, Proceedings,
  Part XXVI}, \bibinfo{organization}{Springer}. pp. \bibinfo{pages}{74--91}.
\bibitem[{Sullivan et~al.(2009)Sullivan, Wood, Iliff, Bonney, Fink and
  Kelling}]{sullivan2009ebird}
\bibinfo{author}{Sullivan, B.L.}, \bibinfo{author}{Wood, C.L.},
  \bibinfo{author}{Iliff, M.J.}, \bibinfo{author}{Bonney, R.E.},
  \bibinfo{author}{Fink, D.}, \bibinfo{author}{Kelling, S.},
  \bibinfo{year}{2009}.
\newblock \bibinfo{title}{ebird: A citizen-based bird observation network in
  the biological sciences}.
\newblock \bibinfo{journal}{Biological conservation} \bibinfo{volume}{142},
  \bibinfo{pages}{2282--2292}.
\bibitem[{Sun et~al.(2014)Sun, Li, Jin and Xie}]{sun2014}
\bibinfo{author}{Sun, C.}, \bibinfo{author}{Li, J.}, \bibinfo{author}{Jin,
  F.F.}, \bibinfo{author}{Xie, F.}, \bibinfo{year}{2014}.
\newblock \bibinfo{title}{Contrasting meridional structures of stratospheric
  and tropospheric planetary wave variability in the northern hemisphere}.
\newblock \bibinfo{journal}{Tellus A: Dynamic Meteorology and Oceanography}
  \bibinfo{volume}{66}, \bibinfo{pages}{25303}.
\bibitem[{Szegedy et~al.(2016)Szegedy, Vanhoucke, Ioffe, Shlens and
  Wojna}]{szegedy2016rethinking}
\bibinfo{author}{Szegedy, C.}, \bibinfo{author}{Vanhoucke, V.},
  \bibinfo{author}{Ioffe, S.}, \bibinfo{author}{Shlens, J.},
  \bibinfo{author}{Wojna, Z.}, \bibinfo{year}{2016}.
\newblock \bibinfo{title}{Rethinking the inception architecture for computer
  vision}, in: \bibinfo{booktitle}{Proceedings of the IEEE conference on
  computer vision and pattern recognition}, pp. \bibinfo{pages}{2818--2826}.
\bibitem[{Tancik et~al.(2022)Tancik, Casser, Yan, Pradhan, Mildenhall,
  Srinivasan, Barron and Kretzschmar}]{tancik2022blocknerf}
\bibinfo{author}{Tancik, M.}, \bibinfo{author}{Casser, V.},
  \bibinfo{author}{Yan, X.}, \bibinfo{author}{Pradhan, S.},
  \bibinfo{author}{Mildenhall, B.}, \bibinfo{author}{Srinivasan, P.P.},
  \bibinfo{author}{Barron, J.T.}, \bibinfo{author}{Kretzschmar, H.},
  \bibinfo{year}{2022}.
\newblock \bibinfo{title}{Block-nerf: Scalable large scene neural view
  synthesis}, in: \bibinfo{booktitle}{Proceedings of the IEEE/CVF Conference on
  Computer Vision and Pattern Recognition}, pp. \bibinfo{pages}{8248--8258}.
\bibitem[{Tancik et~al.(2020)Tancik, Srinivasan, Mildenhall, Fridovich-Keil,
  Raghavan, Singhal, Ramamoorthi, Barron and Ng}]{tancik2020fourier}
\bibinfo{author}{Tancik, M.}, \bibinfo{author}{Srinivasan, P.},
  \bibinfo{author}{Mildenhall, B.}, \bibinfo{author}{Fridovich-Keil, S.},
  \bibinfo{author}{Raghavan, N.}, \bibinfo{author}{Singhal, U.},
  \bibinfo{author}{Ramamoorthi, R.}, \bibinfo{author}{Barron, J.},
  \bibinfo{author}{Ng, R.}, \bibinfo{year}{2020}.
\newblock \bibinfo{title}{Fourier features let networks learn high frequency
  functions in low dimensional domains}, in: \bibinfo{editor}{Larochelle, H.},
  \bibinfo{editor}{Ranzato, M.}, \bibinfo{editor}{Hadsell, R.},
  \bibinfo{editor}{Balcan, M.F.}, \bibinfo{editor}{Lin, H.} (Eds.),
  \bibinfo{booktitle}{Advances in Neural Information Processing Systems},
  \bibinfo{publisher}{Curran Associates, Inc.}. pp.
  \bibinfo{pages}{7537--7547}.
\bibitem[{Tang et~al.(2015)Tang, Paluri, Fei-Fei, Fergus and
  Bourdev}]{tang2015improving}
\bibinfo{author}{Tang, K.}, \bibinfo{author}{Paluri, M.},
  \bibinfo{author}{Fei-Fei, L.}, \bibinfo{author}{Fergus, R.},
  \bibinfo{author}{Bourdev, L.}, \bibinfo{year}{2015}.
\newblock \bibinfo{title}{Improving image classification with location
  context}, in: \bibinfo{booktitle}{Proceedings of the IEEE international
  conference on computer vision}, pp. \bibinfo{pages}{1008--1016}.
\bibitem[{Van~Horn et~al.(2015)Van~Horn, Branson, Farrell, Haber, Barry,
  Ipeirotis, Perona and Belongie}]{van2015building}
\bibinfo{author}{Van~Horn, G.}, \bibinfo{author}{Branson, S.},
  \bibinfo{author}{Farrell, R.}, \bibinfo{author}{Haber, S.},
  \bibinfo{author}{Barry, J.}, \bibinfo{author}{Ipeirotis, P.},
  \bibinfo{author}{Perona, P.}, \bibinfo{author}{Belongie, S.},
  \bibinfo{year}{2015}.
\newblock \bibinfo{title}{Building a bird recognition app and large scale
  dataset with citizen scientists: The fine print in fine-grained dataset
  collection}, in: \bibinfo{booktitle}{Proceedings of the IEEE Conference on
  Computer Vision and Pattern Recognition}, pp. \bibinfo{pages}{595--604}.
\bibitem[{Van~Horn et~al.(2018)Van~Horn, Mac~Aodha, Song, Cui, Sun, Shepard,
  Adam, Perona and Belongie}]{van2018inaturalist}
\bibinfo{author}{Van~Horn, G.}, \bibinfo{author}{Mac~Aodha, O.},
  \bibinfo{author}{Song, Y.}, \bibinfo{author}{Cui, Y.}, \bibinfo{author}{Sun,
  C.}, \bibinfo{author}{Shepard, A.}, \bibinfo{author}{Adam, H.},
  \bibinfo{author}{Perona, P.}, \bibinfo{author}{Belongie, S.},
  \bibinfo{year}{2018}.
\newblock \bibinfo{title}{The inaturalist species classification and detection
  dataset}, in: \bibinfo{booktitle}{Proceedings of the IEEE conference on
  computer vision and pattern recognition}, pp. \bibinfo{pages}{8769--8778}.
\bibitem[{Vaswani et~al.(2017)Vaswani, Shazeer, Parmar, Uszkoreit, Jones,
  Gomez, Kaiser and Polosukhin}]{vaswani2017attention}
\bibinfo{author}{Vaswani, A.}, \bibinfo{author}{Shazeer, N.},
  \bibinfo{author}{Parmar, N.}, \bibinfo{author}{Uszkoreit, J.},
  \bibinfo{author}{Jones, L.}, \bibinfo{author}{Gomez, A.N.},
  \bibinfo{author}{Kaiser, {\L}.}, \bibinfo{author}{Polosukhin, I.},
  \bibinfo{year}{2017}.
\newblock \bibinfo{title}{Attention is all you need}, in:
  \bibinfo{booktitle}{Advances in Neural Information Processing Systems}, pp.
  \bibinfo{pages}{5998--6008}.
\bibitem[{Weyand et~al.(2016)Weyand, Kostrikov and Philbin}]{weyand2016planet}
\bibinfo{author}{Weyand, T.}, \bibinfo{author}{Kostrikov, I.},
  \bibinfo{author}{Philbin, J.}, \bibinfo{year}{2016}.
\newblock \bibinfo{title}{Planet-photo geolocation with convolutional neural
  networks}, in: \bibinfo{booktitle}{European Conference on Computer Vision},
  \bibinfo{organization}{Springer}. pp. \bibinfo{pages}{37--55}.
\bibitem[{Williamson and Browning(1973)}]{williamson1973}
\bibinfo{author}{Williamson, D.}, \bibinfo{author}{Browning, G.},
  \bibinfo{year}{1973}.
\newblock \bibinfo{title}{Comparison of grids and difference approximations for
  numerical weather prediction over a sphere}.
\newblock \bibinfo{journal}{Journal of Applied Meteorology}
  \bibinfo{volume}{12}, \bibinfo{pages}{264--274}.
\bibitem[{Wu et~al.(2020)Wu, Nethery, Sabath, Braun and
  Dominici}]{wu2020exposure}
\bibinfo{author}{Wu, X.}, \bibinfo{author}{Nethery, R.C.},
  \bibinfo{author}{Sabath, B.M.}, \bibinfo{author}{Braun, D.},
  \bibinfo{author}{Dominici, F.}, \bibinfo{year}{2020}.
\newblock \bibinfo{title}{Exposure to air pollution and covid-19 mortality in
  the united states}.
\newblock \bibinfo{journal}{medRxiv} .
\bibitem[{Xiangli et~al.(2022)Xiangli, Xu, Pan, Zhao, Rao, Theobalt, Dai and
  Lin}]{xiangli2022bungeenerf}
\bibinfo{author}{Xiangli, Y.}, \bibinfo{author}{Xu, L.}, \bibinfo{author}{Pan,
  X.}, \bibinfo{author}{Zhao, N.}, \bibinfo{author}{Rao, A.},
  \bibinfo{author}{Theobalt, C.}, \bibinfo{author}{Dai, B.},
  \bibinfo{author}{Lin, D.}, \bibinfo{year}{2022}.
\newblock \bibinfo{title}{Bungeenerf: Progressive neural radiance field for
  extreme multi-scale scene rendering}, in: \bibinfo{booktitle}{Computer
  Vision--ECCV 2022: 17th European Conference, Tel Aviv, Israel, October
  23--27, 2022, Proceedings, Part XXXII}, \bibinfo{organization}{Springer}. pp.
  \bibinfo{pages}{106--122}.
\bibitem[{Xie et~al.(2021)Xie, He, Jia, Bao, Zhou, Ghosh and
  Ravirathinam}]{xie2021statistically}
\bibinfo{author}{Xie, Y.}, \bibinfo{author}{He, E.}, \bibinfo{author}{Jia, X.},
  \bibinfo{author}{Bao, H.}, \bibinfo{author}{Zhou, X.},
  \bibinfo{author}{Ghosh, R.}, \bibinfo{author}{Ravirathinam, P.},
  \bibinfo{year}{2021}.
\newblock \bibinfo{title}{A statistically-guided deep network transformation
  and moderation framework for data with spatial heterogeneity}, in:
  \bibinfo{booktitle}{2021 IEEE International Conference on Data Mining
  (ICDM)}, \bibinfo{organization}{IEEE}. pp. \bibinfo{pages}{767--776}.
\bibitem[{Xu et~al.(2018)Xu, Piao and Gao}]{xu2018encoding}
\bibinfo{author}{Xu, Y.}, \bibinfo{author}{Piao, Z.}, \bibinfo{author}{Gao,
  S.}, \bibinfo{year}{2018}.
\newblock \bibinfo{title}{Encoding crowd interaction with deep neural network
  for pedestrian trajectory prediction}, in: \bibinfo{booktitle}{Proceedings of
  the IEEE Conference on Computer Vision and Pattern Recognition}, pp.
  \bibinfo{pages}{5275--5284}.
\bibitem[{Yan et~al.(2017)Yan, Janowicz, Mai and Gao}]{yan2017itdl}
\bibinfo{author}{Yan, B.}, \bibinfo{author}{Janowicz, K.},
  \bibinfo{author}{Mai, G.}, \bibinfo{author}{Gao, S.}, \bibinfo{year}{2017}.
\newblock \bibinfo{title}{From {ITDL} to {Place2Vec}: Reasoning about place
  type similarity and relatedness by learning embeddings from augmented spatial
  contexts}, in: \bibinfo{booktitle}{Proceedings of the 25th ACM SIGSPATIAL
  International Conference on Advances in Geographic Information Systems},
  \bibinfo{organization}{ACM}. p.~\bibinfo{pages}{35}.
\bibitem[{Yan et~al.(2019a)Yan, Janowicz, Mai and Zhu}]{yan2019spatial}
\bibinfo{author}{Yan, B.}, \bibinfo{author}{Janowicz, K.},
  \bibinfo{author}{Mai, G.}, \bibinfo{author}{Zhu, R.}, \bibinfo{year}{2019}a.
\newblock \bibinfo{title}{A spatially-explicit reinforcement learning model for
  geographic knowledge graph summarization}.
\newblock \bibinfo{journal}{Transactions in GIS} .
\bibitem[{Yan et~al.(2019b)Yan, Ai, Yang and Yin}]{yan2019graph}
\bibinfo{author}{Yan, X.}, \bibinfo{author}{Ai, T.}, \bibinfo{author}{Yang,
  M.}, \bibinfo{author}{Yin, H.}, \bibinfo{year}{2019}b.
\newblock \bibinfo{title}{A graph convolutional neural network for
  classification of building patterns using spatial vector data}.
\newblock \bibinfo{journal}{ISPRS journal of photogrammetry and remote sensing}
  \bibinfo{volume}{150}, \bibinfo{pages}{259--273}.
\bibitem[{Yang and Newsam(2010)}]{yang2010ucmerced}
\bibinfo{author}{Yang, Y.}, \bibinfo{author}{Newsam, S.}, \bibinfo{year}{2010}.
\newblock \bibinfo{title}{Bag-of-visual-words and spatial extensions for
  land-use classification}, in: \bibinfo{booktitle}{Proceedings of the 18th
  SIGSPATIAL international conference on advances in geographic information
  systems}, pp. \bibinfo{pages}{270--279}.
\bibitem[{Zhong et~al.(2020)Zhong, Bepler, Davis and
  Berger}]{zhong2020reconstructing}
\bibinfo{author}{Zhong, E.D.}, \bibinfo{author}{Bepler, T.},
  \bibinfo{author}{Davis, J.H.}, \bibinfo{author}{Berger, B.},
  \bibinfo{year}{2020}.
\newblock \bibinfo{title}{Reconstructing continuous distributions of 3d protein
  structure from cryo-em images}, in: \bibinfo{booktitle}{International
  Conference on Learning Representations}.
\bibitem[{Zhu et~al.(2021)Zhu, Liu, Yao and Fischer}]{zhu2021spatial}
\bibinfo{author}{Zhu, D.}, \bibinfo{author}{Liu, Y.}, \bibinfo{author}{Yao,
  X.}, \bibinfo{author}{Fischer, M.M.}, \bibinfo{year}{2021}.
\newblock \bibinfo{title}{Spatial regression graph convolutional neural
  networks: A deep learning paradigm for spatial multivariate distributions}.
\newblock \bibinfo{journal}{GeoInformatica} , \bibinfo{pages}{1--32}.
\bibitem[{Zhu et~al.(2022)Zhu, Janowicz, Cai and Mai}]{zhu2022reasoning}
\bibinfo{author}{Zhu, R.}, \bibinfo{author}{Janowicz, K.},
  \bibinfo{author}{Cai, L.}, \bibinfo{author}{Mai, G.}, \bibinfo{year}{2022}.
\newblock \bibinfo{title}{Reasoning over higher-order qualitative spatial
  relations via spatially explicit neural networks}.
\newblock \bibinfo{journal}{International Journal of Geographical Information
  Science} \bibinfo{volume}{36}, \bibinfo{pages}{2194--2225}.

\end{thebibliography}

\clearpage

\appendix
\mai{\section{Theoretical Proofs of Each Theorem}  }\label{sec:proof}

\subsection{Proof of Theorem 1} 
\label{sec:proof1}

\begin{proof}
	Given two points $\th_1=(\lon_1,\lat_1), \th_2=(\lon_2,\lat_2) $ on the same sphere $ \coordspasphere^{2} $ with radius $\radius$, we have $\spe{\sphere}_1(\th_i)=[\sin(\lat_i),\cos(\lat_i)\cos(\lon_i),\cos(\lat_i)\sin(\lon_i)]$ for $i=1,2$, 
	the inner product
	\begin{align}
		\begin{split}
			&\langle \spe{\sphere}_1(\th_1),  \spe{\sphere}_1(\th_2) \rangle\\
			&=\sin(\lat_1)\sin(\lat_2)+\cos(\lat_1)\cos(\lon_1)\cos(\lat_2)\cos(\lon_2)\\
			&+\cos(\lat_1)\sin(\lon_1)\cos(\lat_2)\sin(\lon_2)\\
			&=\sin(\lat_1)\sin(\lat_2)+\cos(\lat_1)\cos(\lat_2)\cos(\lon_1-\lon_2)\\
			&=\cos(\ca)=\cos(\sd/\radius),
		\end{split}
	\end{align}
	where $\ca$ is the central angle between $\th_1$and $\th_2$, and the spherical law of cosines is applied to derive the second last equality. 
	So, 
	\begin{align}
		\begin{split}
			&\|\spe{\sphere}_1(\th_1)-\spe{\sphere}_1(\th_2)\|^2\\
			=&\langle  \spe{\sphere}_1(\th_1)- \spe{\sphere}_1(\th_2),  \\ 
			&  \spe{\sphere}_1(\th_1)- \spe{\sphere}_1(\th_2) \rangle\\
			=&2-2\cos(\sd/\radius)\\
			=&4\sin^2(\sd/2\radius).
		\end{split}
	\end{align}
	
	So $\|\spe{\sphere}_1(\th_1)-\spe{\sphere}_1(\th_2)\|=2\sin(\sd/2\radius)$ since $\sd/2\radius\in [0,\frac{\pi}{2}]$.
	By Taylor expansion, $\|\spe{\sphere}_1(\th_1)-\spe{\sphere}_1(\th_2)\|\approx \sd/\radius$ when $\sd$ is small w.r.t. $\radius$.
\end{proof}

\subsection{Proof of Theorem \ref{the:injective}}
\label{sec:proof2}
\begin{proof}
$\forall * \in \{\sphere,\spheregrid,\spheremixscale,\spheregridmixscale\}$, $PE^*_{S}(\th_1)=PE^*_{S}(\th_2)$ implies 
\begin{align}
&\sin(\lat_1)=\sin(\lat_2),	\label{z1=z2}\\
&\cos(\lat_1)\sin(\lon_1)=\cos(\lat_2)\sin(\lon_2),\label{y1=y2}\\
&\cos(\lat_1)\cos(\lon_1)=\cos(\lat_2)\cos(\lon_2),\label{x1=x2}
\end{align}
from $s=0$ terms.
Since $\sin(\lat)$ monotonically increases when $\lat \in   [-{\pi}/{2} , {\pi}/{2}]$, given Equation \ref{z1=z2} we have $\lat_1=\lat_2$. 
If $\lat_1=\lat_2=\pi/2$, then both points are at North Pole, $\lon_1=\lon_2$ equal to whatever longitude defined at North Pole. If $\lat_1=\lat_2=-\pi/2$, it is similar case at South Pole. When $\lat_1=\lat_2\in (-\frac{\pi}{2},\frac{\pi}{2})$, $\cos (\lat_1)=\cos(\lat_2)\neq 0$. Then from Equation  \ref{y1=y2} and \ref{x1=x2}, we have
\begin{align}
	\sin{\lon_1}=\sin(\lon_2),
	\cos(\lon_1)=\cos(\lon_2),
\end{align}
which shows that $\lon_1=\lon_2$. In summary, $\th_1=\th_2$, so $PE^*_{S}$ is injective.\\
If $*=\dft$, 
$PE^*_{S}(\th_1)=PE^*_{S}(\th_2)$ implies  
$\sin(\lat_1)=\sin(\lat_2)$,
	$\cos(\lat_1)=\cos(\lat_2)$,
	$\sin(\lon_1)=\sin(\lon_2)$, and
	$\cos(\lon_1)=\cos(\lon_2)$,
which proves $\th_1=\th_2$ and $PE^*_{\freq}$ is injective directly. 
\end{proof}

\mai{\subsection{Proof of Theorem \ref{thm3}}} 
\label{proof3}
\mai{
\begin{proof}
According to the definition of $\nerf$ encoder  (\ref{equ:nerf}), 
\begin{align}
\begin{split}
&\|\spe{\nerf}_{\nscale}(\th_1)-\spe{\nerf}_{\nscale}(\th_2)\|^2\\
&=\sum_{s=0}^{\nscale-1}\sum_{p\in \{z, x, y\}}      \Big((\sin(2^{s}\pi p_1)-\sin(2^{s}\pi p_2))^2\\
&\qquad+(\cos(2^{s}\pi p_1)-\cos(2^{s}\pi p_2))^2 \Big) \\
&=\sum_{s=0}^{\nscale-1}\sum_{p\in \{z, x, y\}} \Big(\sin^2(2^{s}\pi p_1)+\sin^2(2^{s}\pi p_2)\\
&\qquad-2\sin(2^{s}\pi p_1)\sin(2^{s}\pi p_2)\\
&\qquad+ \cos^2(2^{s}\pi p_1)+\cos^2(2^{s}\pi p_2)\\
&\qquad- 2\cos(2^{s}\pi p_1)\cos(2^{s}\pi p_2) \Big)\\
&=\sum_{s=0}^{\nscale-1} \sum_{p\in \{z, x, y\}} \Big(2-2(\sin(2^{s}\pi p_1)\sin(2^{s}\pi p_2)\\
&\qquad+ \cos(2^{s}\pi p_1)\cos(2^{s}\pi p_2))\Big)\\
&=\sum_{s=0}^{\nscale-1} \sum_{p\in \{z, x, y\}} \Big(2-2\cos(2^{s}\pi (p_1- p_2))\Big)\\
&=\sum_{s=0}^{\nscale-1} \sum_{p\in \{z, x, y\}} 4\sin^2(2^{s-1}\pi (p_1- p_2))\\
&= \sum_{s=0}^{\nscale-1} \Big(4\sin^2(2^{s-1}\pi \Delta\th_z) + 4\sin^2(2^{s-1}\pi \Delta\th_x)  \\
&+ 4\sin^2(2^{s-1}\pi \Delta\th_y)\Big) \\
&= \sum_{s=0}^{\nscale-1} 4\|\vec{Y}_{s}\|^2,
\end{split}
\end{align}
where $\vec{Y}_{s}=[\sin(2^{s-1}\pi \Delta\th_z), \sin(2^{s-1}\pi \Delta\th_x), \sin(2^{s-1}\pi \Delta\th_y)]$.
\end{proof}
}

\end{document}